\documentclass[10pt,journal,compsoc]{IEEEtran}
\ifCLASSOPTIONcompsoc
\usepackage[nocompress]{cite}
\else
\usepackage{cite}
\fi
\usepackage{booktabs} 
\usepackage{multirow}
\usepackage[ruled,linesnumbered]{algorithm2e} 
\usepackage{graphicx}
\usepackage{amssymb}
\usepackage{booktabs}
\usepackage{color}
\usepackage{enumitem}
\usepackage{verbatim}
\usepackage{bbm}
\usepackage{makecell}
\usepackage{array}
\usepackage{multicol}
\usepackage{threeparttable}
\usepackage{xcolor,colortbl}
\usepackage{bm}
\usepackage{bbding}
\usepackage{amsmath,amsfonts}
\usepackage[caption=false,font=normalsize,labelfont=sf,textfont=sf]{subfig}
\usepackage{textcomp}
\usepackage{stfloats}
\usepackage{url}
\usepackage{verbatim}
\usepackage[pagebackref=true,breaklinks=true, colorlinks,bookmarks=false,linkcolor=blue,citecolor=blue]{hyperref}

\SetAlFnt{\small}
\SetAlCapFnt{\small}
\SetAlCapNameFnt{\small}
\SetAlCapHSkip{0pt}
\def\etal{\textit{et al}.}
 
\newcommand{\eg}[0]{{\it e.g.}}
\newcommand{\ie}[0]{{\it i.e.}}

\begin{document}
\title{One-for-All: Towards Universal Domain Translation with a Single StyleGAN}
\author{Yong Du,
Jiahui Zhan, 
Xinzhe Li,  
Junyu Dong,
Sheng Chen,
Ming-Hsuan Yang,
and Shengfeng He
	
	\IEEEcompsocitemizethanks{ 
 \IEEEcompsocthanksitem Yong Du, Xinzhe Li, and Junyu Dong are with the School of Computer Science and Technology, Ocean University of China, Qingdao, China. E-mail: csyongdu@ouc.edu.cn, lixinzhe@stu.ouc.edu.cn, dongjunyu@ouc.edu.cn. 
 
  \IEEEcompsocthanksitem Jiahui Zhan is with the School of Computer Science and Technology, Ocean University of China, and Shanghai Jiao Tong University. E-mail: zhanjiahui@stu.ouc.edu.cn.
		
	\IEEEcompsocthanksitem	Sheng Chen is with the School of Electronics and Computer Science, University of Southampton, Southampton SO17 1BJ, UK. Email: sqc@ecs.soton.ac.uk.

    \IEEEcompsocthanksitem	Ming-Hsuan Yang is with the University of California at Merced, Yonsei University, and Google. E-mail: mhyang@ucmerced.edu.
    
    \IEEEcompsocthanksitem	Shengfeng He is with the School of Computing and Information Systems, Singapore Management University, Singapore. Email: shengfenghe@smu.edu.sg.}}

\IEEEtitleabstractindextext{
\begin{abstract}
In this paper, we propose a novel translation model, UniTranslator, for transforming representations between visually distinct domains under conditions of limited training data and significant visual differences. The main idea behind our approach is leveraging the domain-neutral capabilities of CLIP as a bridging mechanism, while utilizing a separate module to extract abstract, domain-agnostic semantics from the embeddings of both the source and target realms. Fusing these abstract semantics with target-specific semantics results in a transformed embedding within the CLIP space. To bridge the gap between the disparate worlds of CLIP and StyleGAN, we introduce a new non-linear mapper, the CLIP2P mapper. Utilizing CLIP embeddings, this module is tailored to approximate the latent distribution in the StyleGAN's latent space, effectively acting as a connector between these two spaces. The proposed UniTranslator is versatile and capable of performing various tasks, including style mixing, stylization, and translations, even in visually challenging scenarios across different visual domains. Notably, UniTranslator generates high-quality translations that showcase domain relevance, diversity, and improved image quality. UniTranslator surpasses the performance of existing general-purpose models and performs well against specialized models in representative tasks. The source code and trained models will be released to the public. 
\end{abstract}

}

\maketitle

\IEEEdisplaynontitleabstractindextext
\IEEEpeerreviewmaketitle

\IEEEraisesectionheading{\section{Introduction}\label{sec:intro}}
\begin{figure*}
	\centering
	\setlength{\abovecaptionskip}{0cm}
	\centering
	\setlength{\tabcolsep}{0.05em}
	\setlength{\fboxrule}{1pt}
	\setlength{\fboxsep}{0pt}
	\begin{tabular}{cccccccc}		
		\multicolumn{2}{c}{Amedeo Modigliani$\Rightarrow$FFHQ}&
		\multicolumn{2}{c}{Bradypus$\Rightarrow$FFHQ} &
		\multicolumn{2}{c}{AFHQ-wild$\Rightarrow$FFHQ}&
		\multicolumn{2}{c}{LSUN-church $\Rightarrow$ FFHQ} \\	
		\includegraphics[width=.12\linewidth]{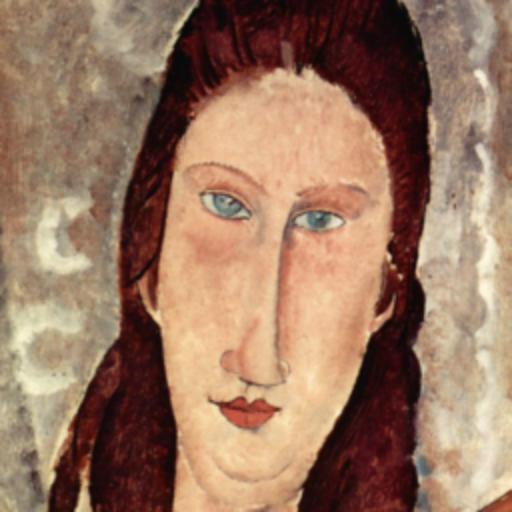}&
		\includegraphics[width=.12\linewidth]{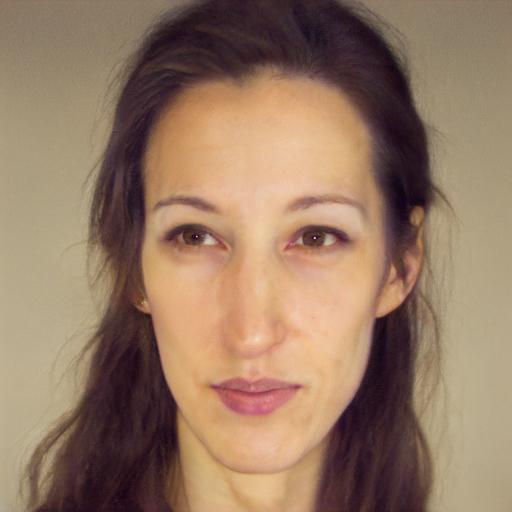}&   	
		\hspace{2mm}\includegraphics[width=.12\linewidth]{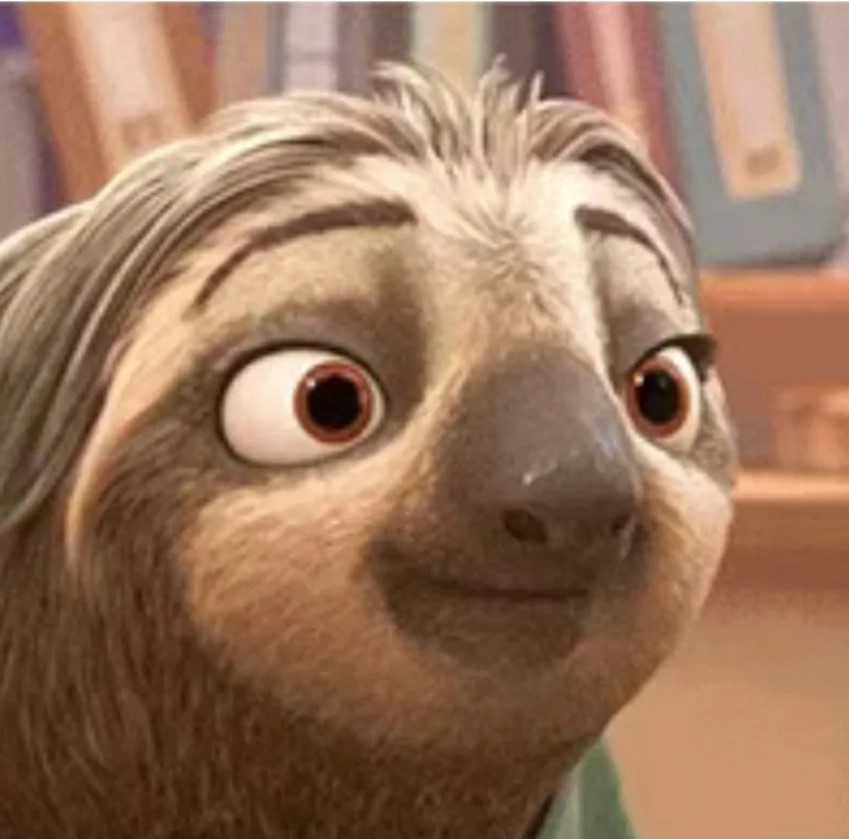}&
		\includegraphics[width=.12\linewidth]{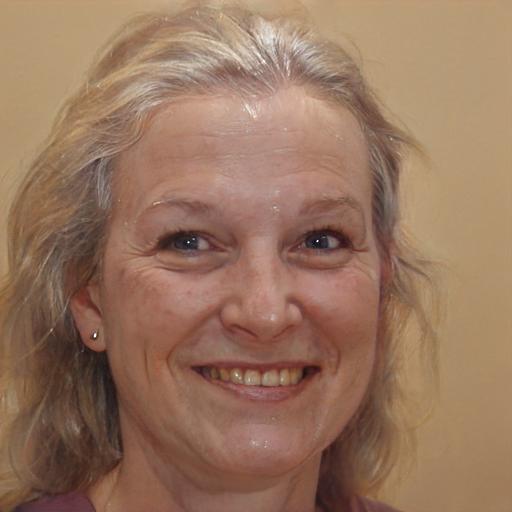}&
		\includegraphics[width=.12\linewidth]{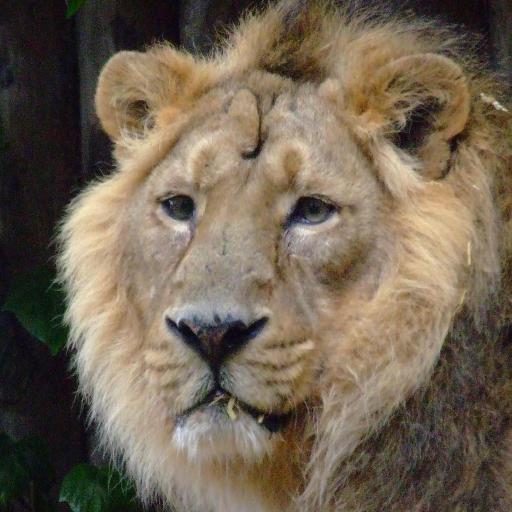}&
		\includegraphics[width=.12\linewidth]{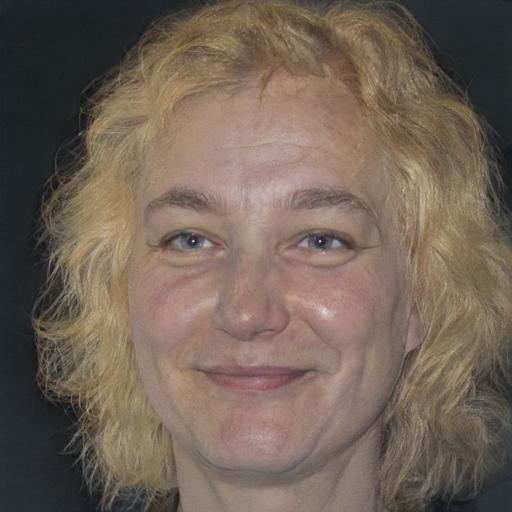}&
		\hspace{2mm}\includegraphics[width=.12\linewidth]{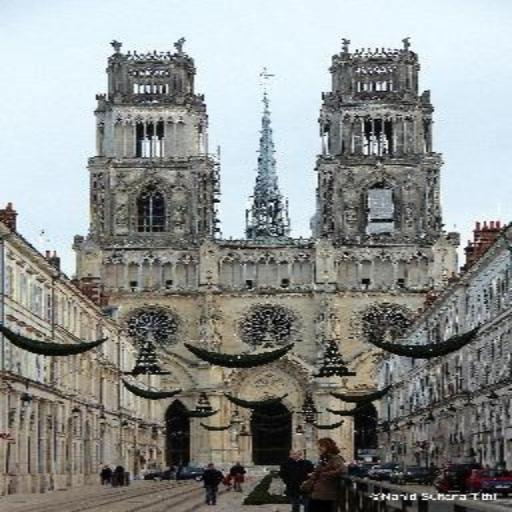}&
		\includegraphics[width=.12\linewidth]{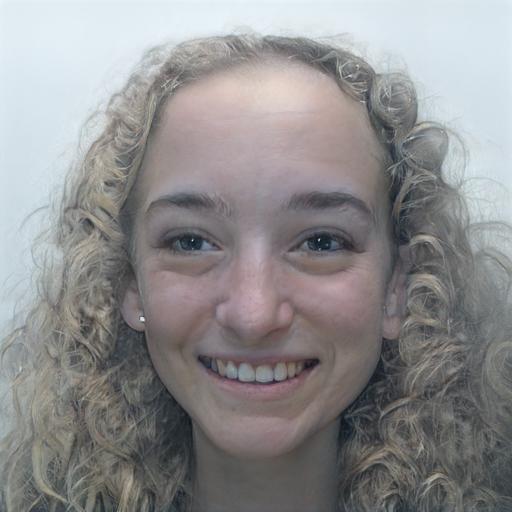}
		\\			
		\multicolumn{2}{c}{AFHQ-cat$\Rightarrow$E621Faces} &
		\multicolumn{2}{c}{D. Johnson$\Rightarrow$AFHQ-dog} &
		\multicolumn{2}{c}{AFHQ-cat$\Rightarrow$Anime}&
		\multicolumn{2}{c}{AFHQ-cat$\Rightarrow$LSUN-church}\\	
		\includegraphics[width=.12\linewidth]{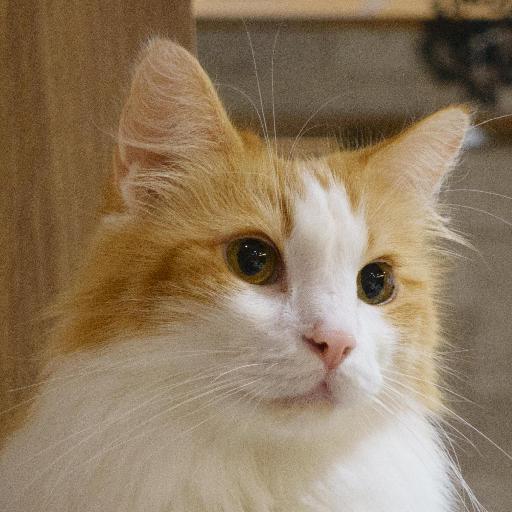}&
		\includegraphics[width=.12\linewidth]{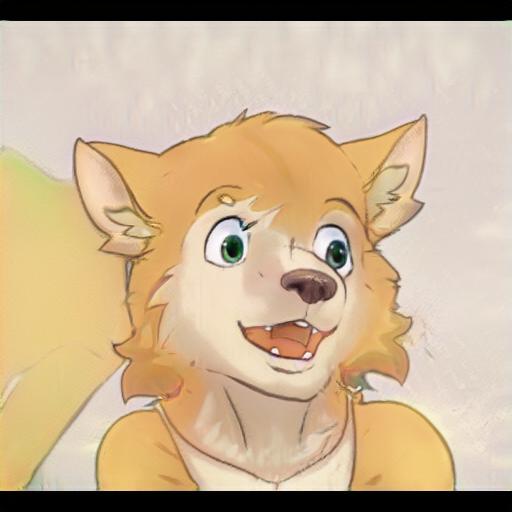}&
		\hspace{2mm}\includegraphics[width=.12\linewidth]{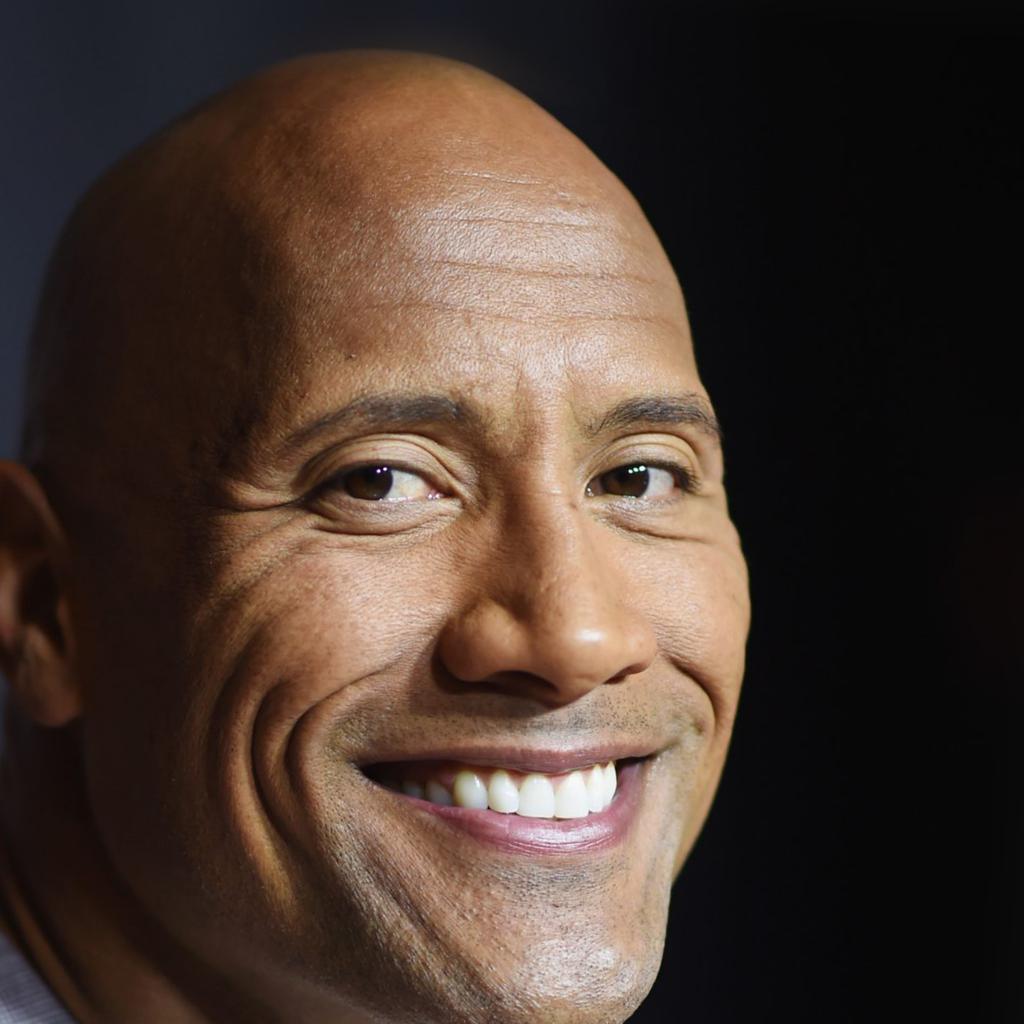}&
		\includegraphics[width=.12\linewidth]{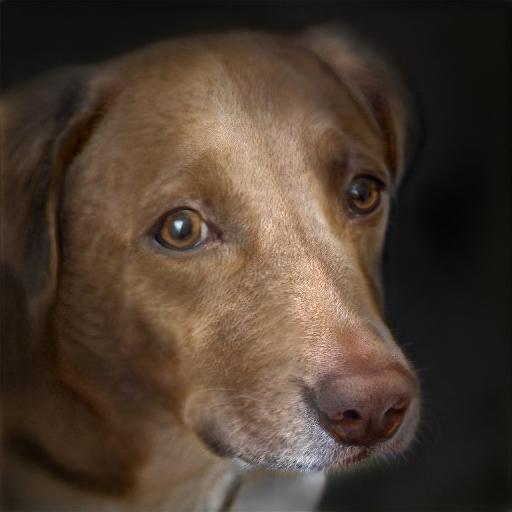}&
		\includegraphics[width=.12\linewidth]{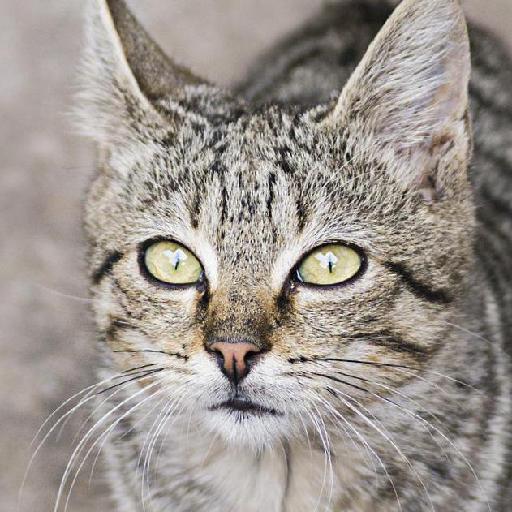}&
		\includegraphics[width=.12\linewidth]{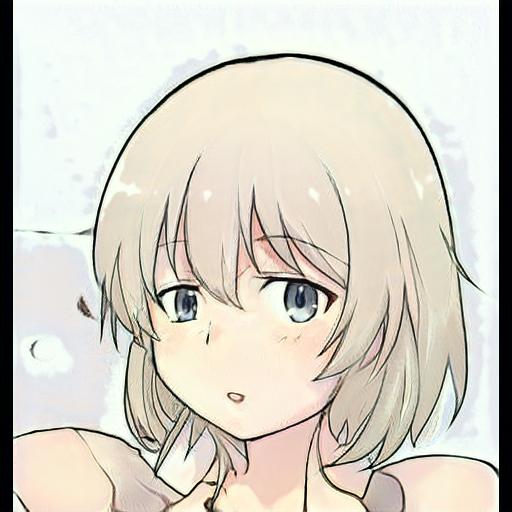}&
		\hspace{2mm}\includegraphics[width=.12\linewidth]{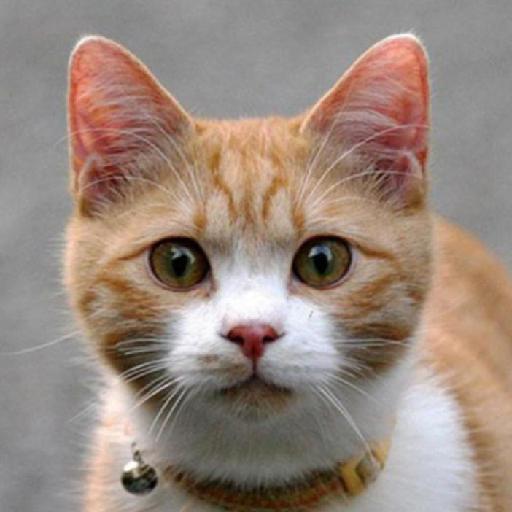}
		&
		\includegraphics[width=.12\linewidth]{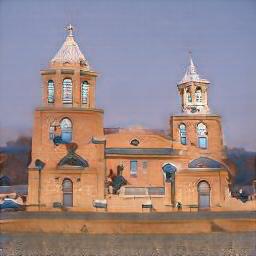}
		\\			
		\multicolumn{8}{c}{$\xrightarrow{\text{\quad\quad\quad\quad\quad\quad\quad\quad\quad\quad\quad\quad\quad\quad\quad\quad\quad\quad\quad\quad\quad\quad\quad\quad\quad\quad\quad\quad\quad\quad\quad\quad\quad\quad\quad\quad\quad\quad\quad\quad\quad\quad\quad\quad\quad\quad\quad\quad\quad\quad\quad\quad\quad\quad\quad\quad\quad\quad\quad\quad\quad\quad\quad\quad\quad\quad\quad\quad\quad}}$}\\
		\multicolumn{2}{c}{\textbf{(a) Adjacent domains}}& \multicolumn{4}{c}{\textbf{(b) Far-off domains}}& \multicolumn{2}{c}{\textbf{(c) Intensely far-off domains}}
	\end{tabular}
	\caption{We introduce UniTranslator, an innovative universal framework for translating across diverse visual domains. It can receive input from any real-world source domain and convert it into a specified target domain, all while ensuring high image quality, domain correspondence, and variability.}
	\label{fig:teaser}
\end{figure*}

\IEEEPARstart{R}{ecent} generative models are developed with a growing emphasis on universality, aiming to enhance the real-world applicability in solving complex challenges~\cite{bao2023one,wang2023detecting,shen2023anything}. Significant advances have been made in visual domain translation~\cite{dalva2022vecgan,chen2022eccv,zhang2022wavelet,ko2022self,kim2022instaformer}, which harnesses the transformation of images by exploiting inherent content correlations across disparate realms. The defining feature of universality for domain translators is their ability to seamlessly convert images from any real-world source domain to a chosen target domain. This pursuit of universality in visual domain translation erodes the barriers segregating different domains and provides invaluable technological support for a wide range of applications. These span from artistic creations such as anthropomorphic or skeuomorphic designs to the entertainment industry, including customized effect generation on various platforms. 

Despite significant progress, existing translators encounter several challenges to achieve universality:
First, most existing techniques necessitate identifying the source and target domains to learn the mappings between them. However, transitioning between domains frequently entails a laborious model retraining process, curtailing their flexibility and applicability. Although recent models such as StarGAN~\cite{choi2018stargan} and StarGAN2~\cite{choi2020stargan} utilize a single model to master many-to-many mappings across all the domains involved, the limitations become apparent in their inability to handle a vast number of domain mappings, thus falling short of achieving true universality.

Second, the training or retraining of existing models typically entails a sizable amount of data. While unsupervised image-to-image translation tasks~\cite{liu2017unsupervised,huang2018multimodal,kim2019u,liu2019few,baek2021rethinking} alleviate the need for paired images, real-world scenarios can still pose formidable challenges, particularly for asymmetric domains. For instance, when leveraging artworks from a specific artist as the source or target domain, the available data may be scant, making it arduous to train the models effectively.
	
\begin{figure*}
	\centering	
	\setlength{\abovecaptionskip}{0cm}
	\centering
	\setlength{\tabcolsep}{0.1em}
	\begin{tabular}{ccccccc}
		\rotatebox[origin=l]{90}{\hspace{7mm}Adjacent}&\includegraphics[width=.16\linewidth]{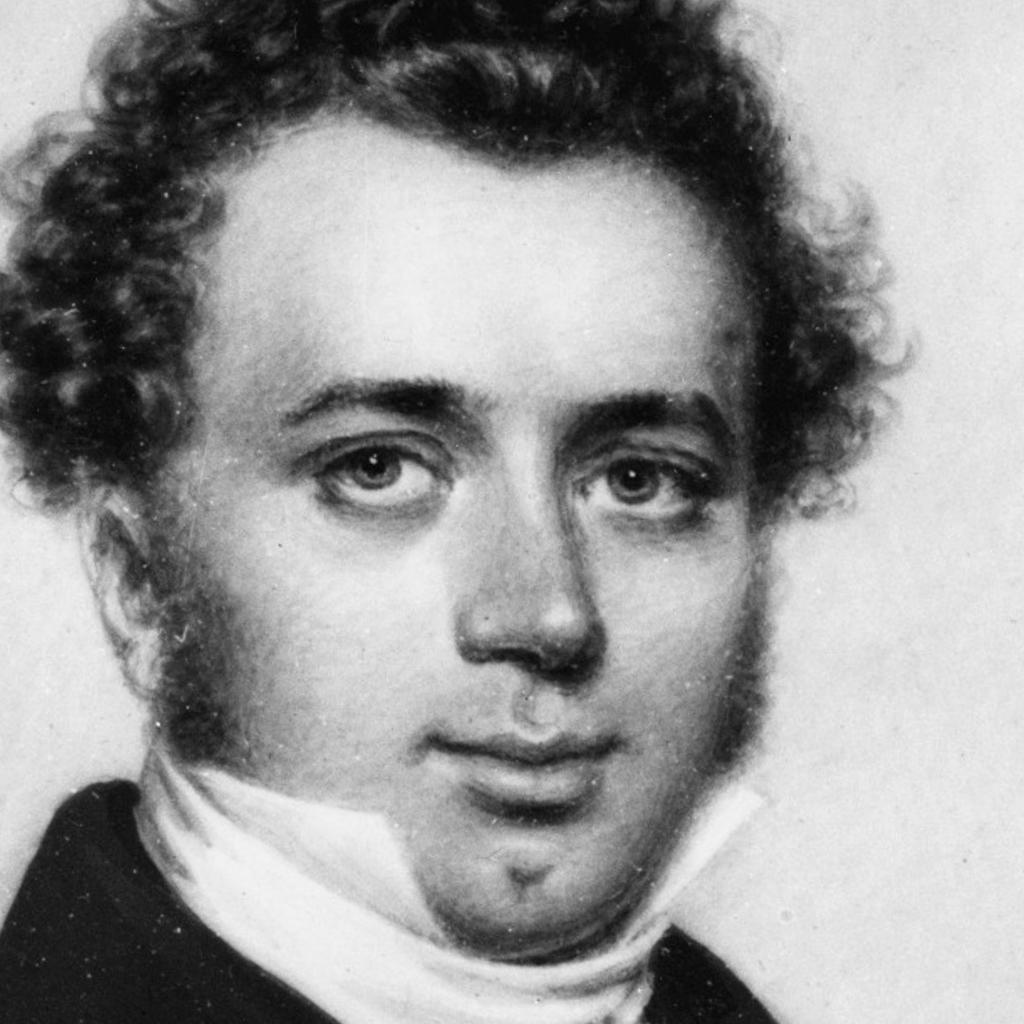} &	
		\includegraphics[width=.16\linewidth]{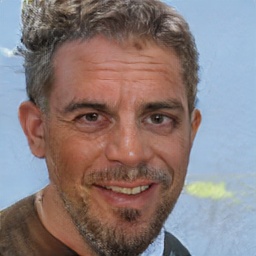} &
		\includegraphics[width=.16\linewidth]{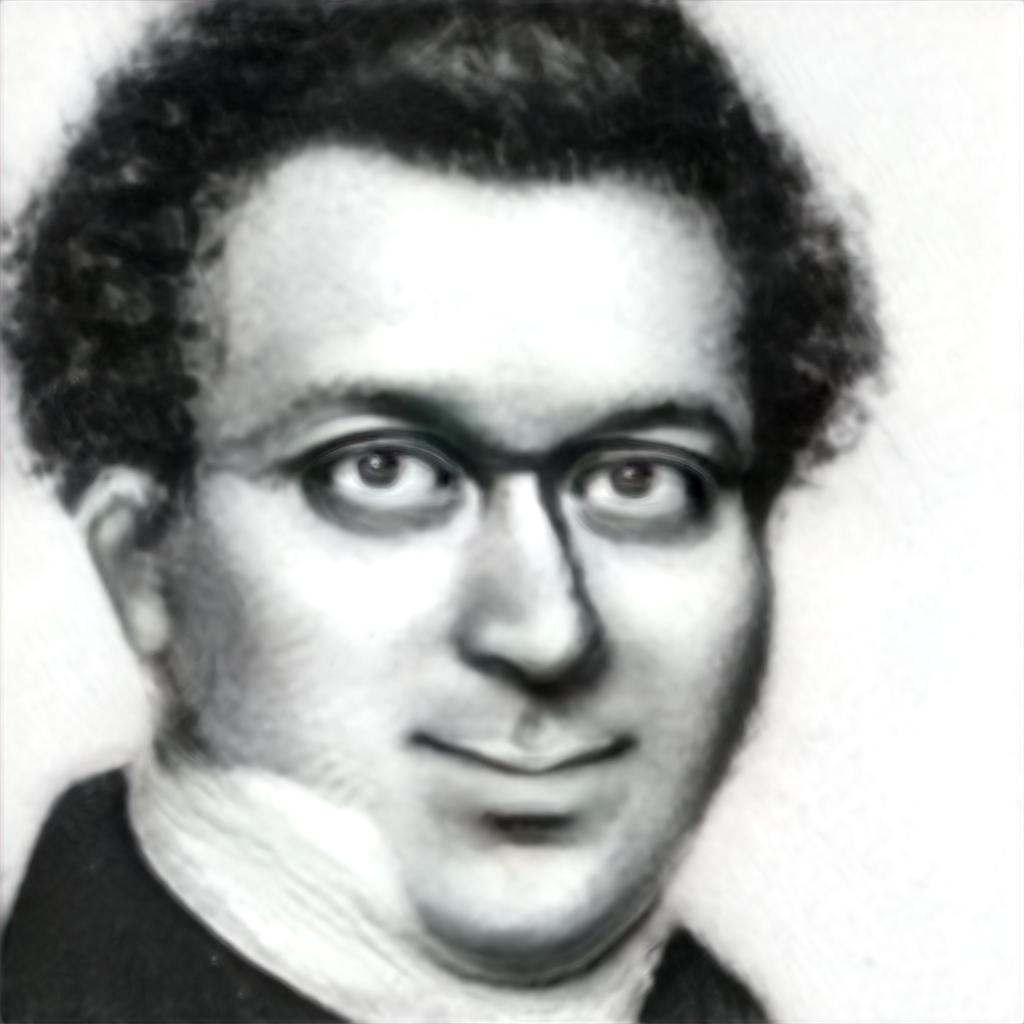}&
		\includegraphics[width=.16\linewidth]{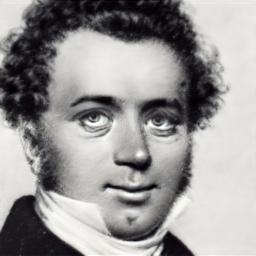}&		
		\includegraphics[width=.16\linewidth]{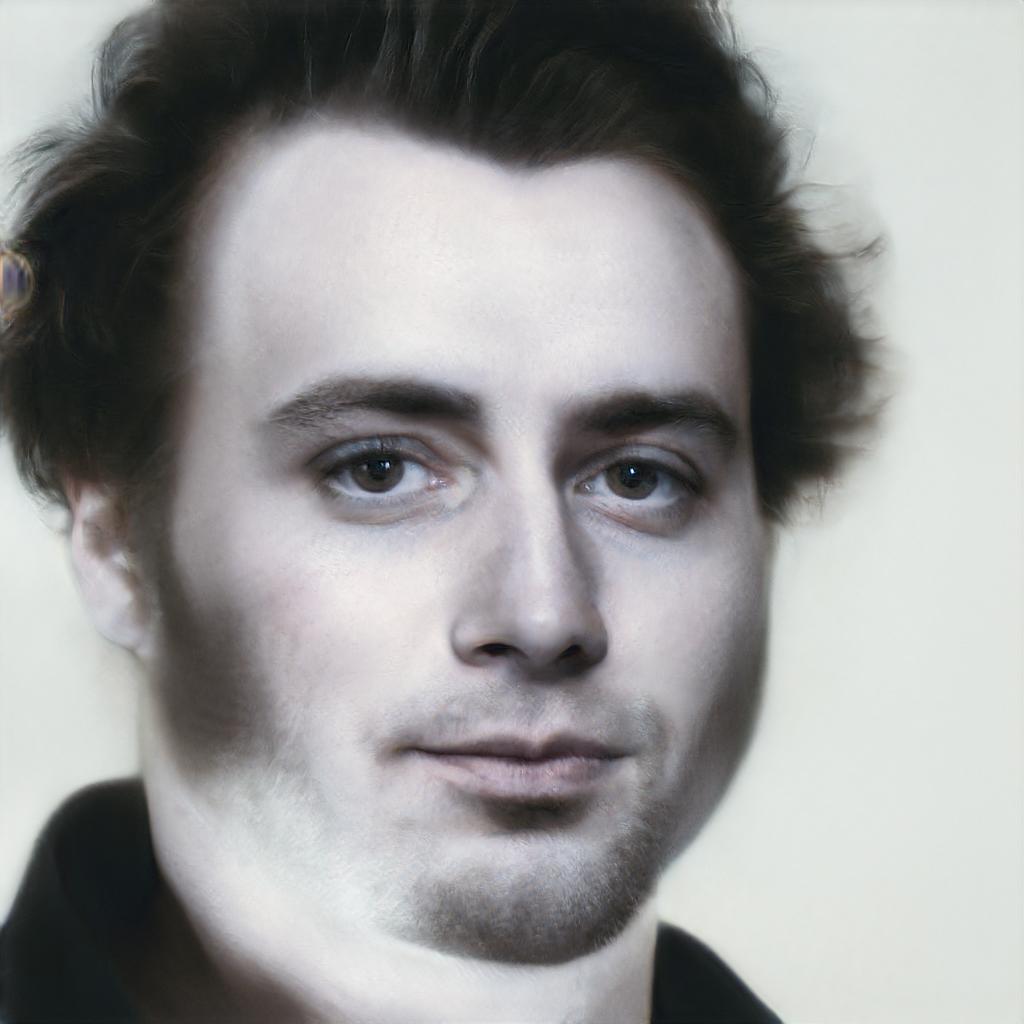}&	
		\includegraphics[width=.16\linewidth]{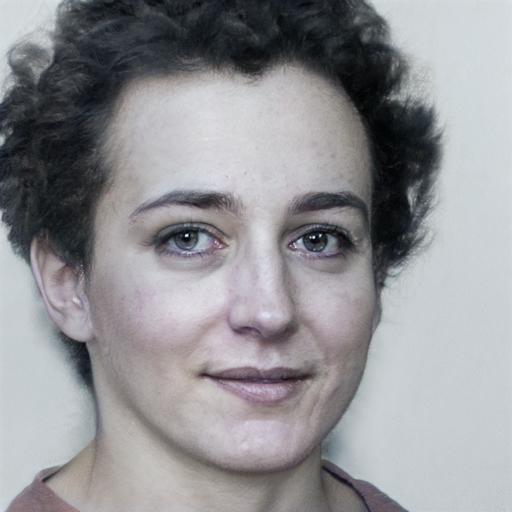} \\
				
		\rotatebox[origin=l]{90}{\hspace{10mm}Far-off}&\includegraphics[width=.16\linewidth]{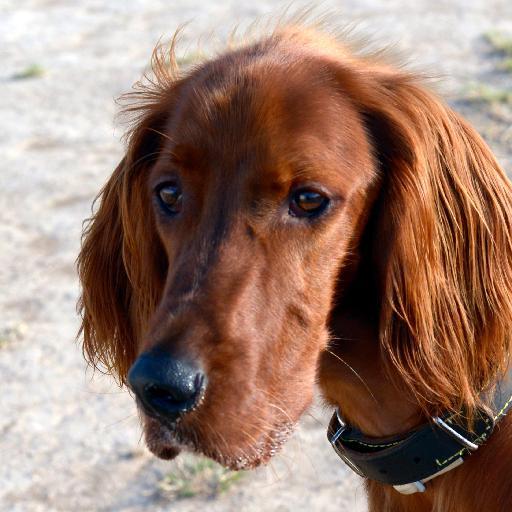} &	
		\includegraphics[width=.16\linewidth]{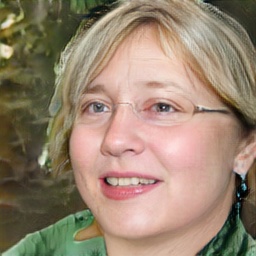} &
		\includegraphics[width=.16\linewidth]{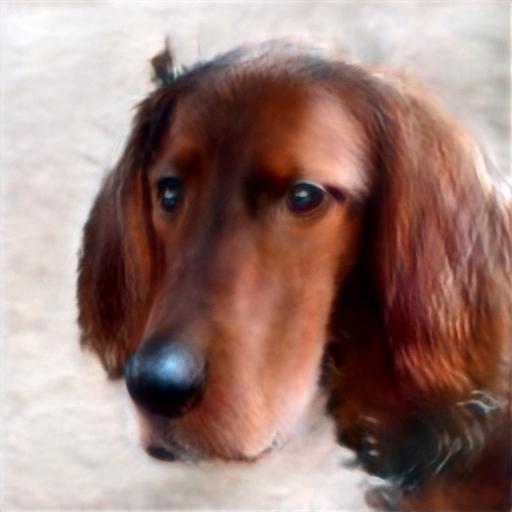}&
		\includegraphics[width=.16\linewidth]{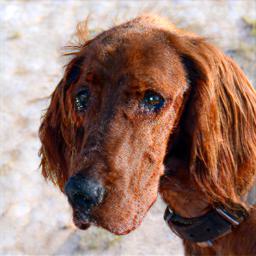}&		
		\includegraphics[width=.16\linewidth]{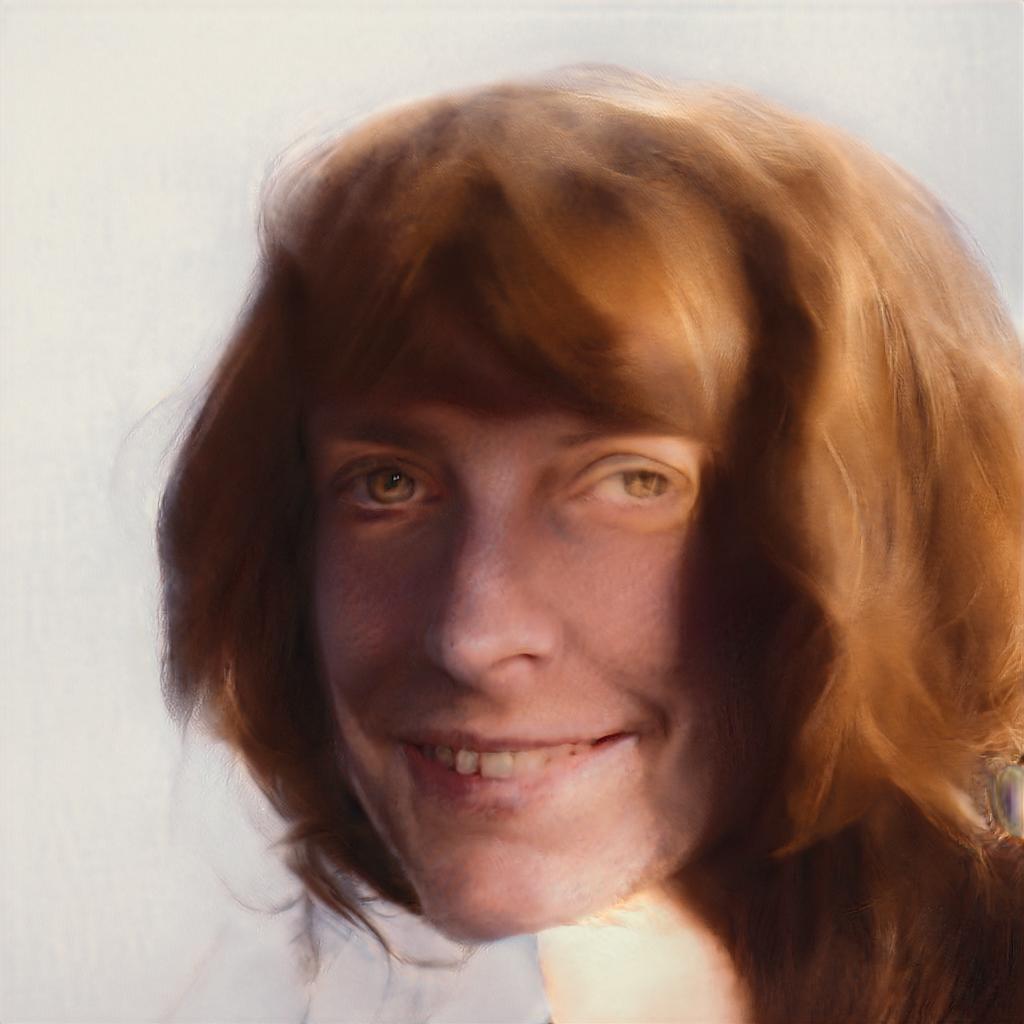}&	
		\includegraphics[width=.16\linewidth]{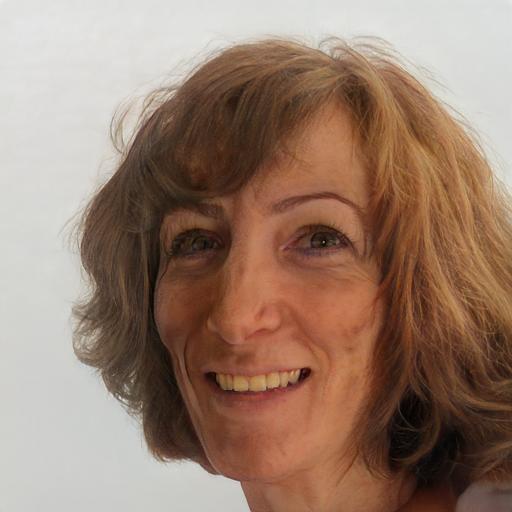} \\	
			
		\rotatebox[origin=l]{90}{\hspace{3mm}Intensely far-off}&\includegraphics[width=.16\linewidth]{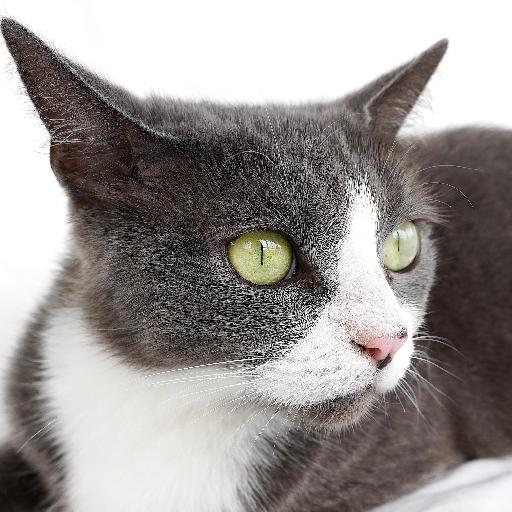} &	
		\includegraphics[width=.16\linewidth]{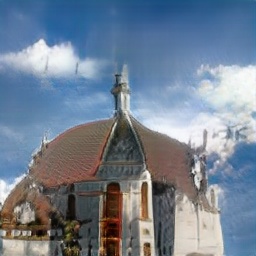} &
		\includegraphics[width=.16\linewidth]{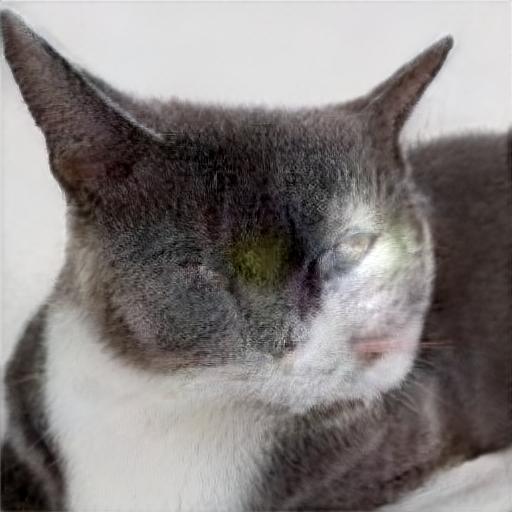}&
		\includegraphics[width=.16\linewidth]{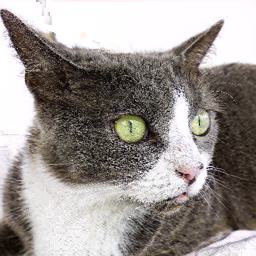}&
		\includegraphics[width=.16\linewidth]{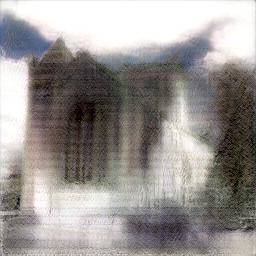}&	
		\includegraphics[width=.16\linewidth]{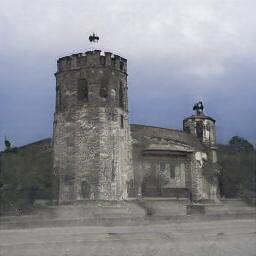} \\
		
		&(a) Source Image & (b) GP-UNIT &(c) DiFa &(d) DiffusionCLIP & (e) PULSE & (f) UniTranslator \\
		&&\cite{yang2022unsupervised}&\cite{zhang2022towards}&		\cite{Kim_2022_CVPR}&\cite{menon2020pulse}&(\textbf{ours})\\	
	\end{tabular}
	\caption{The first row to the last row illustrate visual domain transformations from Metfaces to FFHQ (adjacent domains), AFHQ-dog to FFHQ (far-off domains), and AFHQ-cat to LSUN-church (intensively far-off domains). While GP-UNIT (b) can convert source domain images (a), it suffers from inadequate cross-domain correspondences and compromised image quality. Few-shot (c) or diffusion-based (d) domain adaptation methods display sensitivity to the magnitude of the domain gap. Even in the case of adjacent domains, these methods only result in minor changes to the input image towards the target domain. PULSE (e), lacking decoupling strategies, leaves remnants of source domain patterns when confronted with significant domain gaps. In contrast, UniTranslator (f) consistently achieves high-quality image transformations while upholding domain correspondence despite substantial visual disparities between the domains.}
	\label{fig:intro}
\end{figure*}

Third, few-shot or diffusion-based domain adaptation approaches~\cite{ojha2021few, zhang2022towards, Kim_2022_CVPR, kwon2022diffusion} have been employed for domain translation, building upon inter-domain correlations. These approaches offer the benefits of requiring minimal data for fine-tuning. However, they are primarily suitable for translating between closely related domains like human$\to$sketch. As the scale of the domain gap increases, their effectiveness in accomplishing robust transformations wanes (see Fig.~\ref{fig:intro} (c)-(d)). In real-world translation applications, constraining the input domain to be closely aligned with the target domain would be counterproductive. For example, a platform that generates Internet memes may need to fulfill users' requests to visualize how a cartoon animal might look as a human or to satisfy their curiosity about what a person with lion-like features would resemble, as depicted in Fig.~\ref{fig:teaser} (b). These scenarios involve translating between domains that are far-off but still related, as both belong to the broader category of living beings. In contrast, transforming oil paintings of people into realistic images (Fig.~\ref{fig:teaser} (a)) involves adjacent domains within the same super-category of ``human''.

Moreover, creative gaming applications often demand technology capable of handling even more distant domain translations. For instance, in construction games, users might wish to design buildings that resemble their pets, as shown in Fig.~\ref{fig:teaser} (c). This requires translating between living beings and inanimate objects or artificial structures, representing an intensely far-off domain gap. In such cases, image translation techniques that can manage these extreme domain differences are crucial. Therefore, the primary hurdle to achieving universality is the degree of heterogeneity between the source and target domains. Modeling cross-domain correspondences in terms of shape, appearance, and so on poses challenges, especially when there is a significant chasm between domains.

GP-UNIT~\cite{yang2022unsupervised} can handle domain translation with a large gap among existing translators. It uses generative priors distilled from BigGAN~\cite{brock2019large} to capture coarse-level content correspondences, enabling conversions across highly heterogeneous domains. Nevertheless, GP-UNIT inherits the limitations of BigGAN's latent space, which is not efficiently decoupled, leading to the generation of unrealistic objects. Moreover, due to the intricate division of domains in BigGAN's latent space, the mid-level and fine-level correspondences between various categories of images generated using identical latent codes often fall short. This restriction can affect multi-level cross-domain correspondences in the transformed outcomes with respect to the input (see Fig.~\ref{fig:intro} (b)). As such, it is essential to develop methods to handle domain translation tasks more efficiently, preserving more natural correspondences and ultimately pushing the boundaries of universality in visual domain translation.

In this paper, we propose the UniTranslator, which allows translating a single input image to a target image in a picked domain. It can generate high-quality, visually continuous, diverse translation results even in scenarios with significant visual differences. Our work bears some resemblance to the optimization-based super-resolution method, PULSE~\cite{menon2020pulse}, where an input image serves as a reference to guide the search process in the latent space, and the optimized latent code can then be fed into StyleGAN to generate the proper image. The transformation from input to output in PULSE can be regarded as one type of domain translation, while the generative capability of StyleGAN ensures the quality of the translated result. However, PULSE does not perform well in cross-domain translation due to its lack of efficient mechanisms for handling the inherent correlations between heterogeneous domains. As a consequence, the generated results often exhibit a blending of patterns from both domains (see Fig.~\ref{fig:intro} (e)).

To tackle this issue, we propose a decoupling module in UniTranslator. By integrating descriptive prompts linked to the source image and the intended target domain, this module leverages the language-image alignment proficiency of Contrastive-Language-Image-Pretraining (CLIP)~\cite{radford2021learning} to obtain abstract domain-agnostic semantics. These semantics are then amalgamated with target-specific semantics, refining a target domain embedding that retains cross-domain correlations. However, due to the disparity between the CLIP space and the StyleGAN space, the CLIP embedding may lie beyond the latent domain of StyleGAN, potentially leading to suboptimal conversion results. To overcome this, we devise a mapper that bridges the latent distributions in CLIP and StyleGAN's latent spaces, guided by their statistical properties. This approach effectively translates the meticulously acquired CLIP embedding for the target domain into an appropriate latent code for StyleGAN, thus yielding the desired target image. 
Comprehensive experimentation illustrates UniTranslator's consistent ability to generate highly plausible translation outcomes across diverse visual domains, irrespective of the extent of the domain gap. Moreover, our method exhibits exceptional performance in various real-world applications, including style mixing, stylization, and robust handling of translations even under degraded conditions.

The contributions of this work are:
\begin{itemize}
	\item We introduce UniTranslator for universal visual domain translation. By harnessing the domain-neutral capabilities of CLIP as a bridging conduit, UniTranslator empowers seamless conversions from any real-world source domain to a specified target domain, all accomplished with a single source image.
	\item We propose a decoupling module to efficiently extract cross-domain correspondences and integrate them with target-specific semantics to optimize the expected CLIP embedding.
	\item We design a specialized CLIP2P mapper that connects the CLIP and StyleGAN's spaces. This connection allows us to leverage the powerful generative capability of StyleGAN, leading to high-quality results.
	\item Extensive experiments demonstrate that UniTranslator performs favorably against state-of-the-art models regarding image quality, visual correspondences, and diversity. Furthermore, our method performs effectively in various real-world applications, including style mixing, stylization, and translations, even in challenging scenarios.
\end{itemize}

\section{Related Work}\label{sec:related}
\begin{figure*}[t]
	\centering
	\includegraphics[width=\linewidth]{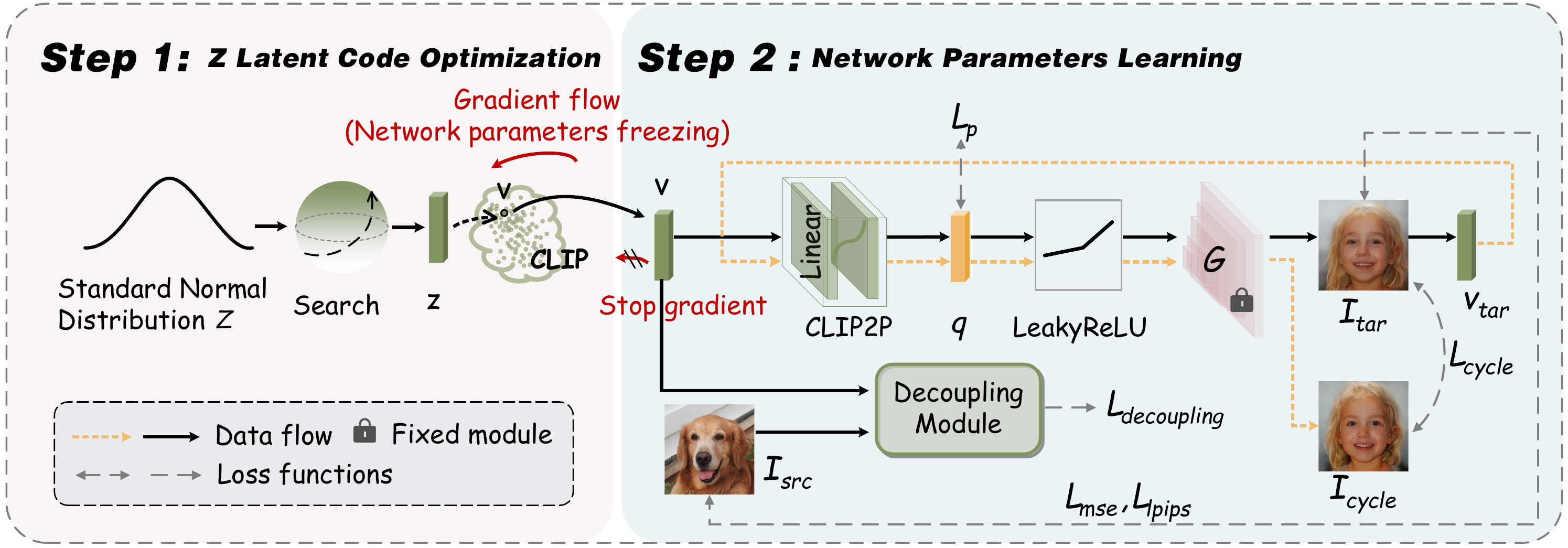}
	\caption{Overview of UniTranslator. It leverages the decoupling module to extract domain-agnostic semantics and integrates them with target-specific information, resulting in a refined CLIP embedding with robust cross-domain correlations. This enhanced CLIP embedding will more effectively guide the search for an optimal $z$ code. Moreover, the CLIP2P mapper is engineered to map the CLIP embedding into $P$ space, reducing the likelihood of it falling outside of StyleGAN's latent space. A demo video is included in the supplementary material. }	
	\label{fig:pipeline}
\end{figure*}

\noindent \textbf{Unsupervised Image-to-Image translation.}
Numerous unsupervised image-to-image translation methods ~\cite{liu2017unsupervised,zhu2017unpaired,huang2018multimodal,liu2019few,choi2018stargan,choi2020stargan,baek2021rethinking,jung2022exploring,zhang2022wavelet,dalva2022vecgan} have been developed to transfer images from the source to target domains. Zhu \etal \cite{zhu2017unpaired} propose the cycle consistency constraints to learn the mapping between the source and target domains without paired data. In \cite{baek2021rethinking}, 
Baek \etal introduce a guiding network for fully unsupervised scenarios. Choi \etal \cite{choi2018stargan,choi2020stargan} develop StarGAN, a model integrating multiple mappings, yet it requires data from multiple domains and cannot handle unseen inputs. On the other hand, Liu \etal \cite{liu2019few} present FUNIT for few-shot translation to unseen classes.

Notably, these methods do not adapt well to situations where the target domain significantly differs from the source domain. GP-UNIT \cite{yang2022unsupervised} specifically focuses on constructing pose mappings between complex domains with notable visual discrepancies. However, it is less effective in dealing with domains beyond the scope of ImageNet or intensely far-off domains. In contrast, our UniTranslator, as a hybrid technique, relies solely on a single reference image from the source domain during training, thereby circumventing limitations imposed by training data. Furthermore, it maintains high-level visual continuity across visual realms, rendering it highly suitable for translating between significantly distant domains.

\noindent \textbf{Few shot/text-driven domain adaptation.}
A plethora of 
few-shot/text-driven domain adaptation
methods have recently been proposed  \cite{ojha2021few,xiao2022few,gal2022stylegan,zhumind,zhang2022towards,zhao2022closer,zhao2022fewshot,mondal2023fewshot} to train a generator using limited examples or text prompts. Ojha \etal \cite{ojha2021few} propose fine-tuning a pre-trained source generator using 10 target images to preserve relative similarities and differences in the source domain while avoiding mode collapse and 
Xiao \etal \cite{xiao2022few} consider spatial structural alignment between domains to enhance generative quality. In \cite{gal2022stylegan}, Gal \etal introduce a text-guided adaptation method that constructs a directional CLIP loss based on a collinear relationship, and Zhu \etal \cite{zhumind} further improve this relationship by employing a GAN inversion method to determine the domain-gap direction. Recently, Zhang \etal \cite{zhang2022towards} develop an attentive style loss and a selective generation strategy to promote the diverse generation and faithful adaptation in a one-shot scenario. 

Despite the significant progress achieved by these methods, their effectiveness might be limited when dealing with substantial inter-domain gaps. In contrast, our approach directly explores the target manifold, ensuring that outcomes reside within the target space while maintaining cross-domain correspondences.

\noindent \textbf{GAN Inversion.}
GAN inversion aims to find a latent code capable of faithfully reconstructing a given real image using a pre-trained generator. These methods can be categorized as encoder-based~\cite{richardson2021encoding,xu2021generative,Kang_2021_ICCV,alaluf2021restyle,bai2022high} and optimization-based  \cite{abdal2019image2stylegan,abdal2020image2stylegan++,abdal2021styleflow,xu2021continuity}. The former directly trains an encoder to map real images to latent codes. For instance, Richardson \etal \cite{richardson2021encoding} and Xu \etal \cite{xu2021generative} concentrate on designing the encoder structure. Tov \etal \cite{tov2021designing} consider the properties of $W$ and $W+$ space, enhancing the editability of reconstructed images, and Alaluf \etal \cite{alaluf2021restyle} propose an iterative feedback mechanism to facilitate the learning process. In \cite{bai2022high}, Bai \etal utilize padding space to enrich the representation capacity of the latent space, thereby refining spatial details. However, learning-based methods typically require a large number of training images and might not be practical when training samples are limited. In contrast, optimization-based methods can infer a single image at a time. Kang \etal~\cite{Kang_2021_ICCV} jointly optimize the extended $f$ and $w+$ latent codes for faithful reconstruction of out-of-range and unaligned real images, and  Abdal \etal \cite{abdal2021styleflow} propose conditional exploration using continuous normalizing flows. Recently, Xu \etal \cite{xu2021continuity} introduce consecutive images to strike a balance between editability and fidelity. 

Motivated by the performance of the optimization-based inversion methods, we propose a new translation paradigm named UniTranslator. This approach entails seeking optimal latent codes guided by our tailored objective.

\section{Method}\label{sec:method}
\subsection{Overview}
The overview of our method is depicted in Fig.~\ref{fig:pipeline} and a demo video in the supplementary material. Given a single source image, UniTranslator aims to:
\begin{itemize}
	\item Discover an optimal embedding corresponding to a target domain image while preserving cross-domain relationships with the source image. 
	\item Map this embedding into StyleGAN's latent space.
\end{itemize}
We begin by navigating the $Z$ space, employing a dual-branch architecture to achieve both goals through a hybrid learning approach. One branch is the decoupling module. Specifically, we utilize the source image as a reference and leverage the domain-neutral capabilities of the CLIP space, which aligns images with prompts of neutral classes. This alignment enables us to extract domain-agnostic information, represented as abstract CLIP embeddings. We achieve this by modeling the relationships among various combinations of embedding components. Merging domain-agnostic semantics with target-specific information guides the optimization process towards a more suitable embedding.

The other branch enhances the quality and diversity of image translation by tapping into the impressive generative capability of StyleGAN. To achieve this goal, we introduce a CLIP2P mapper as a crucial link between the CLIP space and StyleGAN's $P$ space \cite{zhu2020improved}~(the deactivated space of $W$ space). Note that directly converting CLIP embeddings to $W$ space using only a single input image presents significant challenges. Mapping accurately to this complex latent distribution requires designing specific network modules or objectives, which is difficult with limited input data. Instead, we opt to use the $P$ space as an intermediary, leveraging its properties to effectively transfer the desired CLIP embedding into StyleGAN's native latent space, resulting in high-quality output generation. Furthermore, we do not consider $W+$ space due to its higher degrees of freedom, which complicates optimization when limited cues are available from a single image. This increased flexibility could lead to deviations from the StyleGAN target manifold, undermining the quality of the generated images. 

Our UniTranslator operates through a two-step process during each training iteration. We optimize the latent code $z$ in the first step while keeping the network parameters frozen. This step leverages the valuable information stored in the network parameters to guide the search for the optimal $z$ code. In the second step, we fix the discovered $z$ code and update the network parameters. 

\subsection{Decoupling Module}\label{sec:decouple}
\begin{figure}[t]
	\centering
	\includegraphics[width=\linewidth]{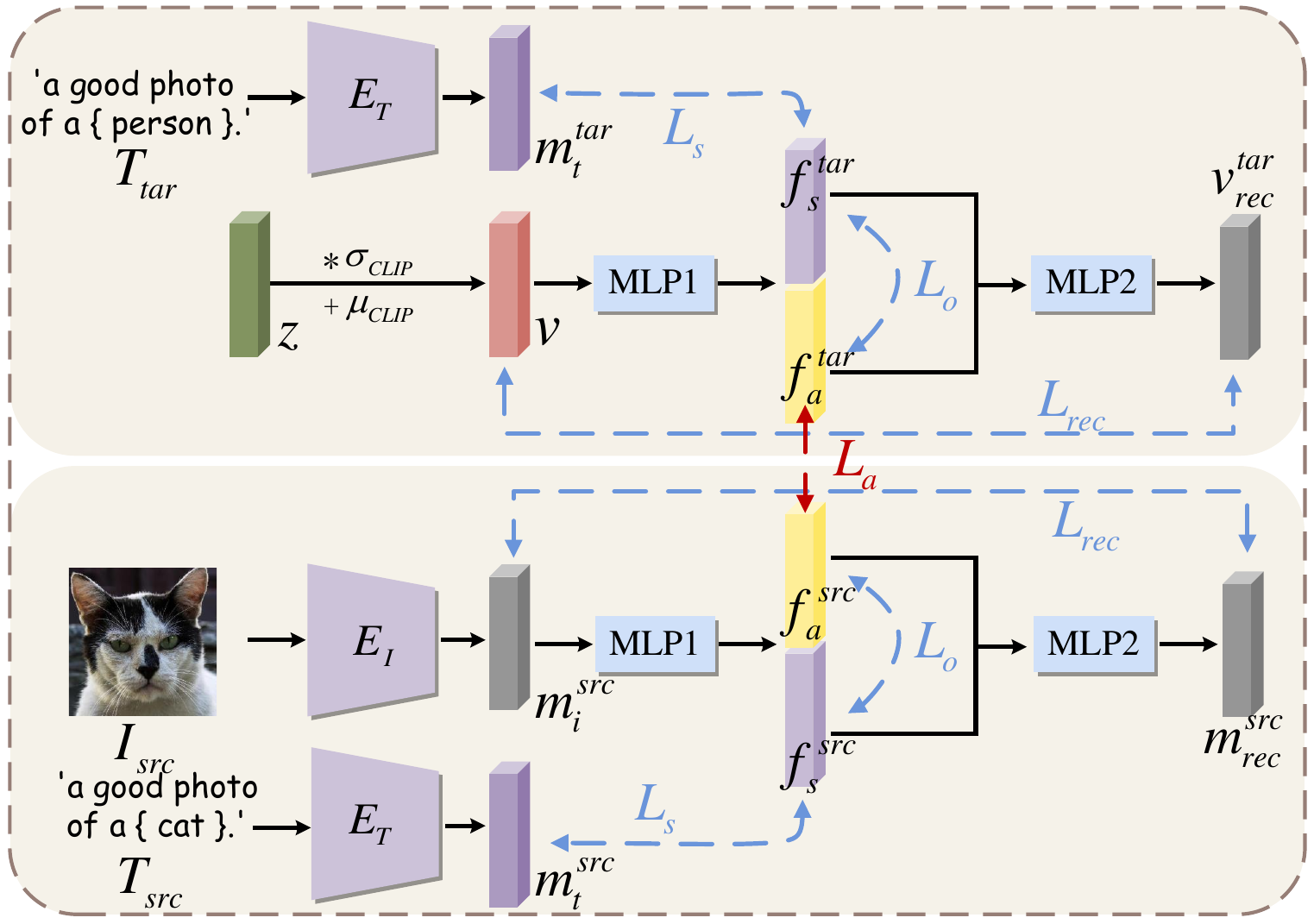}
	\caption{Illustration of the proposed decoupling module.}	
	\label{fig:decouple}
\end{figure}

The main goal of the proposed decoupling module is to use CLIP's image-text alignment capability to extract domain-agnostic information. We initially convert the $z$ code into a CLIP embedding. Following the implicit assumption from Corgi~\cite{Zhou_2023_CVPR} that each neutral domain conforms to a high-dimensional Gaussian distribution, we begin by generating 5,000 target images corresponding to the selected target domain. These images are then processed through CLIP image encoder $E_I$ to acquire their respective embeddings. Subsequently, We calculate statistics for these embeddings, including the standard deviation $\sigma_{CLIP}$ and the mean $\mu_{CLIP}$. Using these statistics, we transform the $z$ code as an embedding $v$ within the CLIP space:
\begin{equation}
	v=\sigma_{CLIP} z+\mu_{CLIP}.
	\label{equ:1}
\end{equation}
Fig.~\ref{fig:decouple} shows a detailed illustration of the module structure. Regarding the module inputs, prompt templates are also required apart from the CLIP embedding $v$ and the provided source domain image $I_{src}$. These inputs are split and fed into two separate streams based on whether they pertain to the source or target domain. Each stream contains two Multi-Layer Perceptrons (MLPs),
with shared parameters across streams.

For the source domain stream (lower part of Fig.~\ref{fig:decouple}), an embedding $m_i^{src}\in{\mathbb R}^{512\times1}$ of the source image is generated by the CLIP image encoder. This embedding is passed through the first MLP (MLP1), generating a 1024-dimensional vector. To extract domain-agnostic information, we divide this vector into two 512-dimensional embeddings $f_s^{src}$ and $f_a^{src}$ to learn domain-specific and domain-agnostic information. It is important to note that these two types of information are expected to be independent.

A similar process is applied to the target domain stream. The main difference is that the input to MLP1 is the CLIP embedding $v$, which is for inverting the target image. As a result, the target domain stream also yields two corresponding embeddings $f_s^{tar}$ and $f_a^{tar}$. The first loss function $\mathcal{L}_o$ is:
\begin{equation}
	\mathcal{L}_{o}=\frac{f_s^{tar} \cdot f_a^{tar}}{|f_s^{tar}||f_a^{tar}|}+\frac{f_s^{src} \cdot f_a^{src}}{|f_s^{src}||f_a^{src}|}.
	\label{equ:Lun}
\end{equation}
Minimizing $\mathcal{L}_o$ enforces that the domain-specific embeddings and their corresponding domain-agnostic embeddings become orthogonal, guaranteeing their independence.

Furthermore, we align $f_s^{tar}$ and $f_s^{src}$ with their respective text embeddings $m_t^{tar}$ and $m_t^{src}$ produced by CLIP's text encoder $E_T$ to ensure that they genuinely carry domain-specific information. This alignment is accomplished using the following loss function $\mathcal{L}_{s}$: 
\begin{equation}
	\mathcal{L}_{s}=(1-\frac{m_t^{tar} \cdot f_s^{tar}}{|m_t^{tar}||f_s^{tar}|})+(1-\frac{m_t^{src} \cdot f_s^{src}}{|m_t^{src}||f_s^{src}|}).
	\label{equ:Lre}
\end{equation}
Similar to the approach by Radford et al. ~\cite{radford2021learning}, we utilize 80 diverse sentence templates, such as 'a photo of the \{\}', 'a photo of my \{\}', and 'a good photo of a \{\}'. The resulting embeddings from these prompts are averaged separately for the source and target domains. We have templates with the neutral class for both domains.
With the impact of the above two loss functions, the remaining embeddings $f_a^{tar}$ and $f_a^{src}$ inherently capture domain-agnostic information. Moreover, we employ $\mathcal{L}_{a}$ to constrain their similarity, thereby aiding in the extraction of cross-domain correspondences:
\begin{equation}
	\mathcal{L}_{a}=||f_a^{tar}-f_a^{src}||_1.
	\label{equ:Lsim}
\end{equation}
We now create a combined embedding comprising $f_s^{tar}$ and $f_a^{tar}$, encompassing target-specific and domain-agnostic information. This concatenated embedding is then fed through the second MLP (MLP2) for reconstructing a 512-dimensional vector $v_{rec}^{tar}$. The last loss term is the reconstruction loss $\mathcal{L}_{rec}$. This process can be formulated as:
\begin{equation}
	\begin{split}
	v_{rec}^{tar}&=\mathrm{MLP2}(f_s^{tar} \oplus f_a^{tar}),
	m_{rec}^{src}=\mathrm{MLP2}(f_s^{src} \oplus f_a^{src}),\\
	\mathcal{L}_{rec}&=||v-v_{rec}^{tar}||_1+||m_i^{src}-m_{rec}^{src}||_1,
	\end{split}
	\label{equ:Lrecon1}
\end{equation}
where $\oplus$ denotes concatenation operation. This loss maintains consistency between the reconstructed vectors and their corresponding image embeddings throughout the network parameters learning step. While optimizing the $z$ code, we aim to use the embedding $v_{rec}^{tar}$, which incorporates both target-specific and domain-agnostic information, to guide the search for an enhanced $z$.

Finally, the learning objective of the decoupling module is formulated as:
\begin{equation}
\mathcal{L}_{decoupling} = \mathcal{L}_o + \mathcal{L}_s +  \mathcal{L}_{a} + \mathcal{L}_{rec}.
\label{equ:Ldecouple}
\end{equation}

Previous latent code optimization techniques~\cite{Roich2022,Pan2023} aim to enhance reconstruction fidelity or editability within StyleGAN's domain or a slightly regularized version to mitigate out-of-domain issues. In contrast, our decoupling module utilizes CLIP's domain-neutral features to maintain essential cross-domain correspondences. This approach liberates the solution space from being restricted to a single domain, allowing us to model relationships between domain-specific and domain-agnostic features across different domains and modalities. Our specialized objective thus establishes robust cross-domain correspondences.

\subsection{CLIP2P Mapper}
\label{sec:clip}
We analyze another branch within the framework where the CLIP2P mapper is situated. First, we introduce the concept of the $P$ space, defined by II2S~\cite{zhu2020improved} as $p=\mathrm{LeakyReLU}_{5.0}(w)$, where $p\in P$ and $w\in W$ represent two latent codes sampled from their respective spaces. The $W$ space can be regarded as the activated counterpart of the $P$ space. Assuming a Gaussian distribution for the $P$ space, such a relationship can be formulated (as adopted by the official code of PULSE):
\begin{equation}
	w=\mathrm{LeakyReLU}_{0.2}(\sigma_{P} z+\mu_{P}),
	\label{equ:pulse}
\end{equation}
where $\sigma_{P}$ and $\mu_{P}$ indicate the standard deviation and mean of the Gaussian distribution. 
With this assumption, we can use the statistics of sampled CLIP embeddings to construct a Gaussian distribution (refer to Eq.~(\ref{equ:1})) and learn the latent distribution in $P$ space through a linear layer that preserves the distribution type, yielding the $w$ latent code as:
\begin{equation}
	w=\mathrm{LeakyReLU_{0.2}}(\mathrm{Linear}(v)).
	\label{equ:tow}
\end{equation}
However, as II2S indicates, the latent distribution in $P$ space is essentially an unimodal distribution resembling a Gaussian but not a true Gaussian distribution. To analyze the deviation from this assumption in transformations across different target domains, we sample 5000 latent codes from the $P$ space of each target domain and calculated the Kullback-Leibler (KL) divergence between the Gaussian distribution established using the statistics of these codes (referred to as $P$ (pseudo)) and the true distribution of these codes (refer to as $P$ (true)). As shown in Table~\ref{tab:kl_real_p_and_gaussian_p}, there is indeed a disparity between the distributions $P$ (true) and $P$ (pseudo), which varies across different target domains. Thus, the $w$ code obtained via Eq.~(\ref{equ:tow}) may lead to unsuccessful transformations in specific target domains (\eg, E621Faces, LSUN-church), as it may fall outside the true latent space of StyleGAN.

\begin{table}[t]
	\setlength{\abovecaptionskip}{0cm}
	\caption{The KL divergence between the true latent distribution in $P$ space and the Gaussian distribution constructed using the statistics of the latent codes sampled from the $P$ space for various target domains.}
	\centering
    \resizebox{0.48\textwidth}{!}{
	\begin{tabular}{c||c||c}
		\toprule
		Target & KL Divergence&KL Divergence \\
		Domain& ($P$ (true) w.r.t $P$ (pseudo))&($P$(pseudo) w.r.t $P$ (true))\\
		\midrule
		FFHQ & 0.6883 & 0.7016 \\
		Anime & 1.0853 & 1.0964 \\
		E621Faces & 1.1810& 1.2037 \\
		LSUN-church & 6.3166 & 6.3641 \\
		\bottomrule
	\end{tabular}}
	\label{tab:kl_real_p_and_gaussian_p}
\end{table}%

We propose adding a nonlinear function $\mathcal{M}(\cdot)$ after the linear layer to learn the unimodal distribution in $P$ space to remedy this. It is defined as:
\begin{equation}
	\mathcal{M}(x) = \left\{
	\begin{array}{rcl}
		e^{h(x-\mu)}-1,       &      & {x > \mu},\\
		-e^{-j(x-\mu)}+1,     &      & {x  \leq \mu}.\\	
	\end{array} \right.
	\label{equ:CLIP2P}
\end{equation}
Here, $h$, $j$, $\mu$ are all learnable parameters, and $e$ is Euler's number. The reasons for this function are threefold. First, while we introduce nonlinearity to ensure that the resulting distribution deviates from a Gaussian shape, it is crucial to maintain the function monotonicity. This guarantees the Gaussian distribution generated by the added linear layer remains unimodal after undergoing such a nonlinear transformation. Second, we introduce three degrees of freedom through parameters $h$, $j$, and $\mu$. Among these, $\mu$ controls the peak position of the resulting unimodal distribution. Meanwhile, $h$ and $j$ control the curvatures on either side of the peak, ensuring the asymmetric distribution. Third, to ensure the continuity of this function, biases of $-1$ and $+1$ are applied at different intervals.

The CLIP2P mapper denoted as CLIP2P$(\cdot)$, comprises a linear layer and the above nonlinear mapping function. It is effectively trained using the distribution $P$ (pseudo) as supervision through the loss term $\mathcal{L}_g$:
\begin{equation}
	\mathcal{L}_{g} = ||\sigma_P z+\mu_P-q||_1,
	\label{equ:Lg}
\end{equation}
where $q=\mathrm{CLIP2P}(v)=\mathcal{M}(\mathrm{Linear}(v))$ indicates the output after applying the CLIP2P mapper. We note the inclusion of $\mathcal{L}_g$ does not alter the output distribution type from the CLIP2P mapper. Instead, it enhances training stability. 

We examine the KL divergence between the true latent distributions in CLIP and $P$ spaces across different target domains. For each target domain, we sample 5000 latent codes. Table~\ref{tab:kl_real_p_and_real_clip} shows that the disparity between these two distributions varies across domains. Consequently, adjusting the hyperparameter related to $\mathcal{L}_{g}$ based on the target domain is necessary although challenging. We set the hyperparameter $\lambda_p$ of $\mathcal{L}_{g}$ as a learnable parameter to address this. This enables us to formulate the final loss function $\mathcal{L}_{p}$ for the CLIP2P mapper as:
\begin{equation}
	\mathcal{L}_{p} = \mathrm{ReLU}(\lambda_p)\mathcal{L}_{g}.
	\label{equ:Lp}
\end{equation}
In practice, we observe that $\lambda_p$ sometimes converges to 0 (\eg, when using FFHQ as the target domain). This suggests that effective learning can occur with less reliance on the supervision from $P$ (pseudo). However, in other scenarios, such as when working with target domains like Anime or LSUN-church, supervision is necessary to constrain parameters learning, and $\lambda_p$ dynamically adjusts to an appropriate non-zero value based on the target domain.

While techniques such as StyleCLIP's latent mapper~\cite{Patashnik2021} also perform latent code mapping, they focus on in-domain mappings within StyleGAN's latent space, which are not designed for cross-domain translation tasks that involve significant domain gaps, such as converting a painting into a photo~\cite{gal2022stylegan}. Our CLIP2P mapper, on the other hand, bridges the CLIP and $P$ spaces by leveraging their unique characteristics. This ensures that the latent code transformed from the CLIP space accurately resides within StyleGAN's target manifold. By doing so, our approach allows for the search of target embeddings with cross-domain correspondences in the open-world CLIP space, while effectively utilizing StyleGAN's generative priors to achieve universal domain translation.

\subsection{Learning Objectives}
Aside from the above-explained loss terms $\mathcal{L}_{decoupling}$ and $\mathcal{L}_p$, we use three other loss terms to train the UniTranslator: $\mathcal{L}_{mse}$, $\mathcal{L}_{lpips}$, and $\mathcal{L}_{cycle}$. 

\begin{table}[t]
	\setlength{\abovecaptionskip}{0cm}
	\caption{The KL divergence between the true latent distributions in the $P$ space and CLIP space for various target domains.} 
	\centering
	\resizebox{0.48\textwidth}{!}{
	\begin{tabular}{c||c||c}
		\toprule
		Target & KL Divergence &KL Divergence\\
		Domain& ($P$ (true) w.r.t CLIP (true))&(CLIP (true) w.r.t $P$ (true))\\
		\midrule
		FFHQ & 0.4754 & 0.4587 \\
		Anime & 1.0571 & 1.0351 \\
		E621Faces & 1.1687 & 1.1141 \\
		LSUN-church & 3.6901 & 4.2885 \\
		\bottomrule
	\end{tabular}}
	\label{tab:kl_real_p_and_real_clip}
\end{table}

In addition to preserving semantic correlations across domains, we maintain visual aspects such as color tone and perceptual relationships in cross-domain translation tasks. As observed in DiffuseIT~\cite{kwon2022diffusion} and Zhu~\etal~\cite{zhumind}, relying solely on CLIP-based semantic alignment is insufficient to ensure color consistency between the input and output images. To address this, we incorporate a loss term $\mathcal{L}_{mse}$, similar to its use in DiffuseIT, to ensure that color matching is maintained before and after translation. This loss enforces a constraint on the Euclidean distance between the source image $I_{src}$ and the target image $I_{tar}$, formulated as: 
\begin{equation}
	\mathcal{L}_{mse} = ||I_{tar}-I_{src}||_2,
	\label{equ:Lmse}
\end{equation}
and
\begin{equation}\label{equ:3}%
	\begin{split}
		I_{tar} &=G(\mathrm{LeakyReLU_{0.2}}(q)),
	\end{split}
\end{equation}
where $G(\cdot)$ denotes the pretrained StyleGAN generator. It is important to note that our approach mitigates the risk of image blurriness by generating images through latent space traversal rather than the traditional feature space-to-output method. With StyleGAN's robust generative capabilities, as long as the latent code remains within the StyleGAN target manifold, constrained by our CLIP2P mapper and $\mathcal L_p$ loss, it produces sharp and high-quality images.

The other term, $\mathcal{L}_{lpips}$~\cite{zhang2018unreasonable}, leverages deep features to guide the perceptual relationship between $I_{tar}$ and $I_{src}$:
\begin{equation}
	\mathcal{L}_{lpips} = ||F(I_{tar})-F(I_{src})||_2,
	\label{equ:Llpips}
\end{equation}
where $F(\cdot)$ represents the perceptual feature extractor.

The last loss term is a cycle loss, denoted as $\mathcal{L}_{cycle}$. In UniTranslator, after generating the translated result $I_{tar}$, we proceed to feed it into CLIP's image encoder, obtaining a CLIP embedding $v_{tar}=E_I(I_{tar})$. Subsequently, we pass $v_{tar}$ as input to the CLIP2P mapper and replicate the remaining process. As such, we have a new output, denoted as $I_{cycle}$. We then impose the constraint $\mathcal{L}_{cycle}$ to ensure consistency between the two images, $I_{tar}$ and $I_{cycle}$. This secures that the CLIP2P mapper only transforms the space type (from CLIP to $P$) without modifying the image semantics:
\begin{equation}
	\begin{split}
	I_{cycle} &= G(\mathrm{LeakyReLU_{0.2}}(\mathrm{CLIP2P}(v_{tar}))),\\
	\mathcal{L}_{cycle} &= ||I_{tar}-I_{cycle}||_2.
    \end{split}
    \label{equ:cycle}
\end{equation}
Note that although our initial assumption regarding the distribution in CLIP's neutral domain as a Gaussian may not strictly hold, the cycle loss can mitigate the risks arising from this violation by constraining the latent vector to carry more semantic information from the CLIP space.

Finally, our total loss function $\mathcal{L}_{total}$ is formulated as:
\begin{equation}
	\begin{aligned}
	\mathcal{L}_{total} = &\lambda_{mse}\mathcal{L}_{mse} + \mathcal{L}_{lpips} +  \mathcal{L}_{decoupling}\\
	& + \mathcal{L}_{cycle}+\mathcal{L}_p,
	\label{equ:Ltotal}
	\end{aligned}
\end{equation}
where $\lambda_{mse}$ denotes a hyperparamter for balancing loss terms.

\section{Experiments}\label{sec:experiment}
\subsection{Implementation Details}
To implement UniTranslator, we use a hybrid learning scheme and outline the workflow in Algorithm~\ref{alg::algorithm}. In each iteration, we start with step 1, optimizing the $z$ code using a spherical optimizer~\cite{menon2020pulse} with a learning rate of 0.4. Then, we move to step 2, where we focus on learning the network parameters using the ADAM optimizer~\cite{kingma2014adam} with exponential decay rates of $\beta_1 = 0$ and $\beta_2 = 0.99$. The learning rate for updating all parameters is set to 0.002, except $\lambda_{p}$, which is assigned a higher value of 0.1. The hyperparameter $\lambda_{mse}$ in our objective is set to 10. Approximately 35 iterations are conducted for each source image, taking between 20 and 45 seconds on a PC with an Nvidia GeForce RTX 3090.

\begin{algorithm}[t]
	\caption{UniTranslator}
	\label{alg::algorithm}	
	\KwData{Initial latent code $z_{(0)}$, hyperparameter $\lambda_p^{(0)}$, and learnable network parameters $\theta_{(0)}$; Statistics $\sigma_{CLIP}$, $\mu_{CLIP}$, $\sigma_{P}$ and $\mu_{P}$; Source Image $I_{src}$; Prompt templates; Hyperparameters $\{\lambda_{i}\}$; Iteration number $N$; Pretrained StyleGAN and CLIP models}	
	\KwResult{Optimal latent code $z_{(N)}$ and Target Image $I_{tar}$}
	\For{$i\leftarrow 1$ \KwTo $N$}{
		\While{Step 1: $z$ latent code optimization}
		{Update the latent code $z_{(i)}$ with $z_{(i-1)}$, $\theta_{(i-1)}$, and $\lambda_{p}^{(i-1)}$ by Eq.~(\ref{equ:Ltotal})\;}
		\While{Step 2: Network parameters learning}
		{Update the learnable parameters $\theta_{(i)}$ and $\lambda_p^{(i)}$ with $z_{(i)}$ by Eq.~(\ref{equ:Ltotal})\;}
	}
	Obtain $I_{tar}$ with $z_{(N)}$ by Eq.~(\ref{equ:3})\;	
\end{algorithm}

\subsection{Evaluated Methods}
We compare UniTranslator with numerous state-of-the-art methods based on learning (\eg, VQ-I2I~\cite{chen2022eccv}, StarGAN2~\cite{choi2020stargan}, and GP-UNIT~\cite{yang2022unsupervised}), one-shot domain adaptation (\eg, DiFa~\cite{zhang2022towards}),  inference (\eg, PULSE~\cite{menon2020pulse}), and diffusion models (\eg, DiffusionCLIP~\cite{Kim_2022_CVPR} and DiffuseIT~\cite{kwon2022diffusion}).

It is worth noting that learning-based methods require a substantial amount of training data from both source and target domains. The domain adaptation method, \eg, DiFa, requires a target image to fine-tune its pre-trained source generator. Diffusion-based methods utilize two text prompts: one for the source domain and another for the target domain. These prompts are used to fine-tune the pre-trained diffusion model or govern the sampling process. On the other hand, inference-based methods such as PULSE and UniTranslator perform direct inference based on the input image (and prompt templates). The settings and hyperparameters of the evaluated methods are according to their official source codes.

\subsection{Datasets}
We use eight source-to-target translation mappings to assess the universality of all the comparisons. These mappings cover a wide range of domain heterogeneity, including cases where domains are `adjacent', such as Metfaces~\cite{karras2020training}$\to$FFHQ~\cite{karras2019style} and AFHQ-cat~\cite{choi2020stargan}$\to$E621Faces~\cite{E621Faces}, as well as cases where domains are `far-off', such as AFHQ-cat$\to$Anime~\cite{danbooru2020}, AFHQ-cat$\to$FFHQ, AFHQ-dog$\to$FFHQ, and AFHQ-wild$\to$FFHQ. To provide a more challenging evaluation, we also include mappings between `intensely far-off' domains, such as LSUN-Church~\cite{yu15lsun}$\to$FFHQ and AFHQ-cat$\to$LSUN-Church. Note that we do not address intra-domain translations, such as male-to-female mapping within the FFHQ domain, as these can be easily achieved with state-of-the-art editing methods~\cite{Roich2022,Alaluf2022}. Instead, our focus is on cross-domain translations, even those involving minimal domain gaps, such as the adjacent domains we target.

For datasets with predefined train-test splits, such as AFHQ and LSUN-church, we use the default training set to train learning-based competitors. In cases where predefined splits are unavailable, as with MetFaces, FFHQ, E621Faces, and Anime, we perform a random 7:3 split between the training and test sets. For fine-tuning the domain adaptation method DiFa, we randomly select a single training image from the target domain. Note that inference-based and diffusion-based methods conduct inference directly on the test set or utilize text prompts without relying on the training images. All reported performances are based on results obtained from the test set.

The datasets used in the evaluations contain images of different resolutions: 1024$\times$1024 (\eg, FFHQ, Metfaces), 512$\times$512 (\eg, AFHQ, Anime, and E621Faces), and 256$\times$256 (\eg, LSUN-church) pixels. Note that VQ-I2I, GP-UNIT, StarGAN2, DiffusionCLIP, and DiffuseIT are limited to generating results at a resolution of 256$\times$256. For DiFa, the generative resolutions align with those of the source domain datasets, as they depend on the source generators. In contrast, PULSE and UniTranslator consistently produce results at the same resolution as the target domain dataset.

\subsection{Evaluation Metrics}
We use various metrics to analyze the quality of the generated results and the perceptual correspondence between inputs and outputs. Specifically, we use the no-reference metric Naturalness Image Quality Evaluator (NIQE) \cite{mittal2012making} to assess image quality, with a particular focus on the perceived realism of the results, including any potential distortions or artifacts. Additionally, in cross-domain translation tasks, the generated images must bear perceptual similarity to the corresponding reference images. We utilize the LPIPS metric~\cite{zhang2018unreasonable}, which relies on deep features. It is calculated for each input-output pair, with an average score reported.

Furthermore, after producing the transformed images for each translation task, we measure the similarity between their CLIP embeddings and those of the target dataset to assess how well the output images fit into the target domain. We also include a user study to support our evaluations, both of which are detailed in the supplementary materials.

It is important to note that while the Fr\'echet Inception Distance (FID) \cite{heusel2017gans} and the Inception Score (IS)~\cite{Salimans2016} are widely used to evaluate image quality, they are not ideally suited for cross-domain translation tasks. A detailed discussion of these metric choices is provided in the supplementary materials.

\subsection{Qualitative Evaluation}
\begin{figure*}[t]
	\centering
	\setlength{\abovecaptionskip}{1mm}
	\centering
	\setlength{\tabcolsep}{0.05em}
	\setlength{\fboxrule}{1pt}
	\setlength{\fboxsep}{0pt}
	\begin{tabular}{ccccccccc}
		\includegraphics[width=.109\linewidth]{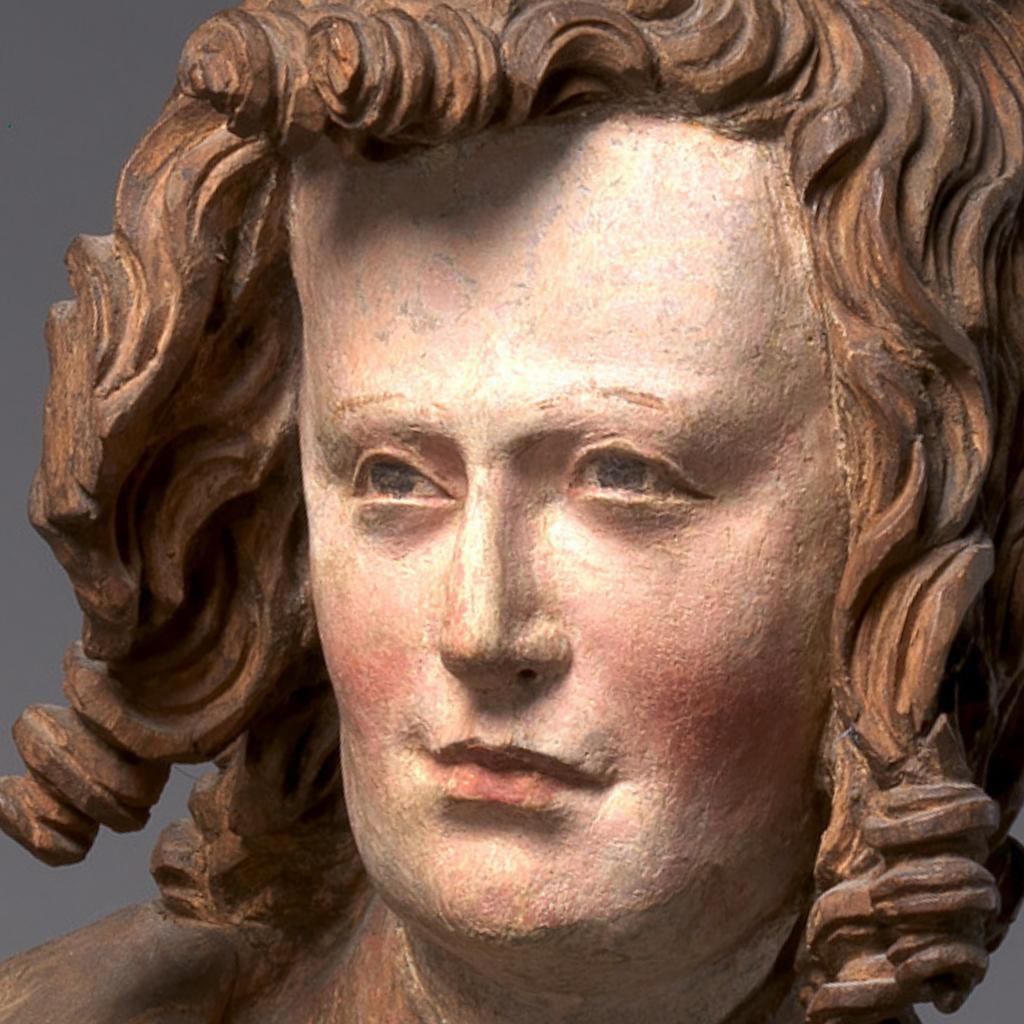} &
		\includegraphics[width=.109\linewidth]{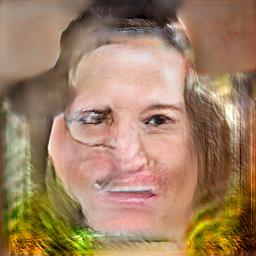} &
		\includegraphics[width=.109\linewidth]{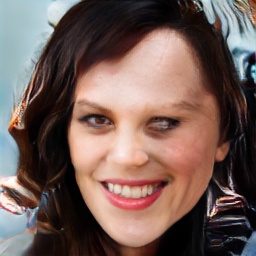} &
		\includegraphics[width=.109\linewidth]{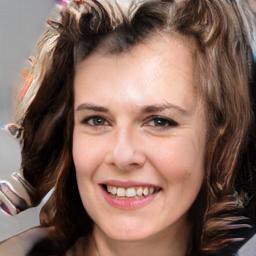} &
		\includegraphics[width=.109\linewidth]{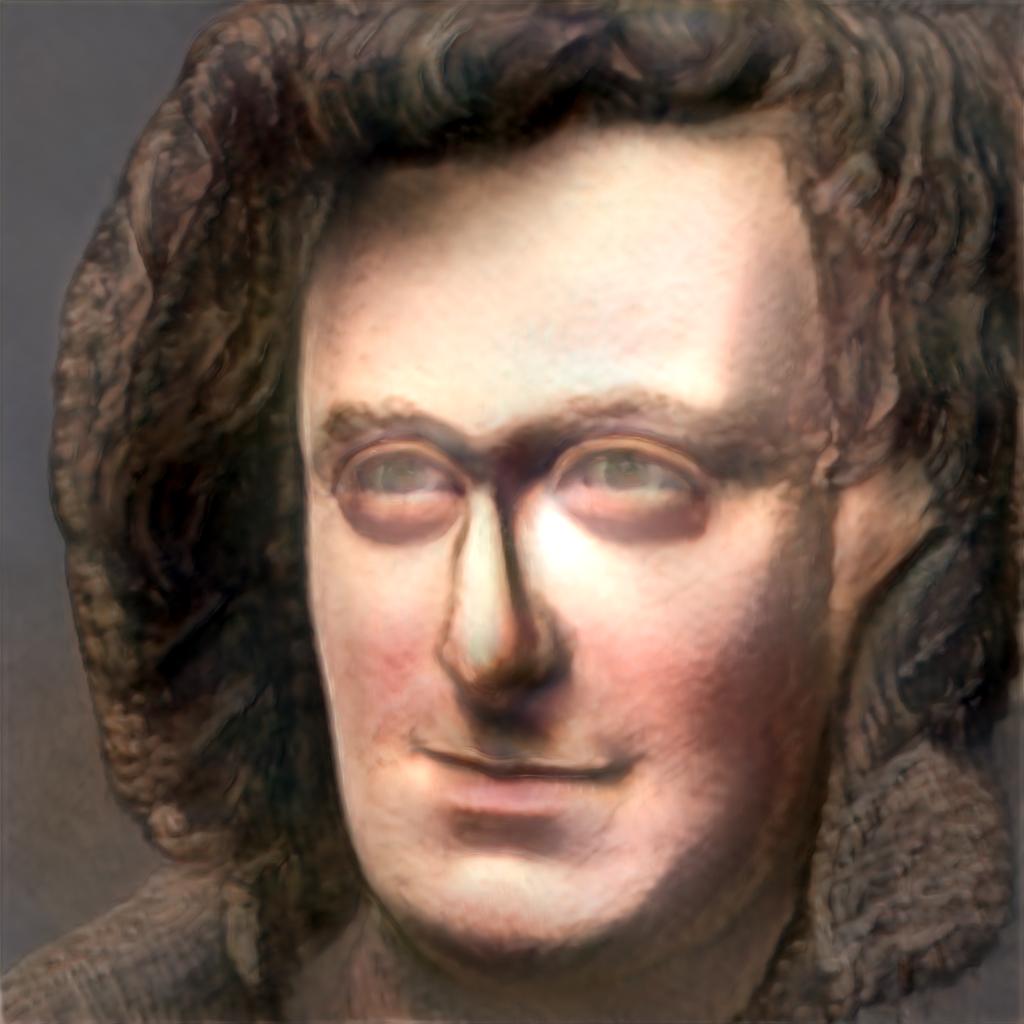} &
		\includegraphics[width=.109\linewidth]{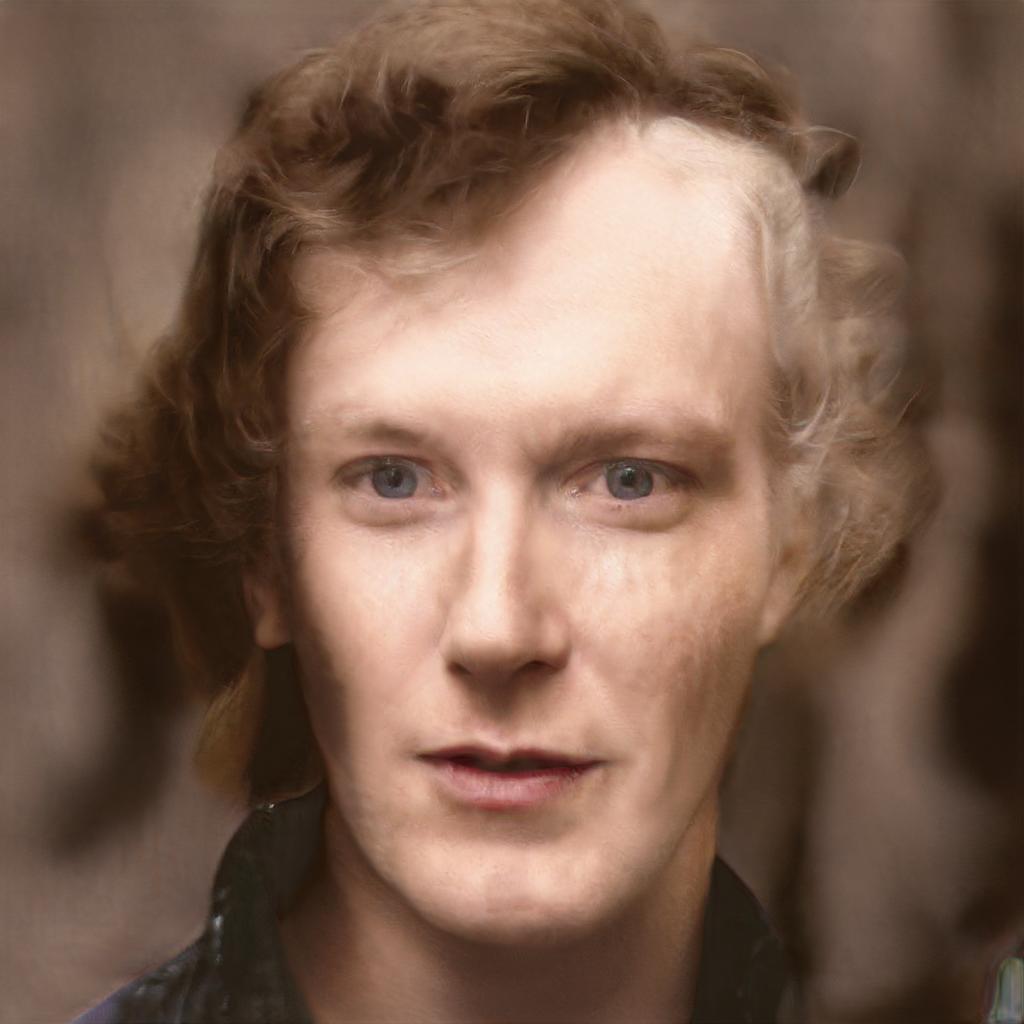} &
		\includegraphics[width=.109\linewidth]{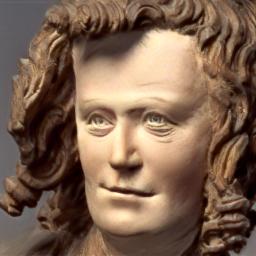} &
		\includegraphics[width=.109\linewidth]{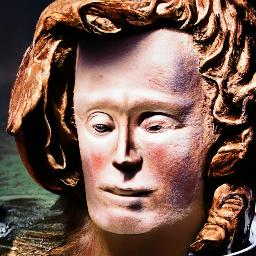} &
		\includegraphics[width=.109\linewidth]{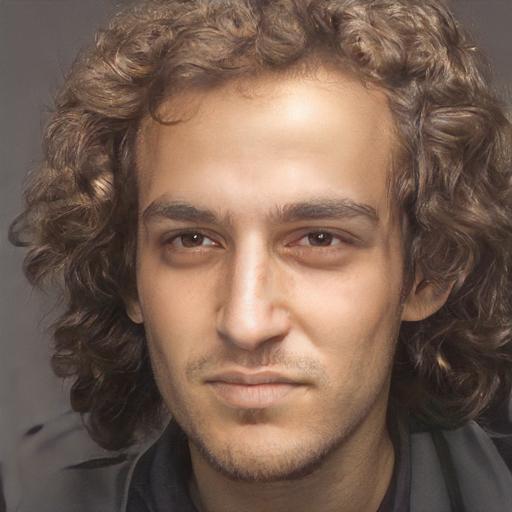}\\
		
		\includegraphics[width=.109\linewidth]{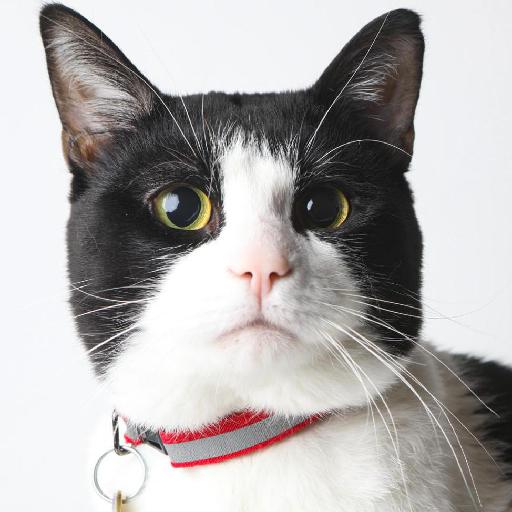} &
		\includegraphics[width=.109\linewidth]{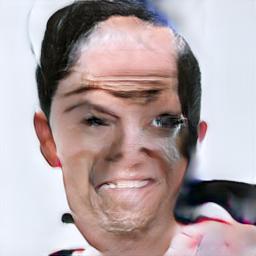} &
		\includegraphics[width=.109\linewidth]{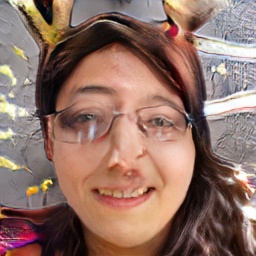} &
		\includegraphics[width=.109\linewidth]{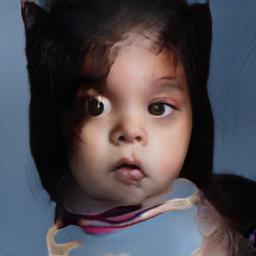} &	
		\includegraphics[width=.109\linewidth]{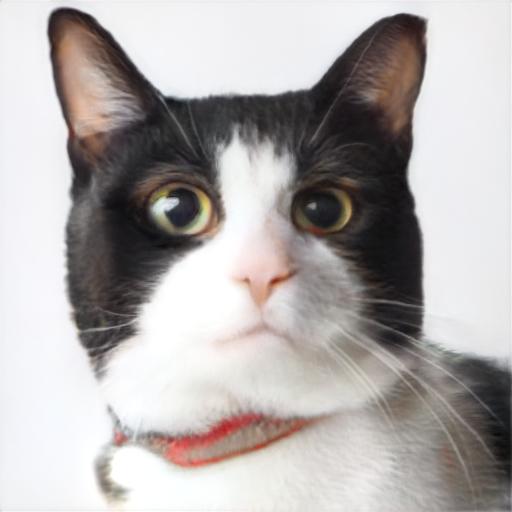} &
		\includegraphics[width=.109\linewidth]{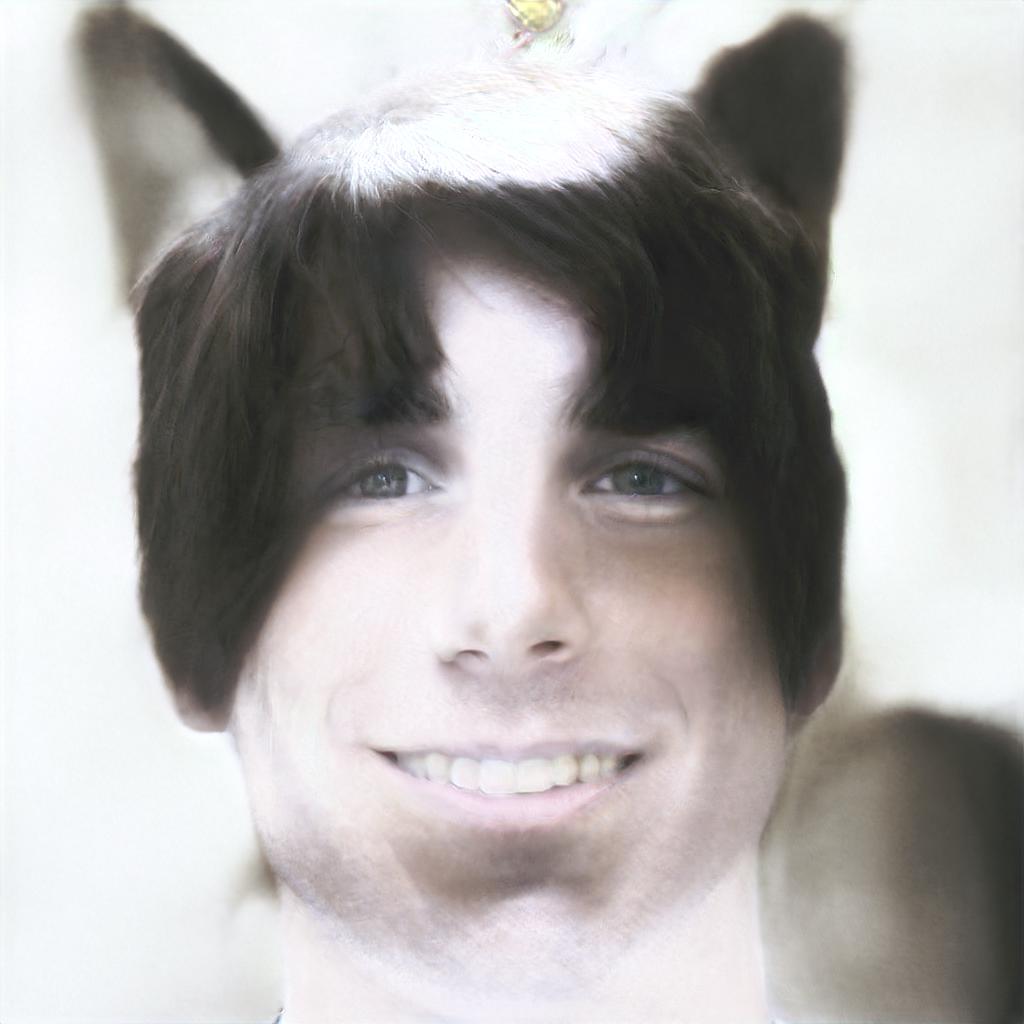} &
		\includegraphics[width=.109\linewidth]{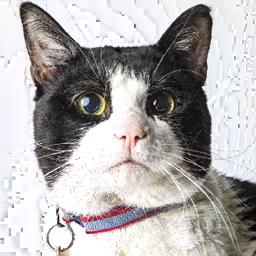} &
		\includegraphics[width=.109\linewidth]{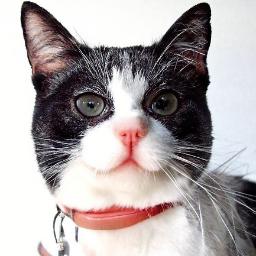} &
		\includegraphics[width=.109\linewidth]{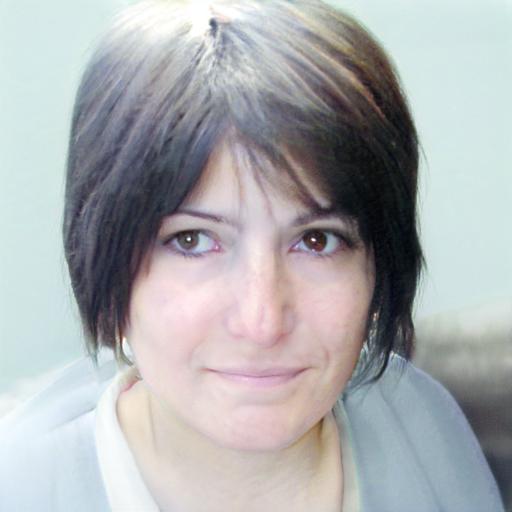}\\	
		\includegraphics[width=.109\linewidth]{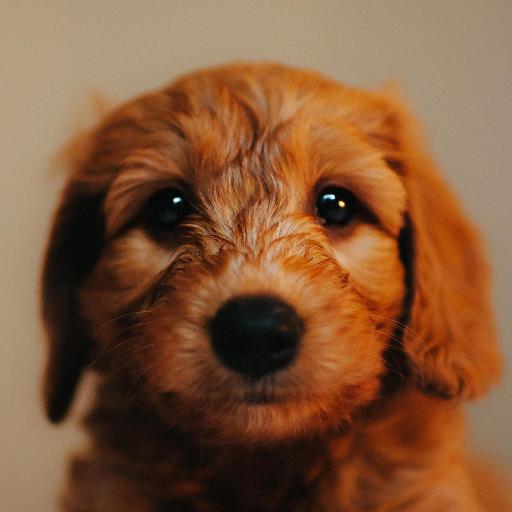} &
		\includegraphics[width=.109\linewidth]{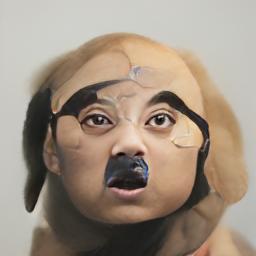} &
		\includegraphics[width=.109\linewidth]{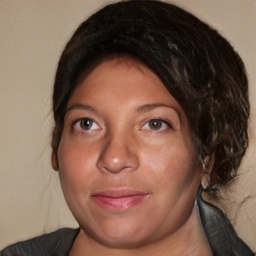} &
		\includegraphics[width=.109\linewidth]{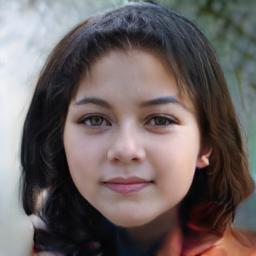} &
		\includegraphics[width=.109\linewidth]{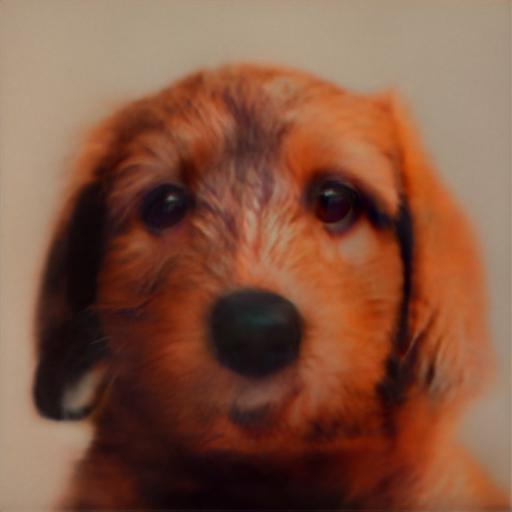} &
		\includegraphics[width=.109\linewidth]{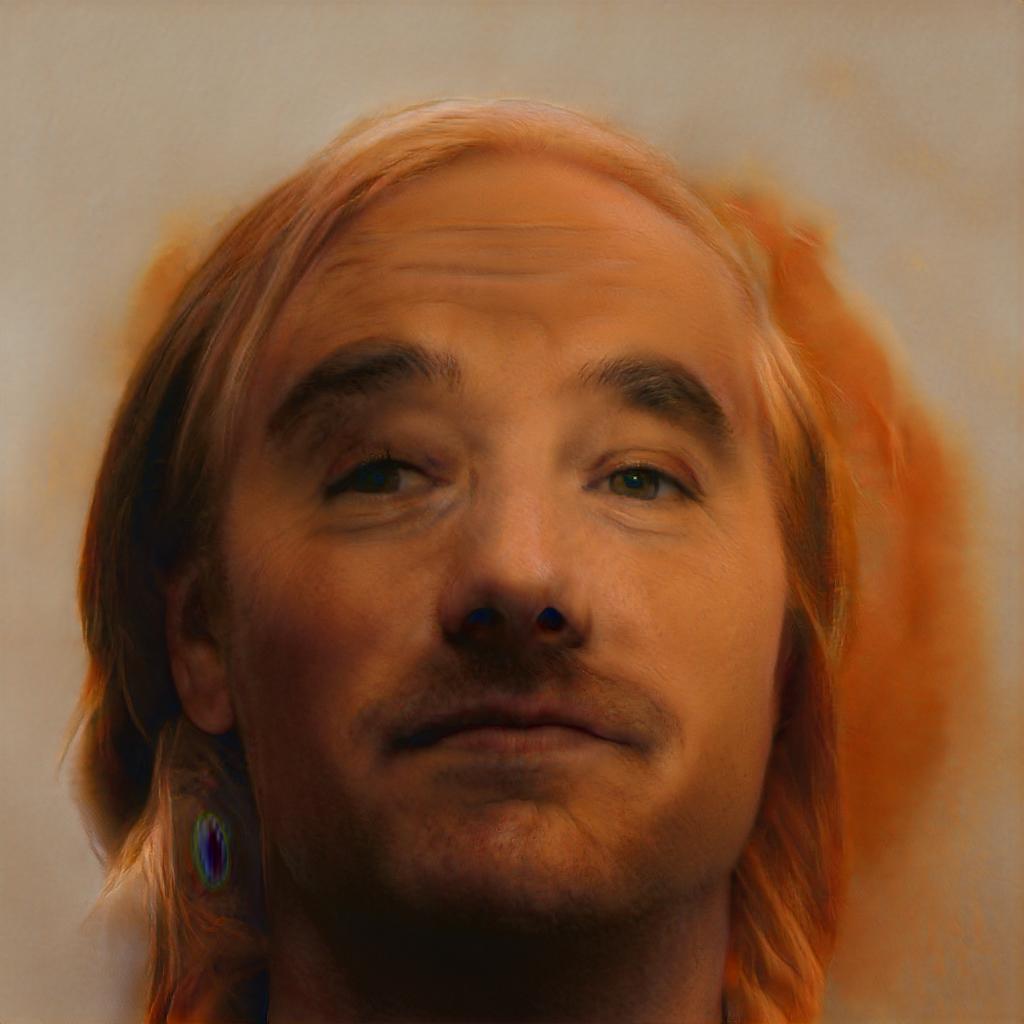} &
		\includegraphics[width=.109\linewidth]{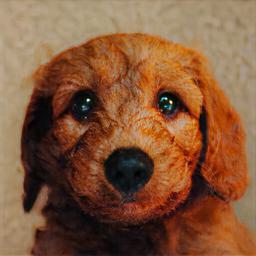} &
		\includegraphics[width=.109\linewidth]{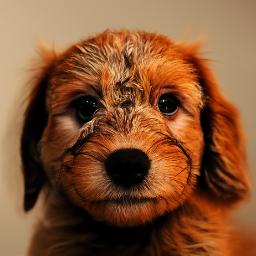} &
		\includegraphics[width=.109\linewidth]{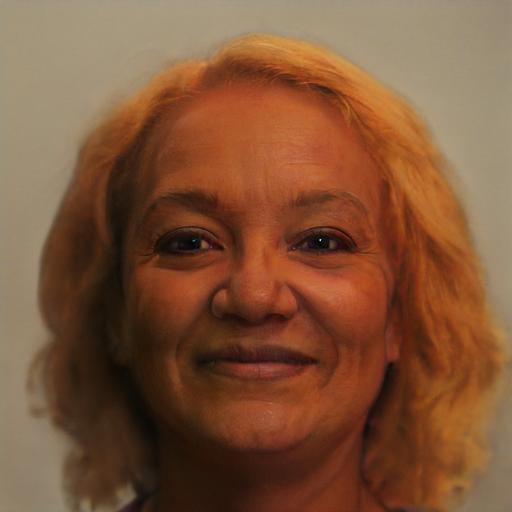}\\	
		\includegraphics[width=.109\linewidth]{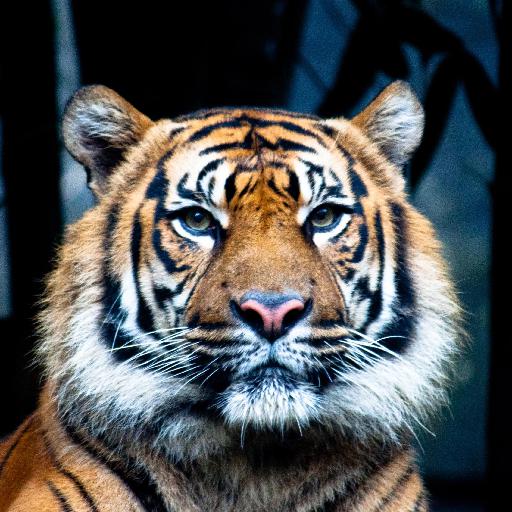} &
		\includegraphics[width=.109\linewidth]{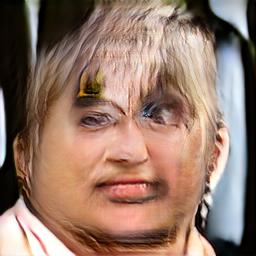} &
		\includegraphics[width=.109\linewidth]{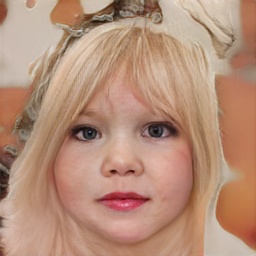} &
		\includegraphics[width=.109\linewidth]{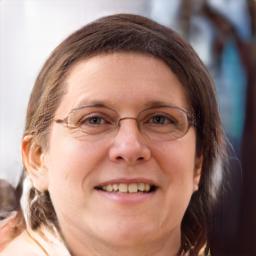} &
		\includegraphics[width=.109\linewidth]{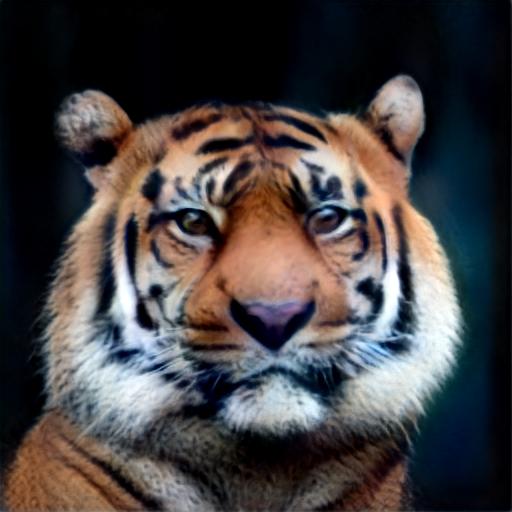} &
		\includegraphics[width=.109\linewidth]{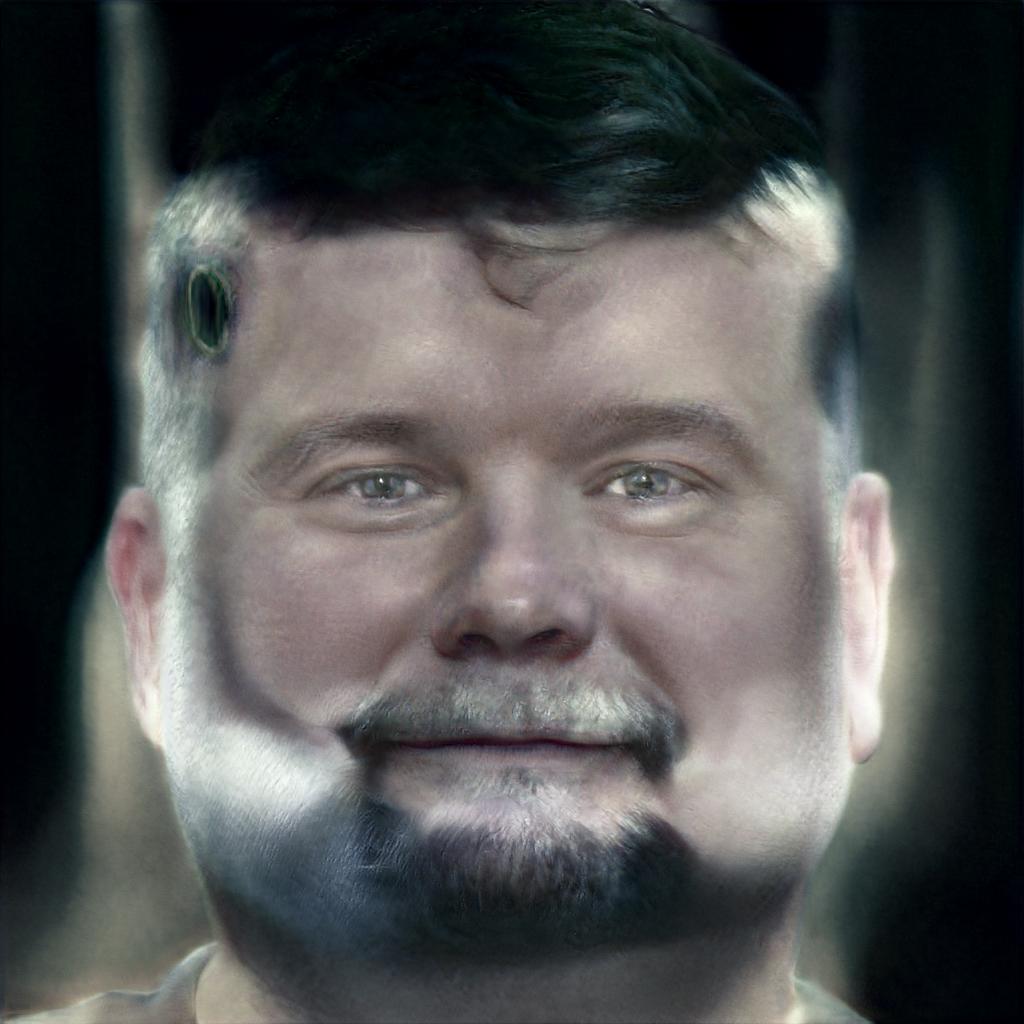} &
		\includegraphics[width=.109\linewidth]{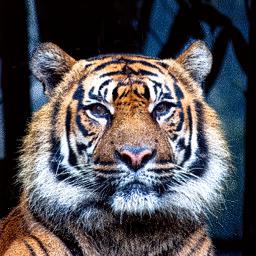} &
		\includegraphics[width=.109\linewidth]{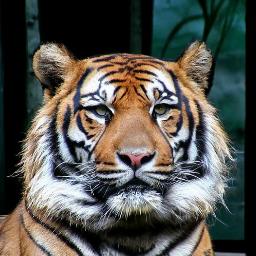} &
		\includegraphics[width=.109\linewidth]{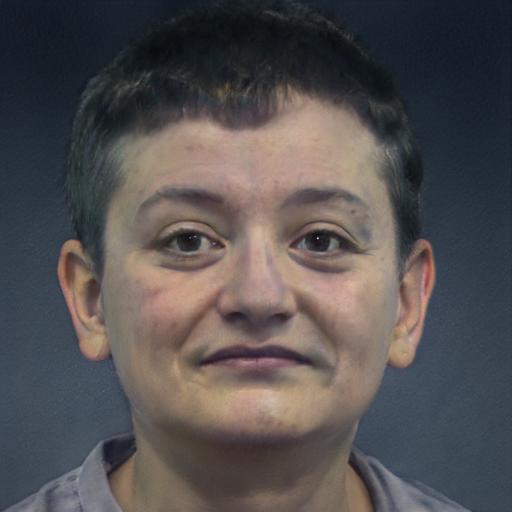}\\	
		
		\includegraphics[width=.109\linewidth]{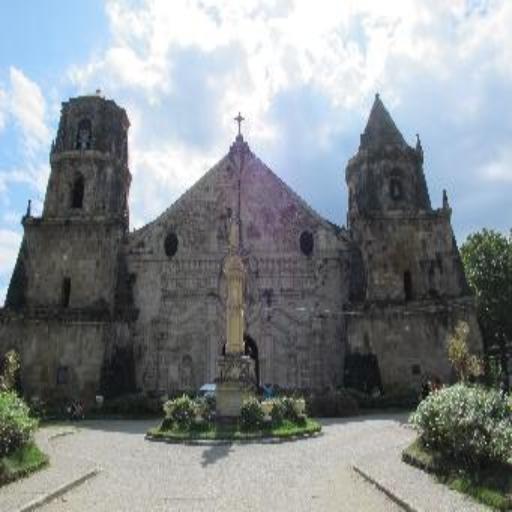} &
		\includegraphics[width=.109\linewidth]{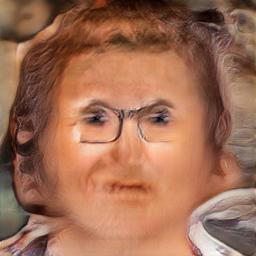} &
		\includegraphics[width=.109\linewidth]{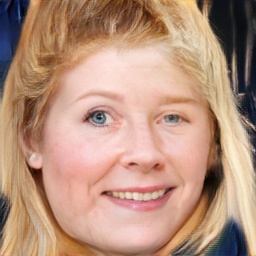} &
		\includegraphics[width=.109\linewidth]{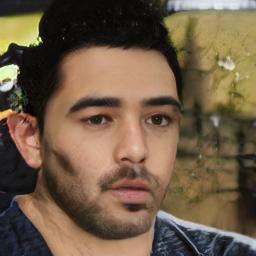} &
		\includegraphics[width=.109\linewidth]{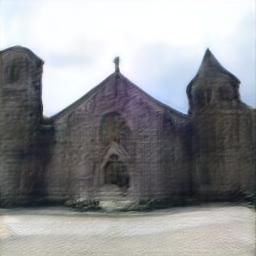} &
		\includegraphics[width=.109\linewidth]{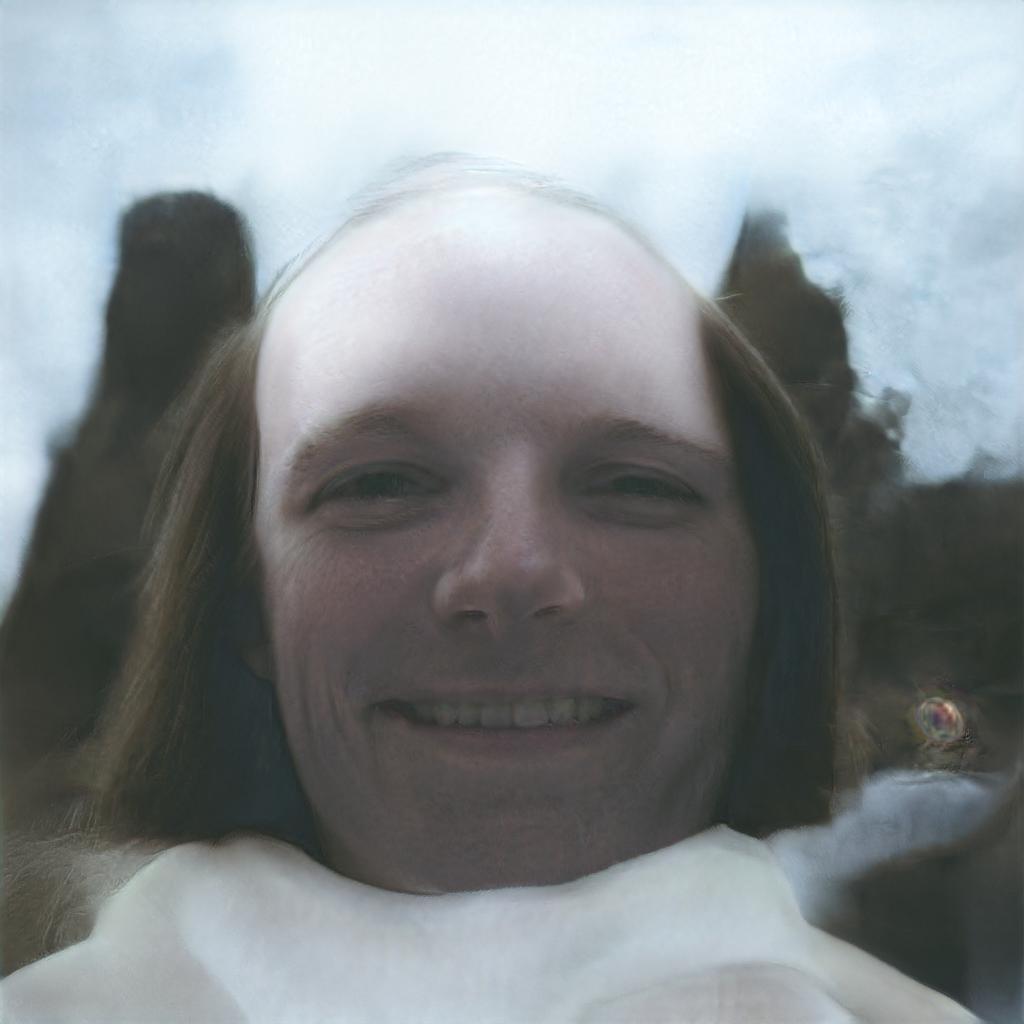} &
		\includegraphics[width=.109\linewidth]{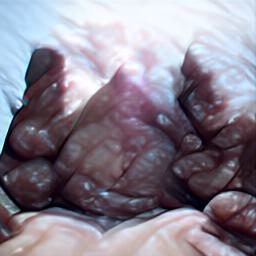} &
		\includegraphics[width=.109\linewidth]{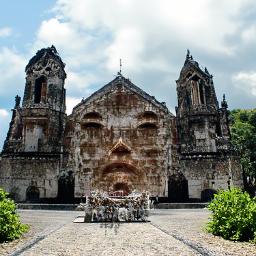} &
		\includegraphics[width=.109\linewidth]{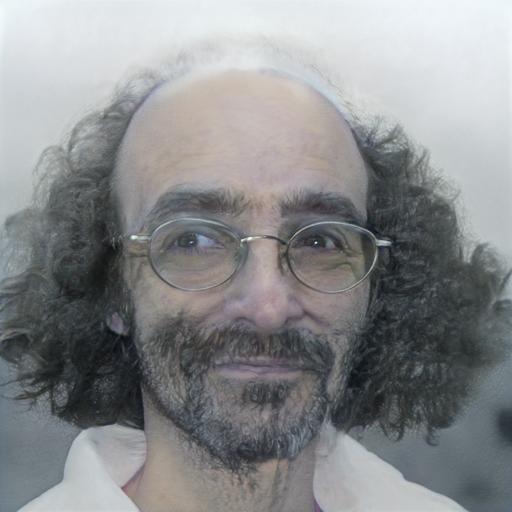}\\
		\small{Source} & \small{VQ-I2I} &\small{GP-UNIT}&\small{StarGAN2}&\small{DiFa}&\small{PULSE}&\footnotesize{DiffusionCLIP}&\small{DiffuseIT}&\small{UniTranslator}\\
		\small{Image}&\small{\cite{chen2022eccv}}&\small{\cite{yang2022unsupervised}}&\small{\cite{choi2020stargan}}&\small{\cite{zhang2022towards}}&\small{\cite{menon2020pulse}}&\small{\cite{Kim_2022_CVPR}}&\small{\cite{kwon2022diffusion}}&(\textbf{ours})\\
	\end{tabular}
	\caption{Qualitative comparison of our UniTranslator with state-of-the-art methods for translating $\mathcal{X}$ to FFHQ. Note that significant issues encountered by other methods, including severe distortions (VQ-I2I), poor cross-domain correspondences (GP-UNIT and StarGAN2), presence of source domain patterns (VQ-I2I and PULSE), and even the inability to generate target domain patterns (Difa, DiffusionCLIP, and DiffuseIT).}
	\label{fig:comparsion1}
\end{figure*}

\begin{figure*}[t]
	\centering
	\setlength{\abovecaptionskip}{0cm}
	\centering
	\setlength{\tabcolsep}{0.05em}	
	\begin{tabular}{ccccccccc}
		\includegraphics[width=.109\linewidth]{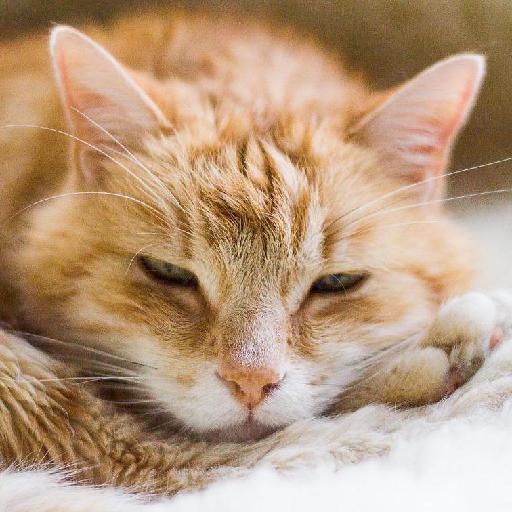} &
		\includegraphics[width=.109\linewidth]{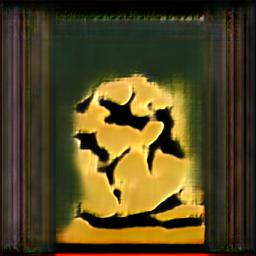} &
		\includegraphics[width=.109\linewidth]{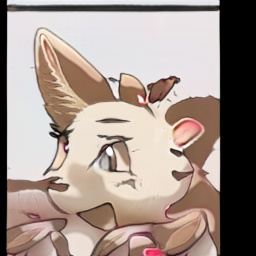} &
		\includegraphics[width=.109\linewidth]{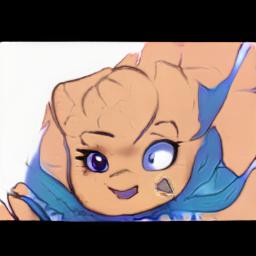} &
		\includegraphics[width=.109\linewidth]{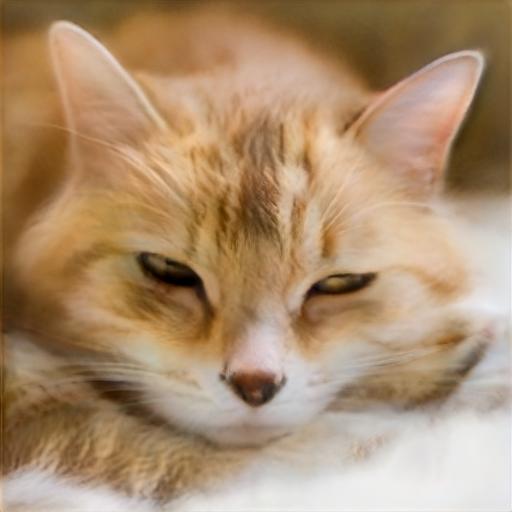} &
		\includegraphics[width=.109\linewidth]{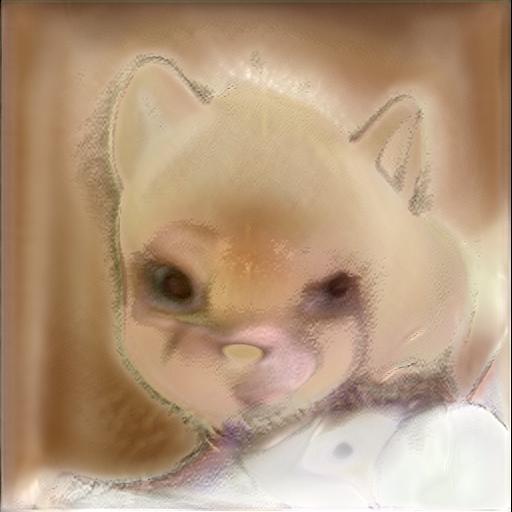} &
		\includegraphics[width=.109\linewidth]{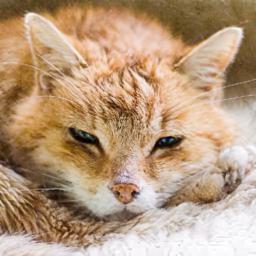} &
		\includegraphics[width=.109\linewidth]{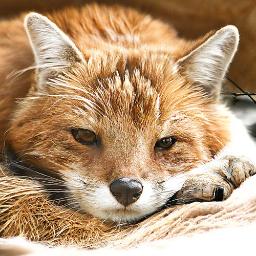} &
		\includegraphics[width=.109\linewidth]{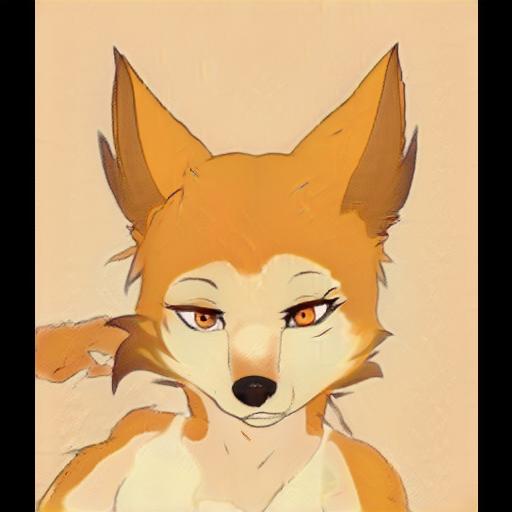} \\		
		\includegraphics[width=.109\linewidth]{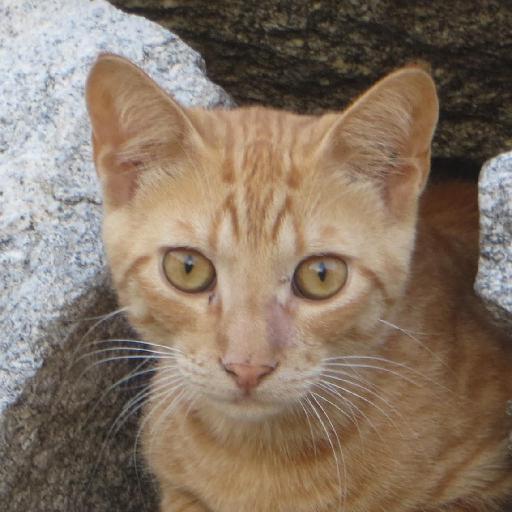} &
		\includegraphics[width=.109\linewidth]{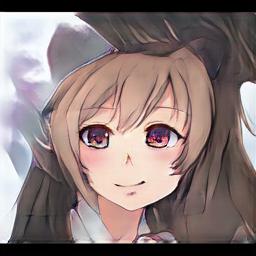} &
		\includegraphics[width=.109\linewidth]{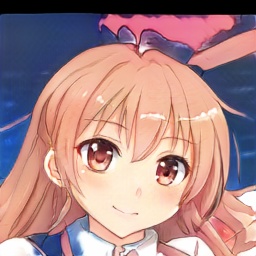} &
		\includegraphics[width=.109\linewidth]{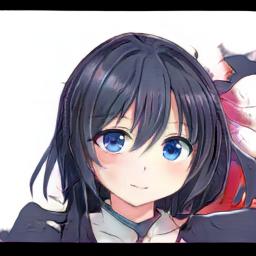} &
		\includegraphics[width=.109\linewidth]{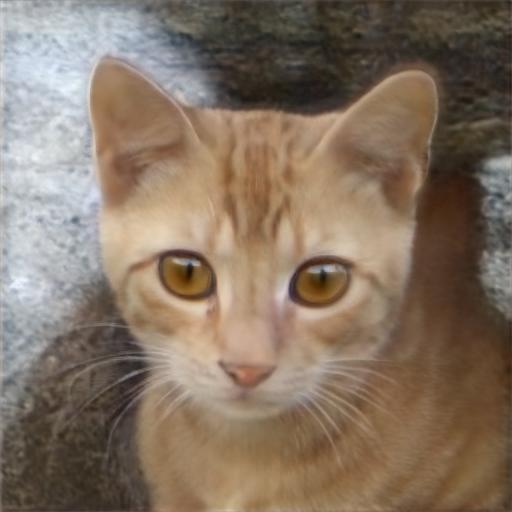} &
		\includegraphics[width=.109\linewidth]{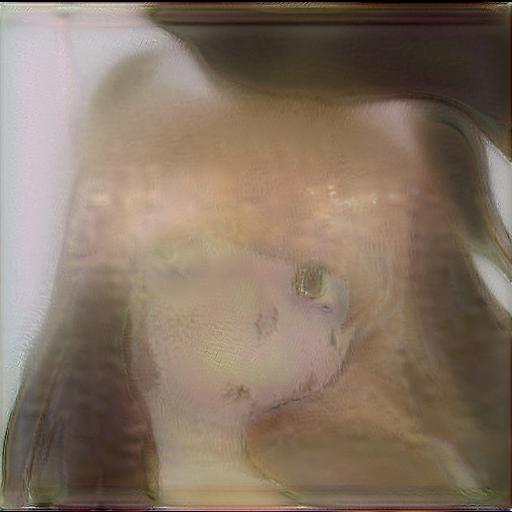} &
		\includegraphics[width=.109\linewidth]{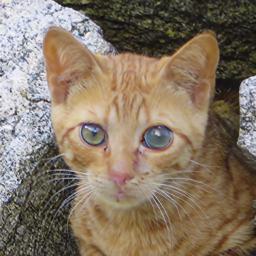} &
		\includegraphics[width=.109\linewidth]{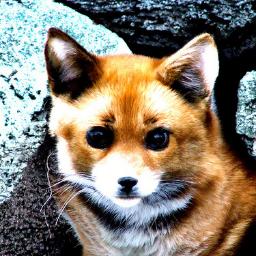} &
		\includegraphics[width=.109\linewidth]{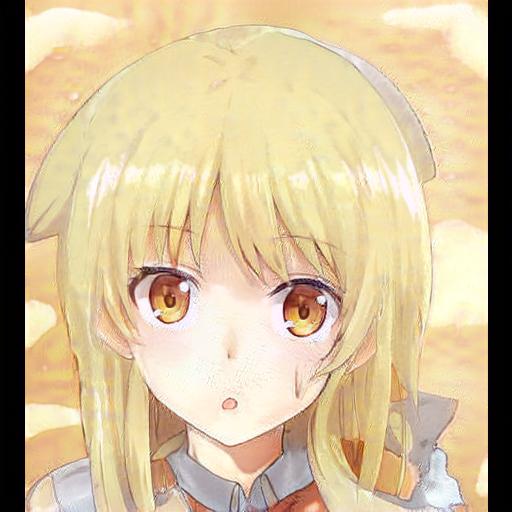} \\	
		\includegraphics[width=.109\linewidth]{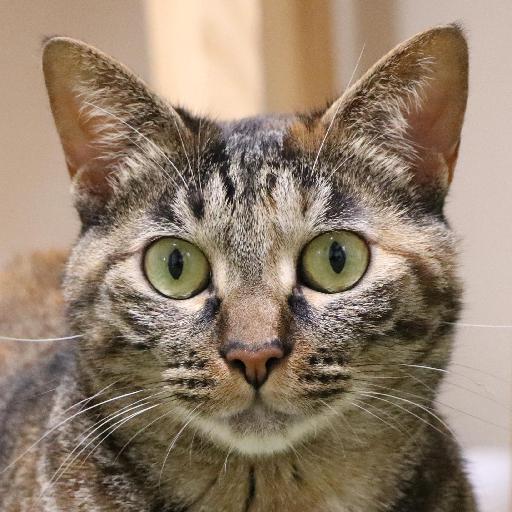} &
		\includegraphics[width=.109\linewidth]{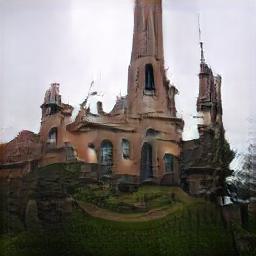} &
		\includegraphics[width=.109\linewidth]{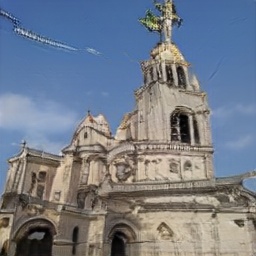} &
		\includegraphics[width=.109\linewidth]{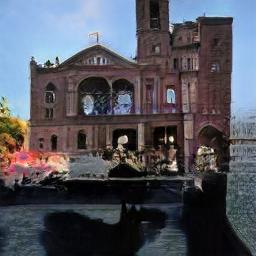} &
		\includegraphics[width=.109\linewidth]{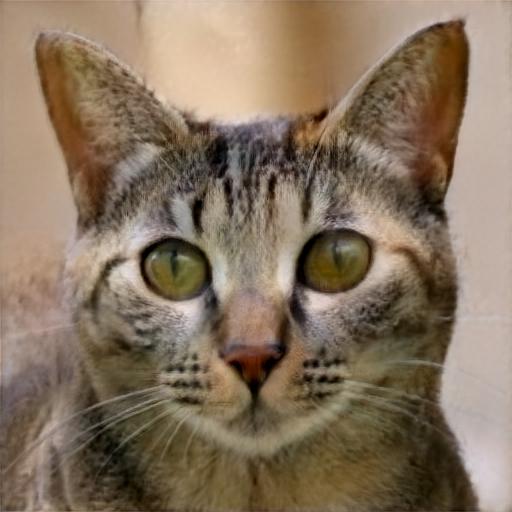} &
		\includegraphics[width=.109\linewidth]{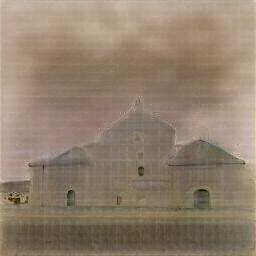} &
		\includegraphics[width=.109\linewidth]{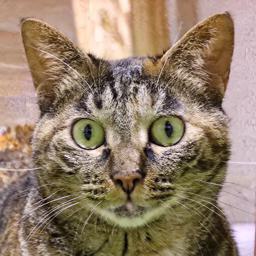} &
		\includegraphics[width=.109\linewidth]{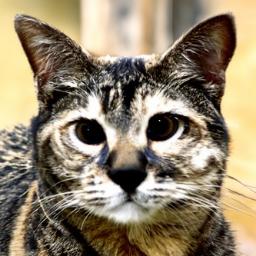} &
		\includegraphics[width=.109\linewidth]{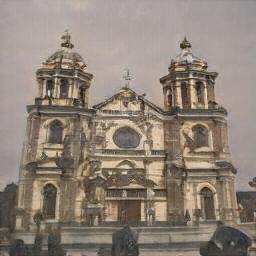} \\
		
		\small{Source}&\small{VQ-I2I}&\small{GP-UNIT}&\small{StarGAN2}&\small{DiFa}&\small{PULSE}&\footnotesize{DiffusionCLIP}&\small{DiffuseIT}&\small{UniTranslator}\\
		\small{Image}&\small{\cite{chen2022eccv}}&\small{\cite{yang2022unsupervised}}&\small{\cite{choi2020stargan}}&\small{\cite{zhang2022towards}}&\small{\cite{menon2020pulse}}&\small{\cite{Kim_2022_CVPR}}&\small{\cite{kwon2022diffusion}}&(\textbf{ours})\\
	\end{tabular}
	\caption{Comparison with the competitors.
		First row: AFHQ-cat$\to$E621Faces; Second row: AFHQ-cat$\to$	Anime;Third row:AFHQ-cat$\to$LSUN-church.}
	\label{fig:comparsion2}
\end{figure*}
\begin{figure*}[t]    	
	\centering
	\setlength{\abovecaptionskip}{0cm}
	\centering
	\setlength{\tabcolsep}{0.05em}
	\begin{tabular}{cccccccccccccccc}		
		\includegraphics[width=.059\linewidth]{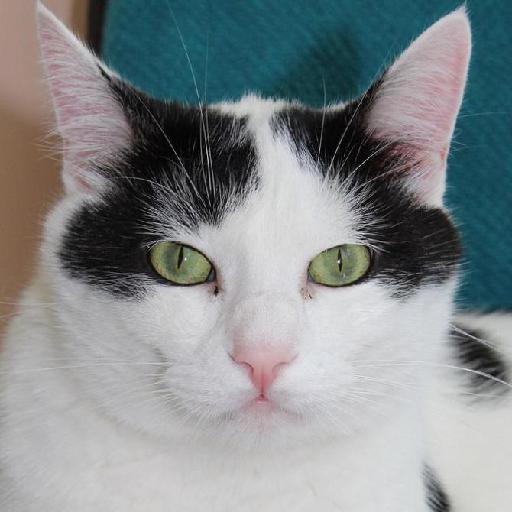} &
		\hspace{0.1mm}
		\includegraphics[width=.059\linewidth]{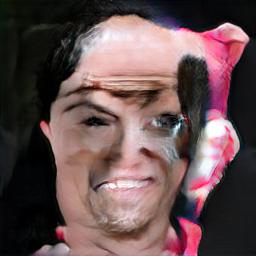} &
		\includegraphics[width=.059\linewidth]{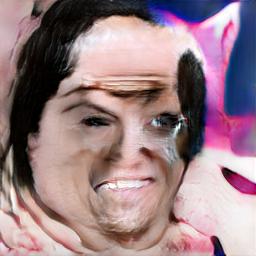} &
		\includegraphics[width=.059\linewidth]{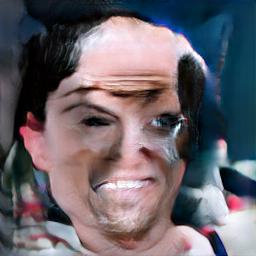} &
		\hspace{0.1mm}
		\includegraphics[width=.059\linewidth]{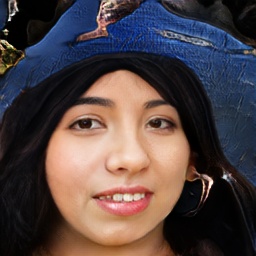} &
		\includegraphics[width=.059\linewidth]{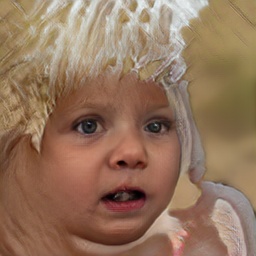} &
		\includegraphics[width=.059\linewidth]{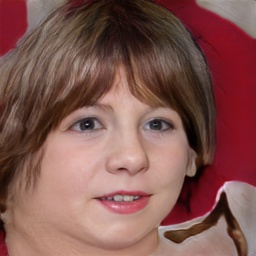} &
		\hspace{0.1mm}		
		\includegraphics[width=.059\linewidth]{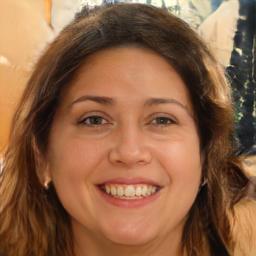} &
		\includegraphics[width=.059\linewidth]{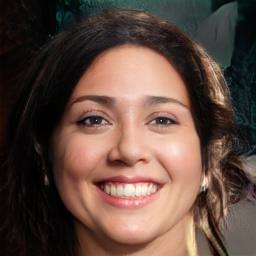} &
		\includegraphics[width=.059\linewidth]{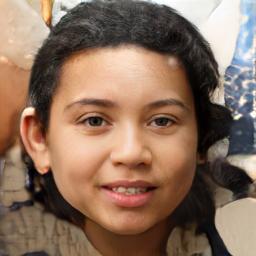} &
		\hspace{0.1mm}	
		\includegraphics[width=.059\linewidth]{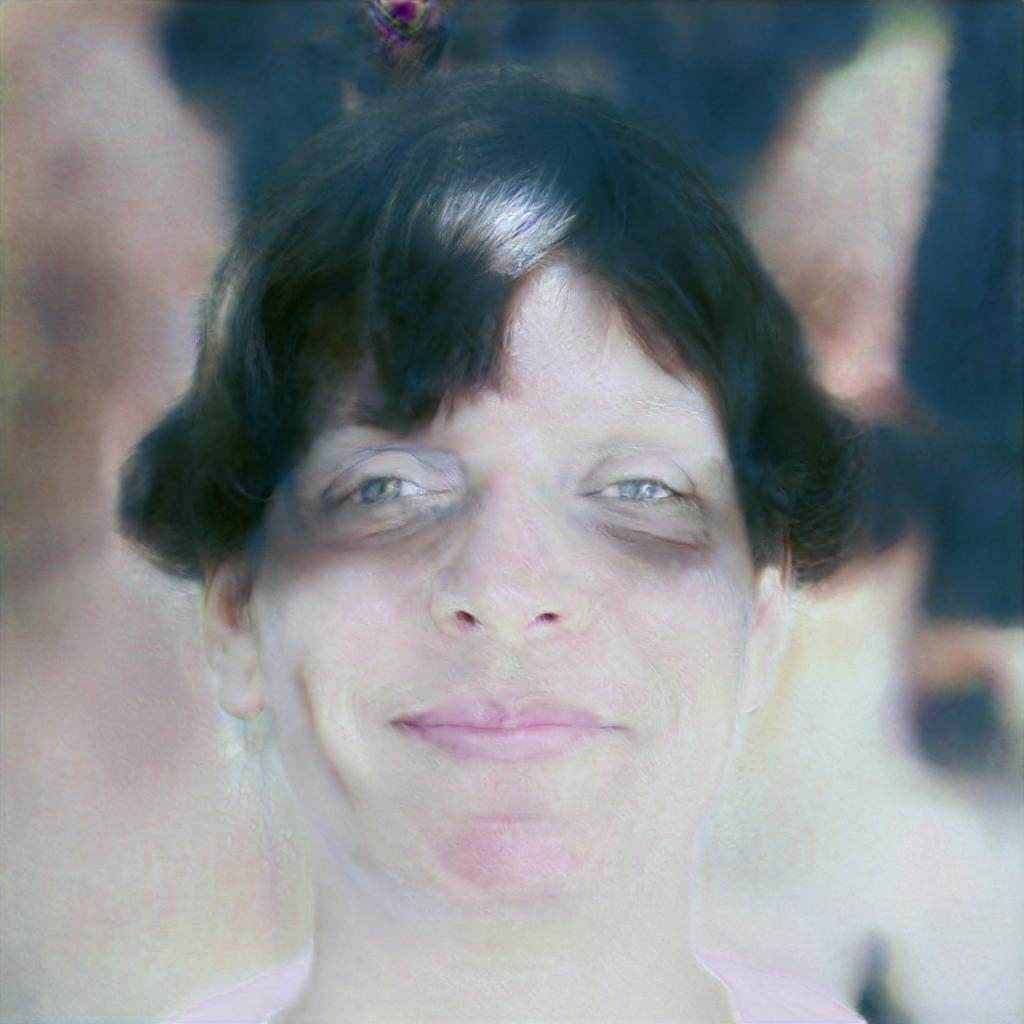} &
		\includegraphics[width=.059\linewidth]{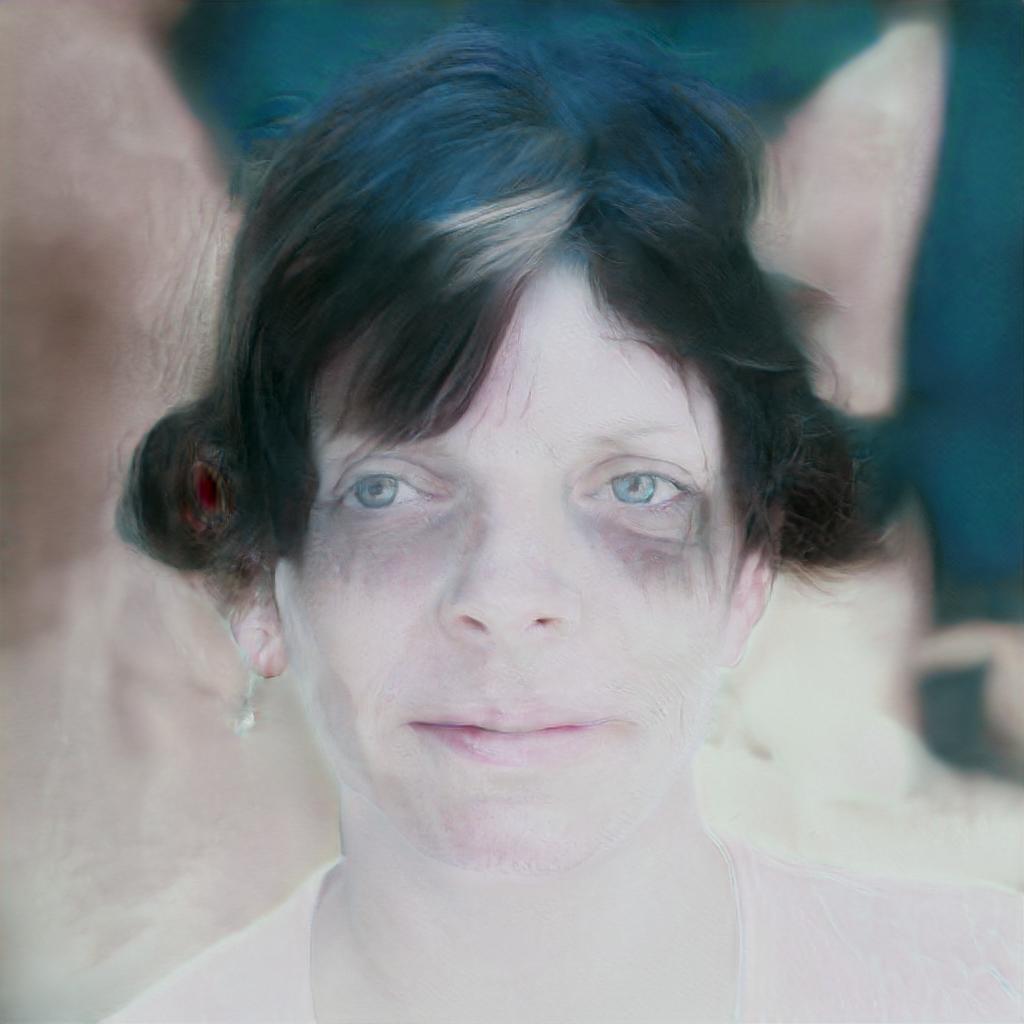} &
		\includegraphics[width=.059\linewidth]{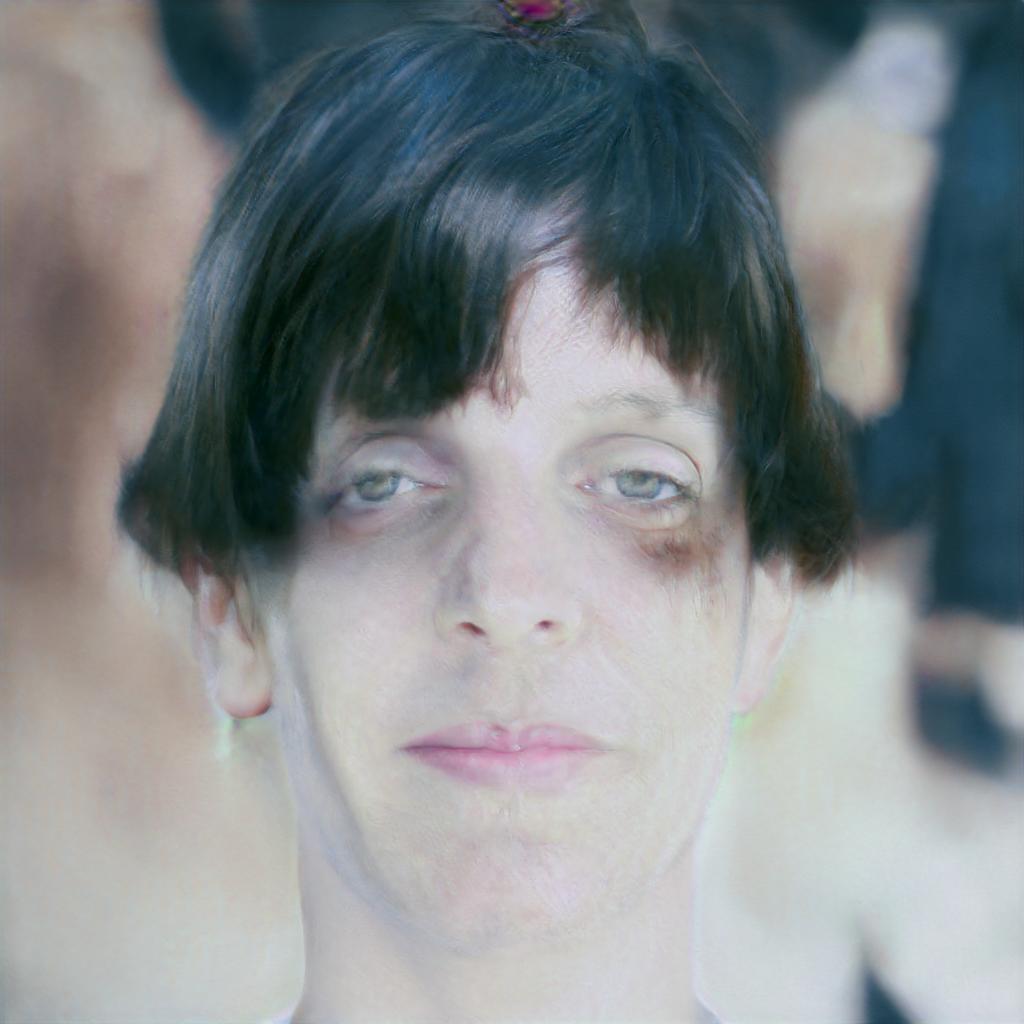} &
		\hspace{0.1mm}
		\includegraphics[width=.059\linewidth]{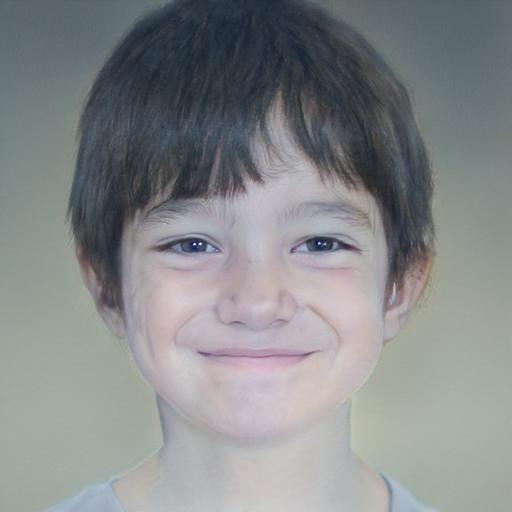} &
		\includegraphics[width=.059\linewidth]{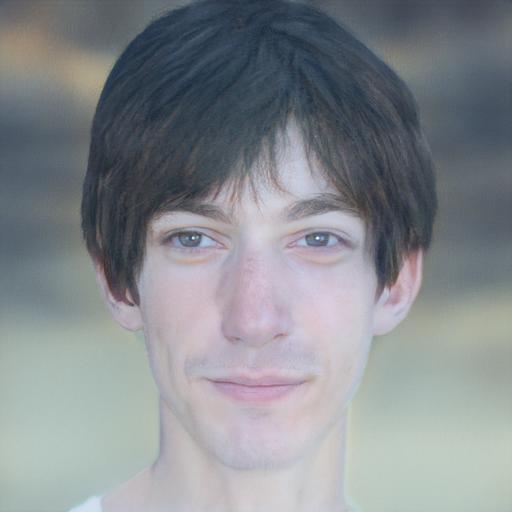} &
		\includegraphics[width=.059\linewidth]{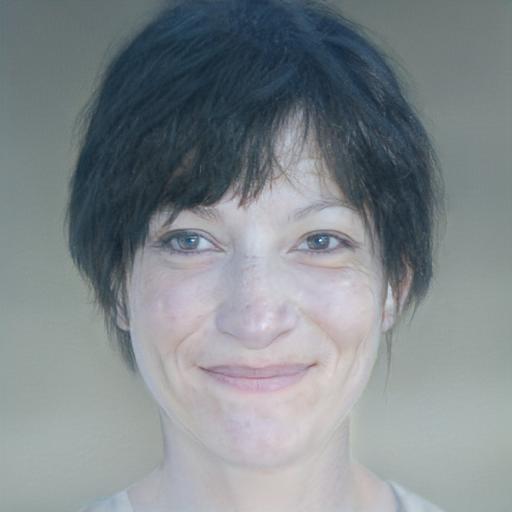}
		\\			
		\includegraphics[width=.059\linewidth]{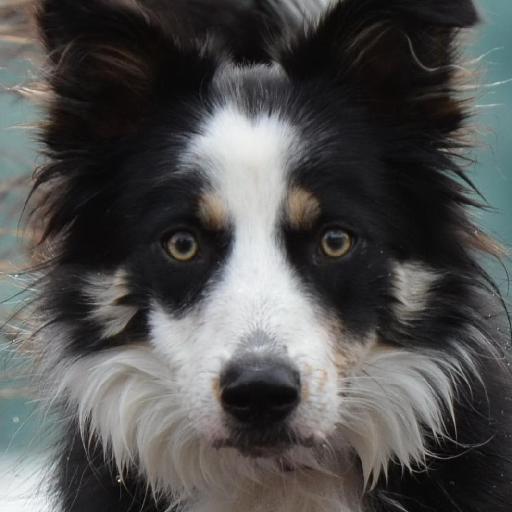} &
		\hspace{0.1mm}
		\includegraphics[width=.059\linewidth]{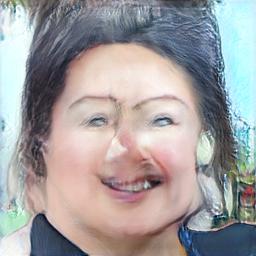} &
		\includegraphics[width=.059\linewidth]{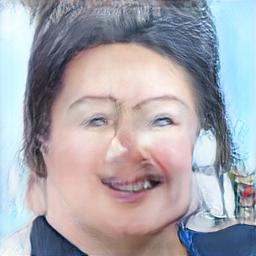} &
		\includegraphics[width=.059\linewidth]{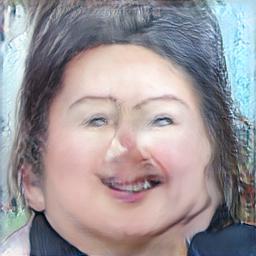} &
		\hspace{0.1mm}
		\includegraphics[width=.059\linewidth]{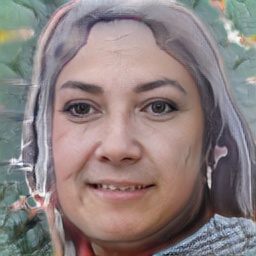} &
		\includegraphics[width=.059\linewidth]{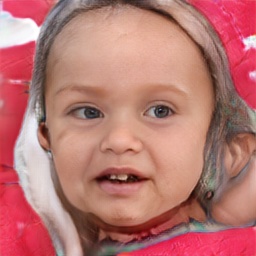} &
		\includegraphics[width=.059\linewidth]{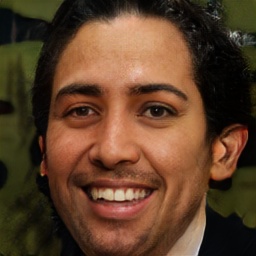} &
		\hspace{0.1mm}
		\includegraphics[width=.059\linewidth]{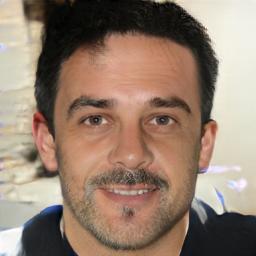} &
		\includegraphics[width=.059\linewidth]{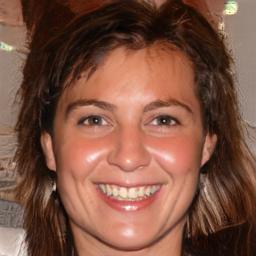} &
		\includegraphics[width=.059\linewidth]{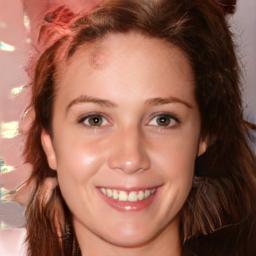} &
		\hspace{0.1mm}	
		\includegraphics[width=.059\linewidth]{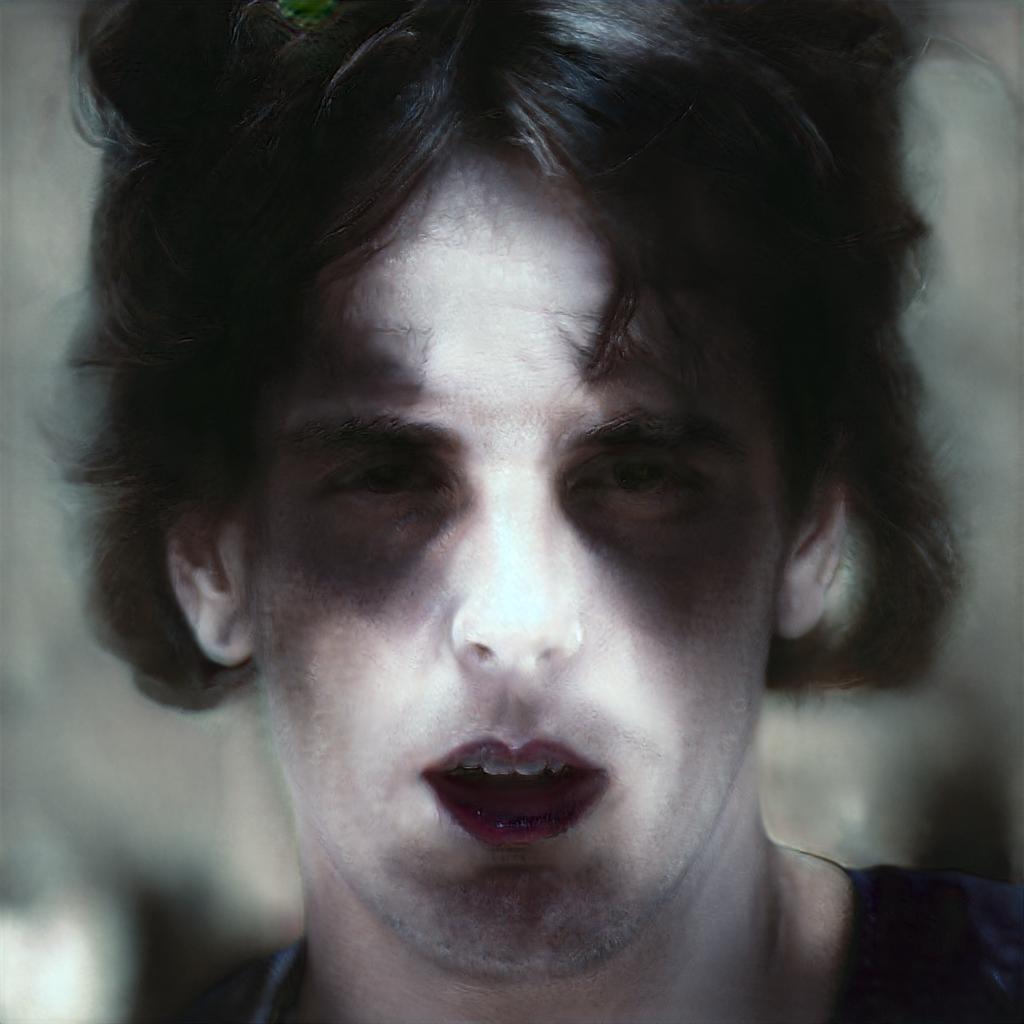} &
		\includegraphics[width=.059\linewidth]{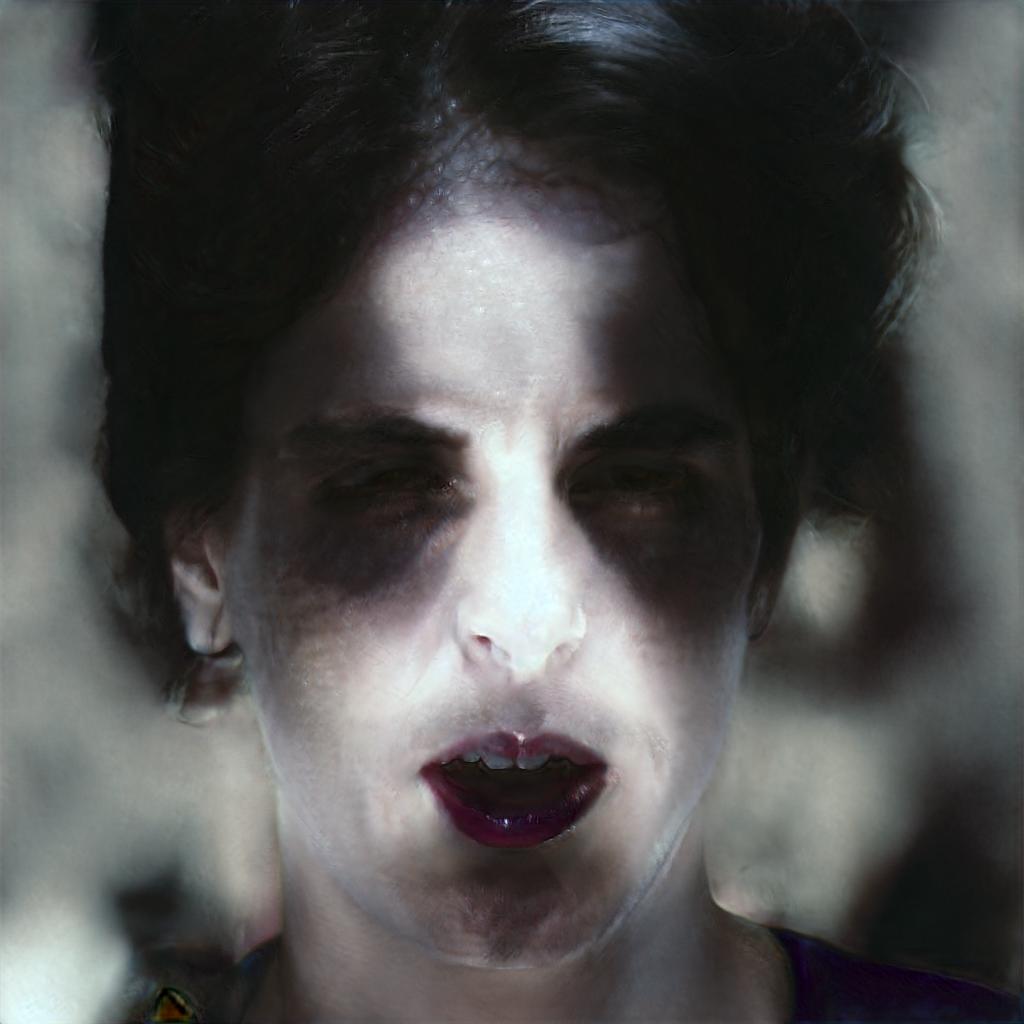} &
		\includegraphics[width=.059\linewidth]{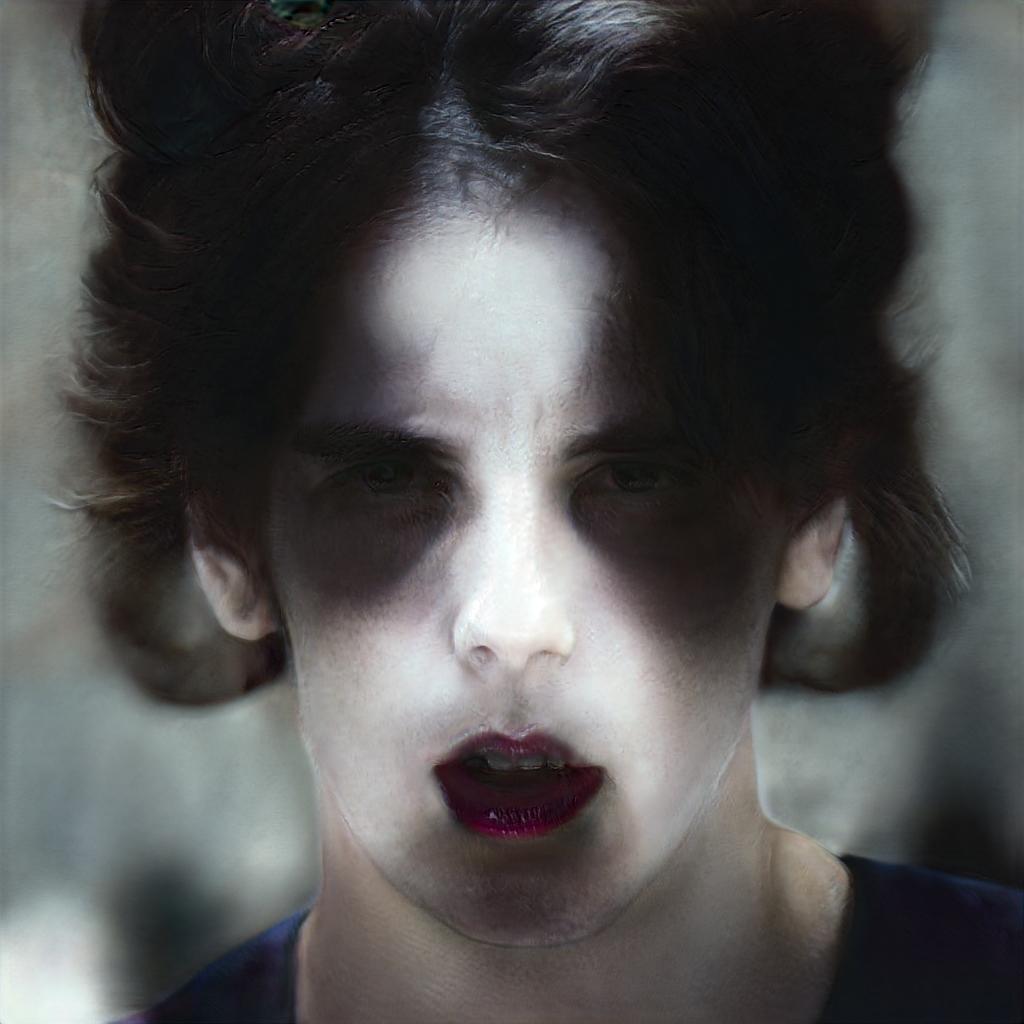} &
		\hspace{0.1mm}
		\includegraphics[width=.059\linewidth]{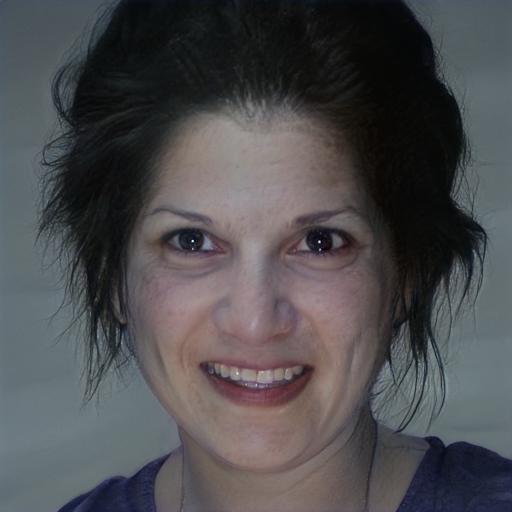} &
		\includegraphics[width=.059\linewidth]{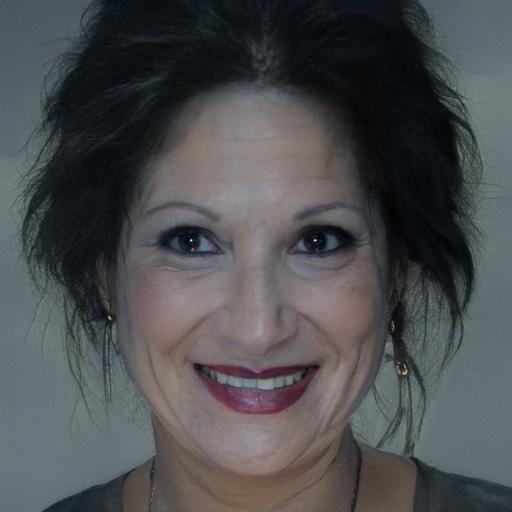} &
		\includegraphics[width=.059\linewidth]{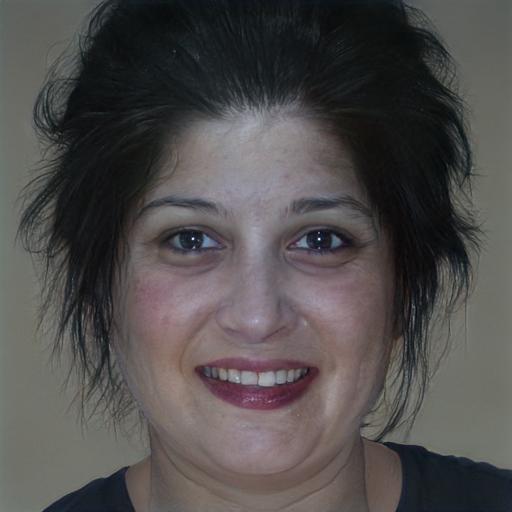}
		\\			
		\includegraphics[width=.059\linewidth]{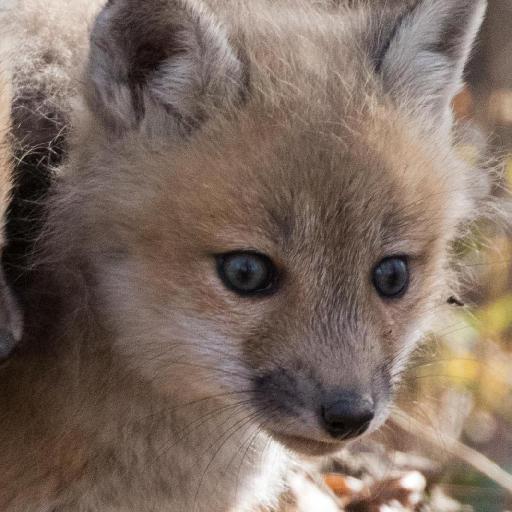} &
		\hspace{0.1mm}
		\includegraphics[width=.059\linewidth]{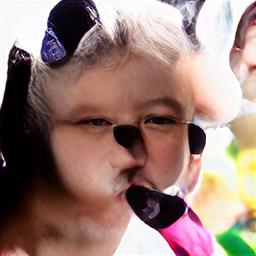} &
		\includegraphics[width=.059\linewidth]{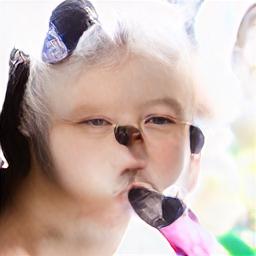} &
		\includegraphics[width=.059\linewidth]{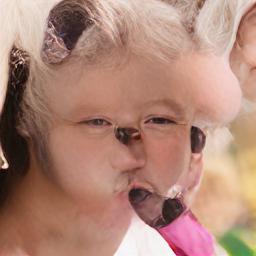} &
		\hspace{0.1mm}
		\includegraphics[width=.059\linewidth]{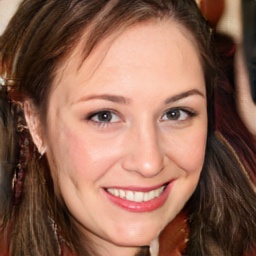} &
		\includegraphics[width=.059\linewidth]{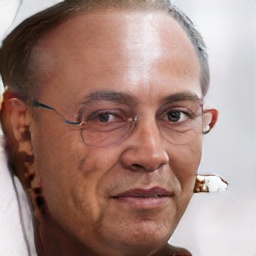} &
		\includegraphics[width=.059\linewidth]{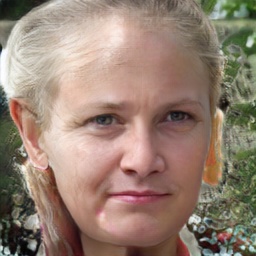} &
		\hspace{0.1mm}
		\includegraphics[width=.059\linewidth]{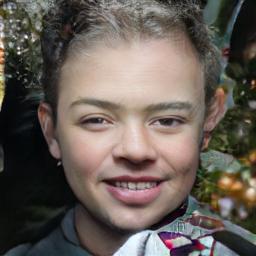} &
		\includegraphics[width=.059\linewidth]{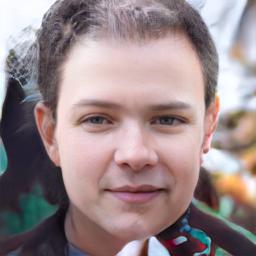} &
		\includegraphics[width=.059\linewidth]{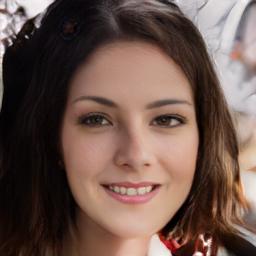} &
		\hspace{0.1mm}	
		\includegraphics[width=.059\linewidth]{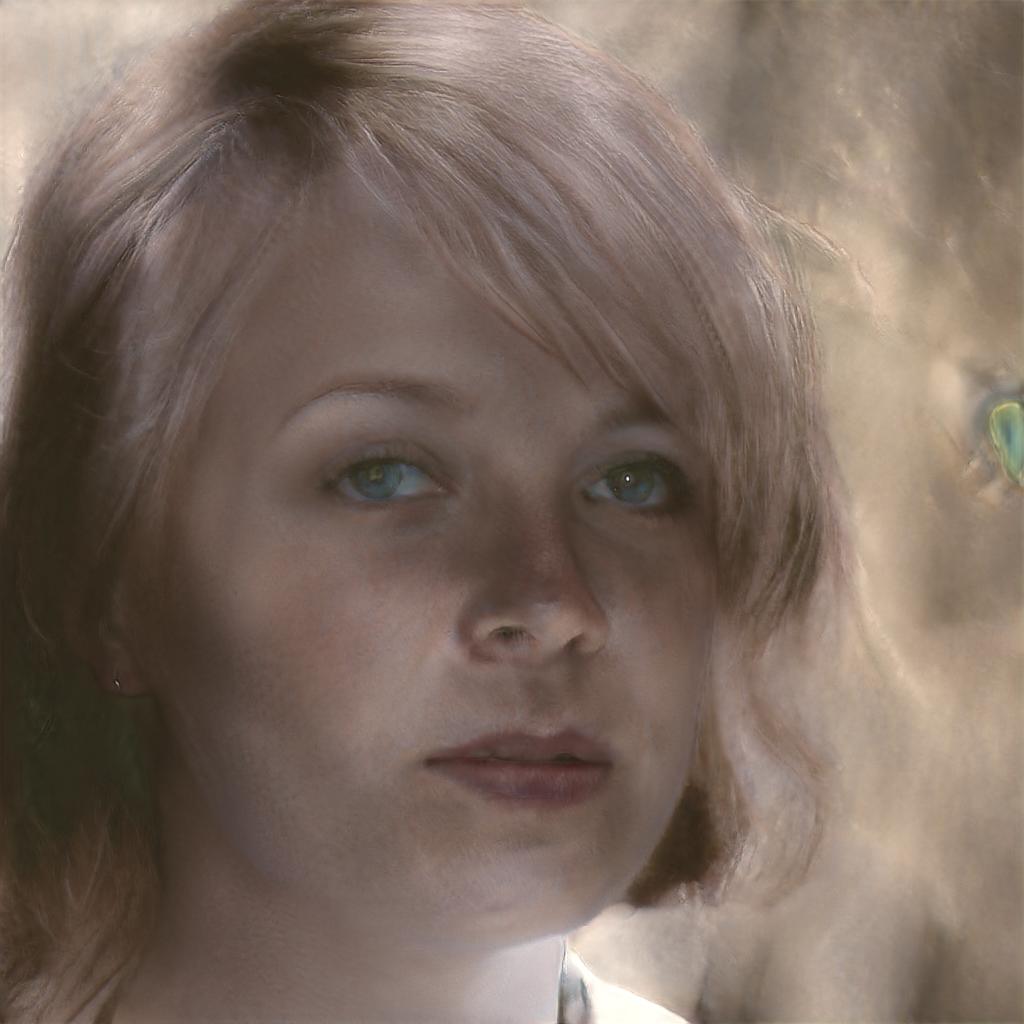} &
		\includegraphics[width=.059\linewidth]{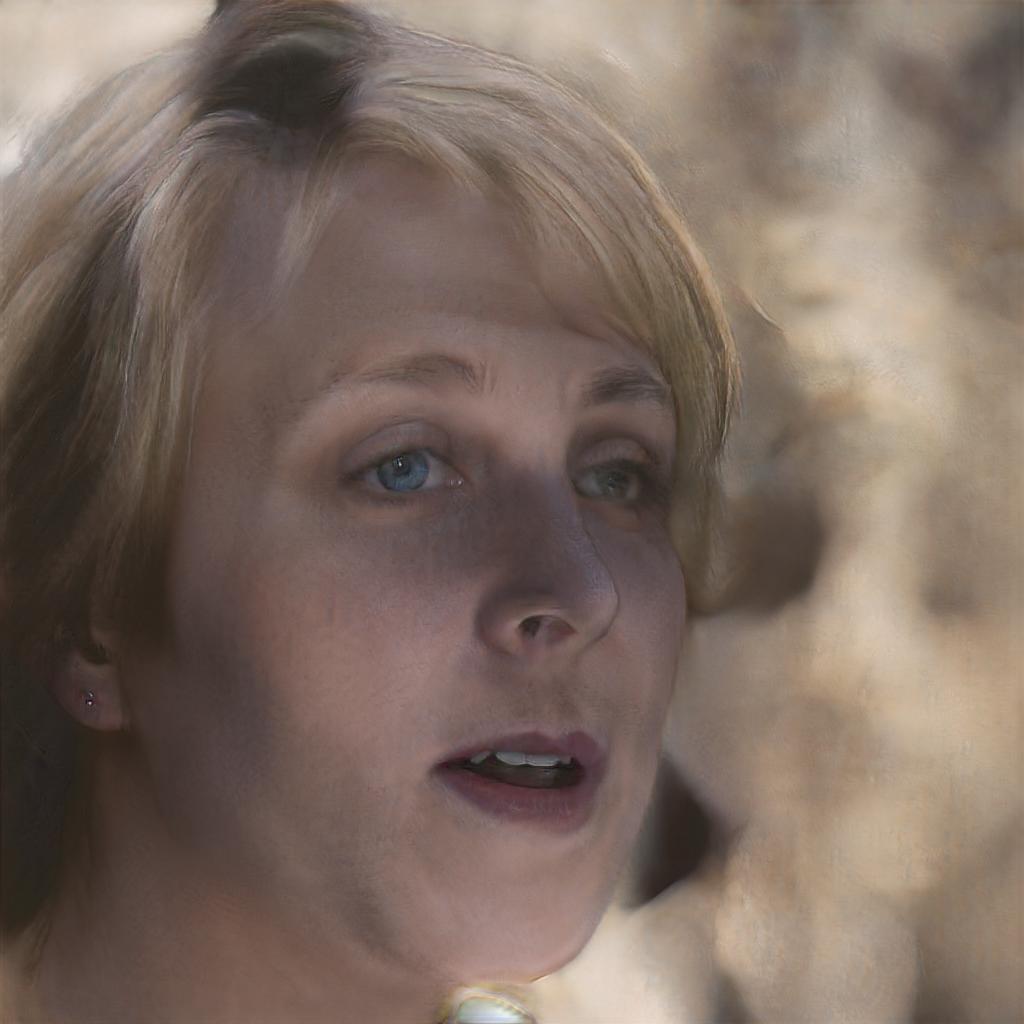} &
		\includegraphics[width=.059\linewidth]{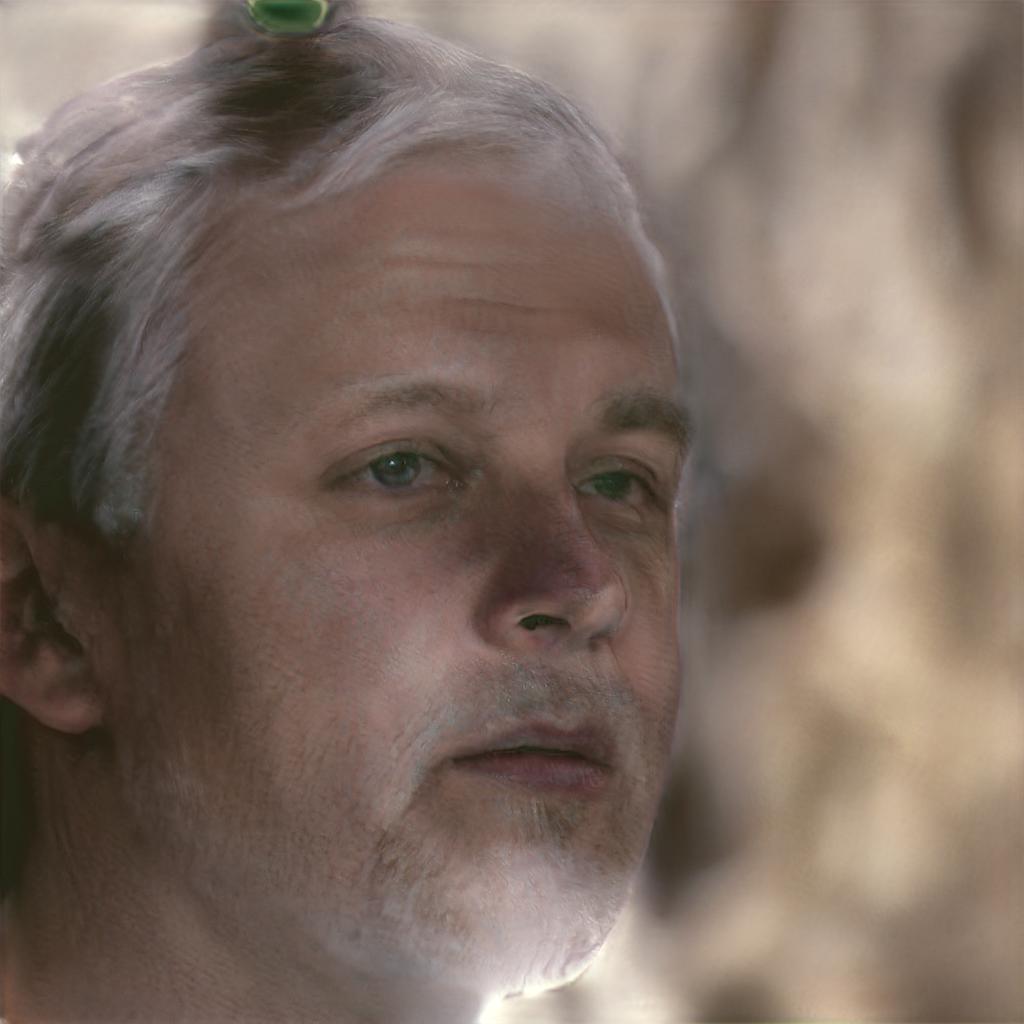} &
		\hspace{0.1mm}
		\includegraphics[width=.059\linewidth]{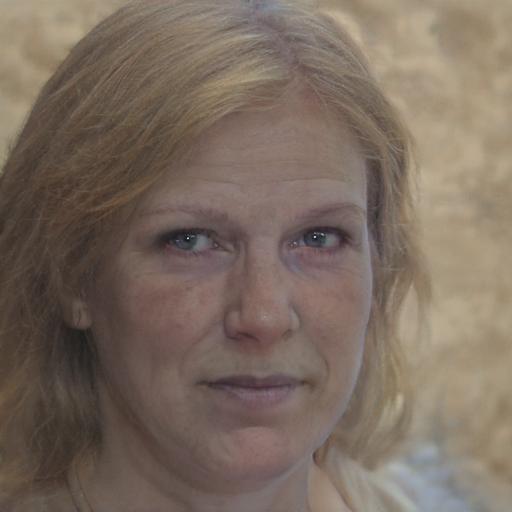} &
		\includegraphics[width=.059\linewidth]{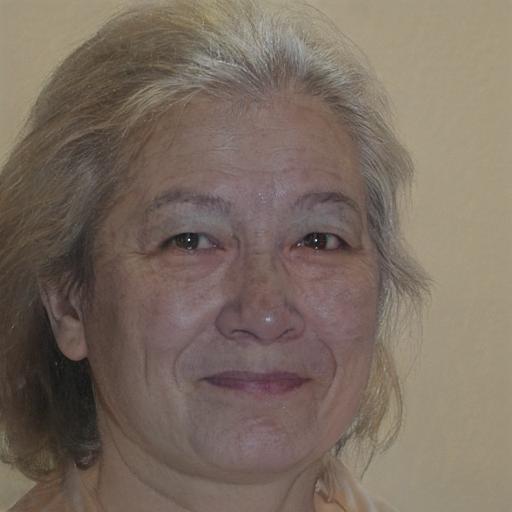} &
		\includegraphics[width=.059\linewidth]{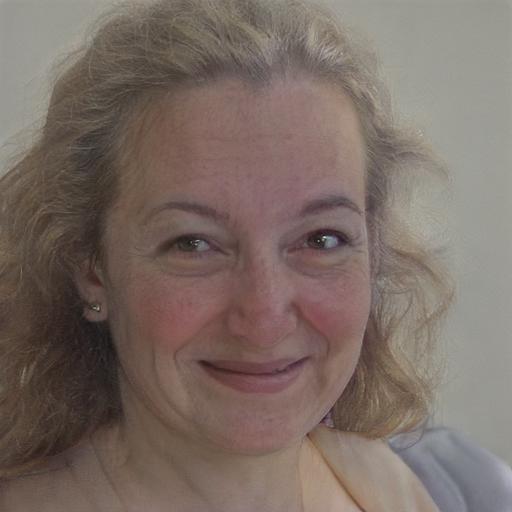}
		\\	
		\includegraphics[width=.059\linewidth]{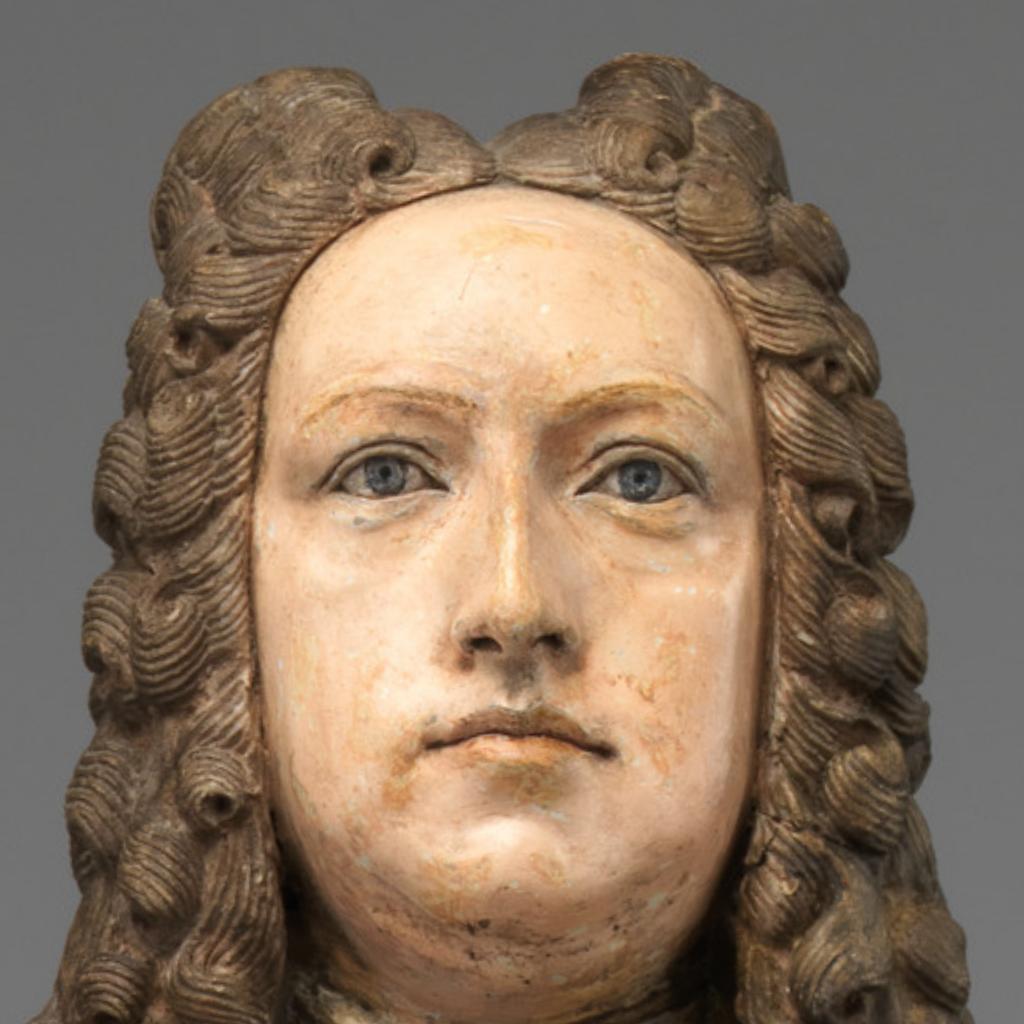} &
		\hspace{0.1mm}
		\includegraphics[width=.059\linewidth]{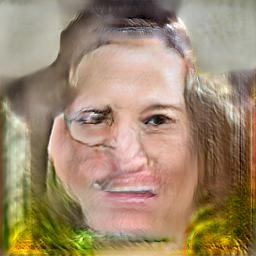} &
		\includegraphics[width=.059\linewidth]{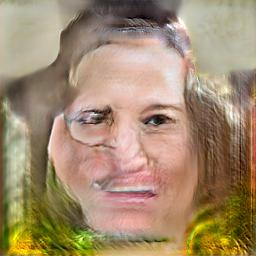} &
		\includegraphics[width=.059\linewidth]{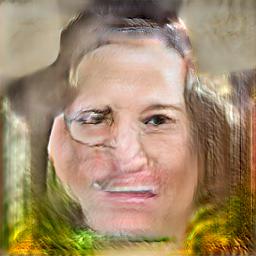} &
		\hspace{0.1mm}
		\includegraphics[width=.059\linewidth]{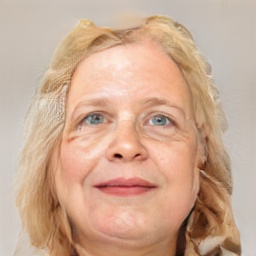} &
		\includegraphics[width=.059\linewidth]{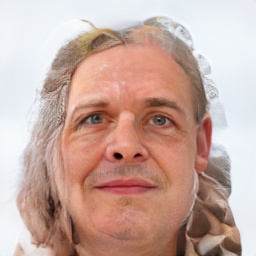} &
		\includegraphics[width=.059\linewidth]{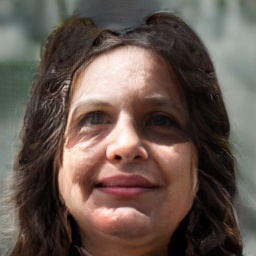} &
		\hspace{0.1mm}
		\includegraphics[width=.059\linewidth]{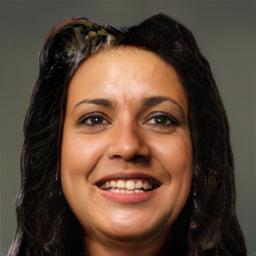} &
		\includegraphics[width=.059\linewidth]{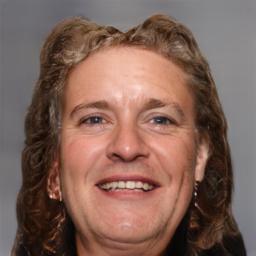} &
		\includegraphics[width=.059\linewidth]{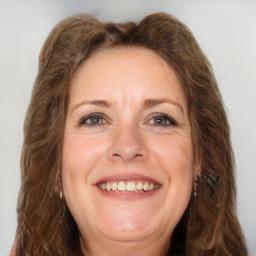} &
		\hspace{0.1mm}
		\includegraphics[width=.059\linewidth]{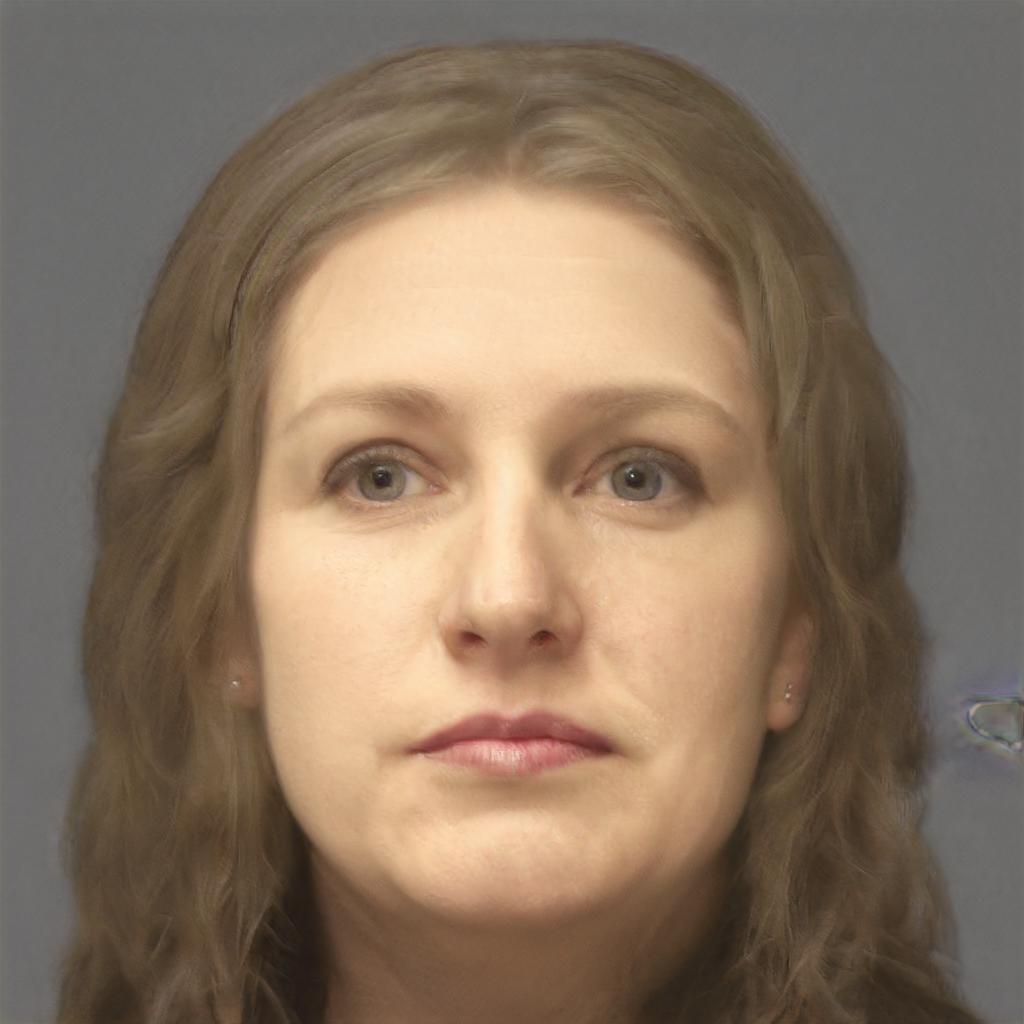} &
		\includegraphics[width=.059\linewidth]{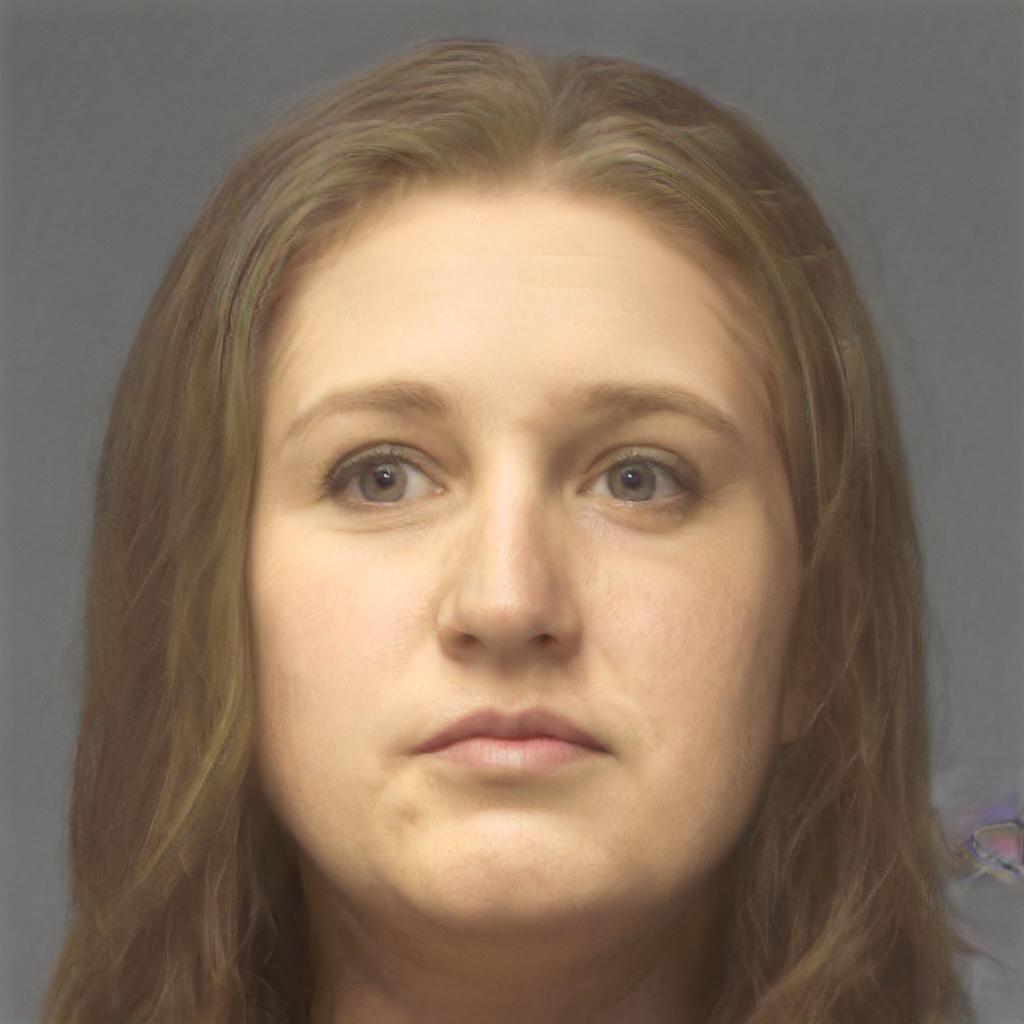} &
		\includegraphics[width=.059\linewidth]{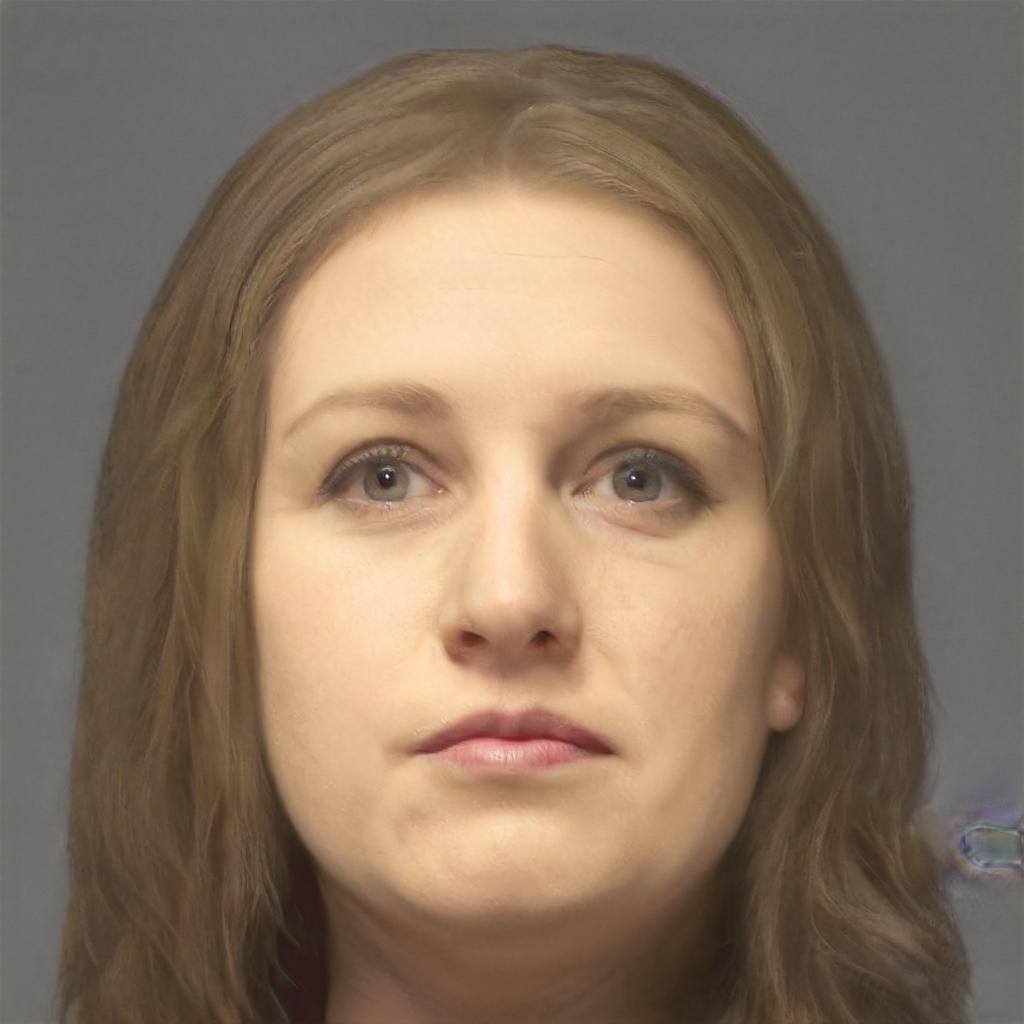} &
		\hspace{0.1mm}
		\includegraphics[width=.059\linewidth]{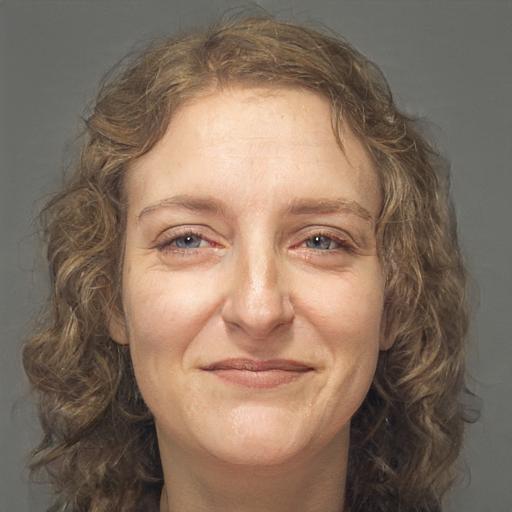} &
		\includegraphics[width=.059\linewidth]{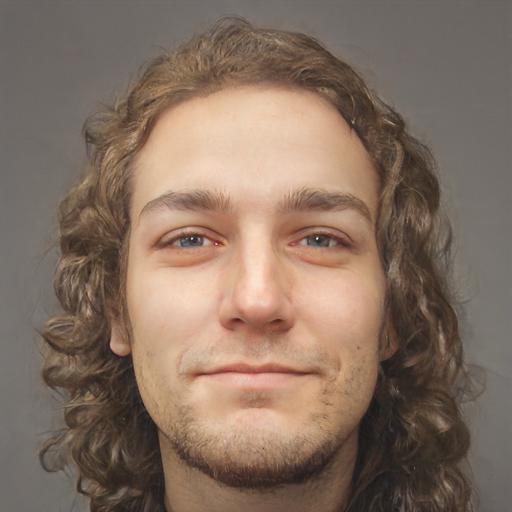} &
		\includegraphics[width=.059\linewidth]{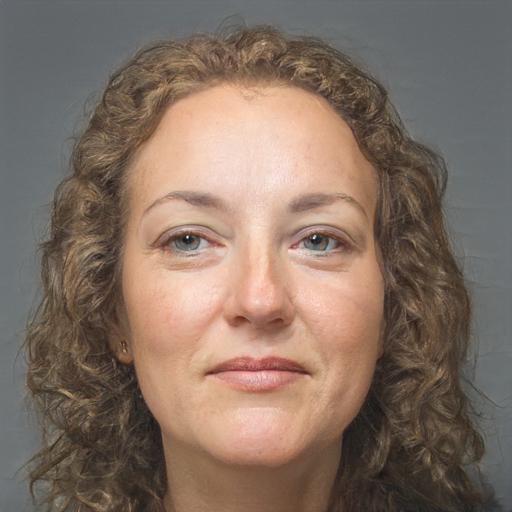}
		\\			
		Source & \multicolumn{3}{c}{VQ-I2I}
		&\multicolumn{3}{c}{GP-UNIT}
		&\multicolumn{3}{c}{StarGAN2}	
		& \multicolumn{3}{c}{PULSE}
		& \multicolumn{3}{c}{UniTranslator (\textbf{ours})}\\
		Image&\multicolumn{3}{c}{\cite{chen2022eccv}}&\multicolumn{3}{c}{\cite{yang2022unsupervised}}&\multicolumn{3}{c}{\cite{choi2020stargan}}&\multicolumn{3}{c}{\cite{menon2020pulse}}
	\end{tabular}
	\caption{Diversity comparison. Given one input, UniTranslator generates diverse and sensible results through multiple optimizations, while alternative methods often yield single-modal or lower-quality outputs.}
	\label{fig:diversity}
\end{figure*}
 
Qualitative evaluation results, as shown in Fig. \ref{fig:comparsion1} and Fig. \ref{fig:comparsion2}, demonstrate the limitations of learning-based methods (VQ-I2I, StarGAN2, and GP-UNIT) in achieving high-quality translations. While these methods succeed in translating certain patterns into target domains, the synthesized images are not less appealing. This issue primarily arises because their feature extraction mechanisms excel mainly within visually similar domains. Specifically, VQ-I2I and StarGAN2 overlook high-level correspondences between domains, restricting their capability to handle translations across visually distinct domains. GP-UNIT, which leverages the generative prior from BigGAN~\cite{brock2019large}, can synthesize images in complex domains with substantial visual disparities. However, due to the inadequate disentanglement of BigGAN's latent space, GP-UNIT cannot effectively establish effective correspondences with the source domain concerning contours and color tones, resulting in less realistic outcomes. Additionally, GP-UNIT exhibits a lack of generalization when applied to domains beyond the training range of its pose encoder, further diminishing its performance. These limitations significantly undermine the universality of learning-based methods. In contrast, our UniTranslator performs well by requiring only a single source domain image, disregarding the source domain's range, and extracting domain-agnostic information through a decoupling module. It achieves robust correspondences with the source domain across various aspects, including poses, contours, and color tones. Furthermore, by harnessing the better-disentangled generative prior of StyleGAN, UniTranslator consistently generates highly realistic outputs.

The domain adaptation method, DiFa, and diffusion-based methods, DiffusionCLIP and DiffuseIT, do not translate the images well, as the outputs closely resemble the sources. These methods utilize CLIP losses applied to pre-trained source models for cross-domain translations. However, the qualitative results underscore the insufficiency of this approach in bridging substantial domain gaps. In contrast, our approach uses the CLIP2P mapper to navigate the StyleGAN target manifold, ensuring that the output results reside within the target domain. 

It is worth noting that although PULSE also leverages the generative prior of StyleGAN, it often produces ambiguous translations. Its guidance for traversal relies solely on pixel-wise constraints applied to downsampled images, a strategy well-suited for super-resolution methods but overly simplistic for cross-domain translations. On the other hand, our approach considers both the establishment of cross-domain correspondences and the quality of the generated results, thereby translating images effectively. Additionally, due to UniTranslator's minimal increase in learnable parameters, it steadily translates images through optimization without excessive overheads.

\subsection{Diversity}
The correspondences become increasingly abstract and less rigid as the gap widens between the source and target domains. This natural relaxation also affects the accompanying target-specific information, rendering it similarly more flexible. The confluence of these two types of information can lead to multiple valid solutions, as the inherent multiplicity of reasonable outputs for a given input in cross-domain translation. From the visual perspective, the rationality of the results must take into account cross-domain correspondences, while diversity stems from the flexibility in both cross-domain correspondences and target-specific information.

Fig.~\ref{fig:diversity} shows comparisons that encapsulate this aspect of diversity. Our approach generates diverse yet reasonable results through multiple inferences. The results show that the extracted domain-agnostic information possesses flexibility without sacrificing reasonableness, in line with the fact that strict cross-domain correspondences may not always be guaranteed (such as when inferring a person's facial age based on an animal's face). In contrast, other methods often generate low-quality outputs with poor alignments with the source images or translated images with minimal variation. 

\begin{table*}[t]
	\setlength{\abovecaptionskip}{0cm}
	\caption{Quantitative comparison of our UniTranslator with state-of-the-art methods. We use NIQE and LPIPS Scores to assess the quality and perceptual similarity of the generated images. Both metrics are the lower, the better, and the best results are highlighted in bold with underline.}
	\centering
	\resizebox{1\textwidth}{!}{
		\begin{tabular}{l||l||c|c||c||c||c|c||c|c||c|c||c|c||c|c||c|c}
			\toprule
			\multirow{2}*{Type} & \multirow{2}*{Mapping} & \multicolumn{2}{c||}{VQ-I2I} & \multicolumn{2}{c||}{GP-UNIT} & \multicolumn{2}{c||}{StarGAN2}& \multicolumn{2}{c||}{DiFa} & \multicolumn{2}{c||}{PULSE} & \multicolumn{2}{c||}{DiffusionCLIP} &\multicolumn{2}{c||}{DiffuseIT}& \multicolumn{2}{c}{\textbf{UniTranslator}} \\		
			\cmidrule(lr){3-4}
			\cmidrule(lr){5-6}
			\cmidrule(lr){7-8}		
			\cmidrule(lr){9-10}
			\cmidrule(lr){11-12}
			\cmidrule(lr){13-14}
			\cmidrule(lr){15-16}
			\cmidrule(lr){17-18}
			&&NIQE&LPIPS&NIQE&LPIPS&NIQE&LPIPS&NIQE&LPIPS&NIQE&LPIPS&NIQE&LPIPS&NIQE&LPIPS&NIQE&LPIPS\\
			\midrule			
			Adjacent & Metfaces$\to$FFHQ & 4.81 & 0.68 & 4.44 & 0.62 & 4.47 & 0.60 & 7.47 & - & 5.07 & 0.50 &5.65 & - & 5.79 &0.54 &\textbf{\underline{4.25}} & \textbf{\underline{0.18}}\\			
			
			Adjacent & AFHQ-cat$\to$E621Faces & 7.09 & 0.80 & 6.49 & 0.76 & 5.91 & 0.75 & 5.10 & - & 7.41 & 0.64 & \textbf{\underline{3.75}} & - &6.12 &0.45 &4.94 & \textbf{\underline{0.26}}\\
			
			Far-off & AFHQ-cat$\to$Anime & 6.06 & 0.74 & 5.80 & 0.75 & 6.02 & 0.75 & 4.87 & - & 5.03 & 0.63 & \textbf{\underline{4.07}} & - &5.59 &0.44 &4.25 & \textbf{\underline{0.26}}\\
			
			Far-off & AFHQ-cat$\to$FFHQ & 4.44 & 0.72 & 4.57 & 0.76 & 4.70 & 0.72 & 4.83 & - & 4.72 & 0.61 & 4.89  & - &6.74 &0.45 &\textbf{\underline{3.69}} & \textbf{\underline{0.23}}\\
			
			Far-off & AFHQ-dog$\to$FFHQ & 4.83 & 0.66 & 4.68 & 0.76 & 4.78 & 0.68 & 6.37 & - & 4.81 & 0.58 & 4.47 & - &5.88 &0.41 &\textbf{\underline{4.12}} & \textbf{\underline{0.22}}\\
			
			Far-off & AFHQ-wild$\to$FFHQ & 4.23 & 0.72 & 4.40 & 0.76 & 4.64 & 0.72 & 5.18 & - & 4.63 & 0.64 & 5.86 & - &5.98 &0.45 &\textbf{\underline{4.17}} & \textbf{\underline{0.23}}\\
			
			Intensely far-off& LSUN-church$\to$FFHQ & 5.11 & 0.80 & 4.43 & 0.81 & 4.47 & 0.81 & 5.69 & - & 4.82 & 0.65 & 5.86 & - &6.09 &0.45 &\textbf{\underline{3.98}} & \textbf{\underline{0.23}}\\
			
			Intensely far-off& AFHQ-cat$\to$LSUN-church & 4.69 & 0.72 & 4.23 & 0.76 & 4.57 & 0.76 & 4.40 & - & 6.01 & 0.67 & 4.33 & - &5.69 &0.44 &\textbf{\underline{4.12}} & \textbf{0.26} \\
			\midrule
			/&Average&5.16&0.73&4.88&0.75&4.95&0.72&5.49&-&5.31&0.62&4.86&-&5.99&0.45 &\textbf{\underline{4.19}}& \textbf{\underline{0.23}}\\			
			\bottomrule
		\end{tabular}}
	\label{tab:comparsion_NIQE_LPIPS}
\end{table*}

	\begin{figure*}[t]
	\centering
	\setlength{\abovecaptionskip}{0cm}
	\centering
	\setlength{\tabcolsep}{0.05em}
	\setlength{\fboxrule}{1pt}
	\setlength{\fboxsep}{0pt}
	\begin{tabular}{cccccccc}			
		\includegraphics[width=.12\linewidth]{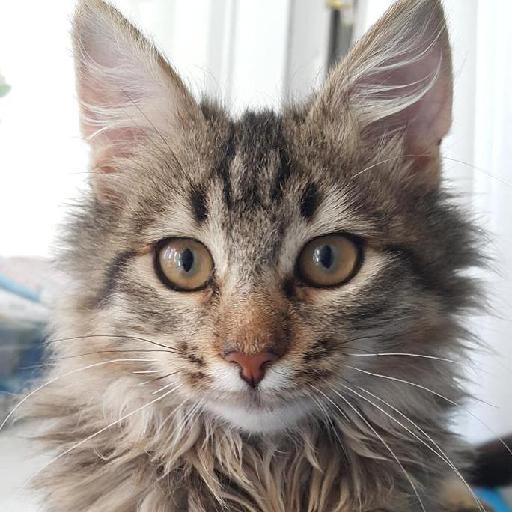}&
		\includegraphics[width=.12\linewidth]{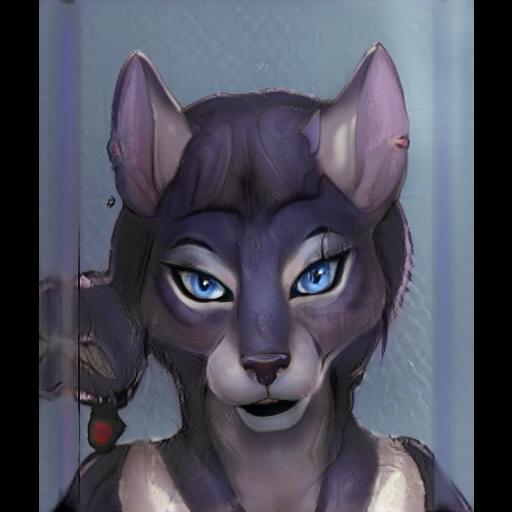}&
		\includegraphics[width=.12\linewidth]{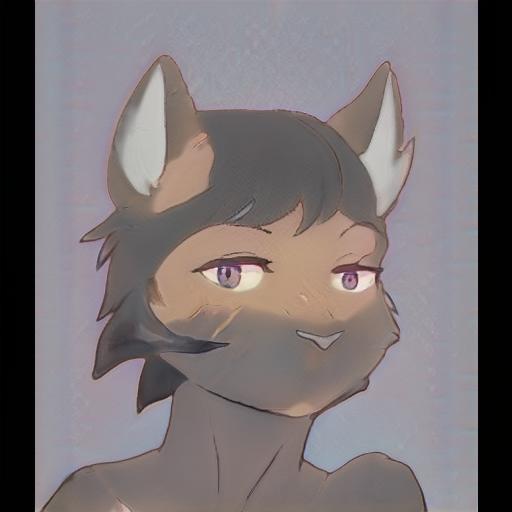}&
		\includegraphics[width=.12\linewidth]{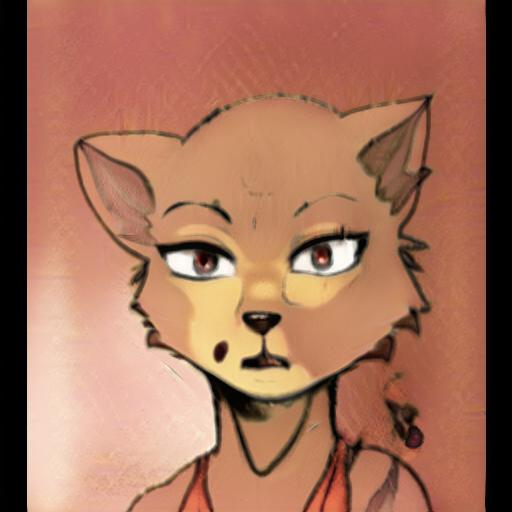}&
		\includegraphics[width=.12\linewidth]{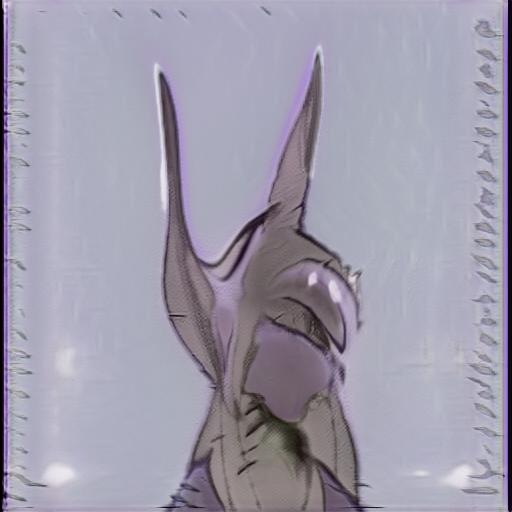}&
		\includegraphics[width=.12\linewidth]{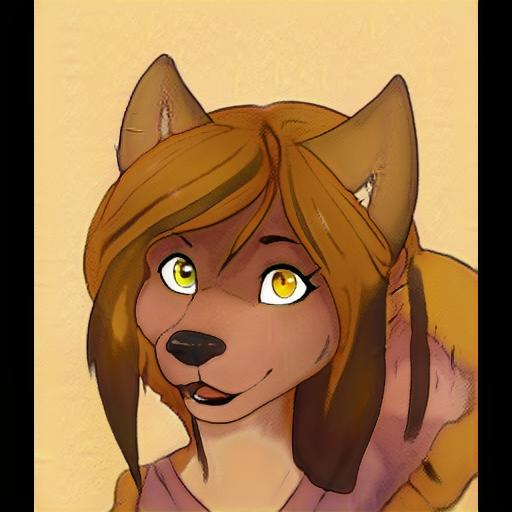}&		
		\includegraphics[width=.12\linewidth]{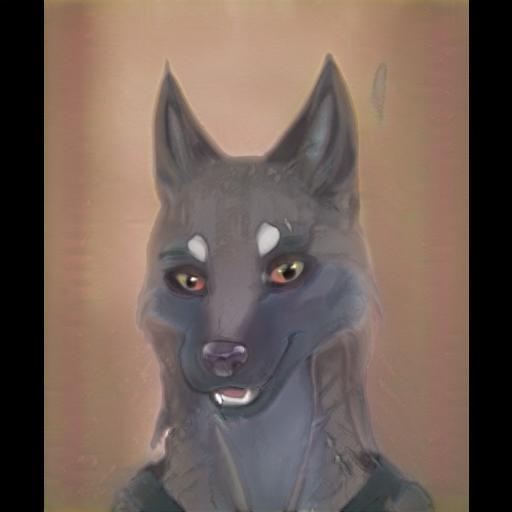}&
		\includegraphics[width=.12\linewidth]{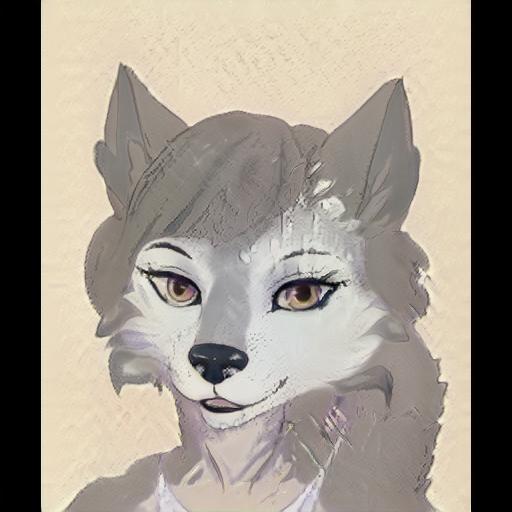}		
		\\		
		\includegraphics[width=.12\linewidth]{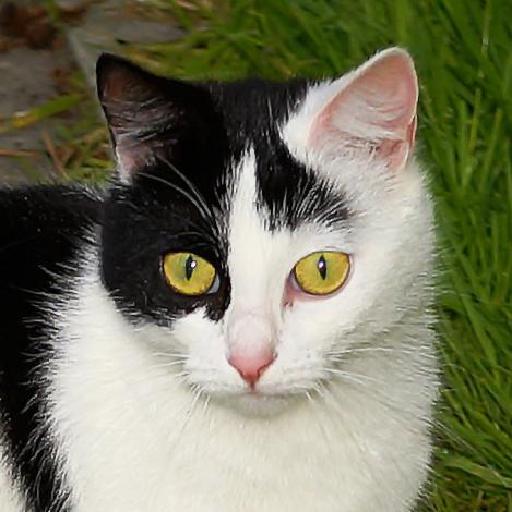}&
		\includegraphics[width=.12\linewidth]{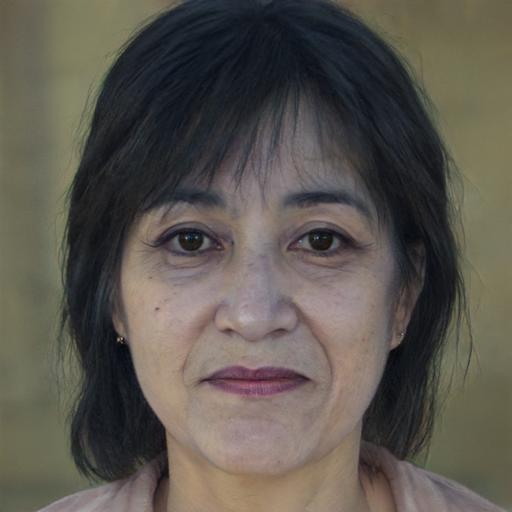}&
		\includegraphics[width=.12\linewidth]{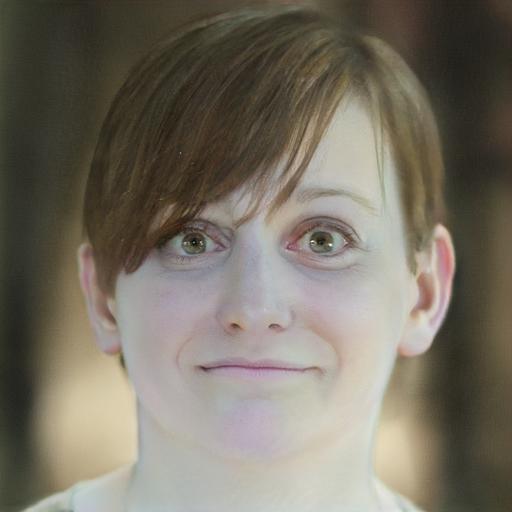}&
		\includegraphics[width=.12\linewidth]{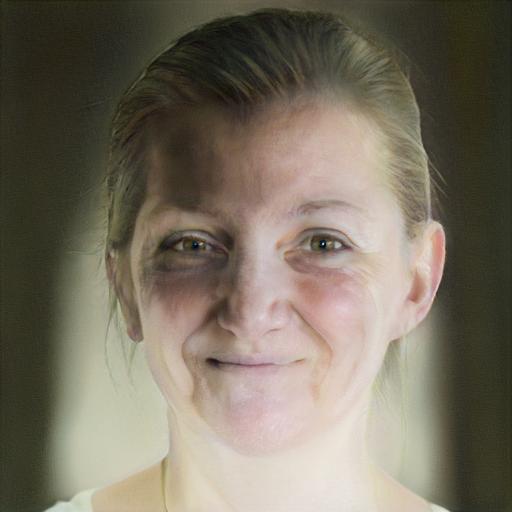}&
		\includegraphics[width=.12\linewidth]{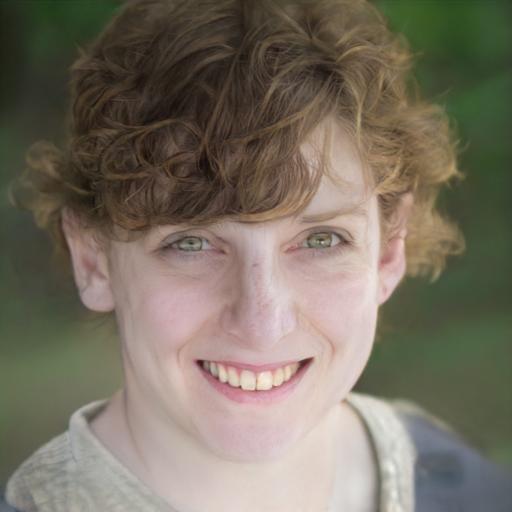}&
		\includegraphics[width=.12\linewidth]{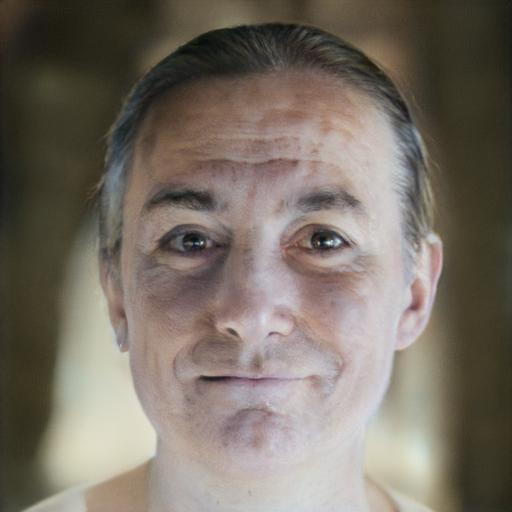}&	
		\includegraphics[width=.12\linewidth]{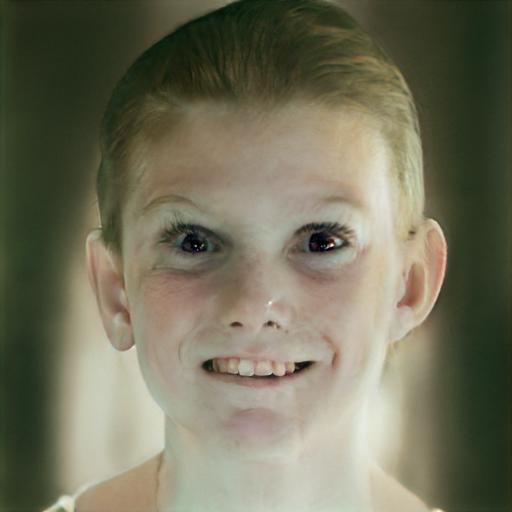}&
		\includegraphics[width=.12\linewidth]{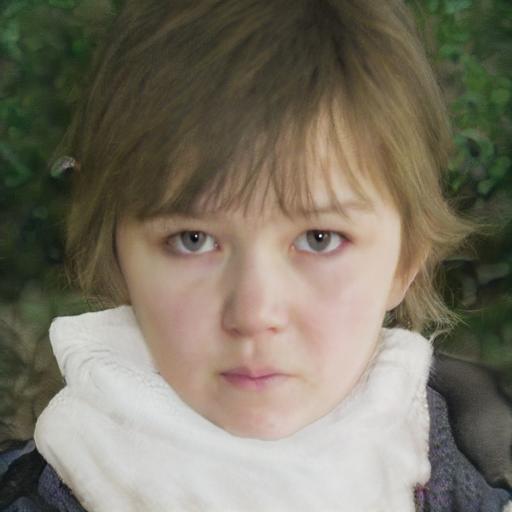}		
		\\		
		\includegraphics[width=.12\linewidth]{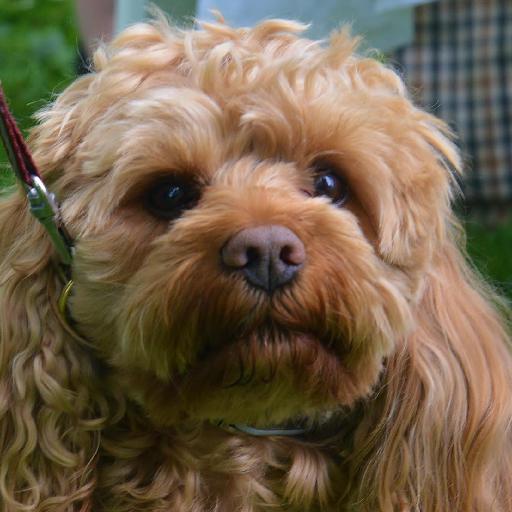}&
		\includegraphics[width=.12\linewidth]{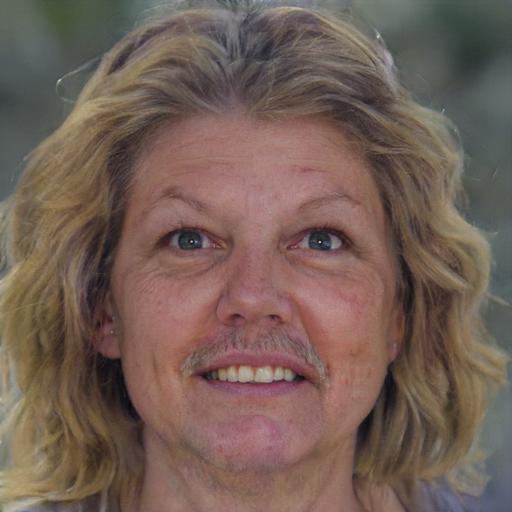}&
		\includegraphics[width=.12\linewidth]{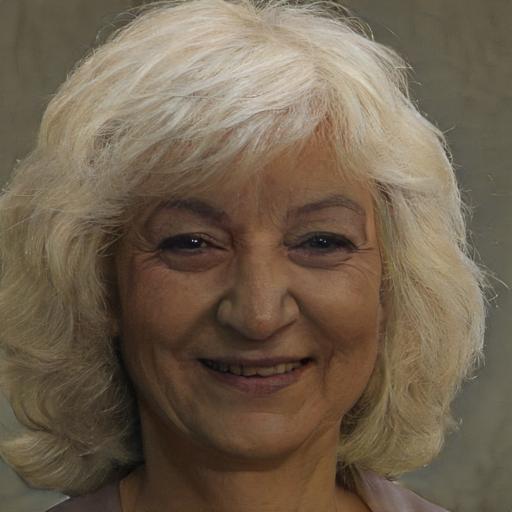}&
		\includegraphics[width=.12\linewidth]{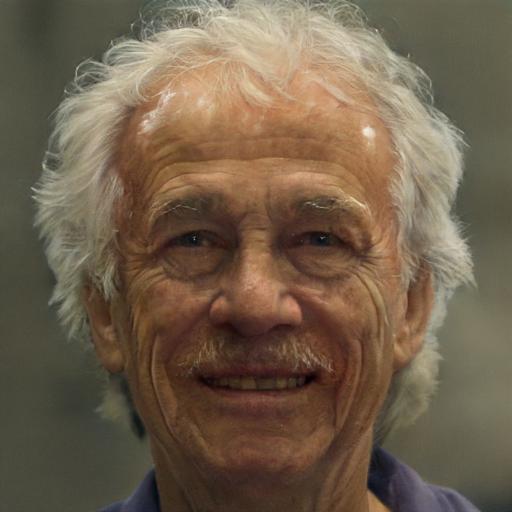}&
		\includegraphics[width=.12\linewidth]{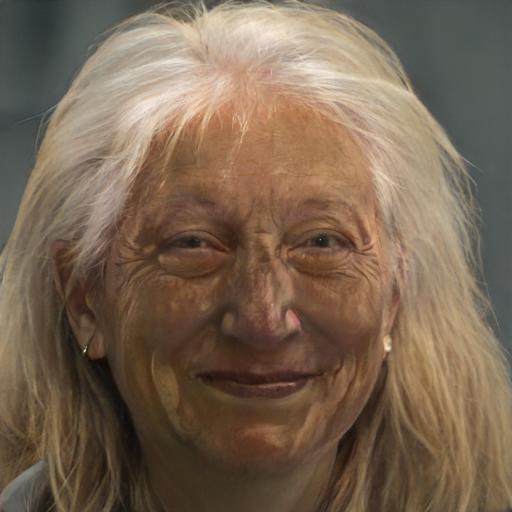}&
		\includegraphics[width=.12\linewidth]{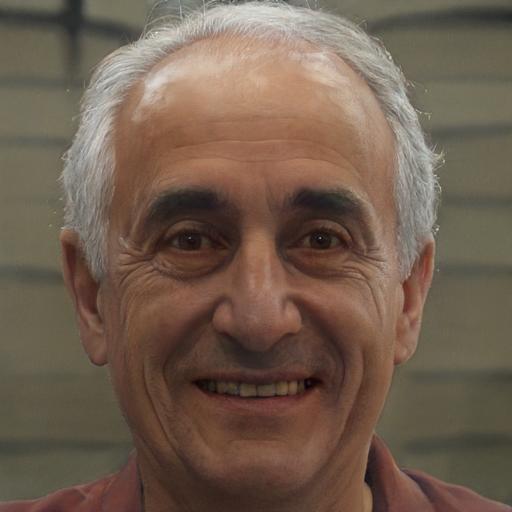}&	
		\includegraphics[width=.12\linewidth]{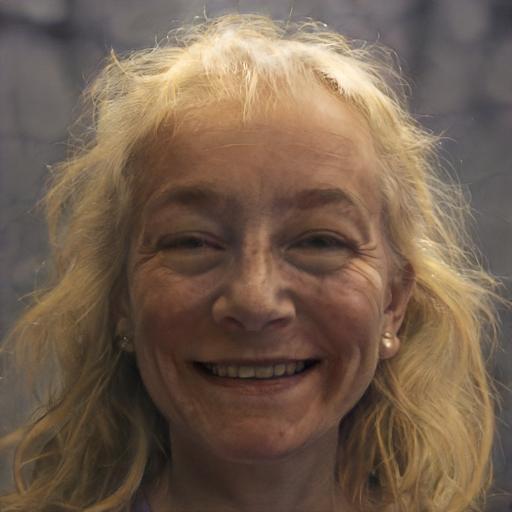}&
		\includegraphics[width=.12\linewidth]{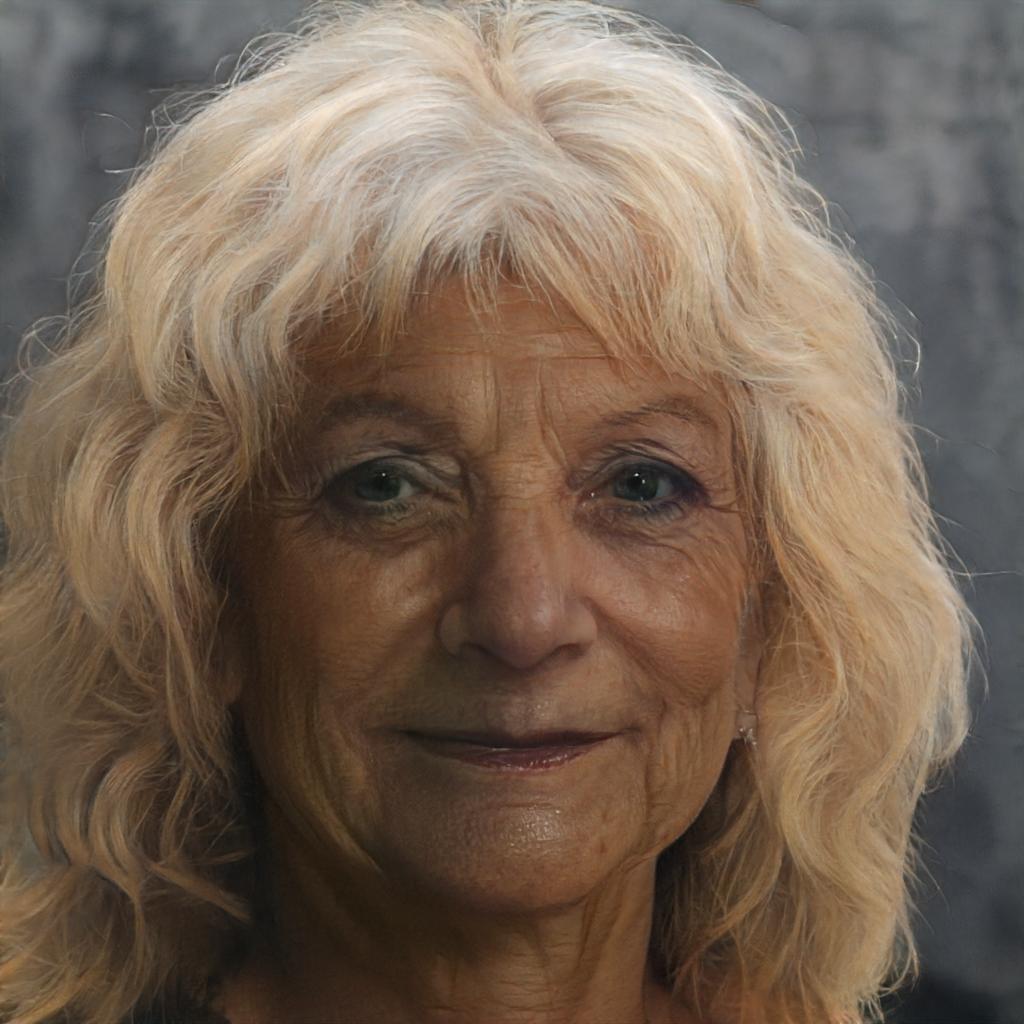}		
		\\		
		\includegraphics[width=.12\linewidth]{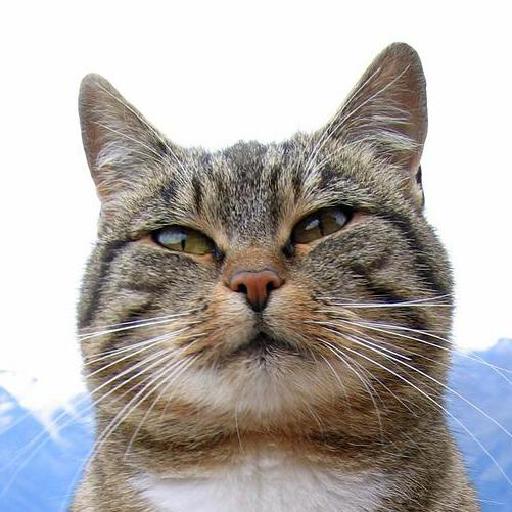}&
		\includegraphics[width=.12\linewidth]{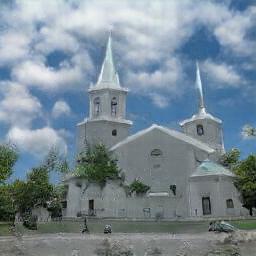}&
		\includegraphics[width=.12\linewidth]{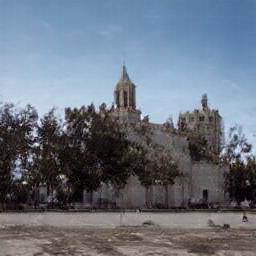}&
		\includegraphics[width=.12\linewidth]{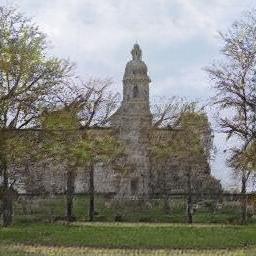}&
		\includegraphics[width=.12\linewidth]{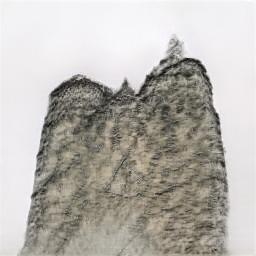}&
		\includegraphics[width=.12\linewidth]{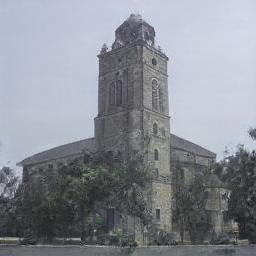}&
		\includegraphics[width=.12\linewidth]{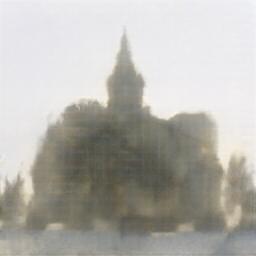}&
		\includegraphics[width=.12\linewidth]{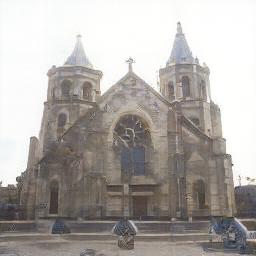}		
		\\		
		Source &
		w/o $\mathcal{L}_{mse}$ &
		w/o $\mathcal{L}_{lpips}$ &
		w/o $\mathcal{L}_{cycle}$ &
		w/o $\mathcal{L}_p$ &
		\footnotesize{w/o Decoupling} &
		\footnotesize{w/o Nonlinear} &
		Full 		
		\\
		Image&&&&&\footnotesize{Module}&\footnotesize{Mapping $\mathcal M(\cdot)$}&			
	\end{tabular}
	\caption{Ablation studies on network modules (\ie, ``w/o Decoupling Module'', ``w/o Nonlinear Mapping $\mathcal M(\cdot)$'') and loss function designs (\ie, ``w/o $\mathcal{L}_{mse}$'', ``w/o $\mathcal{L}_{lpips}$'', ``w/o $\mathcal{L}_{cycle}$'' and ``w/o $\mathcal{L}_p$''). Each of these components contributes to the final quality of the results. (First row: AFHQ-cat$\to$E621Faces; Second row: AFHQ-cat$\to$FFHQ; Third row: AFHQ-dog$\to$FFHQ; Last row: AFHQ-cat$\to$LSUN-church.)}
	\label{fig:ablation}
\end{figure*}

\subsection{Quantitative evaluation}
Table~\ref{tab:comparsion_NIQE_LPIPS} shows the NIQE and LPIPS scores of the evaluation results. Our method consistently performs favorably across various configurations and mapping tasks. These results demonstrate our method's ability to translate high-quality images reliably while preserving strong perceptual correlations between the sources and targets. We do not include the LPIPS score for DiFa and DiffusionCLIP, as their outputs predominantly remain within the source domain. As evident in Fig.~\ref{fig:comparsion1} and Fig.~\ref{fig:comparsion2}, their translated outputs resemble the input images. Thus, assessing the perceptual correspondences of such results within the scope of cross-domain translation tasks becomes inherently inconsequential. Furthermore, we compute the similarity of CLIP embeddings between the generated results and the target dataset for each mapping task, assessing the degree to which the outcomes belong to the target domain. More details are provided in the supplementary material.

\subsection{Ablation Study}
\noindent \textbf{Effectiveness of CLIP2P Mapper.}
To evaluate the effect of our CLIP2P mapper, we conduct a qualitative ablation study by removing the nonlinear mapping $\mathcal M(\cdot)$ and using the output of the linear layer as the $q$ code in Eq. (\ref{equ:Lg}). The resulting images are displayed in the 7th column of Fig. \ref{fig:ablation}. Removing this function noticeably degrades image quality in the E621Faces and LSUN-church domains, while the effect on image quality in the FFHQ domain is relatively minor.

These results can be attributed to the differences between the distributions $P$ (true) and $P$ (pseudo) across various target domains, as observed in Table~\ref{tab:kl_real_p_and_gaussian_p}. Domains like FFHQ, where the corresponding $P$ (pseudo) and $P$ (true) distributions are closer, only require an additional linear layer to enable the $q$ code to approximate the $p$ code, even without the nonlinear mapping. Conversely, when the distribution $P$ (pseudo) significantly deviates from the distribution $P$ (true), learning the $q$ code from the clip embedding $v$ becomes more complex, necessitating our nonlinear mapping. Notably, even in the FFHQ domain, using nonlinear mapping can yield improved visual results, as it can more accurately approximate the distribution $P$ (true).

Furthermore, we remove the linear layer from the CLIP2P mapper while keeping the nonlinear mapping for analysis. However, in practice, we encounter abnormally high loss values during the early iterations, resulting in training instability. This suggests that the small number of additional parameters provided by the linear layer is necessary to fit the required mapping better. We exclude the results of completely removing the entire CLIP2P mapper due to apparent compatibility issues between the CLIP and $P$ spaces.

\begin{figure*}[t]
	\centering
	\includegraphics[width=\linewidth]{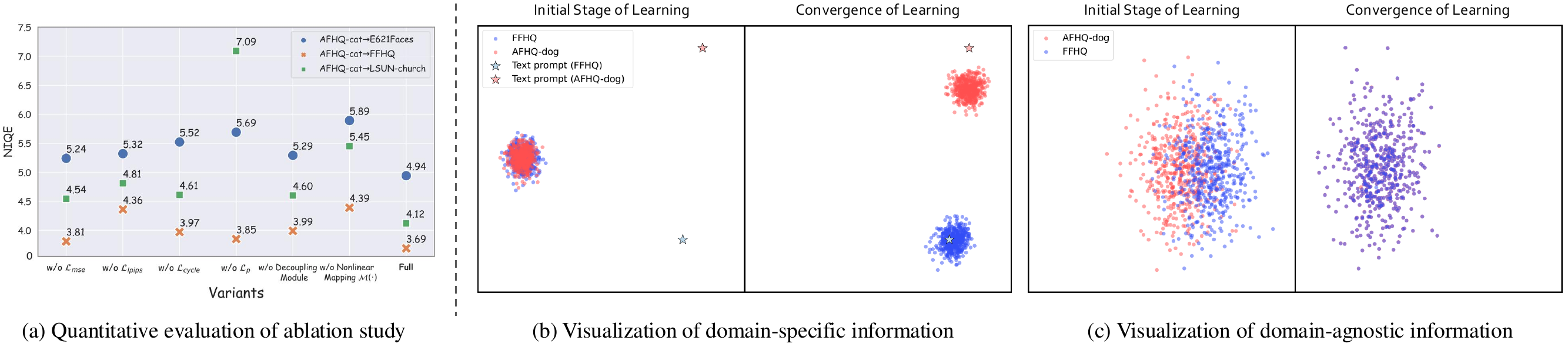}
	\caption{Quantitative evaluation of ablation study and visualized features generated by the decoupling module.}	
	\label{fig:NIQE_ab}
\end{figure*}

In addition to the qualitative ablation study, we conduct a quantitative analysis, as shown in Fig.~\ref{fig:NIQE_ab} (a), comparing the proposed method with its variations across different types of mappings. The results demonstrate that the most significant improvement in translation quality comes from the integration of our novel mapping function, highlighting its pivotal role in enhancing the overall system performance. An exception is the extreme case without the $\mathcal{L}_p$ loss in the LSUN-church target domain, which will be discussed in the context of effectiveness of loss functions. All observations indicate that our CLIP2P mapper significantly expands the boundaries of cross-domain translation universality.

\noindent \textbf{Effectiveness of Decoupling Module.}
To demonstrate the decoupling module's importance, we analyze this component removed, and the qualitative results are presented in the 6th column of Fig.~\ref{fig:ablation}. A comparison with our full-featured model reveals that the decoupling module effectively leverages semantic information to construct correlation mappings related to pose, coarse outlines, and specific elements of fine attributes. 

For example, in the 2nd row of Fig.~\ref{fig:ablation}, our method generates a white scarf on the girl's neck, corresponding to the white patch on the cat's chest in the source image. In contrast, the images generated without the decoupling module lack such a vivid correspondence. In the 3rd row, while other configurations tend to produce long curls to match the dog's curly fur, the absence of the decoupling module results in short hair. Even in scenarios characterized by a significant gap between source and target domains, such as the translation from a cat to a church, the decoupling module plays a crucial role by successfully matching the cat's pointed ears with the spires of the church. 

Beyond its significant advantage in maintaining cross-domain correspondences, the decoupling module also enhances translation quality across source and target domains with varying levels of heterogeneity, as shown in Fig.~\ref{fig:NIQE_ab} (a). This improvement is likely attributed to the effective utilization of the natural statistical properties inherent in the source image.

We also visualize domain-specific and domain-agnostic information to understand the decoupling module's effect. Using the translation of AFHQ-dog to FFHQ as an example, we sample 500 images from the AFHQ-dog domain and translate them to the FFHQ domain. We apply PCA to reduce the dimensions of domain-specific and domain-agnostic features learned by the decoupling module during the initial and convergence stages of learning, visualizing the results in Fig.~\ref{fig:NIQE_ab} (b) and (c). Here, red dots represent source domain features, while blue dots indicate target domain features. Additionally, in Fig.~\ref{fig:NIQE_ab} (b), we mark the average embeddings (also PCA-reduced) of the source and target prompt templates as red and blue stars. It is demonstrated that as training progresses, domain-specific information progressively shifts towards its respective text prompts, eventually clustering around them. Meanwhile, domain-agnostic information converges from a scattered state. All of these outcomes provide compelling evidence for the effectiveness of the decoupling module.

\noindent \textbf{Effectiveness of Loss Functions.}
We analyze the roles of each loss term in the proposed method. First,  we remove $\mathcal{L}_{p}$ from our total objective. As depicted in Fig.~\ref{fig:ablation}, when the target domain is E621Faces or LSUN-church, removing $\mathcal{L}_{p}$ causes the generated results to deviate almost entirely from the target domain. When the target domain is FFHQ, the absence of $\mathcal{L}_{p}$ results in a degradation of image quality, although the generated results remain near the face domain.

Furthermore, we analyze the effect of excluding $\mathcal{L}_{cycle}$. As illustrated in Fig.~\ref{fig:ablation}, the absence of $\mathcal{L}_{cycle}$ affects the semantic correspondences between the source and target domains. In this scenario, step 2 may alter the semantics carried by the $v$ code, not just its distribution. The conflict with the function of step 1 results in the ineffectiveness of the hybrid learning strategy, leading to suboptimal results.

Finally, we examine the effects of $\mathcal{L}_{mse}$ and $\mathcal{L}_{lpips}$. As shown in Fig.~\ref{fig:ablation}, the omission of $\mathcal{L}_{mse}$ influences the color space consistency with regard to the input image (\eg, skin tones), while the exclusion of $\mathcal{L}_{lpips}$ leads to a failure in establishing perceptual relationships (\eg, determining the scale of the church).

Referring to Fig.~\ref{fig:NIQE_ab} (a), the quantitative impact of each loss term is evident, aligning with our qualitative ablation analysis. While all loss terms contribute to performance improvement, it is particularly noteworthy that in the LSUN-church target domain, the $\mathcal{L}_{p}$ loss assumes an even more critical role. As shown in Table~\ref{tab:kl_real_p_and_real_clip}, the gap between the distributions in $P$ and CLIP spaces is more pronounced in the LSUN-church domain than in other target domains. Therefore, utilizing $\mathcal{L}_{p}$ is vital in such domains, guiding the CLIP embedding towards the $p$ latent code. All the evidence underscores the importance of these components in our approach to enhancing the quality of cross-domain image translations.

\begin{figure}[t]
	\centering
	\setlength{\abovecaptionskip}{0cm}
	\centering
	\setlength{\tabcolsep}{0.05em}
	\setlength{\fboxrule}{1pt}
	\setlength{\fboxsep}{0pt}
	\begin{tabular}{cccccccc}
		
		&\includegraphics[width=.15\linewidth]{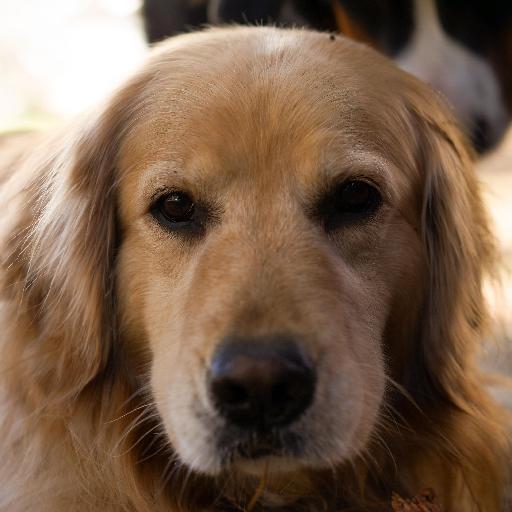}
		& & & & &
		\includegraphics[width=.15\linewidth]{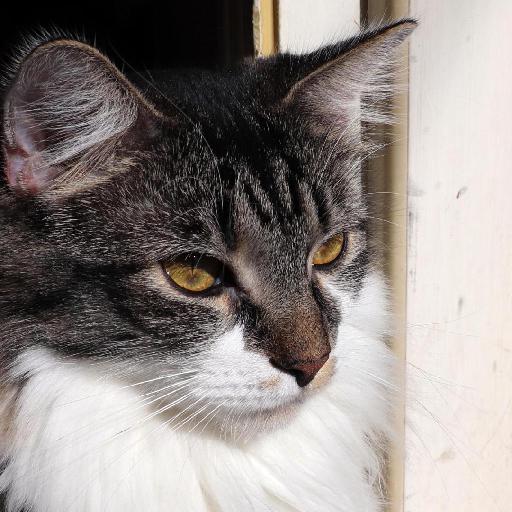}&
		\\	
		&Source A&&&&&Source B&\\
		\toprule
		&Result A&\multicolumn{4}{c}{$\longleftarrow$\quad\quad\quad Interpolation\quad\quad\quad$\longrightarrow$}&Result B&\\
		&\fcolorbox{red}{red}{\includegraphics[width=.15\linewidth]{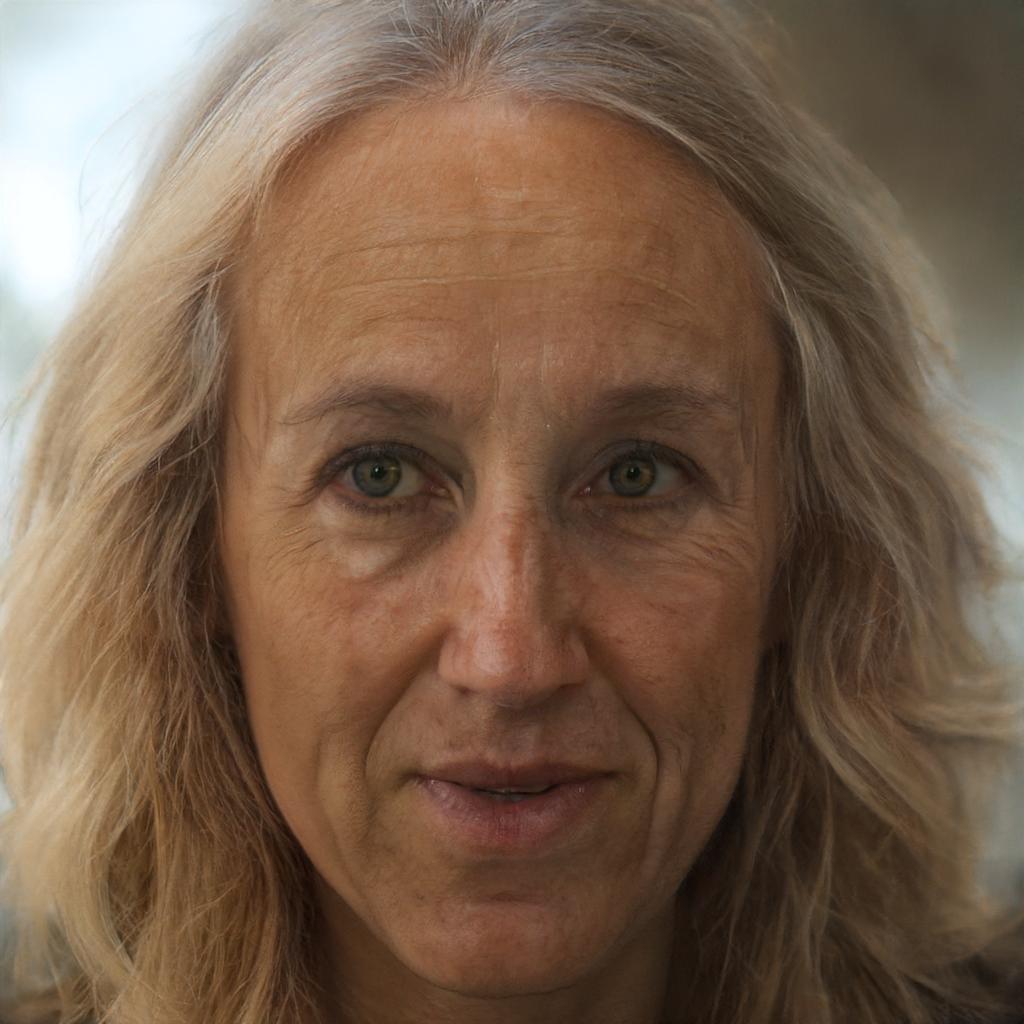}} &
		\includegraphics[width=.15\linewidth]{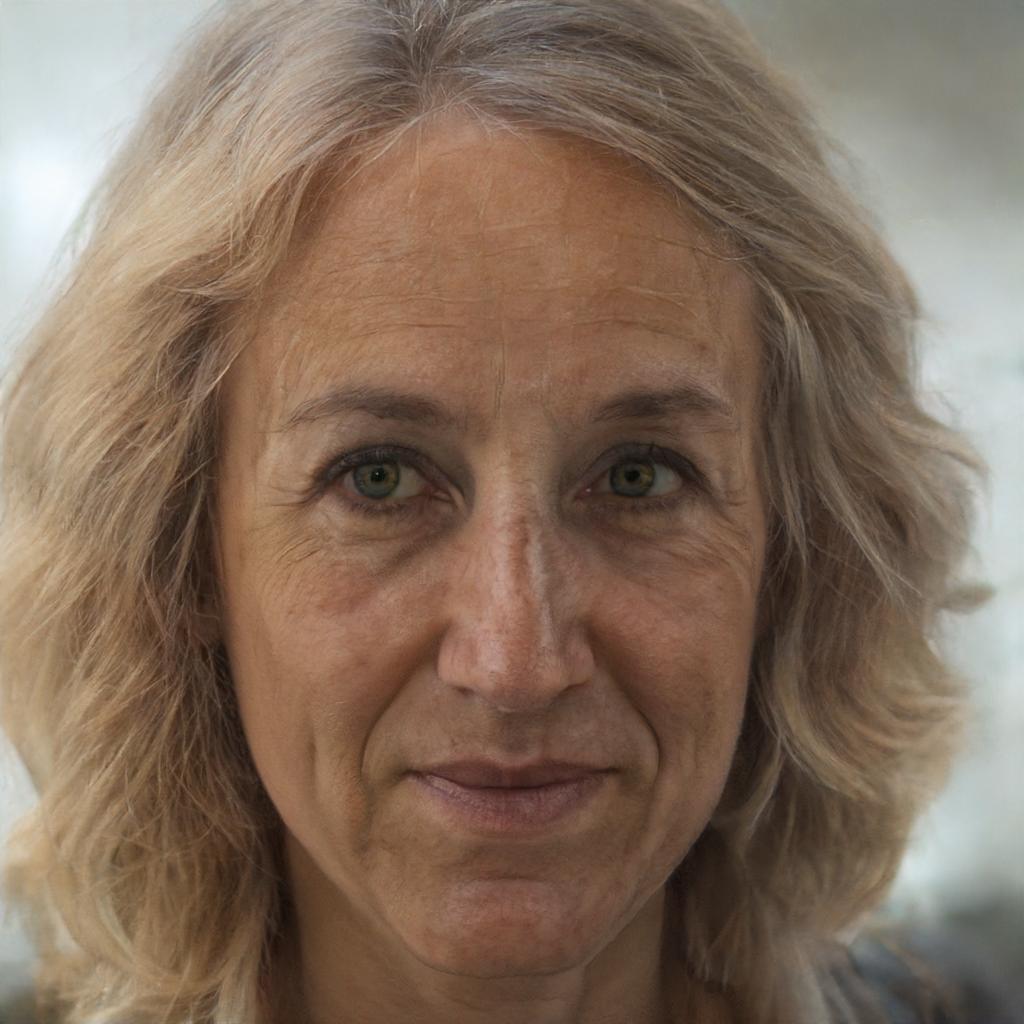} &
		\includegraphics[width=.15\linewidth]{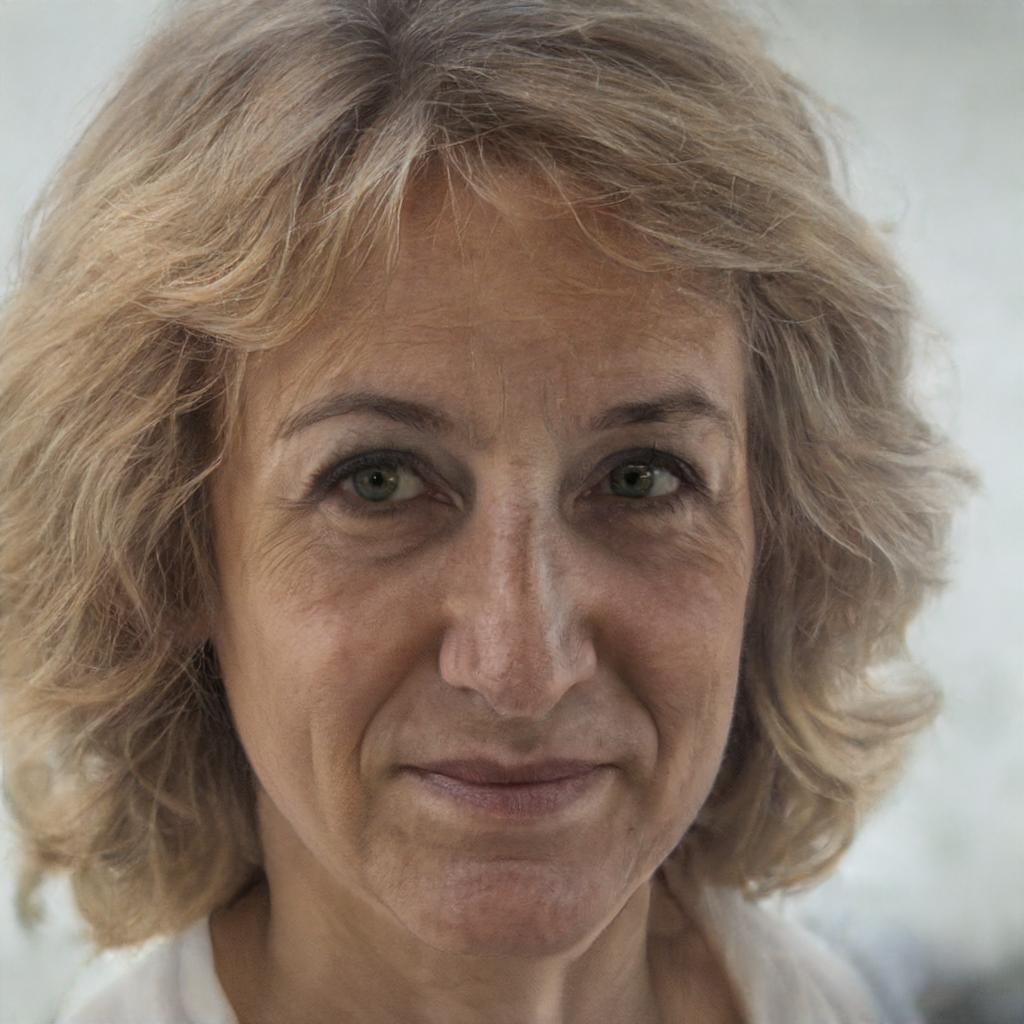} &
		\includegraphics[width=.15\linewidth]{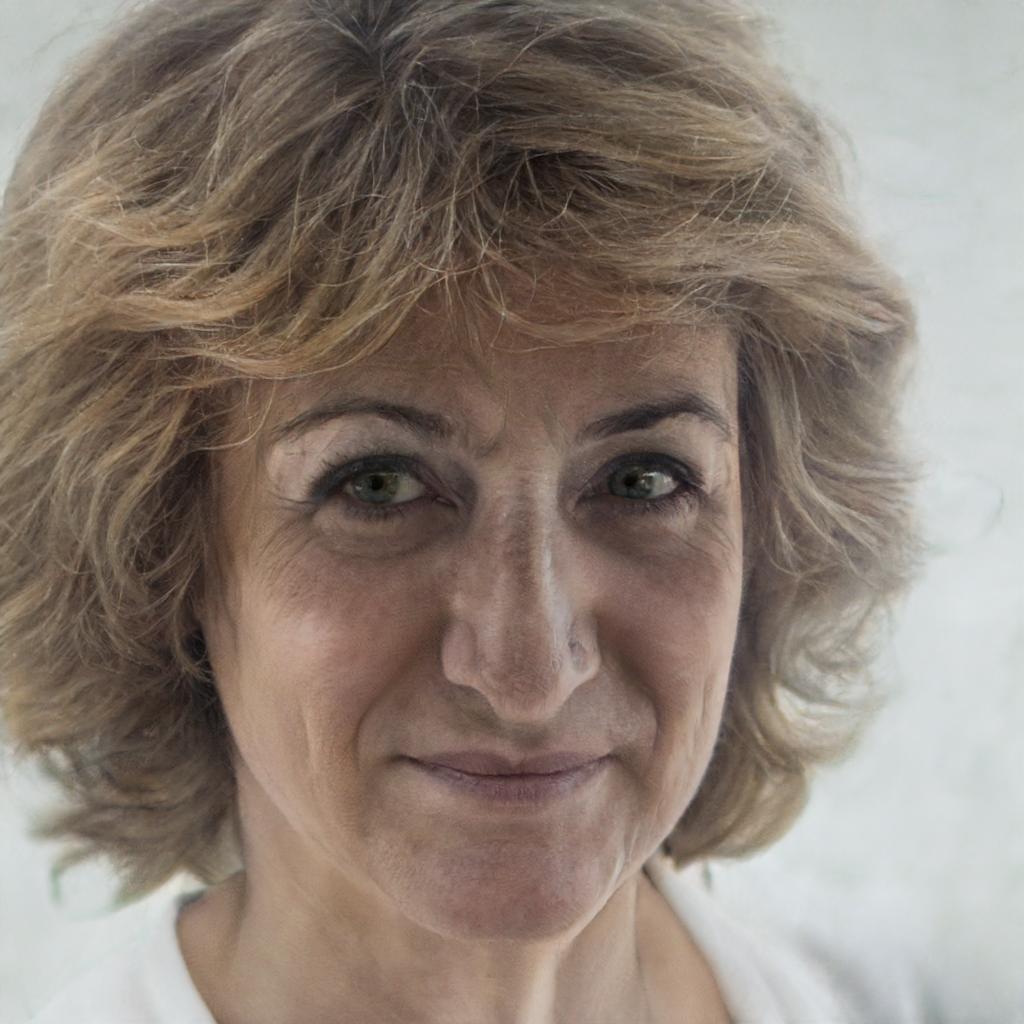} &
		\includegraphics[width=.15\linewidth]{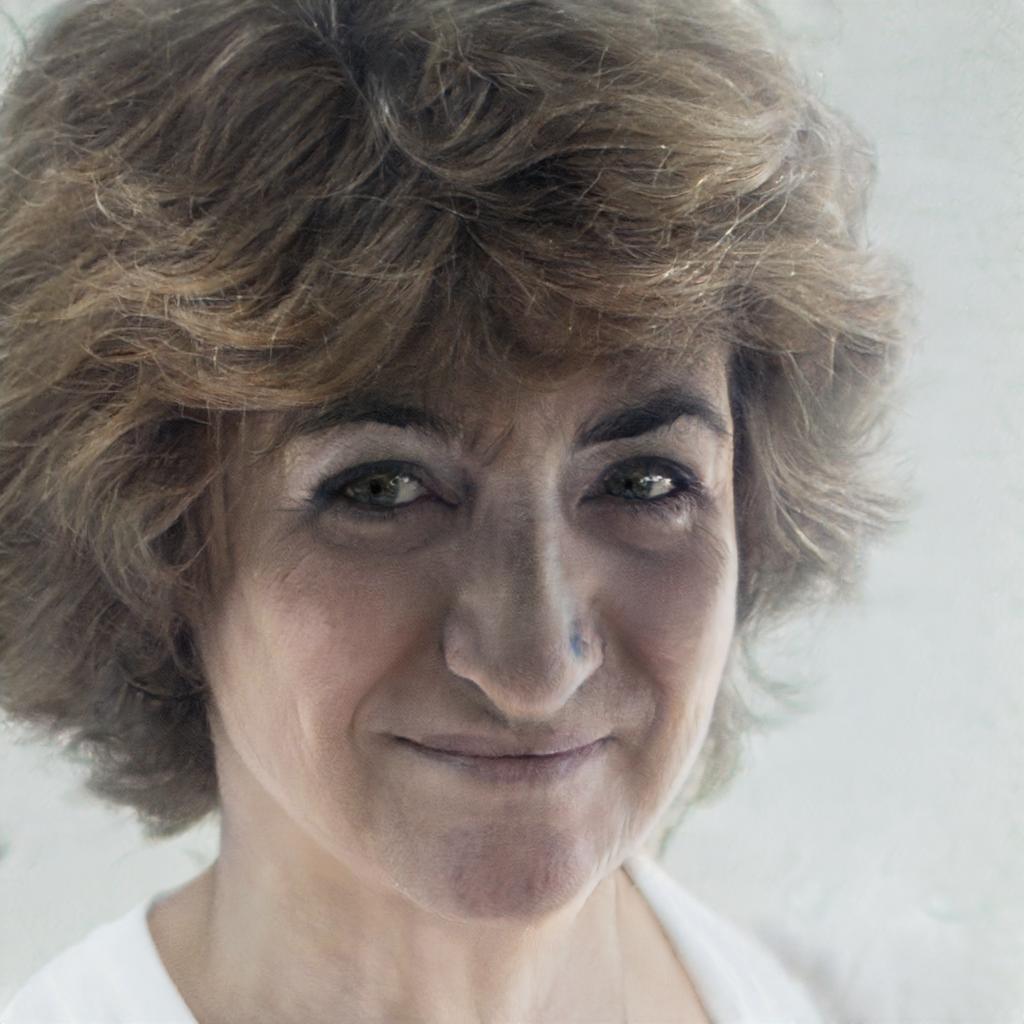} &
		\fcolorbox{red}{red}{\includegraphics[width=.15\linewidth]{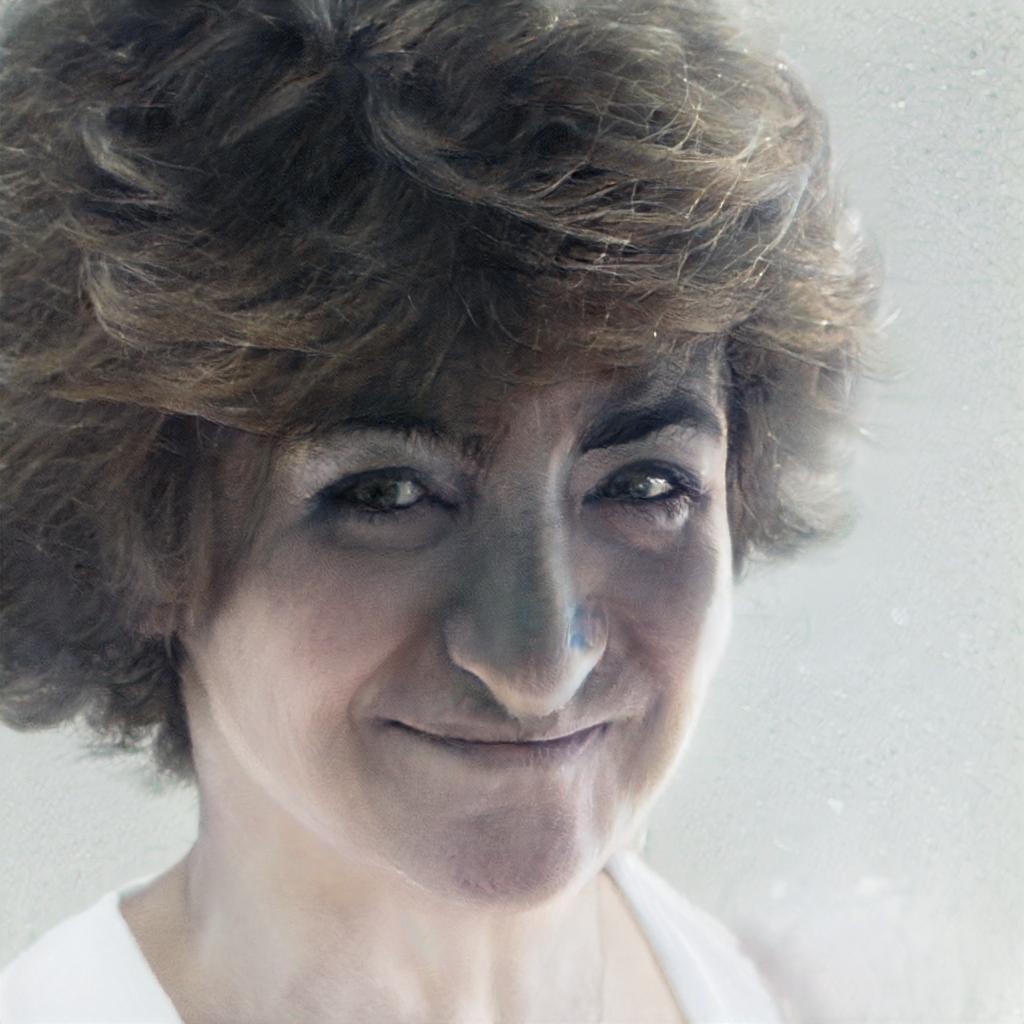}}
		&
		\\	
		\rotatebox[origin=l]{90}{\hspace{4mm}$\longrightarrow$}
		&
		\includegraphics[width=.15\linewidth]{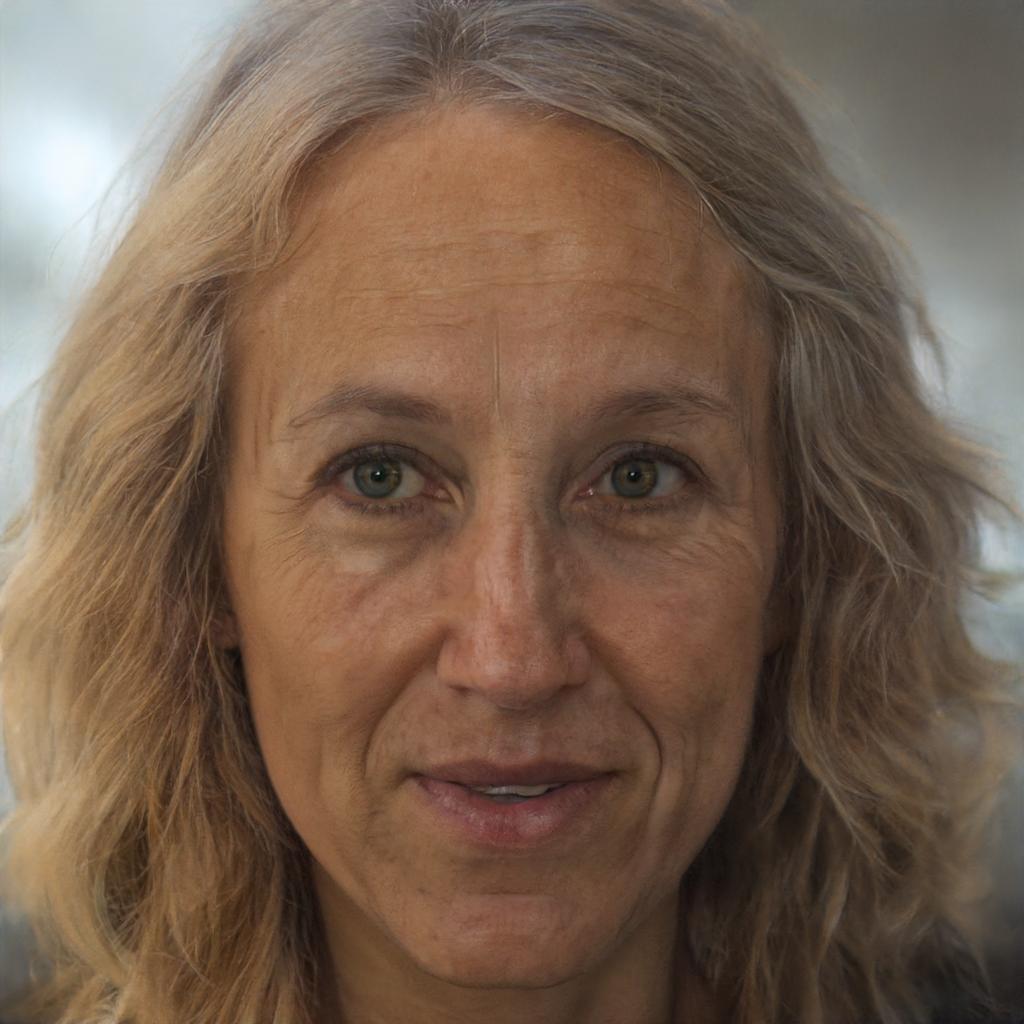} &
		\includegraphics[width=.15\linewidth]{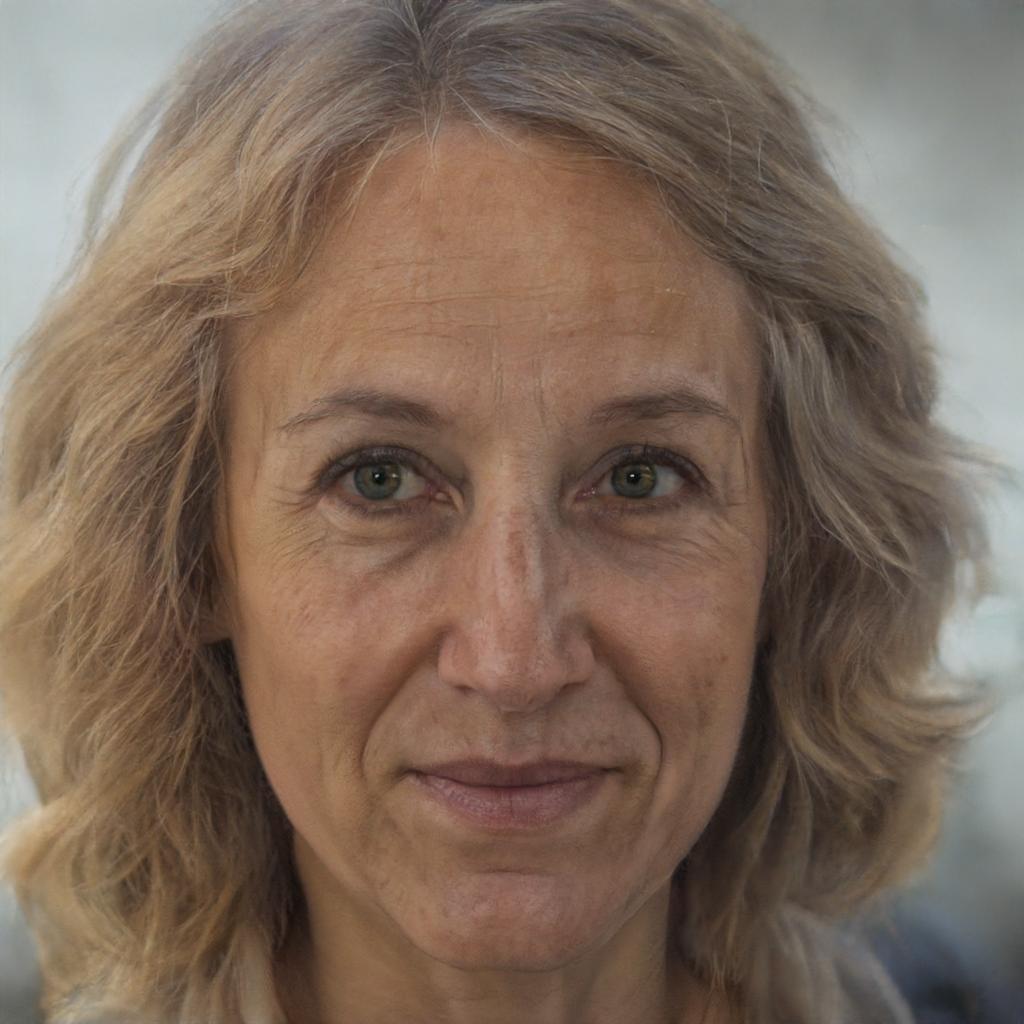} &
		\includegraphics[width=.15\linewidth]{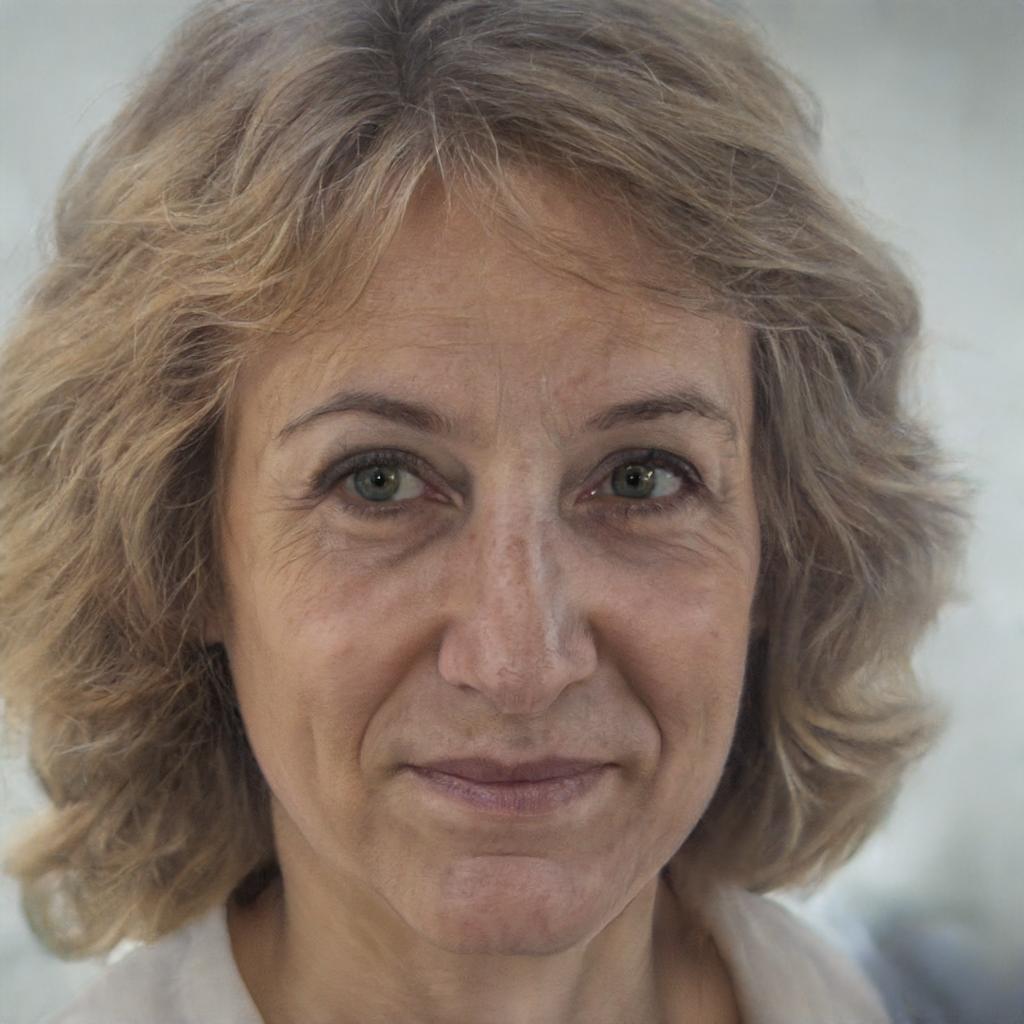} &
		\includegraphics[width=.15\linewidth]{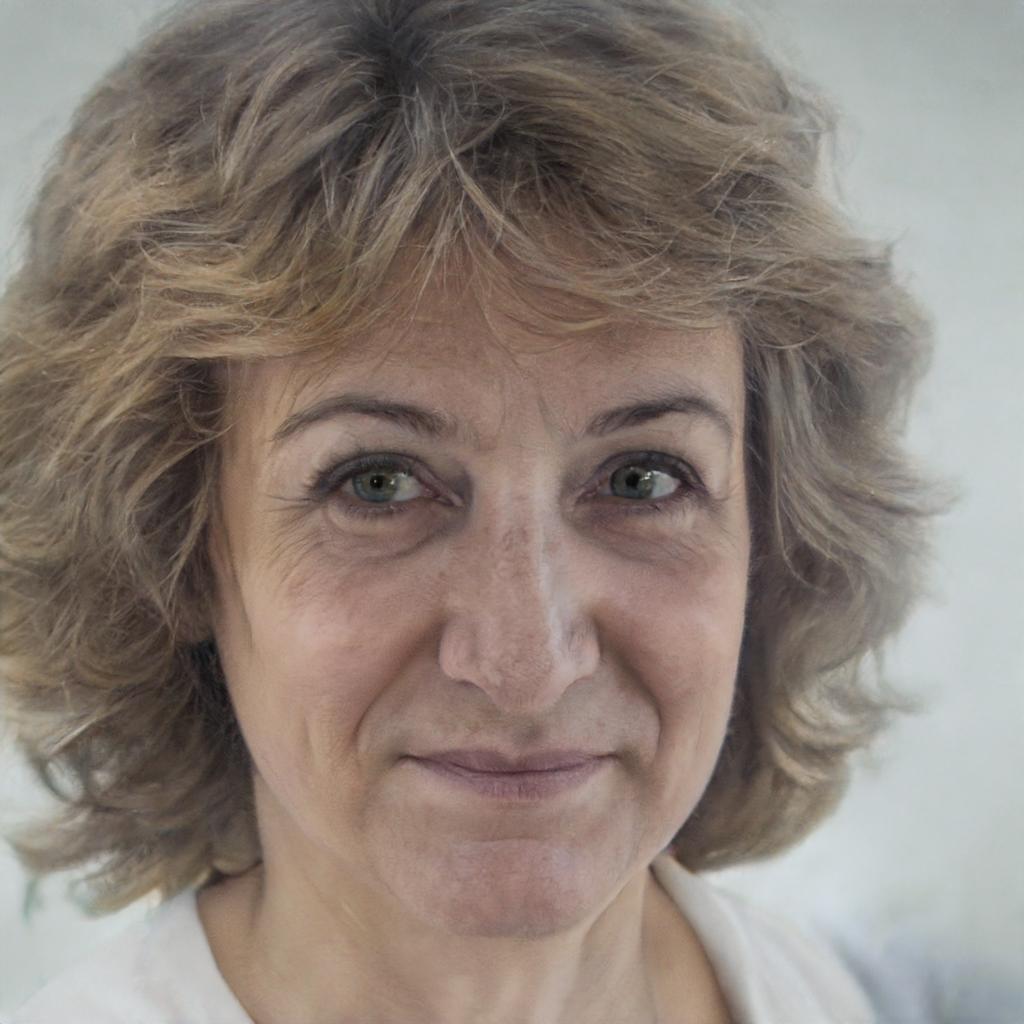} &
		\includegraphics[width=.15\linewidth]{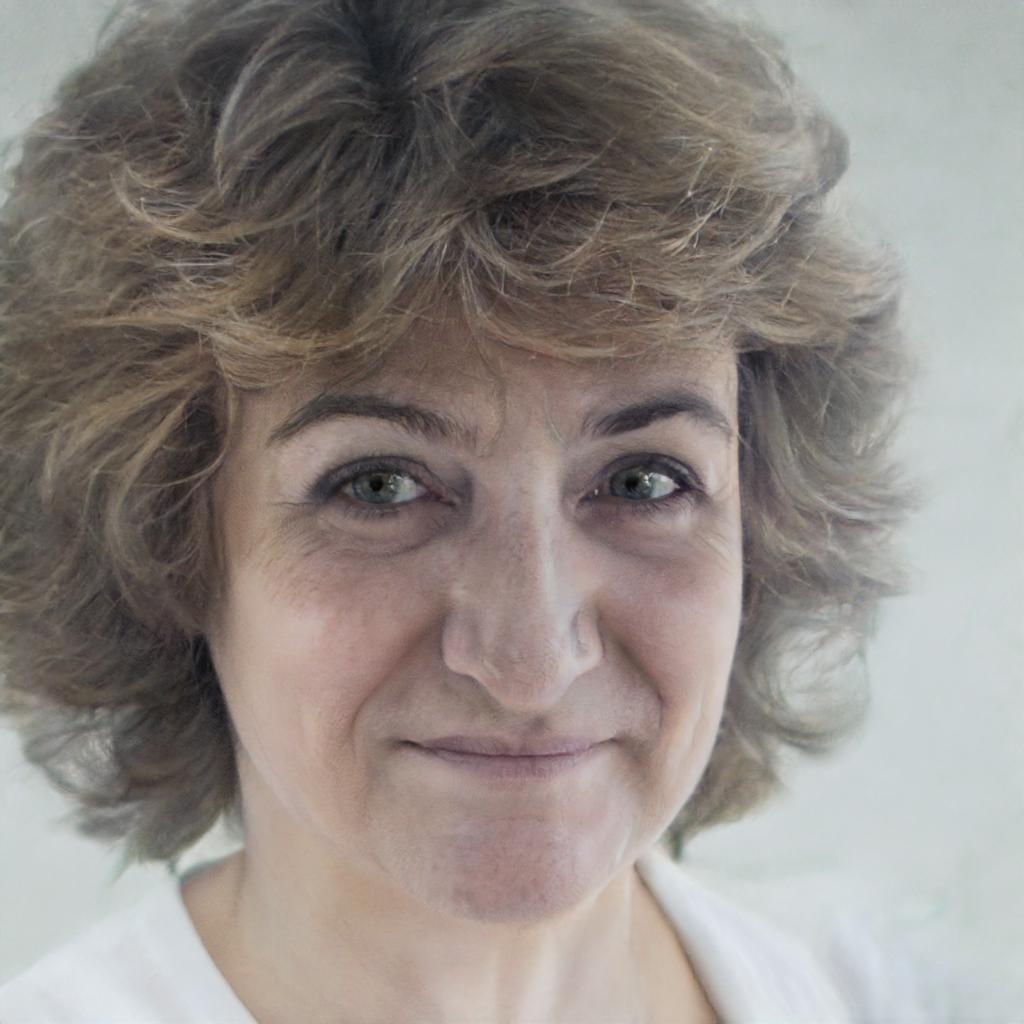} &
		\includegraphics[width=.15\linewidth]{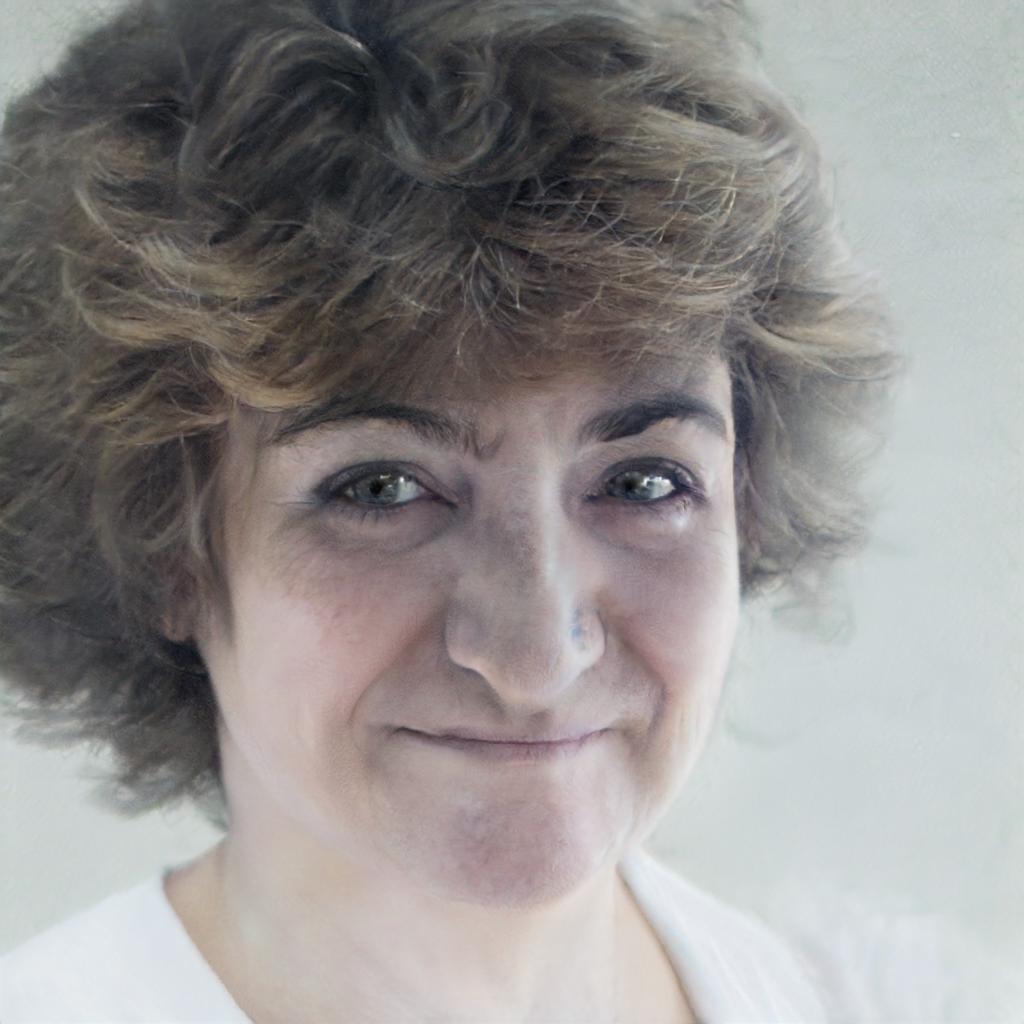}
		&\rotatebox[origin=l]{90}{\hspace{4mm}$\longrightarrow$}
		\\	
		\rotatebox[origin=l]{90}{\hspace{-1mm}polation}&
		\includegraphics[width=.15\linewidth]{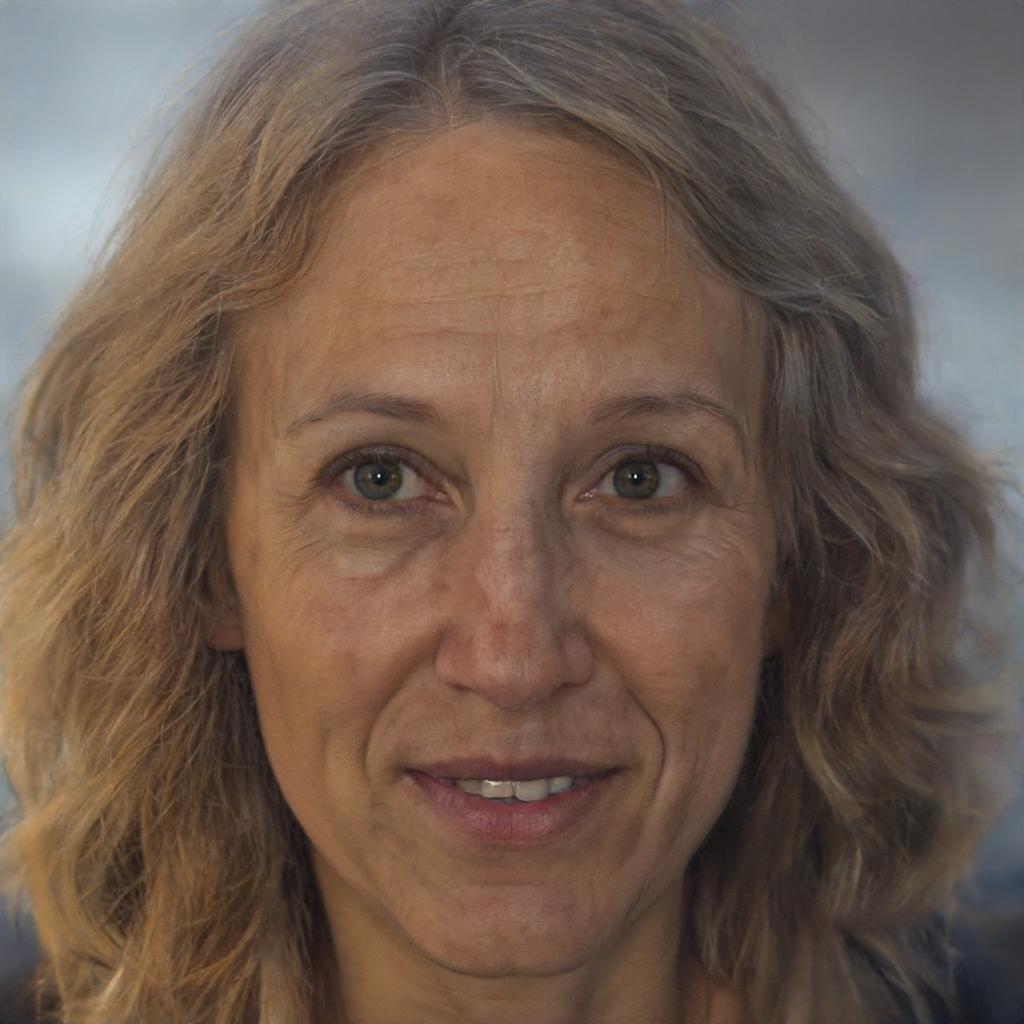} &
		\includegraphics[width=.15\linewidth]{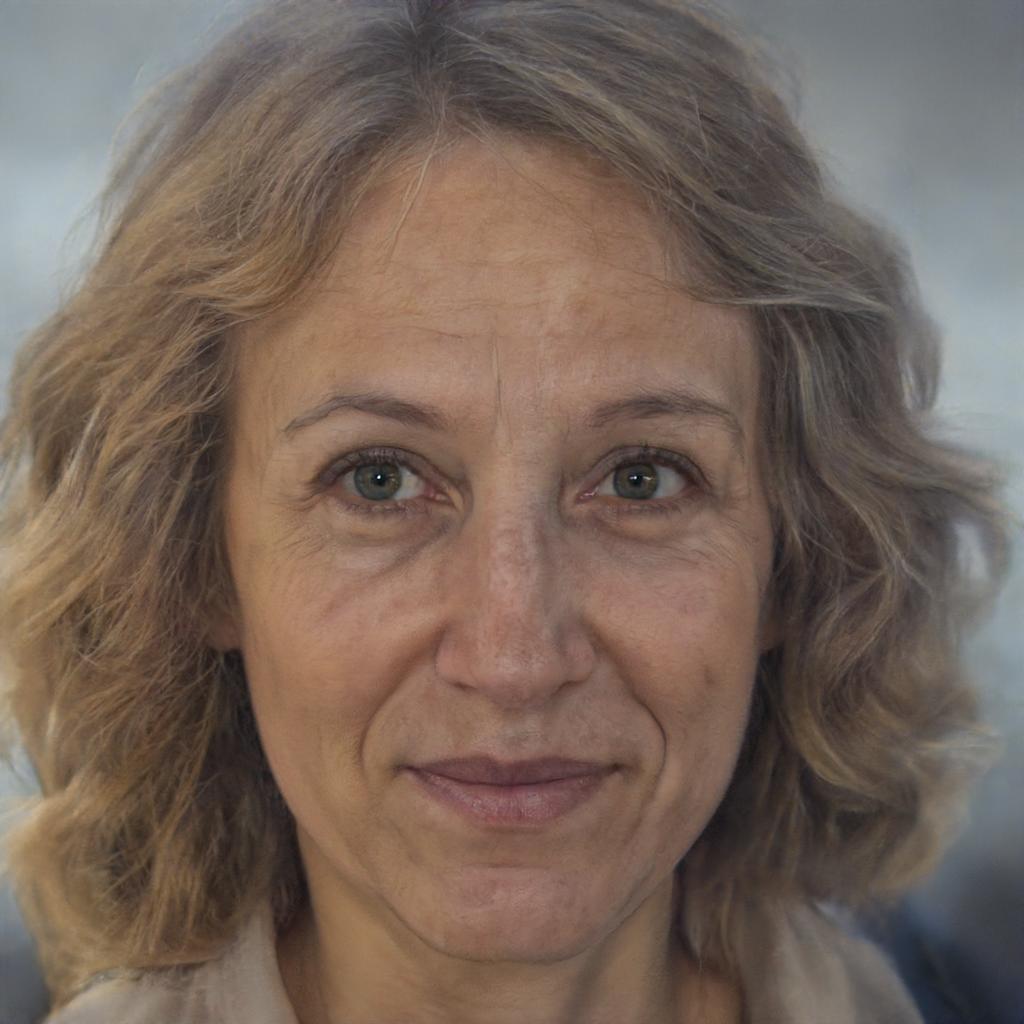} &
		\includegraphics[width=.15\linewidth]{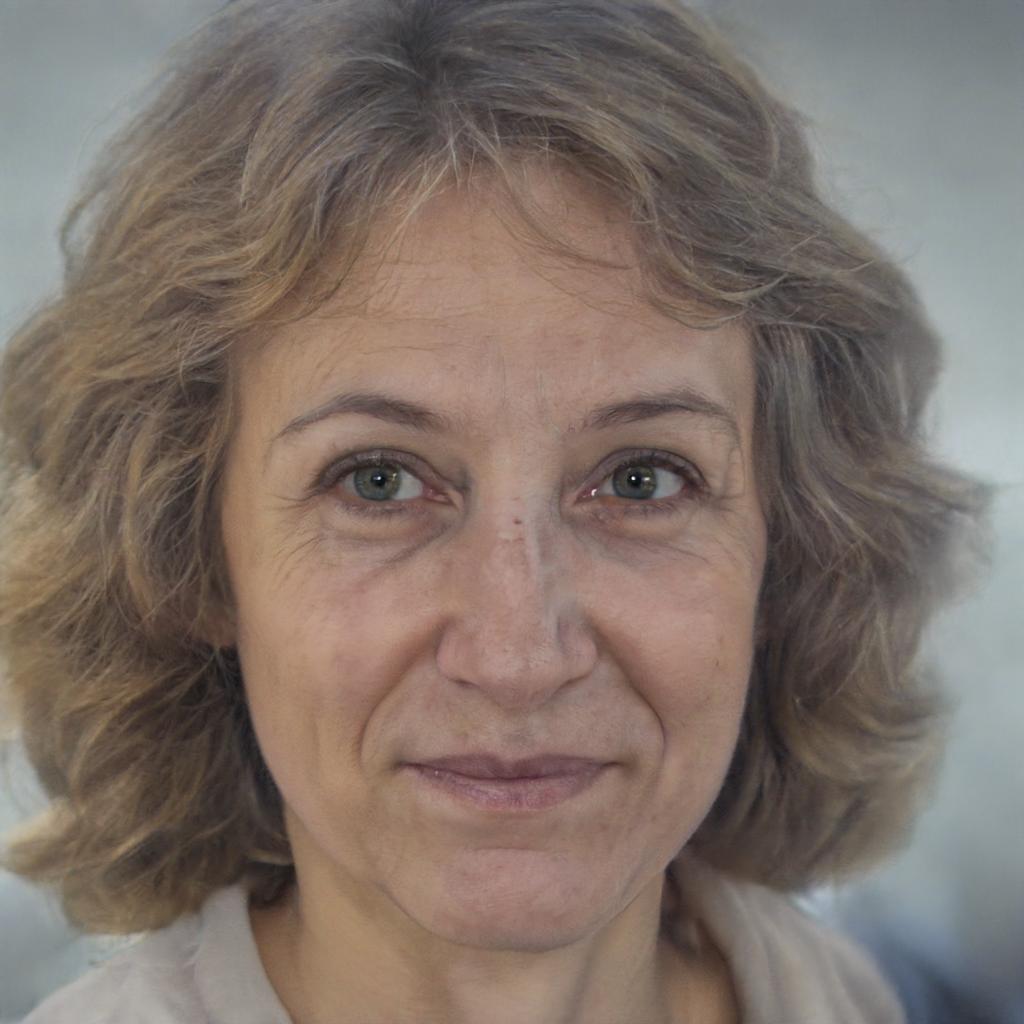} &
		\includegraphics[width=.15\linewidth]{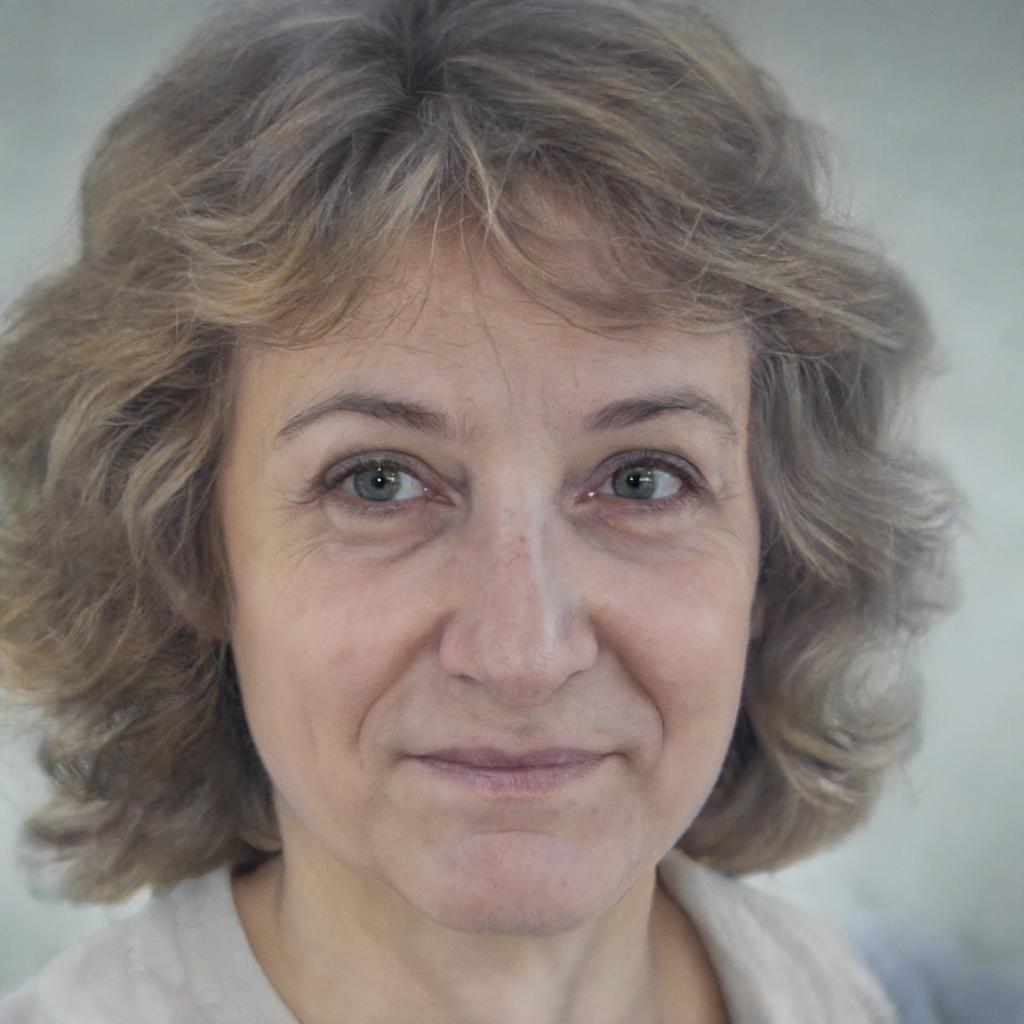} &
		\includegraphics[width=.15\linewidth]{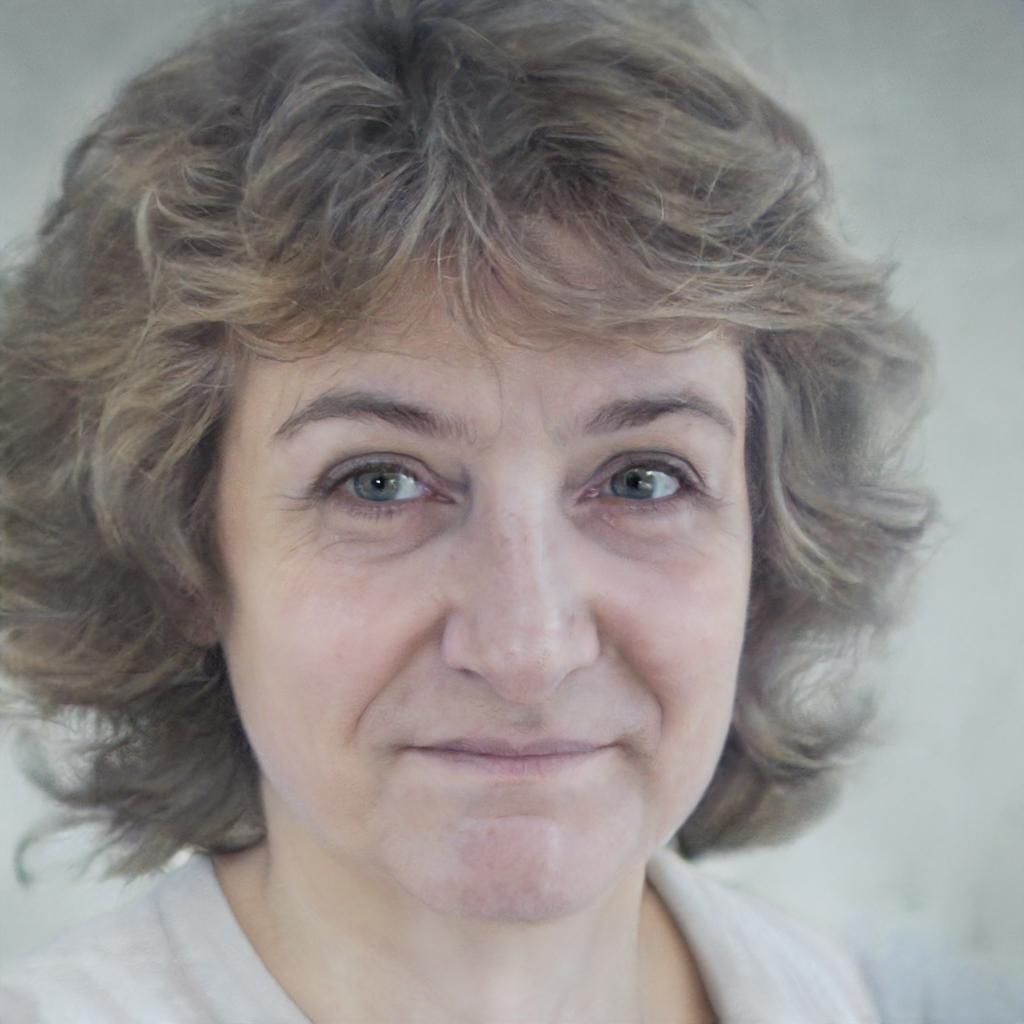} &
		\includegraphics[width=.15\linewidth]{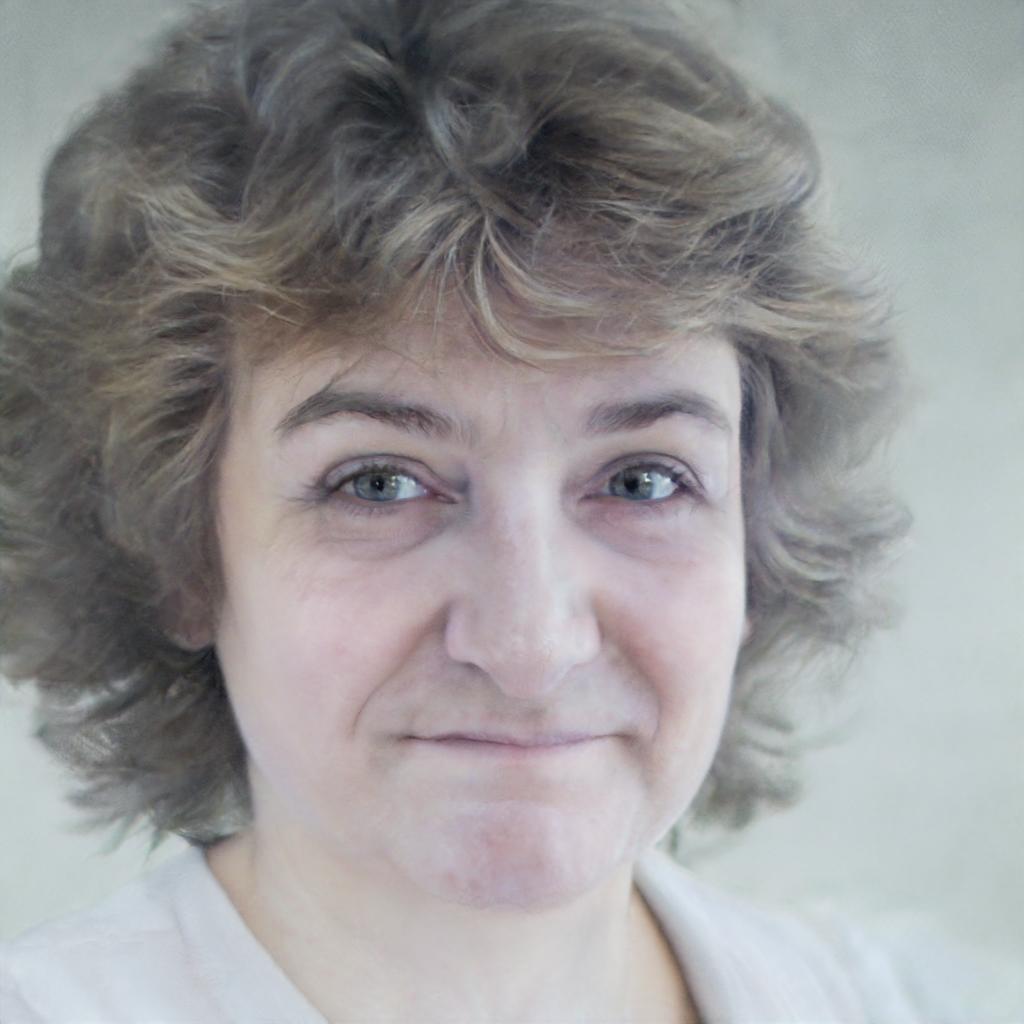}
		&\rotatebox[origin=l]{90}{\hspace{-1mm}polation}
		\\	
		\rotatebox[origin=l]{90}{\hspace{4.5mm}Inter-}&
		\includegraphics[width=.15\linewidth]{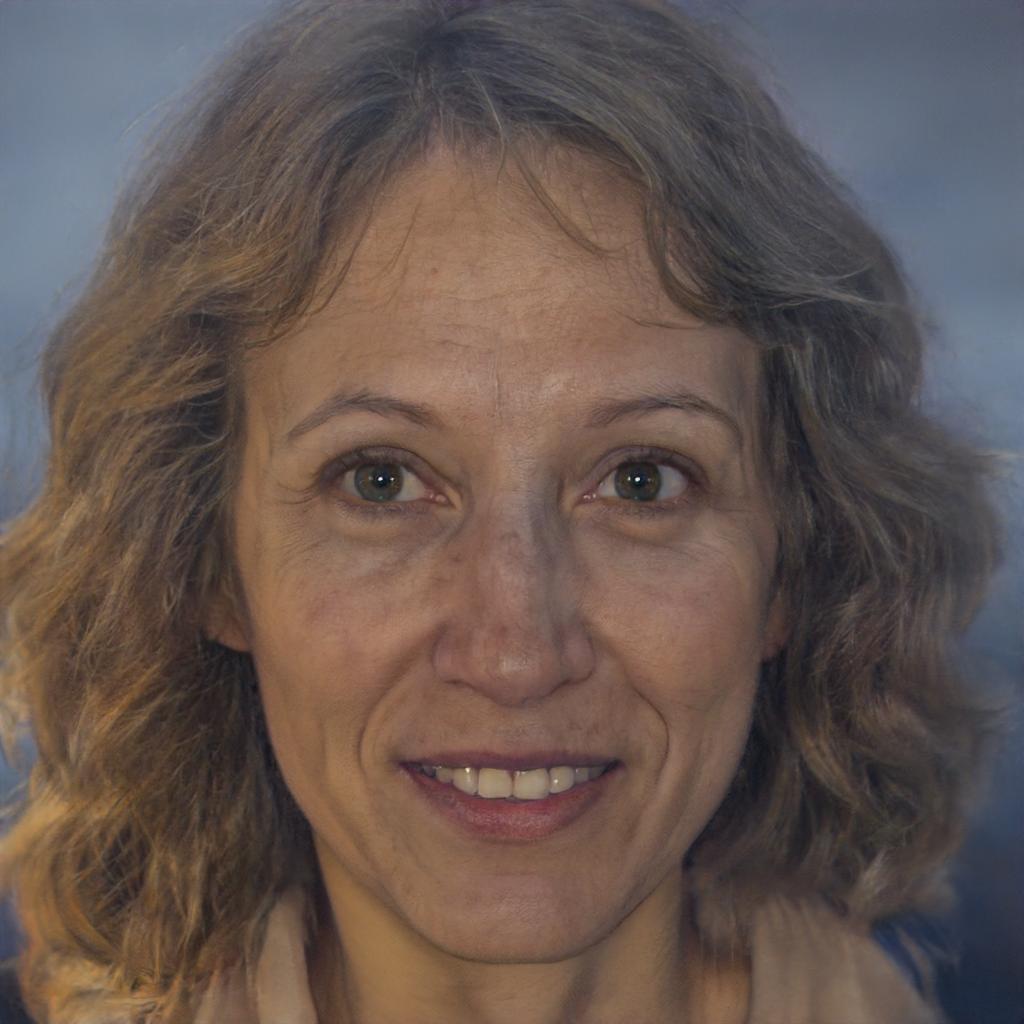} &
		\includegraphics[width=.15\linewidth]{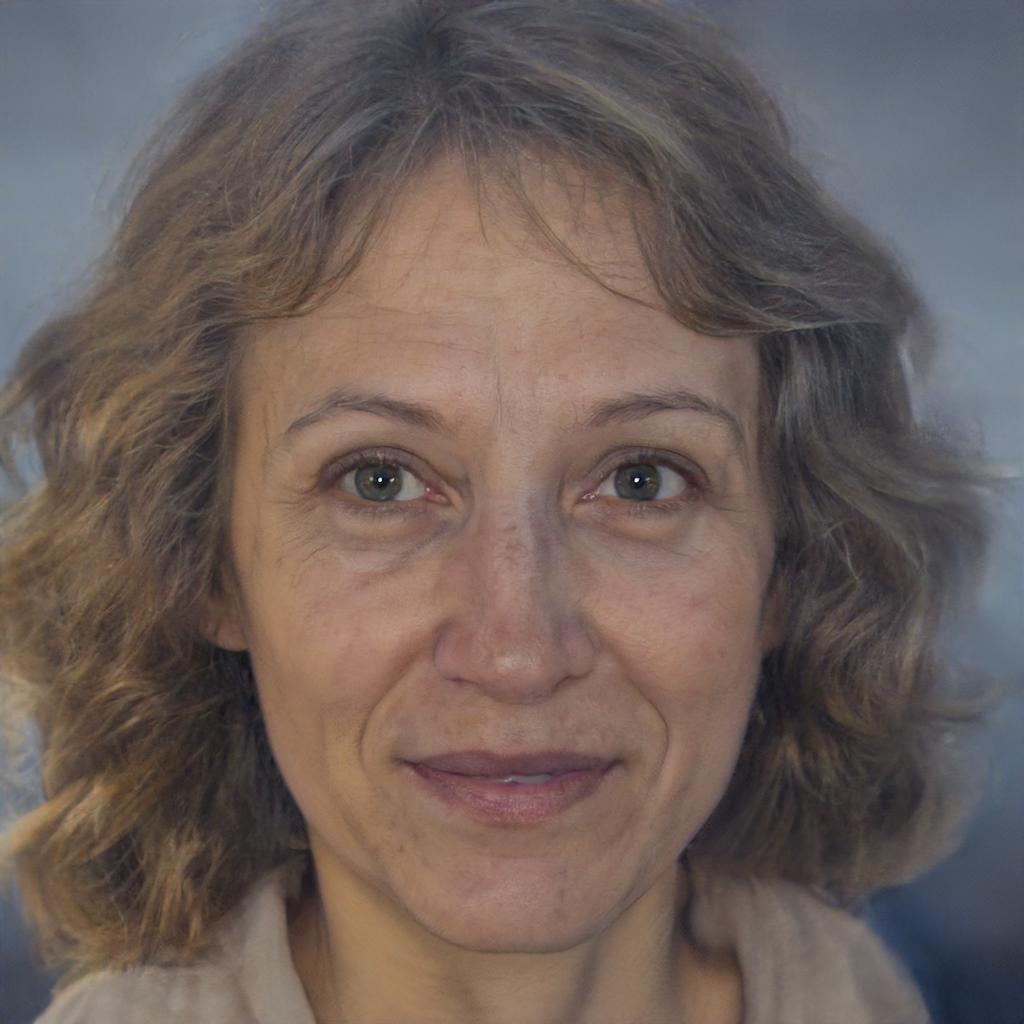} &
		\includegraphics[width=.15\linewidth]{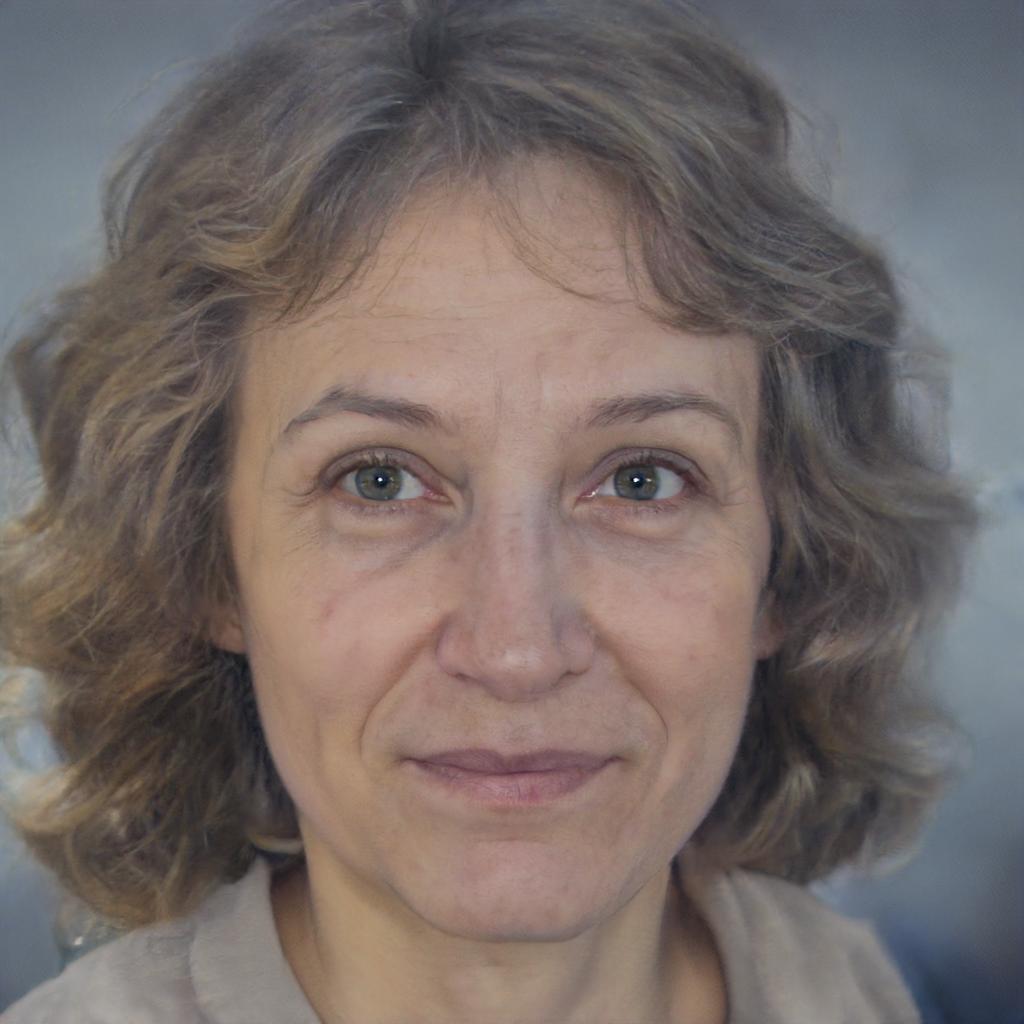} &
		\includegraphics[width=.15\linewidth]{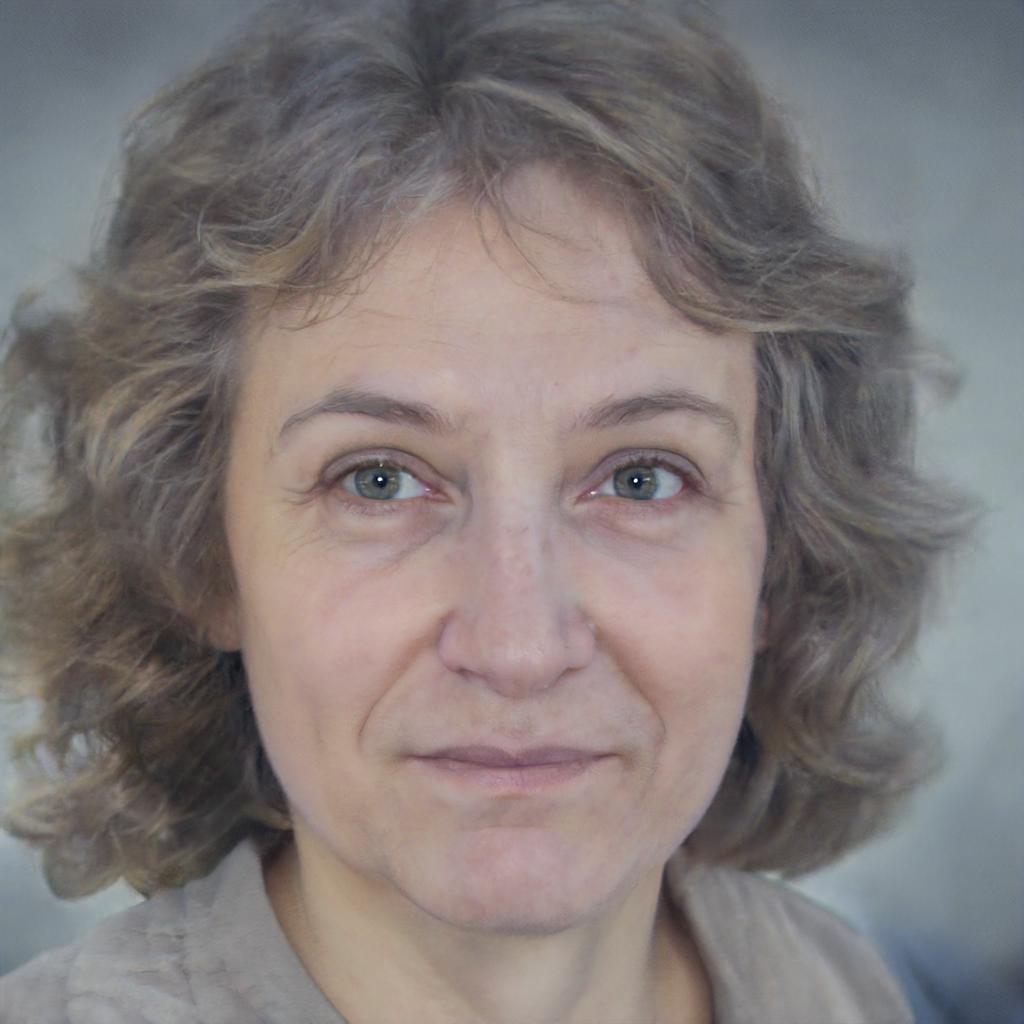} &
		\includegraphics[width=.15\linewidth]{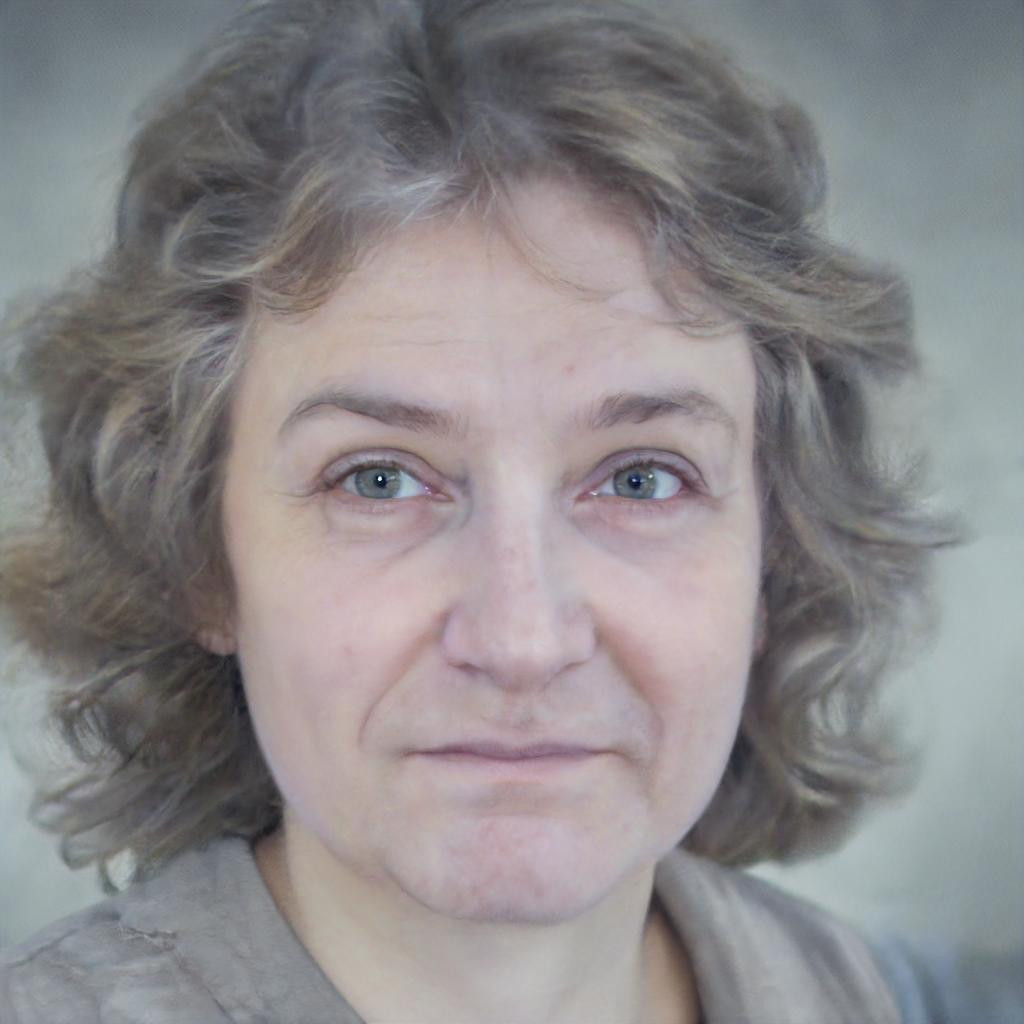} &
		\includegraphics[width=.15\linewidth]{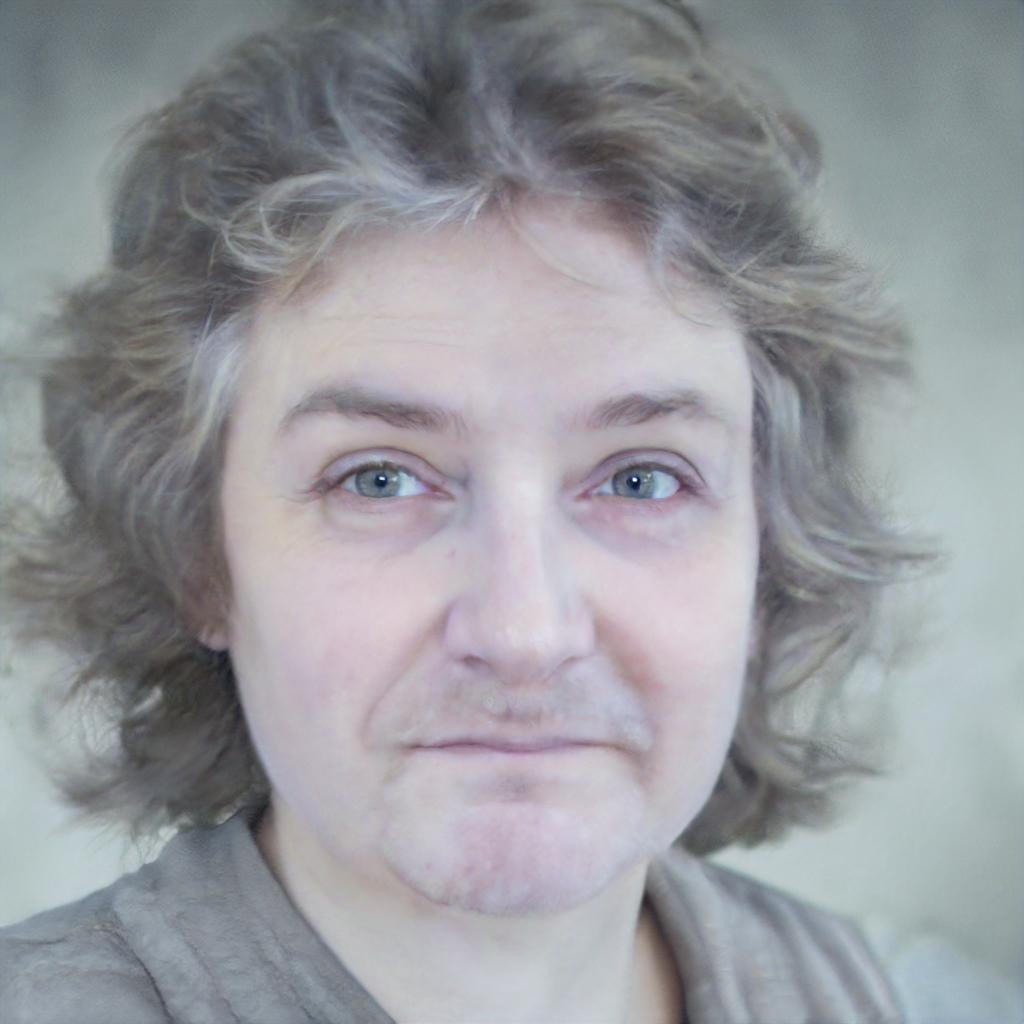}
		&\rotatebox[origin=l]{90}{\hspace{4.5mm}Inter-}
		\\	
		\rotatebox[origin=l]{90}{\hspace{4mm}$\longleftarrow$}&
		\includegraphics[width=.15\linewidth]{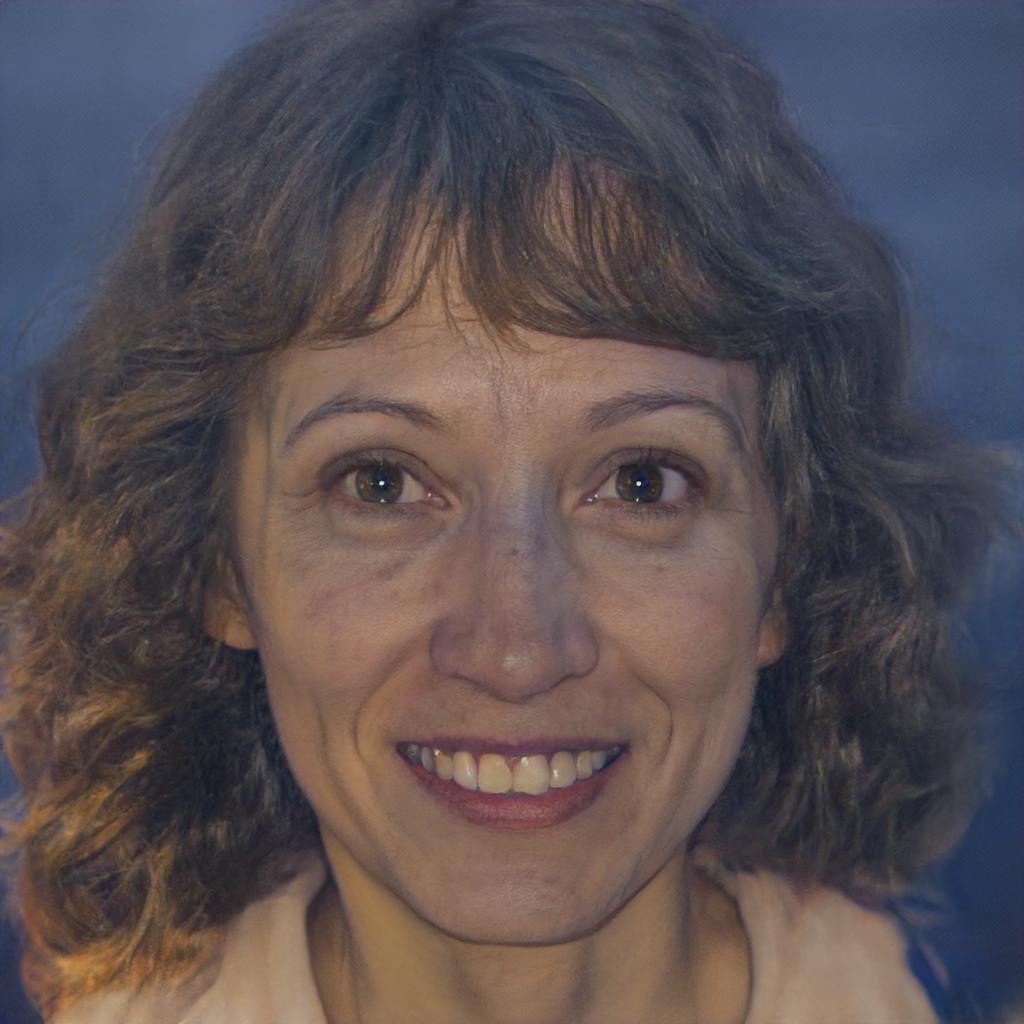} &		\includegraphics[width=.15\linewidth]{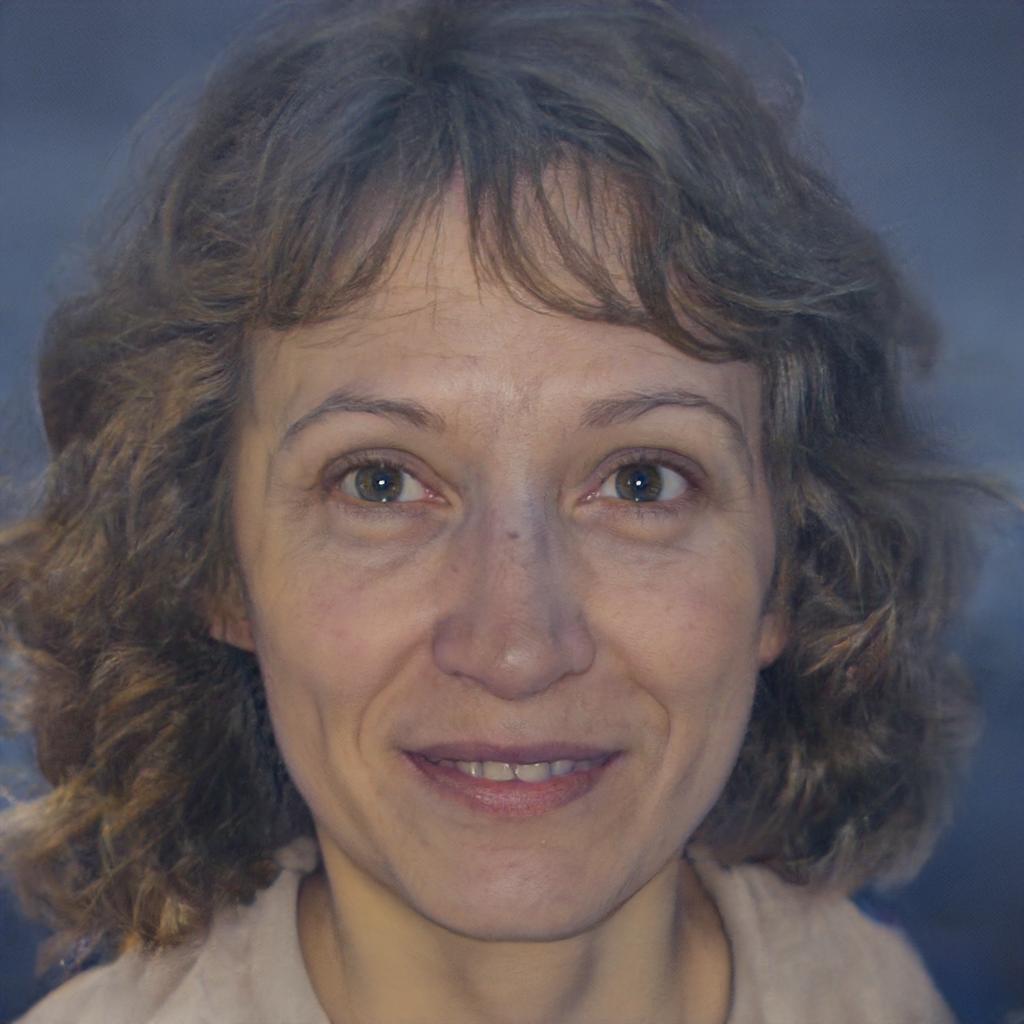} &
		\includegraphics[width=.15\linewidth]{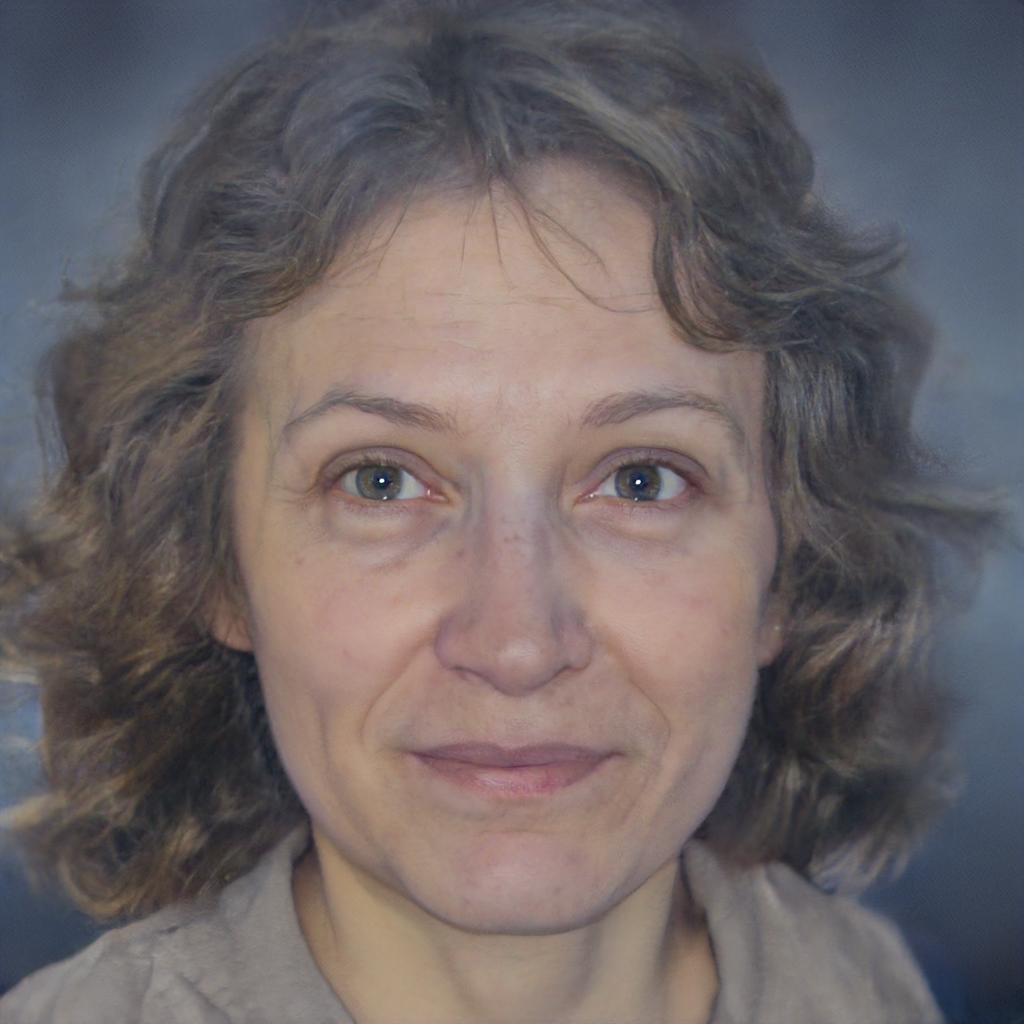} &
		\includegraphics[width=.15\linewidth]{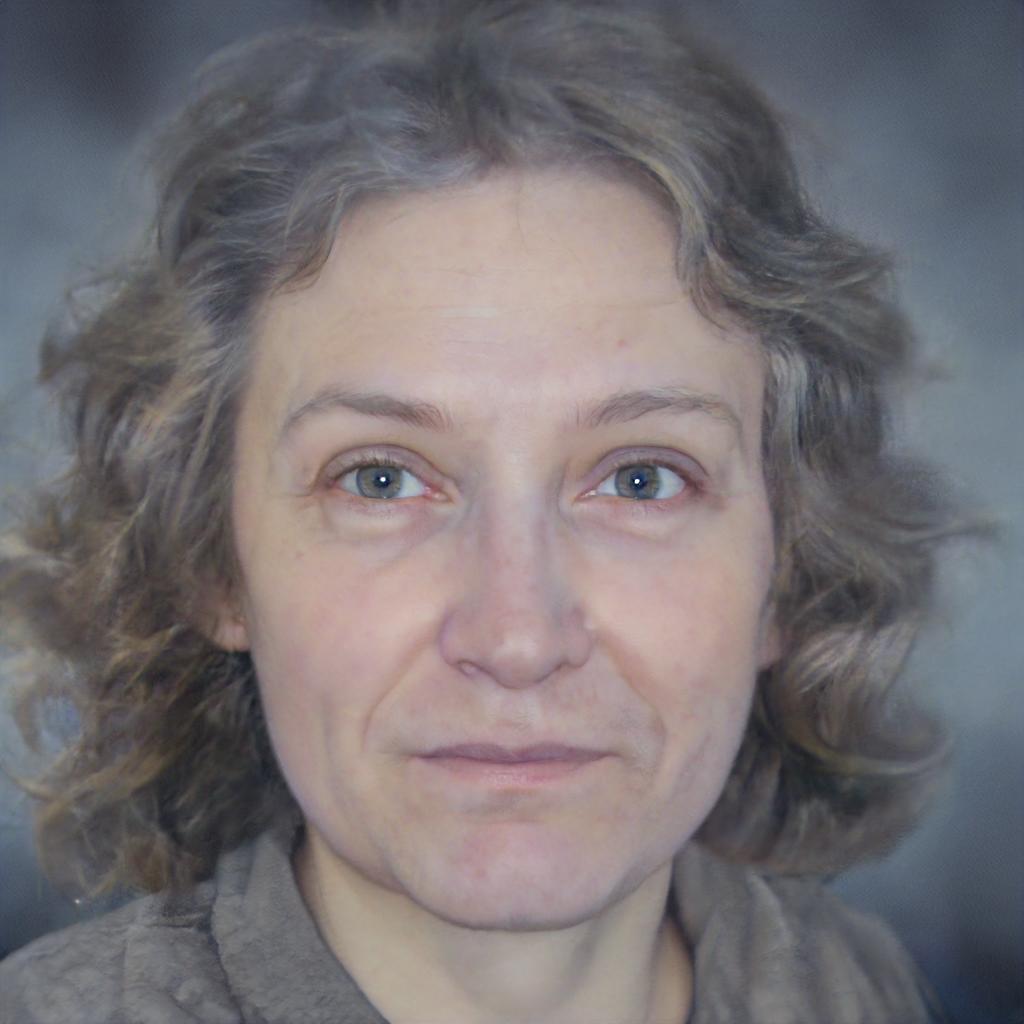} &
		\includegraphics[width=.15\linewidth]{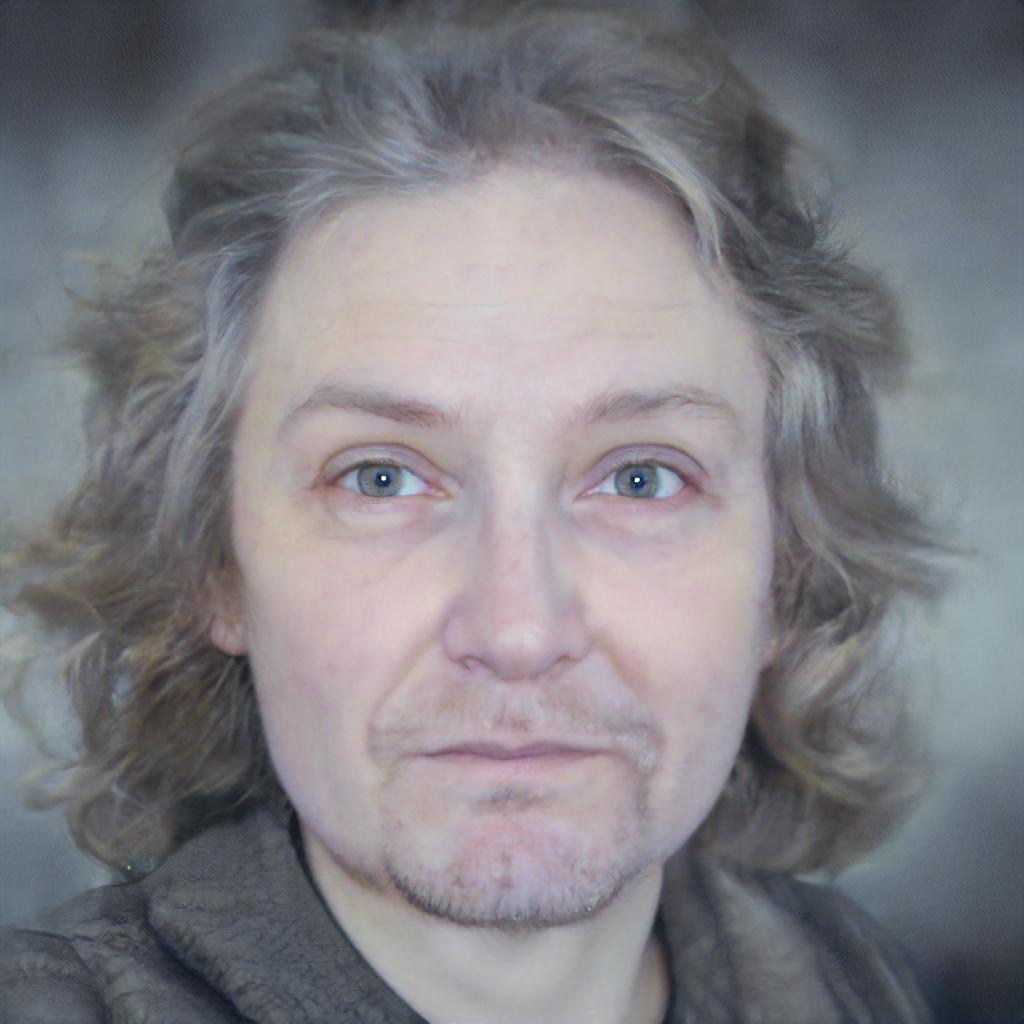} &
		\includegraphics[width=.15\linewidth]{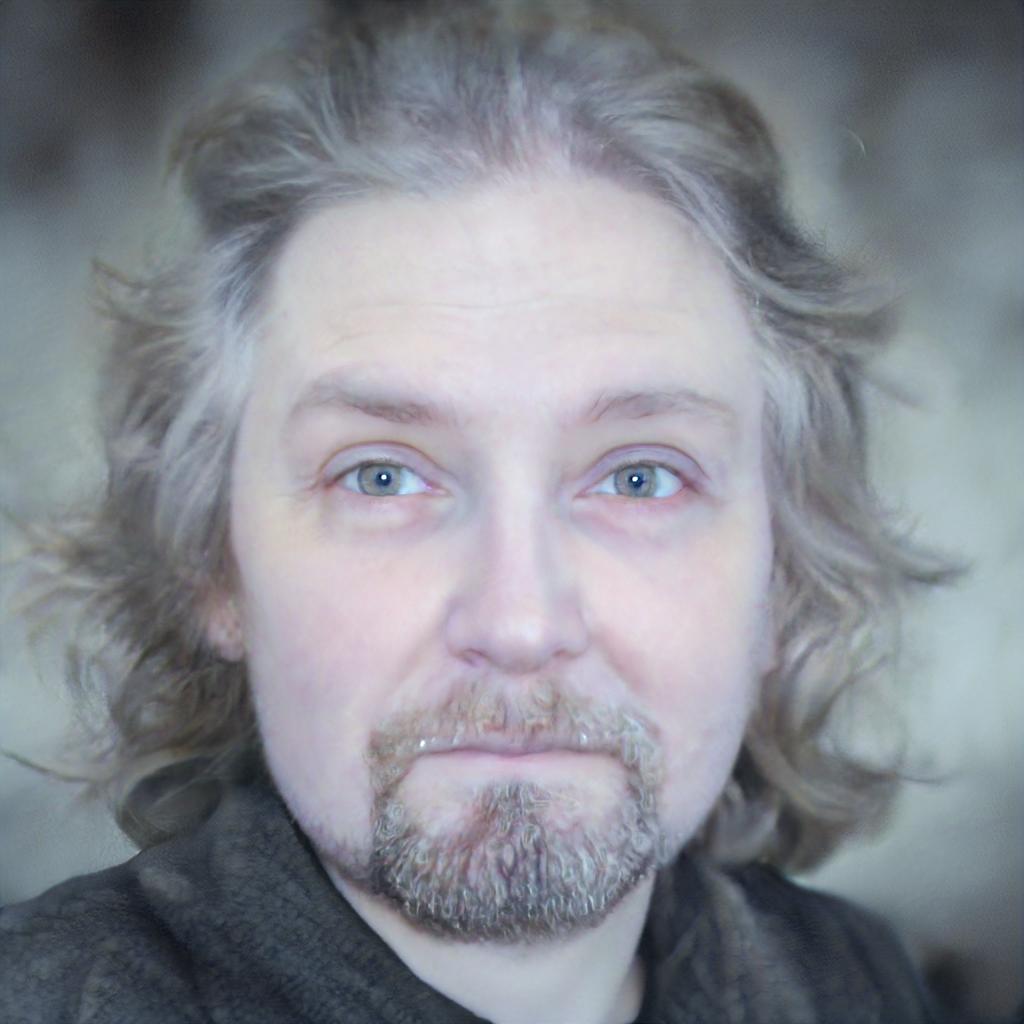}
		&\rotatebox[origin=l]{90}{\hspace{4mm}$\longleftarrow$}
		\\	
		&
		\fcolorbox{red}{red}{\includegraphics[width=.15\linewidth]{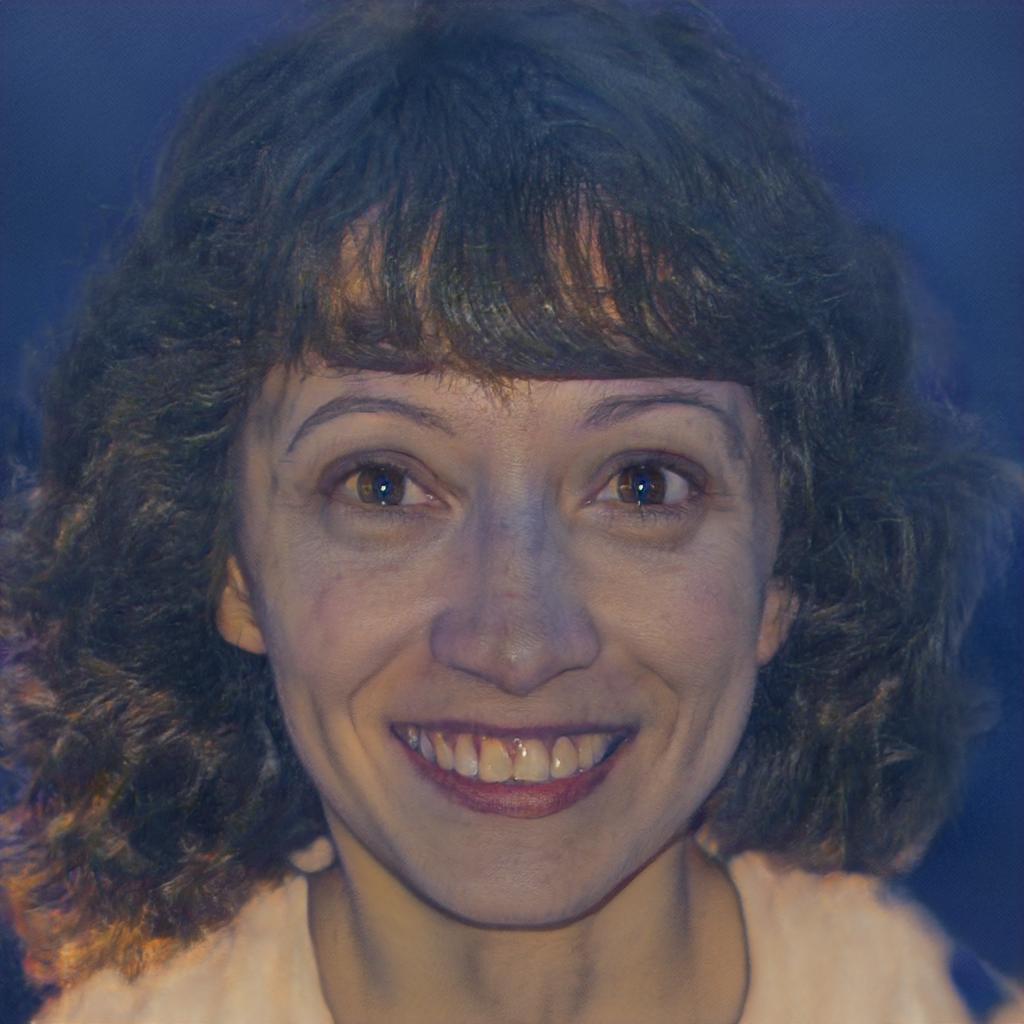}} &
		\includegraphics[width=.15\linewidth]{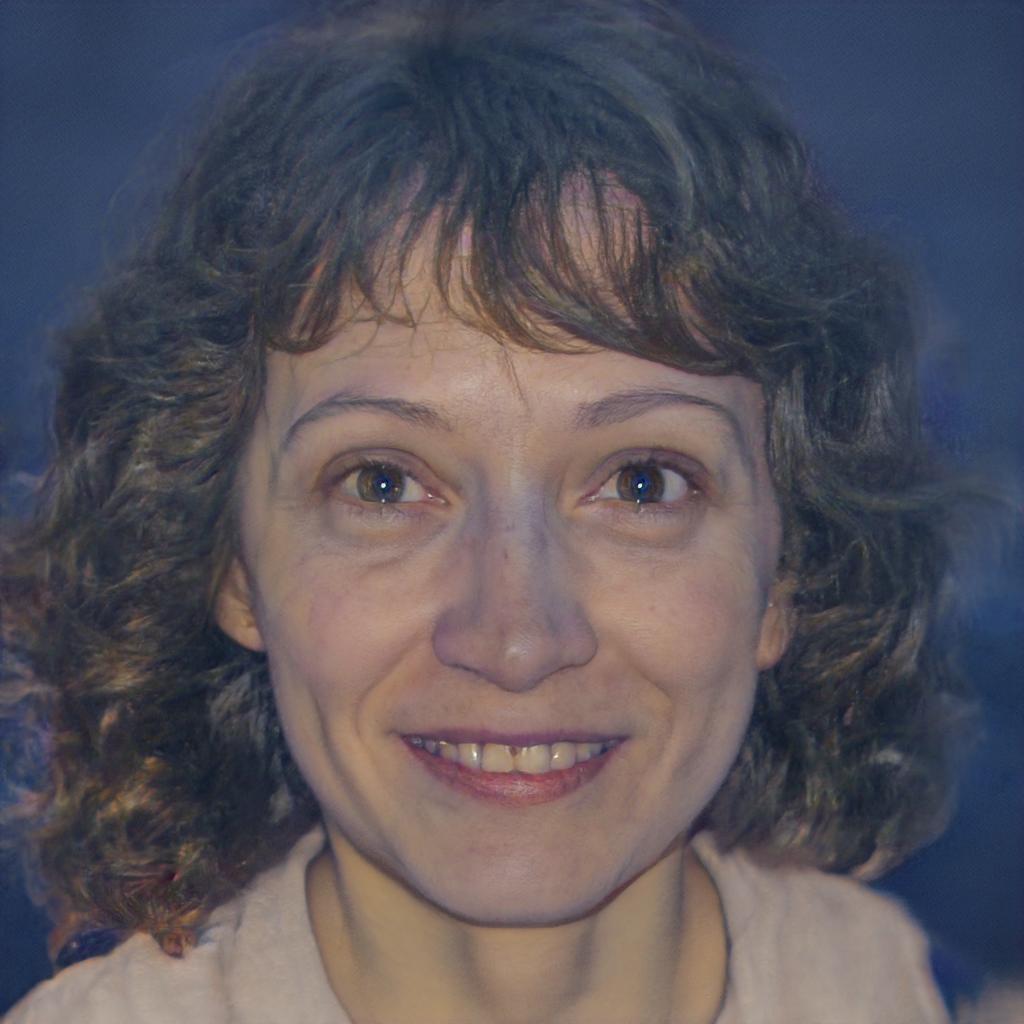} &
		\includegraphics[width=.15\linewidth]{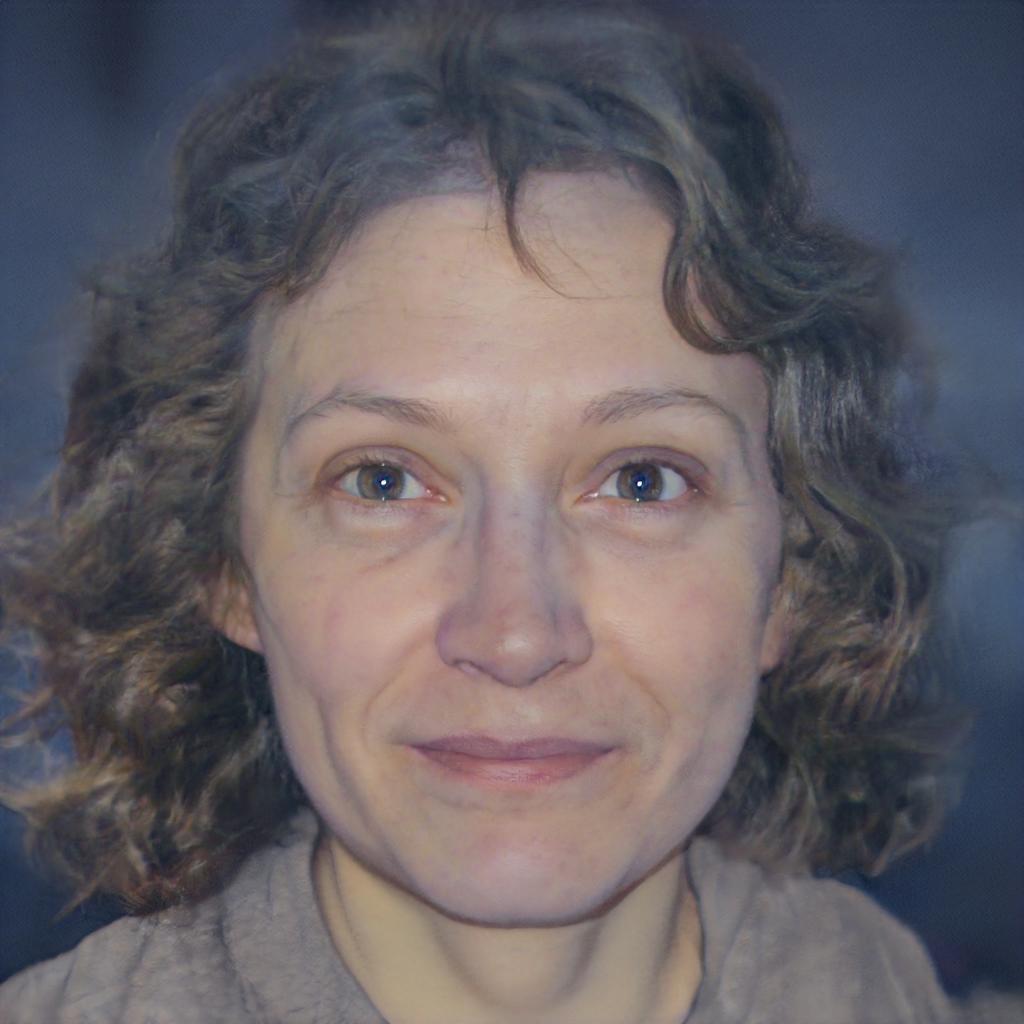} &
		\includegraphics[width=.15\linewidth]{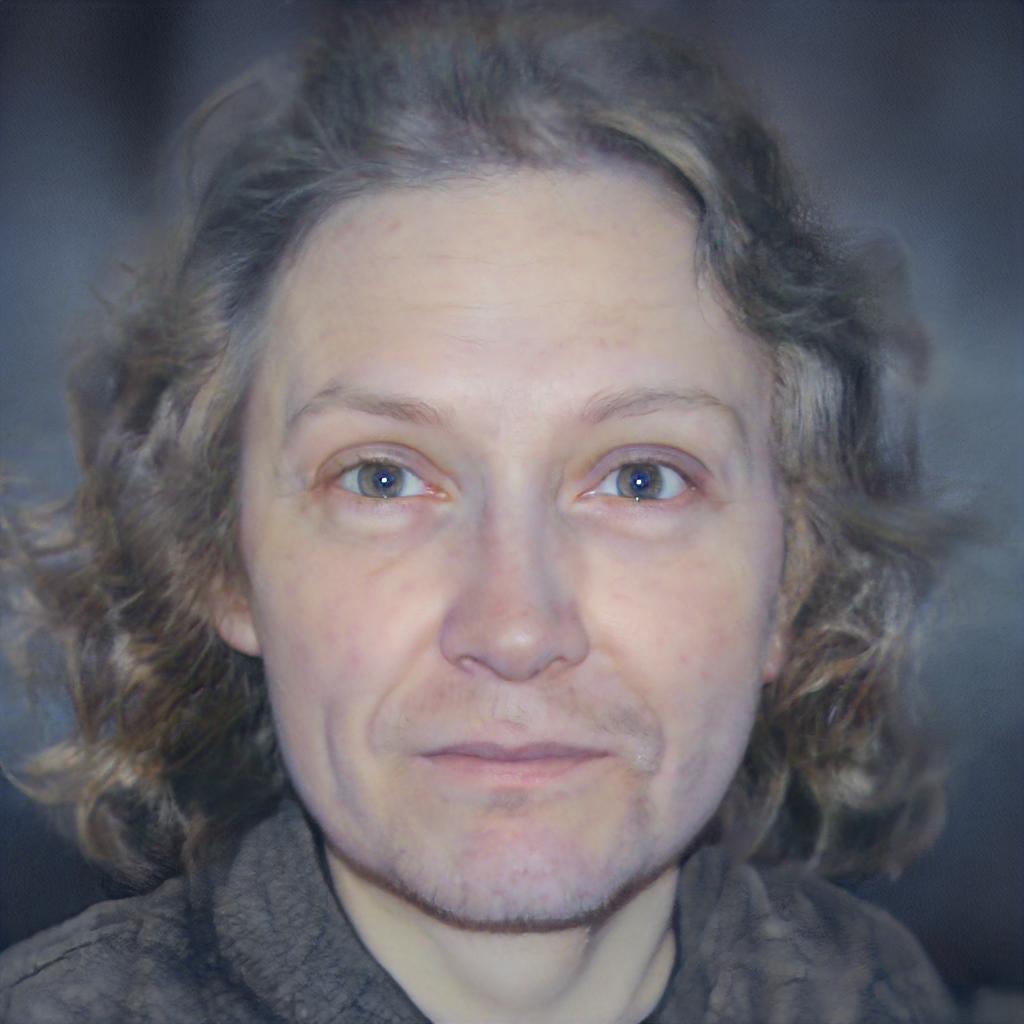} &
		\includegraphics[width=.15\linewidth]{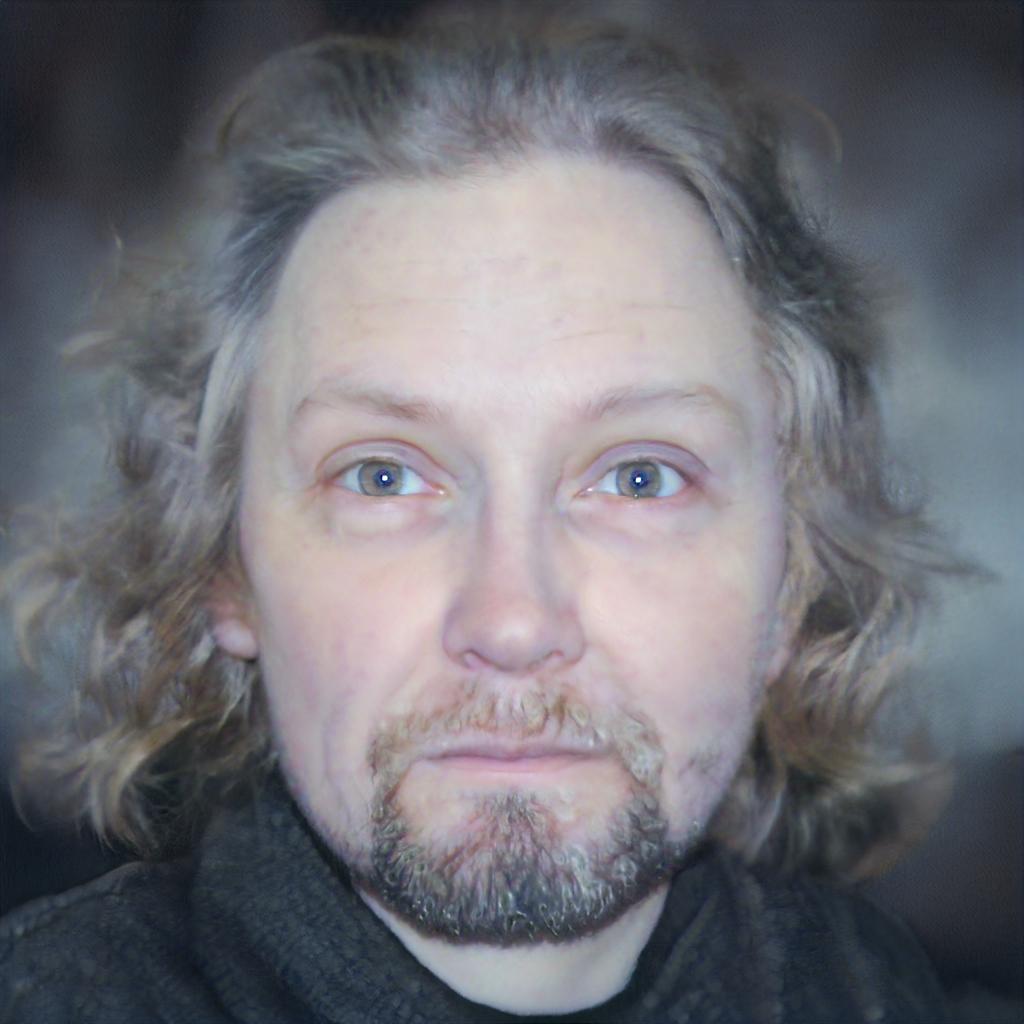} &
		\fcolorbox{red}{red}{\includegraphics[width=.15\linewidth]{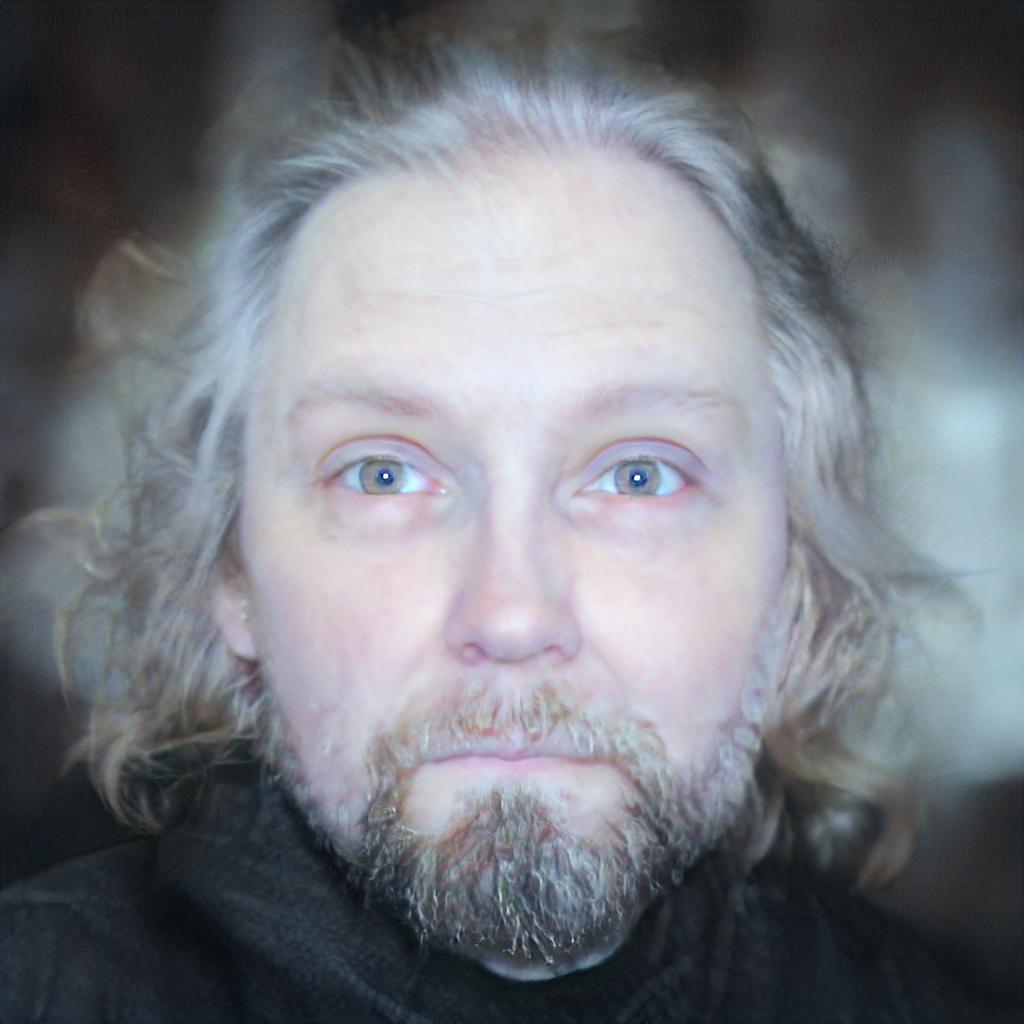}}
		&
		\\
		&Result C&\multicolumn{4}{c}{$\longleftarrow$\quad\quad\quad Interpolation\quad\quad\quad$\longrightarrow$}&Result D&\\	
		\bottomrule
		&Source C&&&&&Source D&\\
		&\includegraphics[width=.15\linewidth]{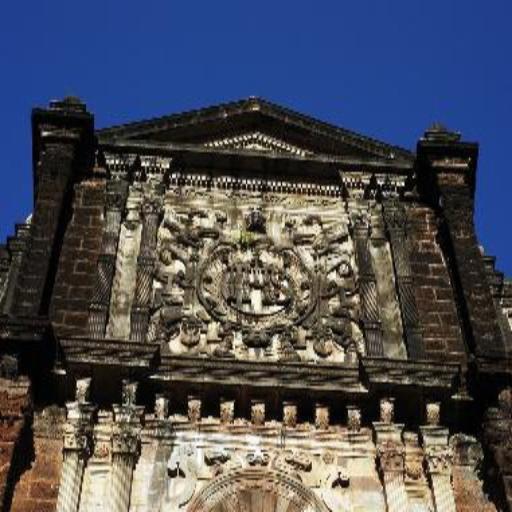}
		& & & & &
		\includegraphics[width=.15\linewidth]{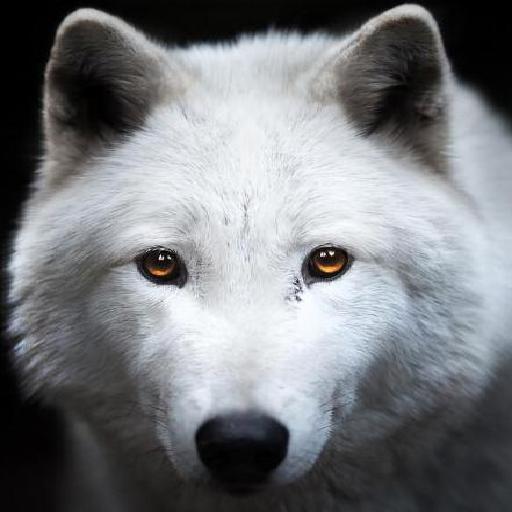}&
		
	\end{tabular}
	\caption{We take four input images from different source domains, transform them into the target FFHQ domain, and save the corresponding $w$ latent codes. Subsequently, we conduct smooth interpolation between these latent codes. The images in the corners indicate the source images and the 6$\times$6 images in the middle depict the interpolated results.}
	\label{fig:interpolation}
\end{figure}

\begin{figure}[t]
	\centering
	\includegraphics[width=\linewidth]{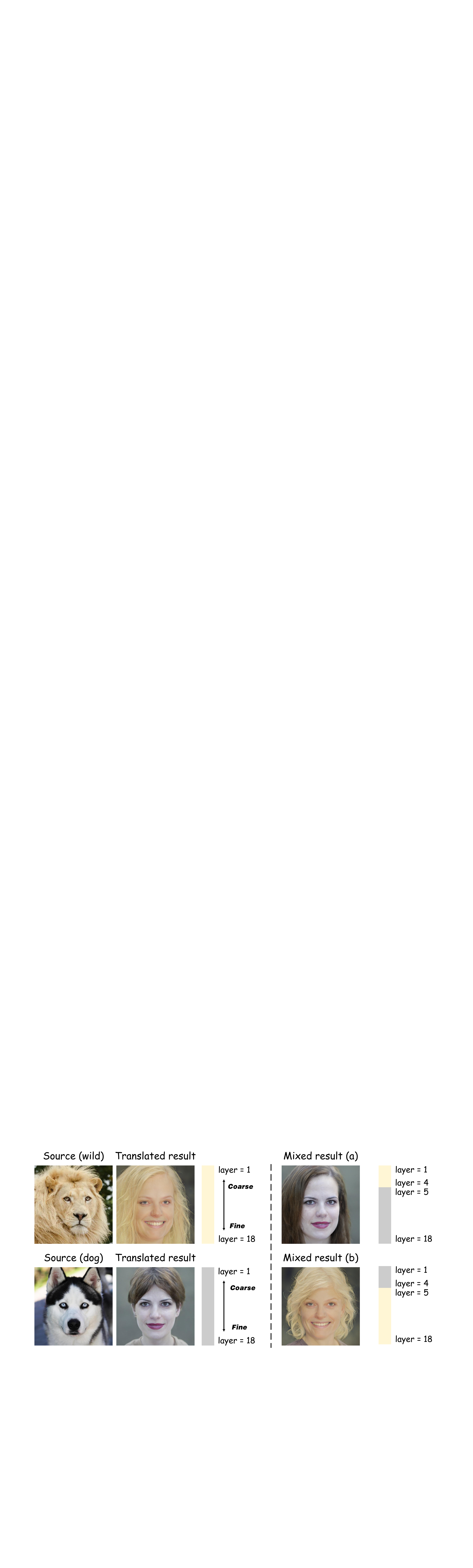}
	\caption{Style mixing. Mixed result (a) is generated by injecting coarse wild style into the dog style, while mixed result (b) is produced by injecting fine wild style into the dog style.}
	\label{fig:stylemixing}
\end{figure}

\begin{figure}[t]
	\centering
	\setlength{\tabcolsep}{0.05em}
	\begin{tabular}{ccccc}
		\includegraphics[width=.19\linewidth]{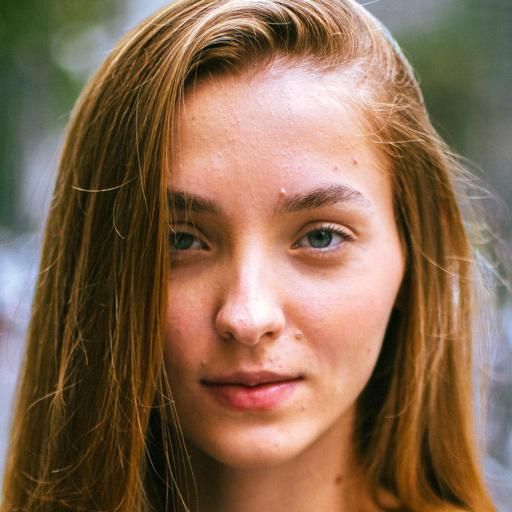} &
		\includegraphics[width=.19\linewidth]{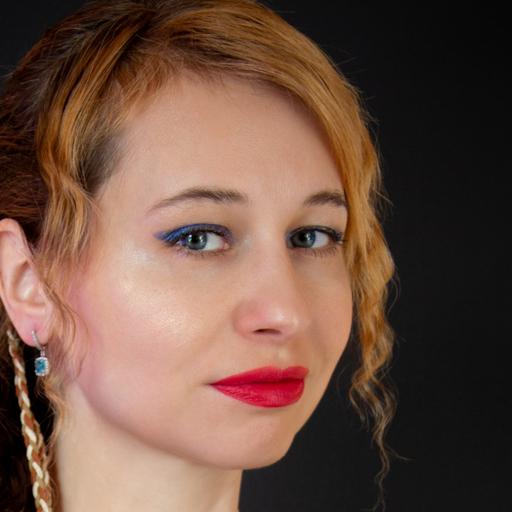} &
		\includegraphics[width=.19\linewidth]{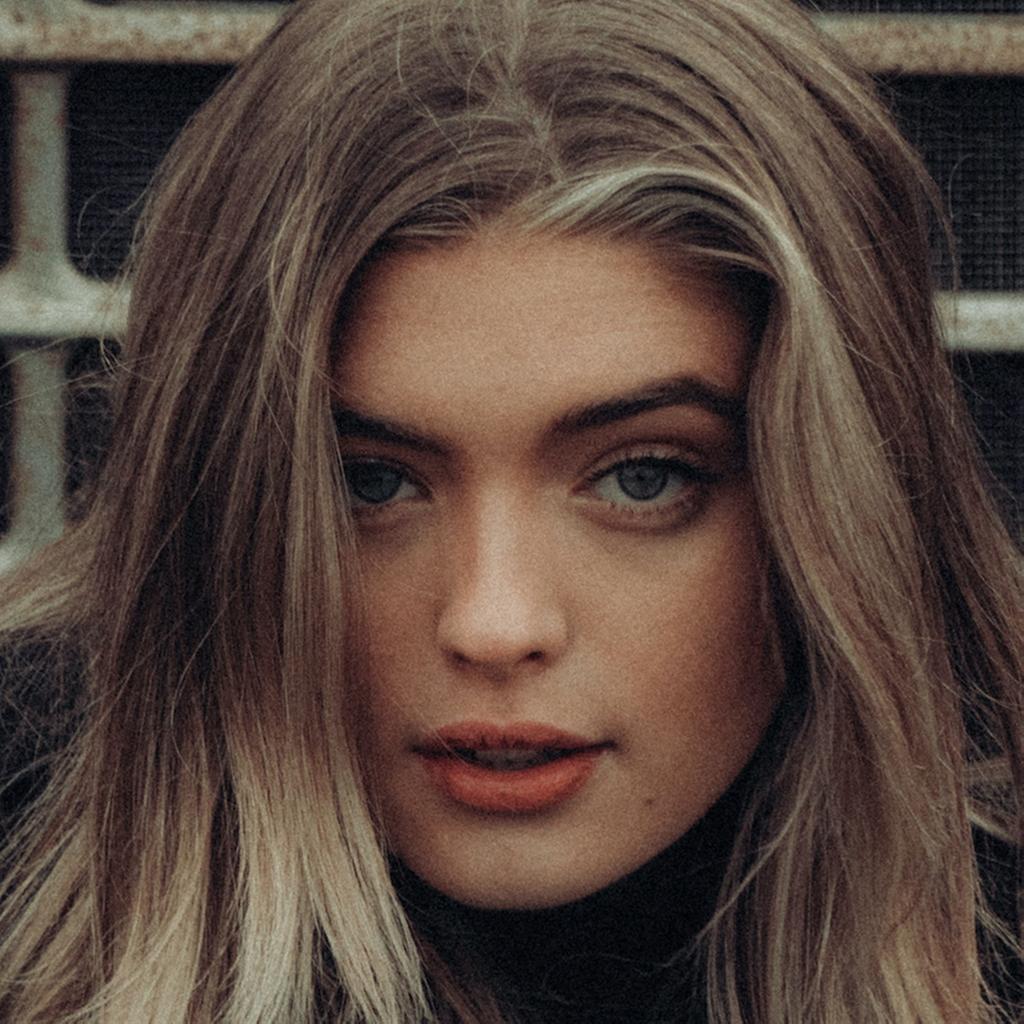} &
		\includegraphics[width=.19\linewidth]{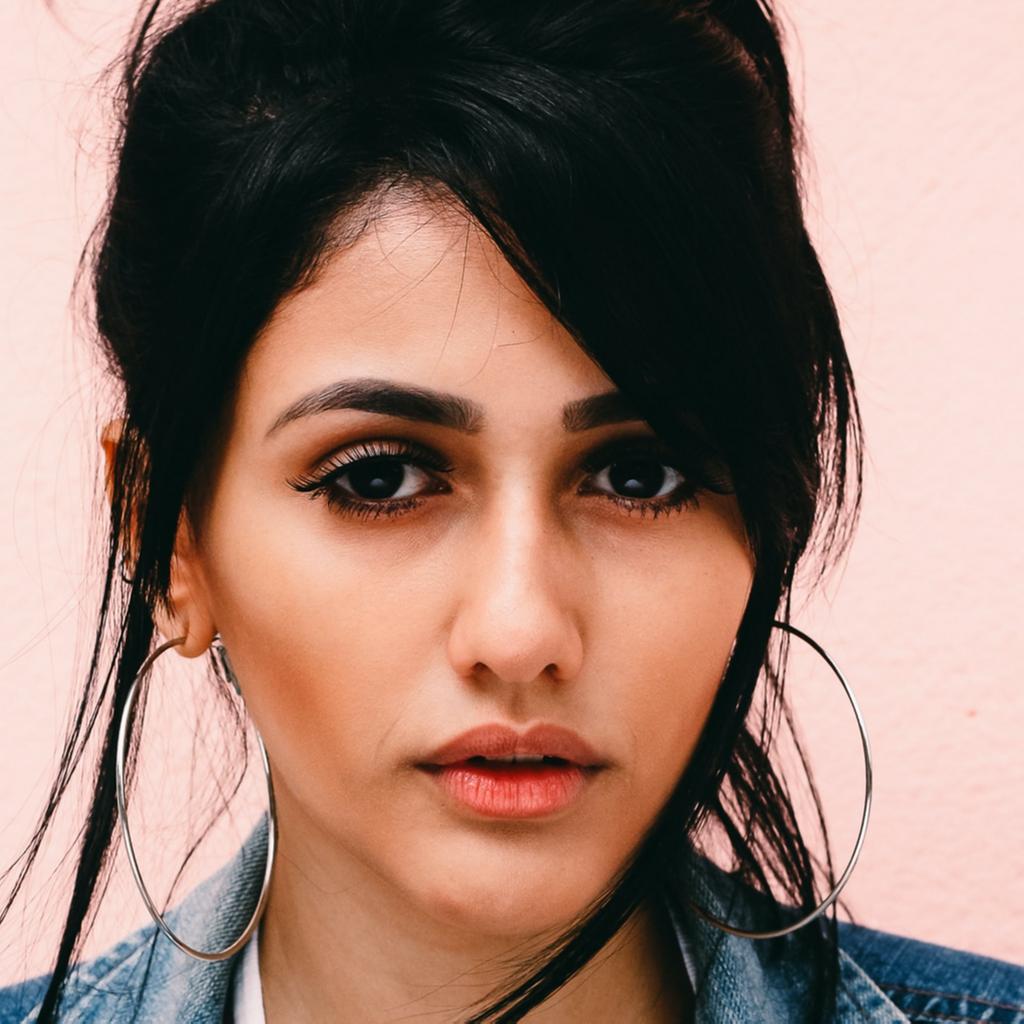} &
		\includegraphics[width=.19\linewidth]{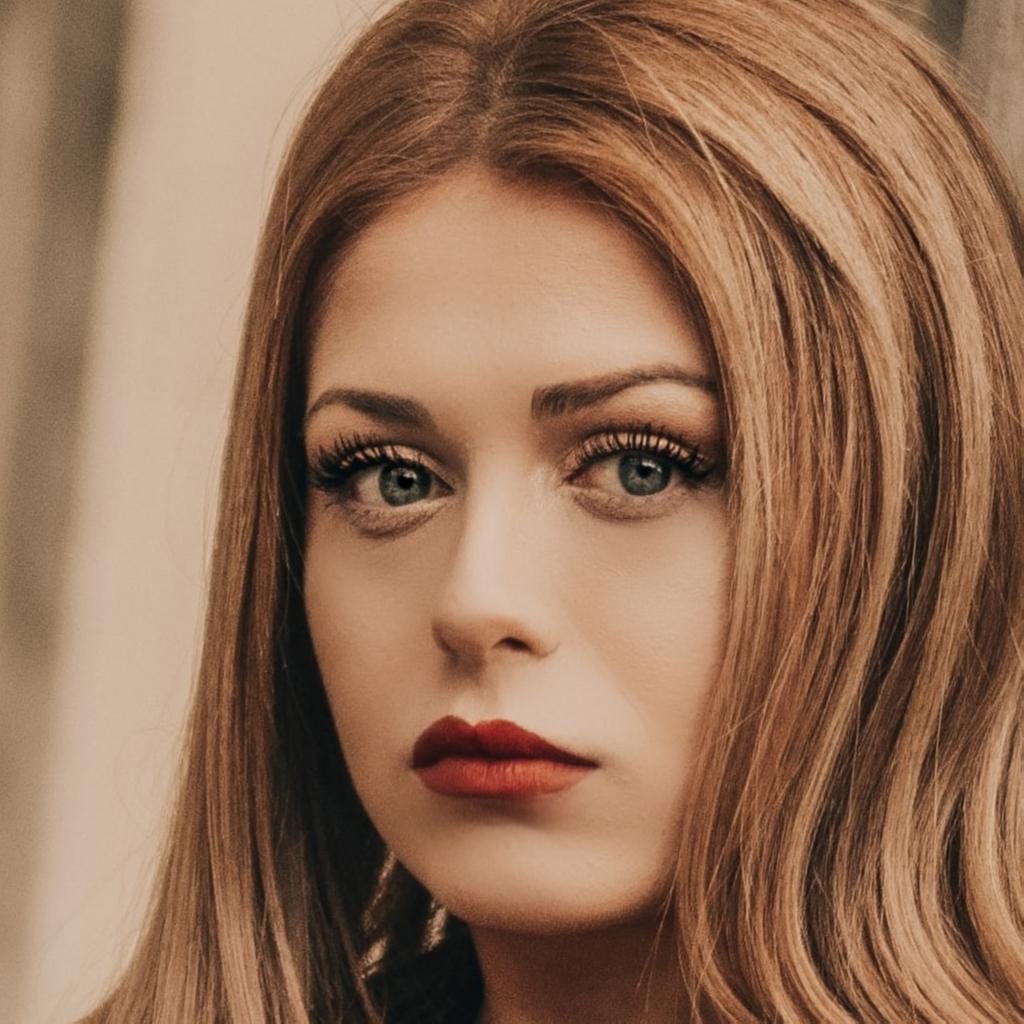} \\		
	    \includegraphics[width=.19\linewidth]{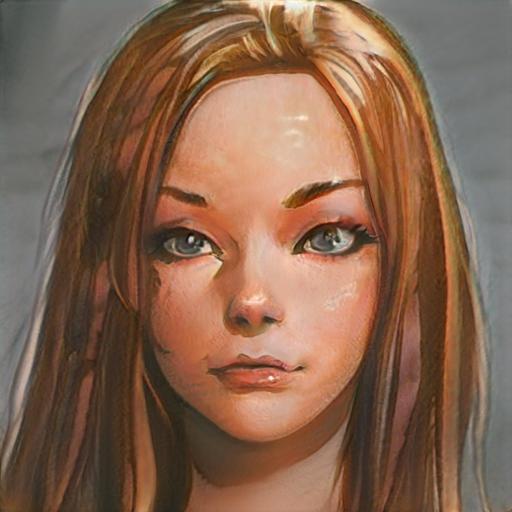} &
	    \includegraphics[width=.19\linewidth]{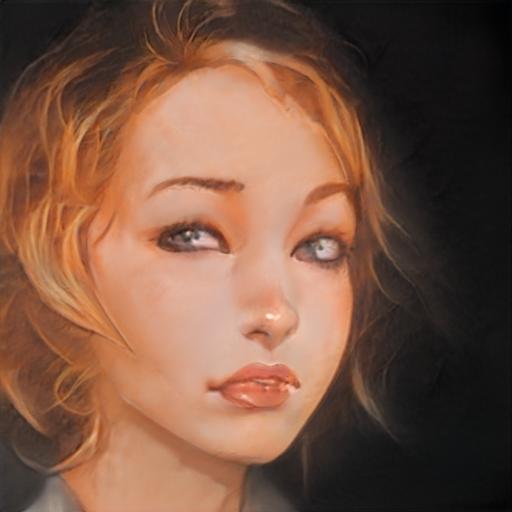} &
		\includegraphics[width=.19\linewidth]{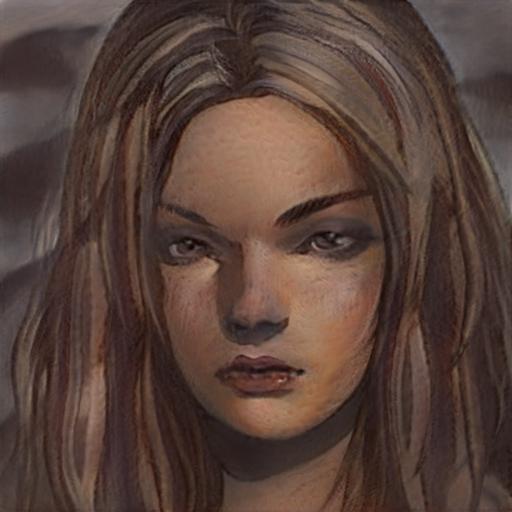} &
		\includegraphics[width=.19\linewidth]{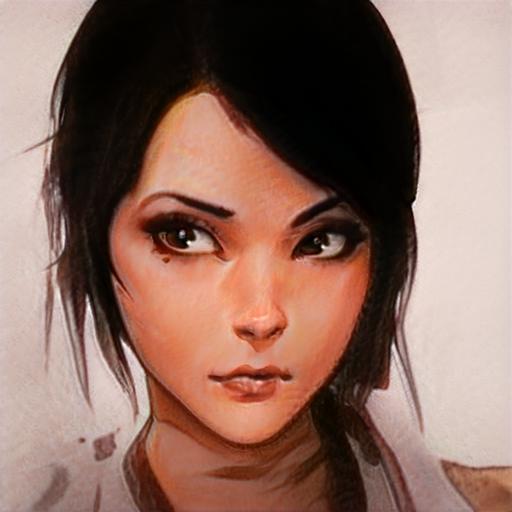} &
		\includegraphics[width=.19\linewidth]{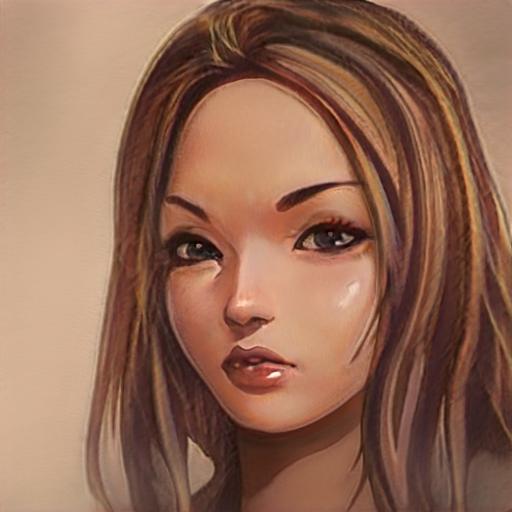} \\
		\includegraphics[width=.19\linewidth]{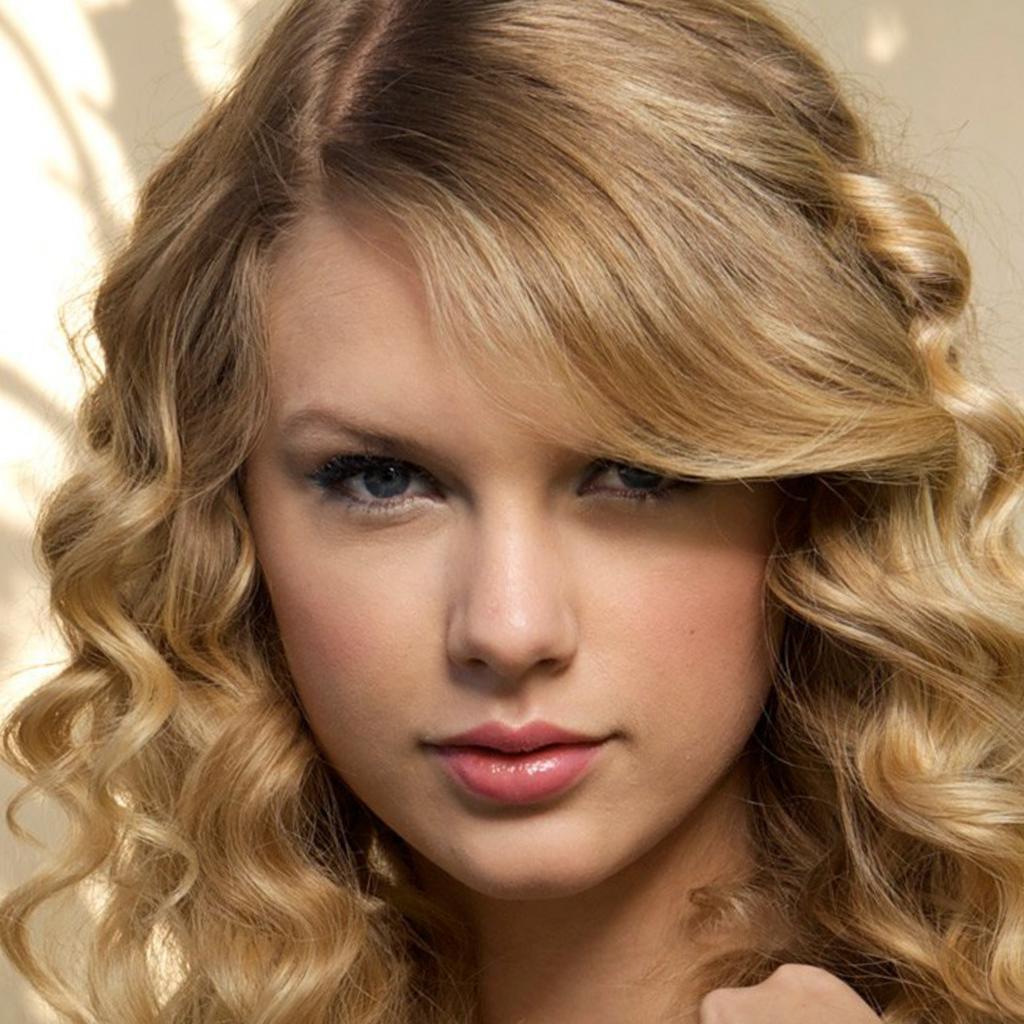} &
		\includegraphics[width=.19\linewidth]{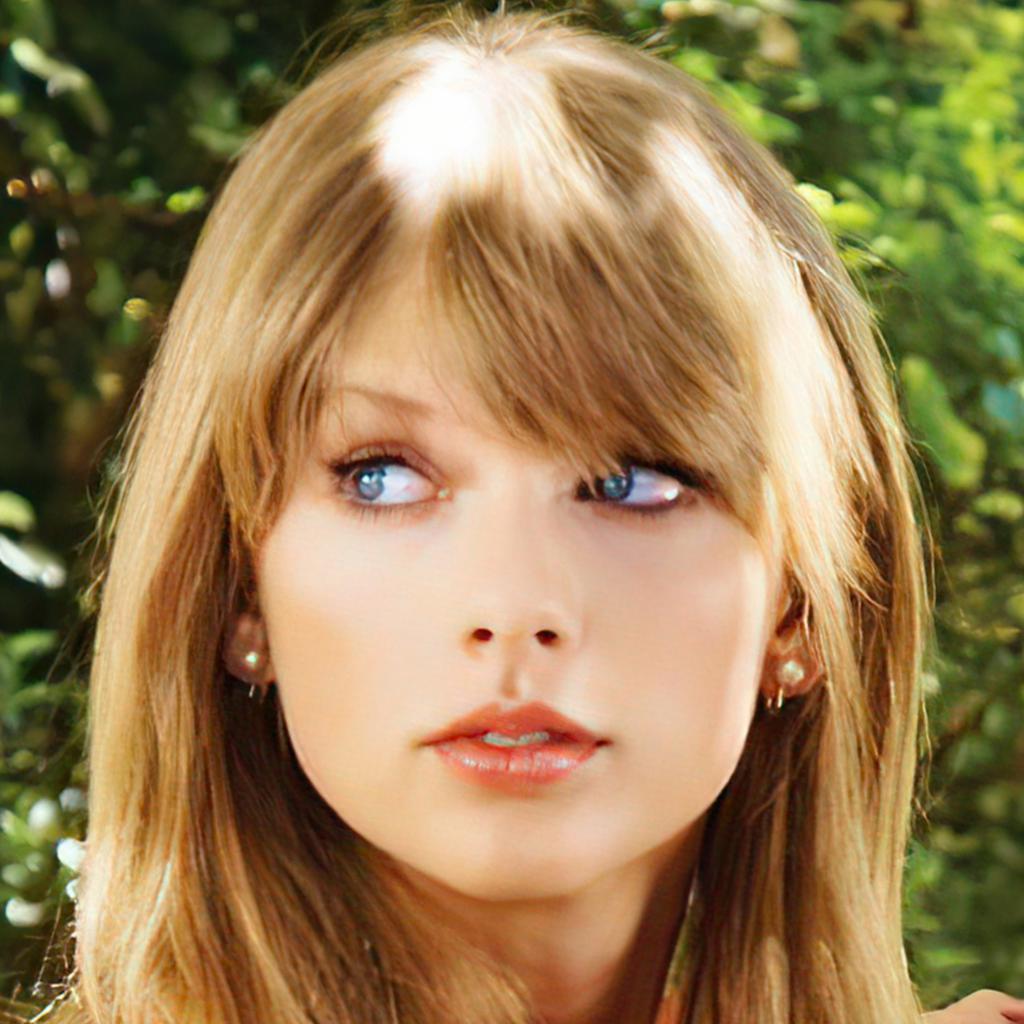} &
		\includegraphics[width=.19\linewidth]{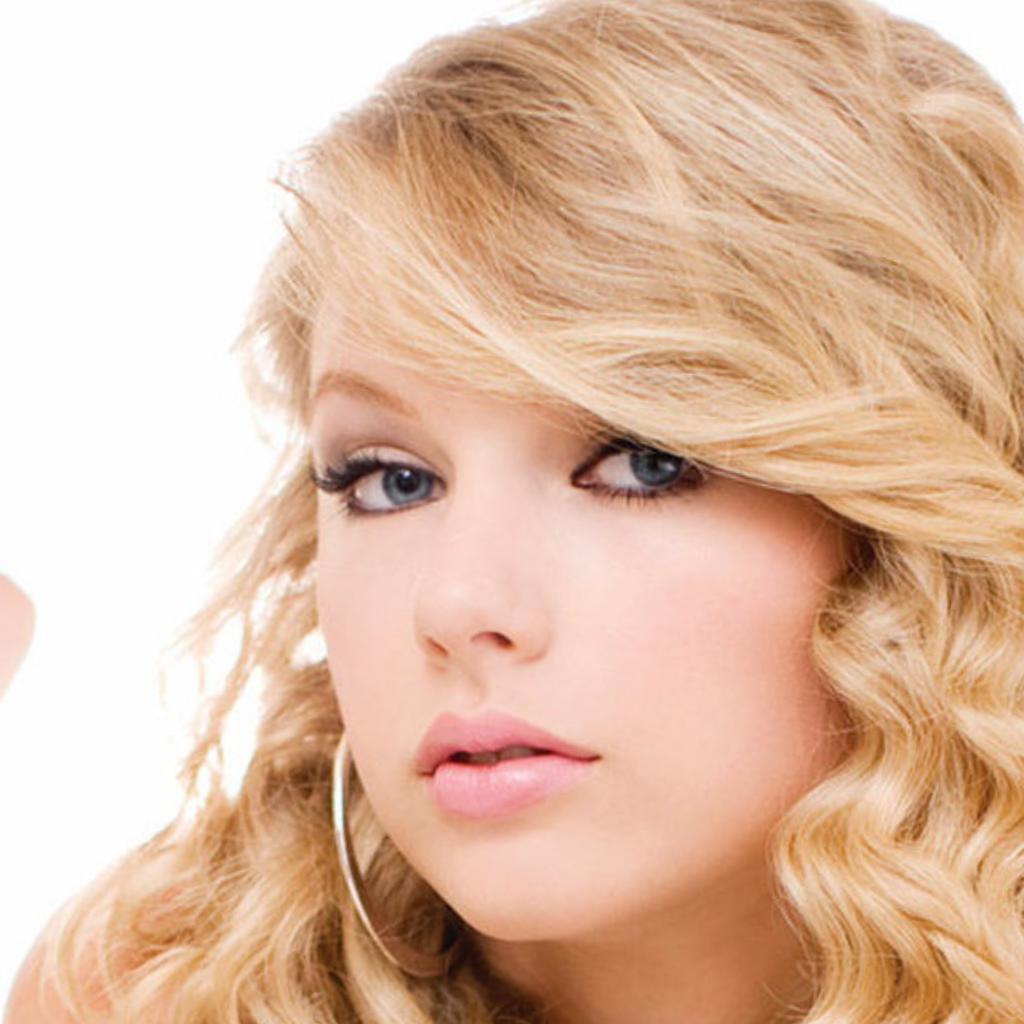} &
		\includegraphics[width=.19\linewidth]{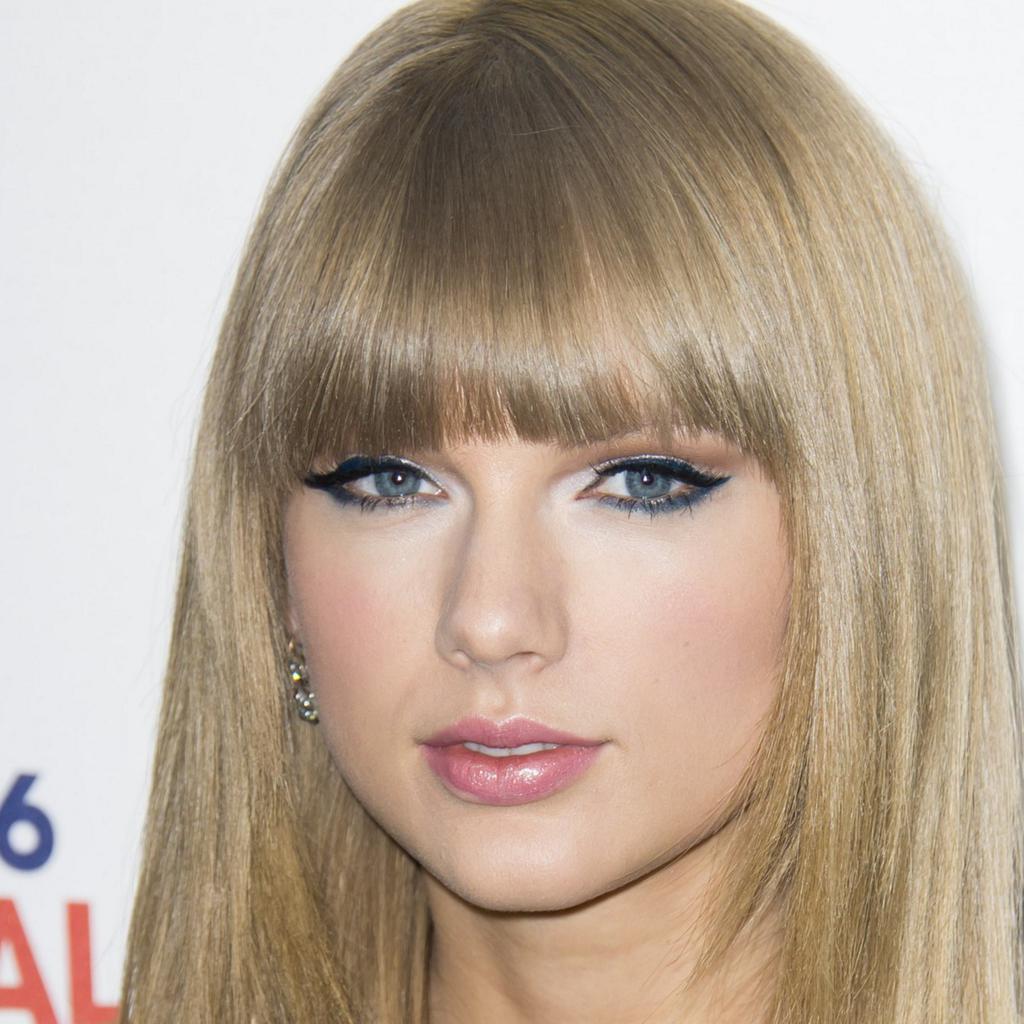}&
		\includegraphics[width=.19\linewidth]{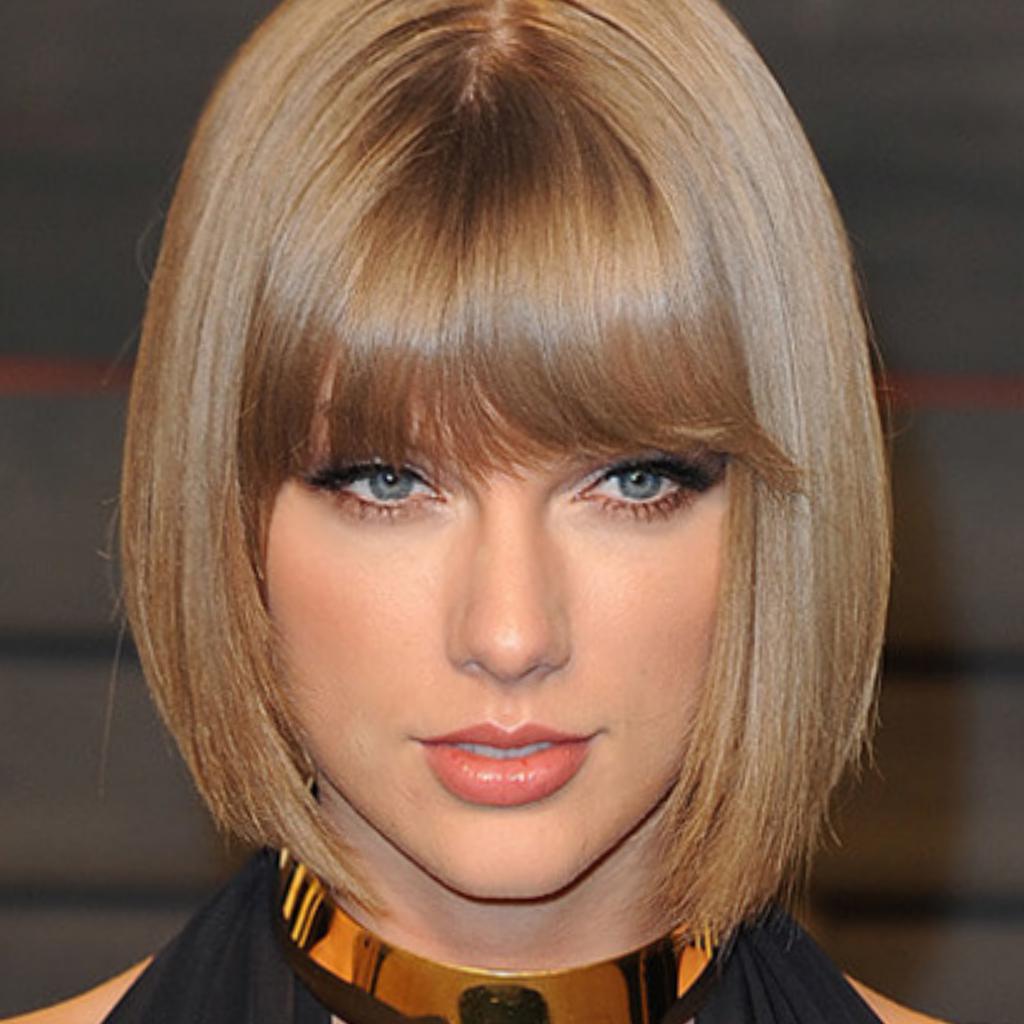}\\
		\includegraphics[width=.19\linewidth]{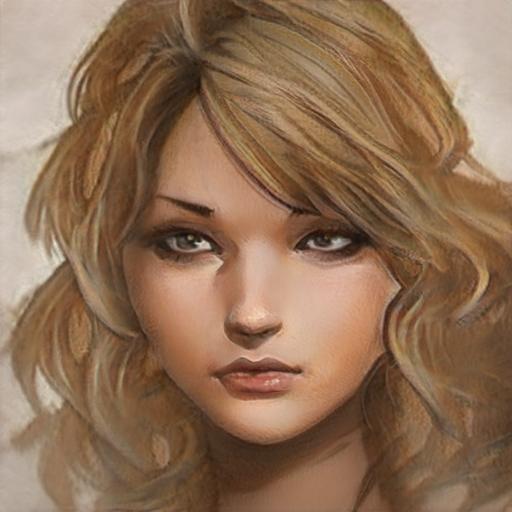} &
		\includegraphics[width=.19\linewidth]{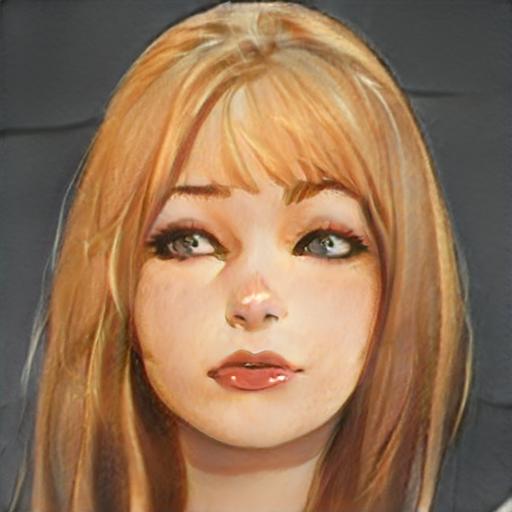} &
		\includegraphics[width=.19\linewidth]{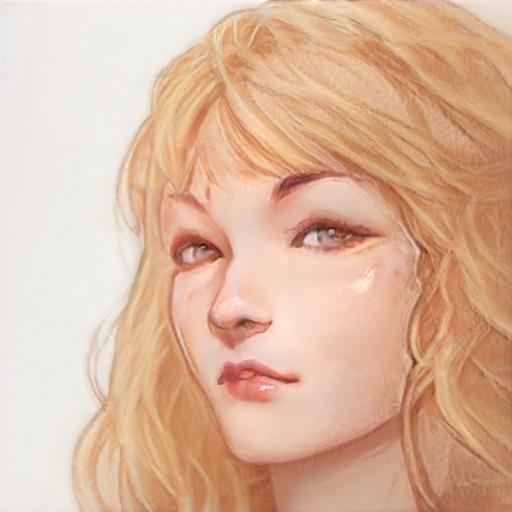} &
        \includegraphics[width=.19\linewidth]{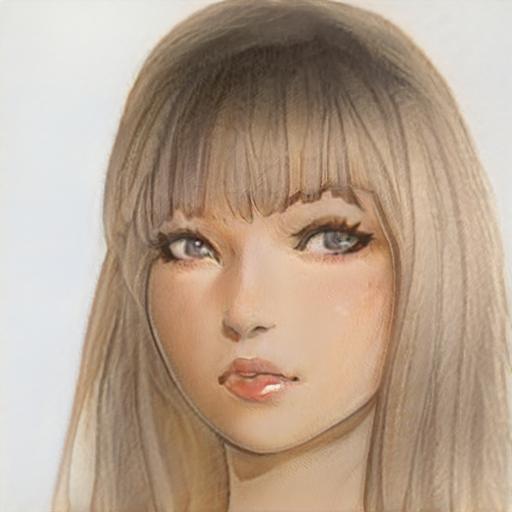}&
		\includegraphics[width=.19\linewidth]{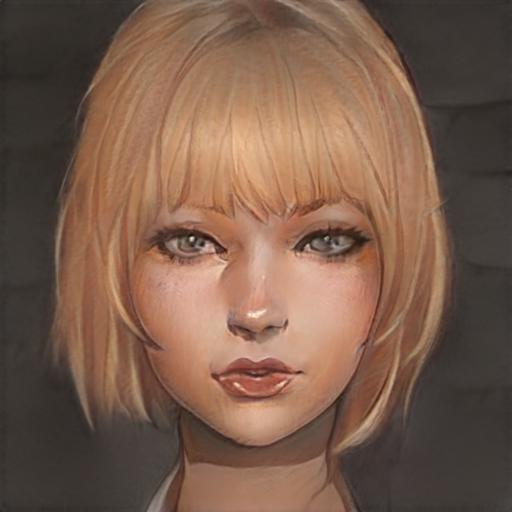} \\
	\end{tabular}
	\caption{Real images stylization. We transform real portraits into Ilya Kuvshinov's style. The first and third rows represent the real images, while the second and fourth rows showcase the stylized images.}
	\label{fig:stylization}
\end{figure}

\begin{figure*}[t]
	\centering
	\setlength{\abovecaptionskip}{0cm}
	\centering
	\setlength{\tabcolsep}{0.05em}
	\setlength{\fboxrule}{1pt}
	\setlength{\fboxsep}{0pt}
	\begin{tabular}{cccccccc}
		
	    \includegraphics[width=.12\linewidth]{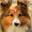} &
		\includegraphics[width=.12\linewidth]{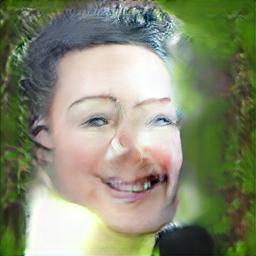} &
		\includegraphics[width=.12\linewidth]{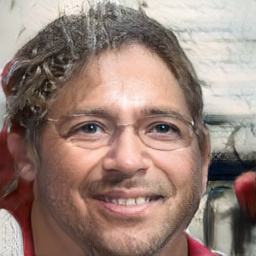} &
		\includegraphics[width=.12\linewidth]{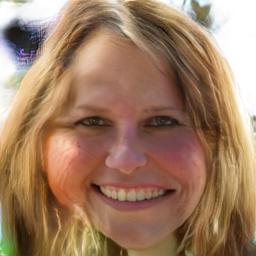} &
		\includegraphics[width=.12\linewidth]{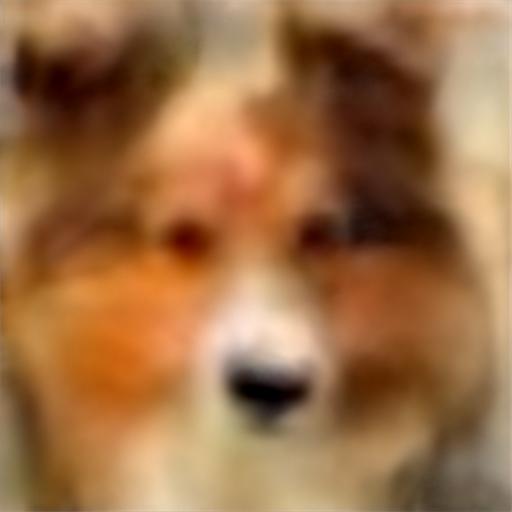} &
		\includegraphics[width=.12\linewidth]{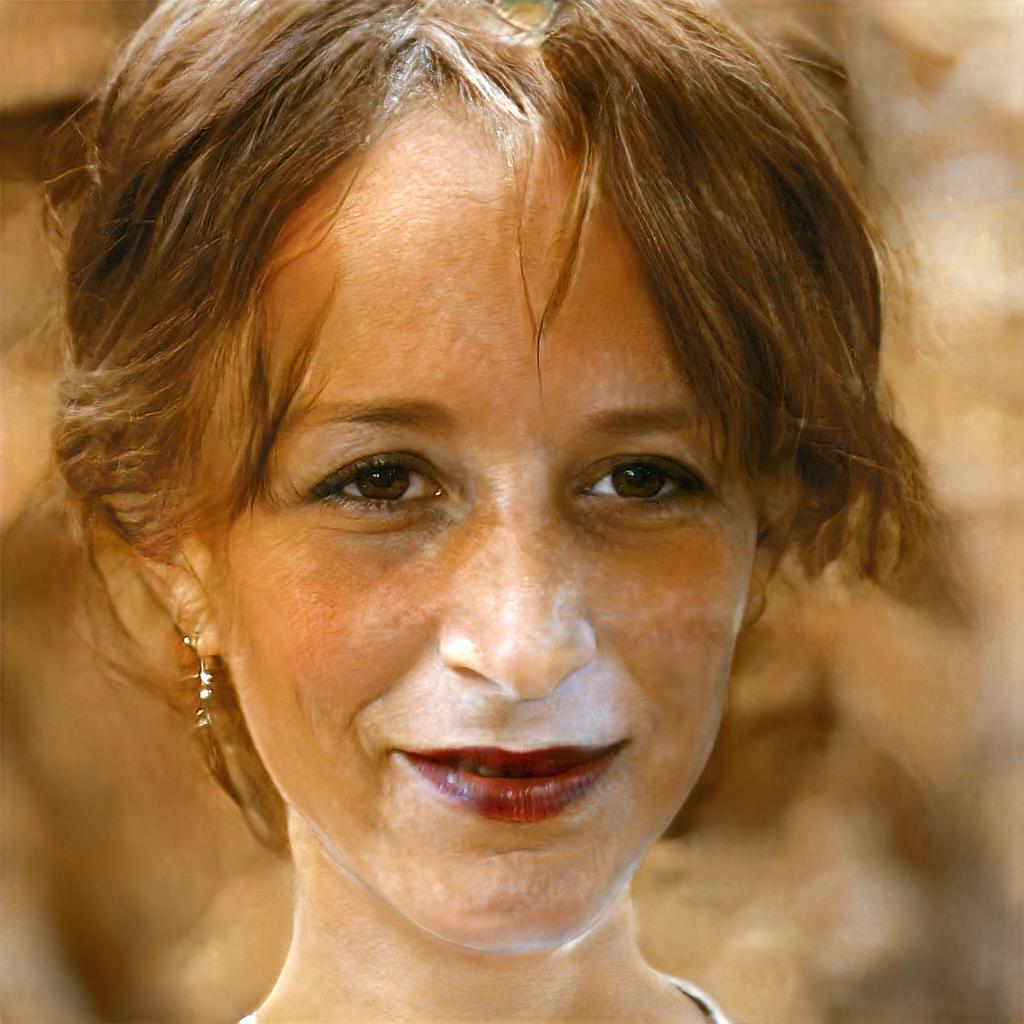} &
		\includegraphics[width=.12\linewidth]{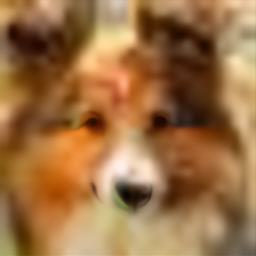} &
		\includegraphics[width=.12\linewidth]{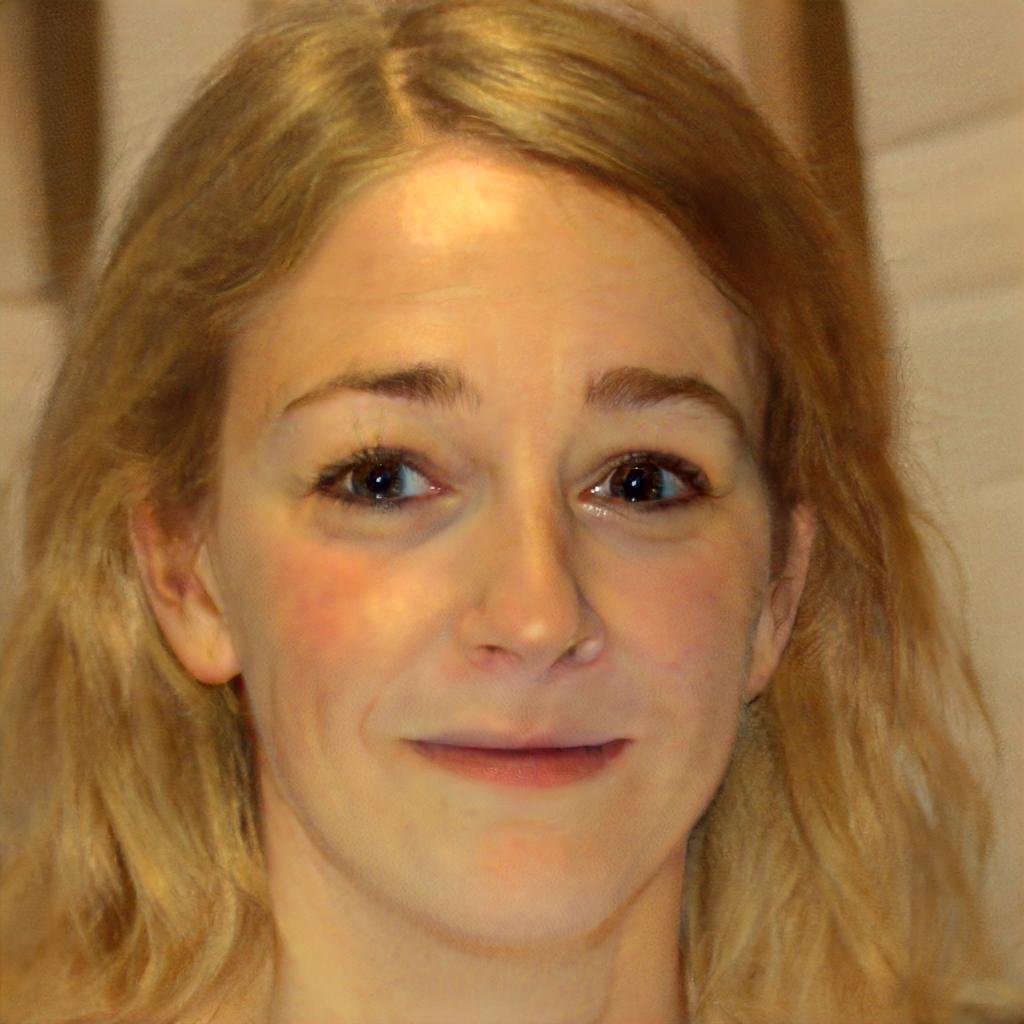}
		\\	
		\includegraphics[width=.12\linewidth]{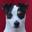} &
		\includegraphics[width=.12\linewidth]{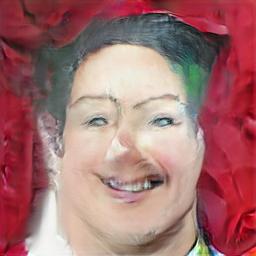} &
		\includegraphics[width=.12\linewidth]{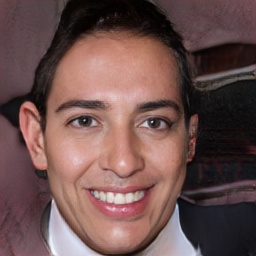} &
		\includegraphics[width=.12\linewidth]{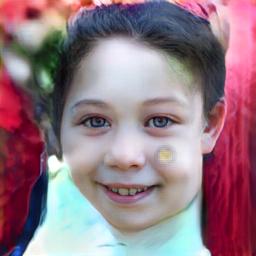} &
		\includegraphics[width=.12\linewidth]{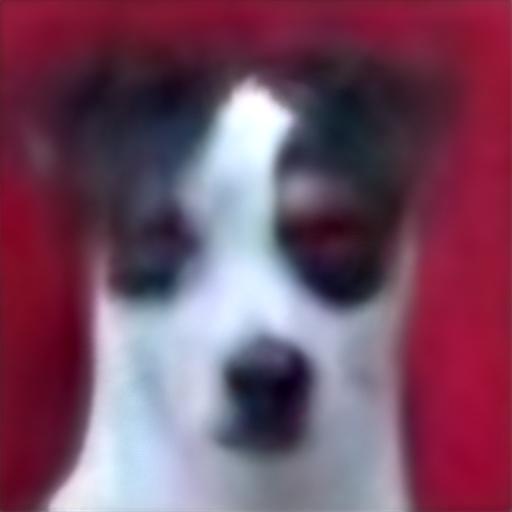} &
		\includegraphics[width=.12\linewidth]{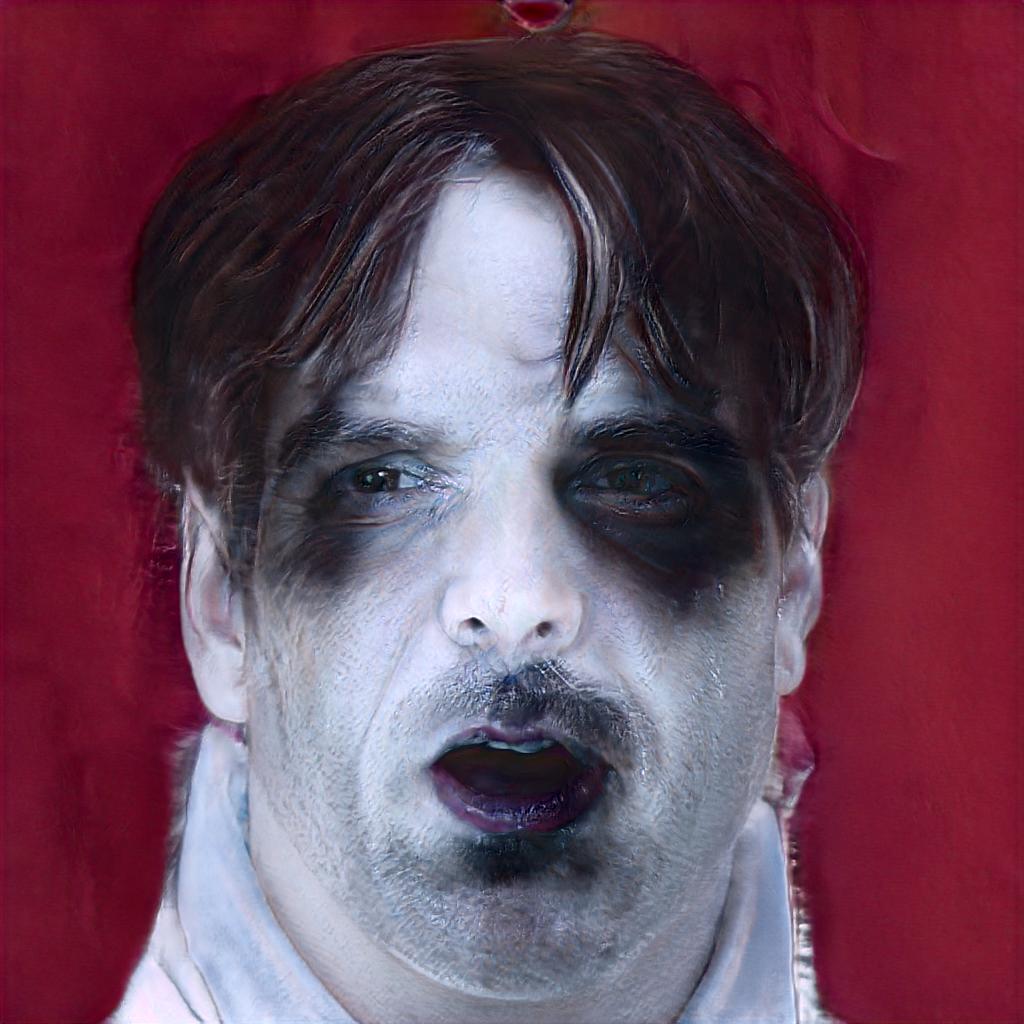} &
		\includegraphics[width=.12\linewidth]{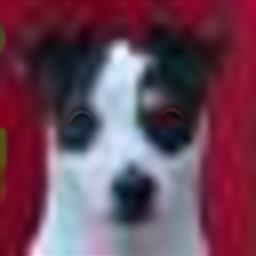} &
		\includegraphics[width=.12\linewidth]{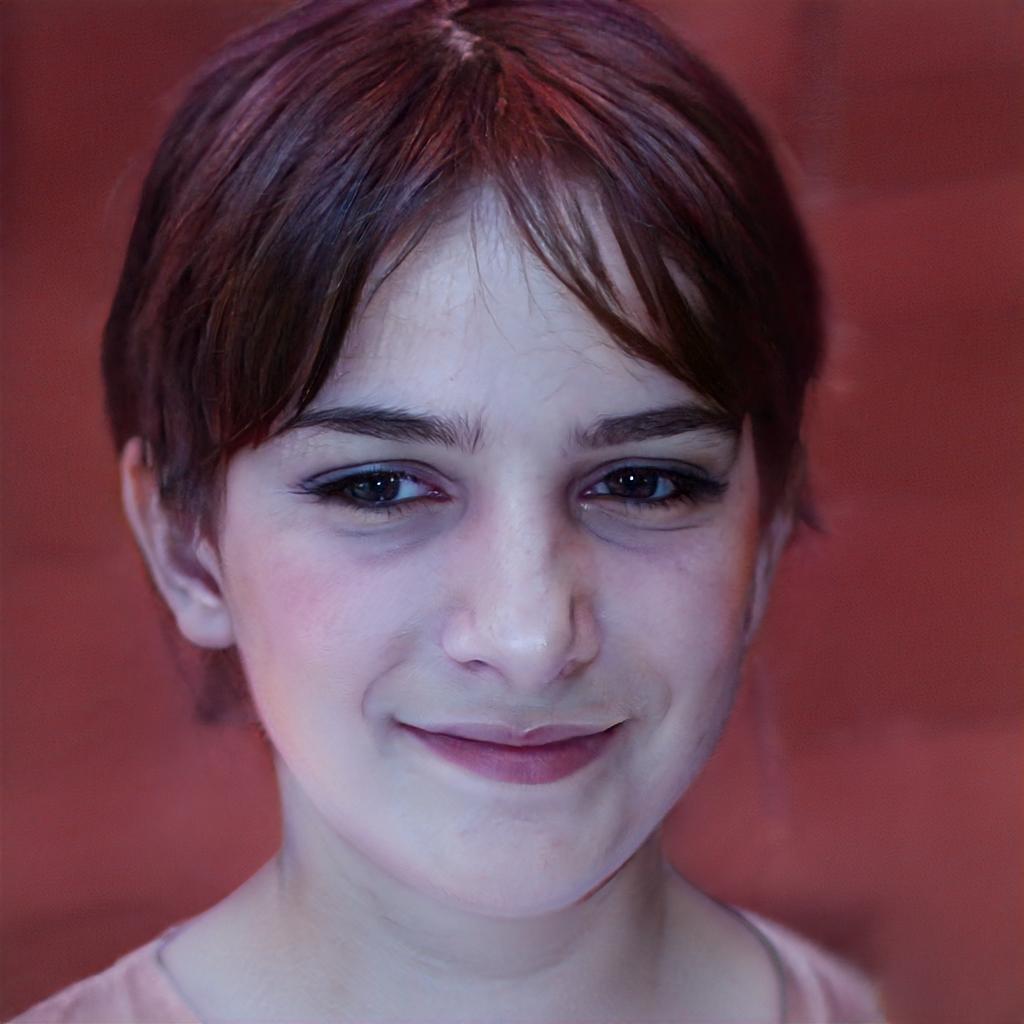}
		\\		
		LR, 32$\times$32&8$\times$&8$\times$&8$\times$&16$\times$&32$\times$&8$\times$&\textbf{Ours}, 32$\times$\\	
     	\includegraphics[width=.12\linewidth]{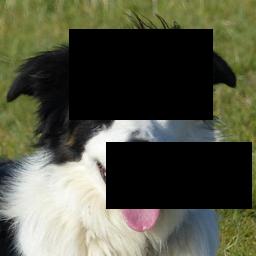} &
		\includegraphics[width=.12\linewidth]{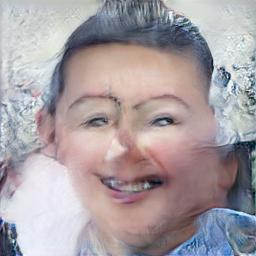} &
		\includegraphics[width=.12\linewidth]{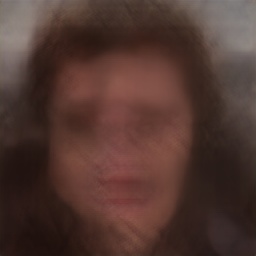} &
		\includegraphics[width=.12\linewidth]{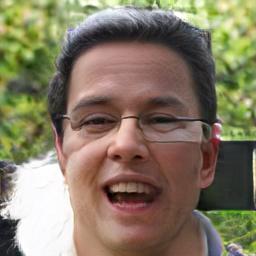} &
		\includegraphics[width=.12\linewidth]{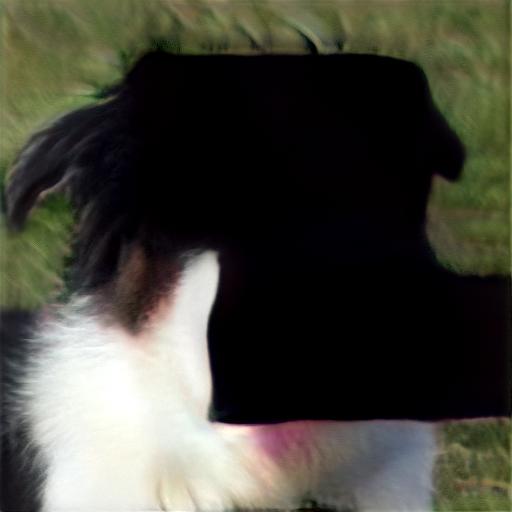} &
		\includegraphics[width=.12\linewidth]{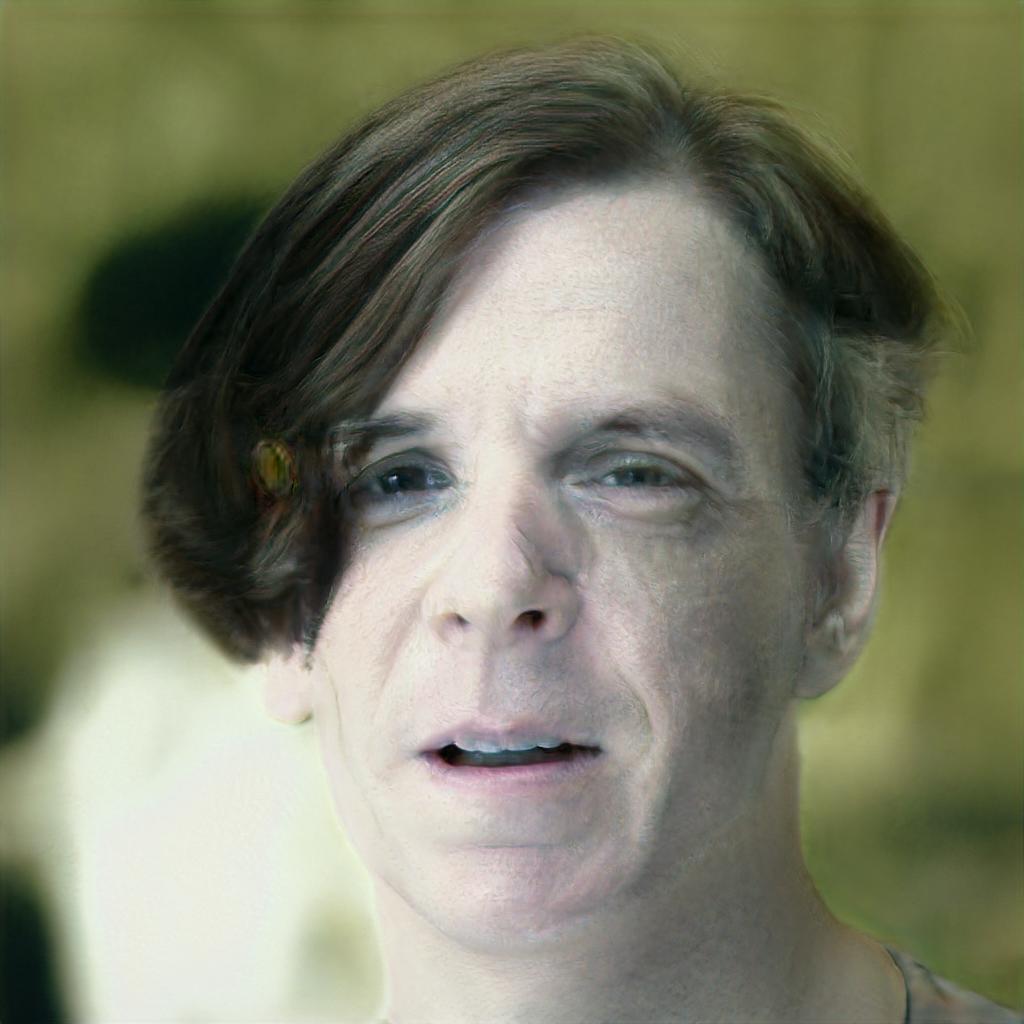} &
		\includegraphics[width=.12\linewidth]{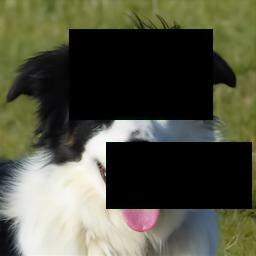} &
		\includegraphics[width=.12\linewidth]{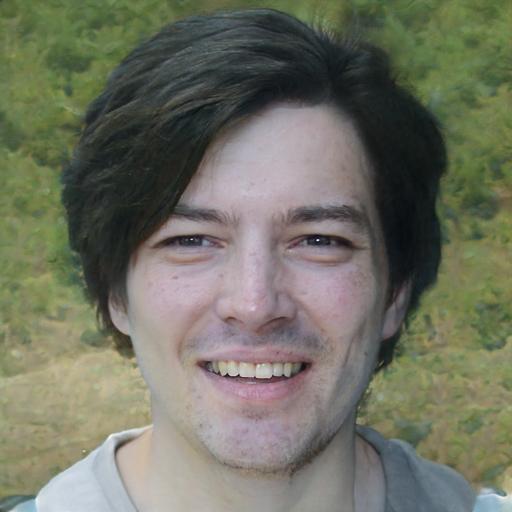}
		\\
		\includegraphics[width=.12\linewidth]{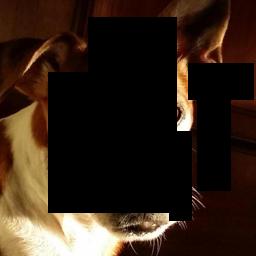} &
		\includegraphics[width=.12\linewidth]{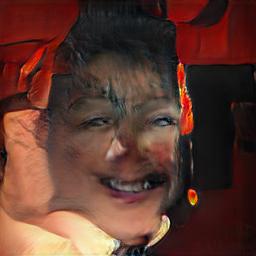} &
		\includegraphics[width=.12\linewidth]{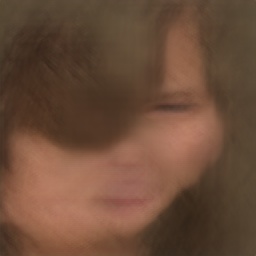} &
		\includegraphics[width=.12\linewidth]{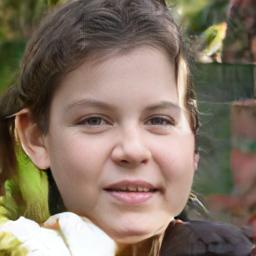} &
		\includegraphics[width=.12\linewidth]{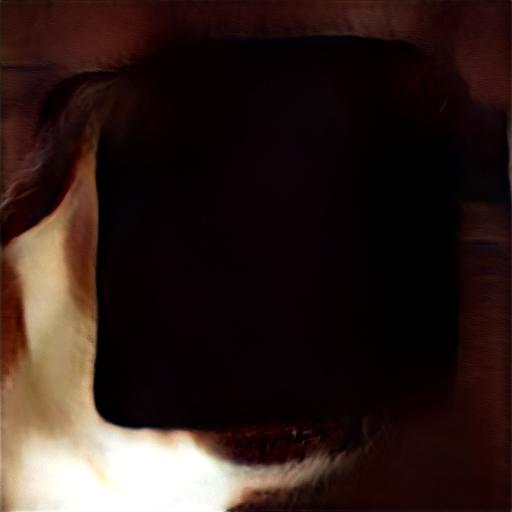} &
		\includegraphics[width=.12\linewidth]{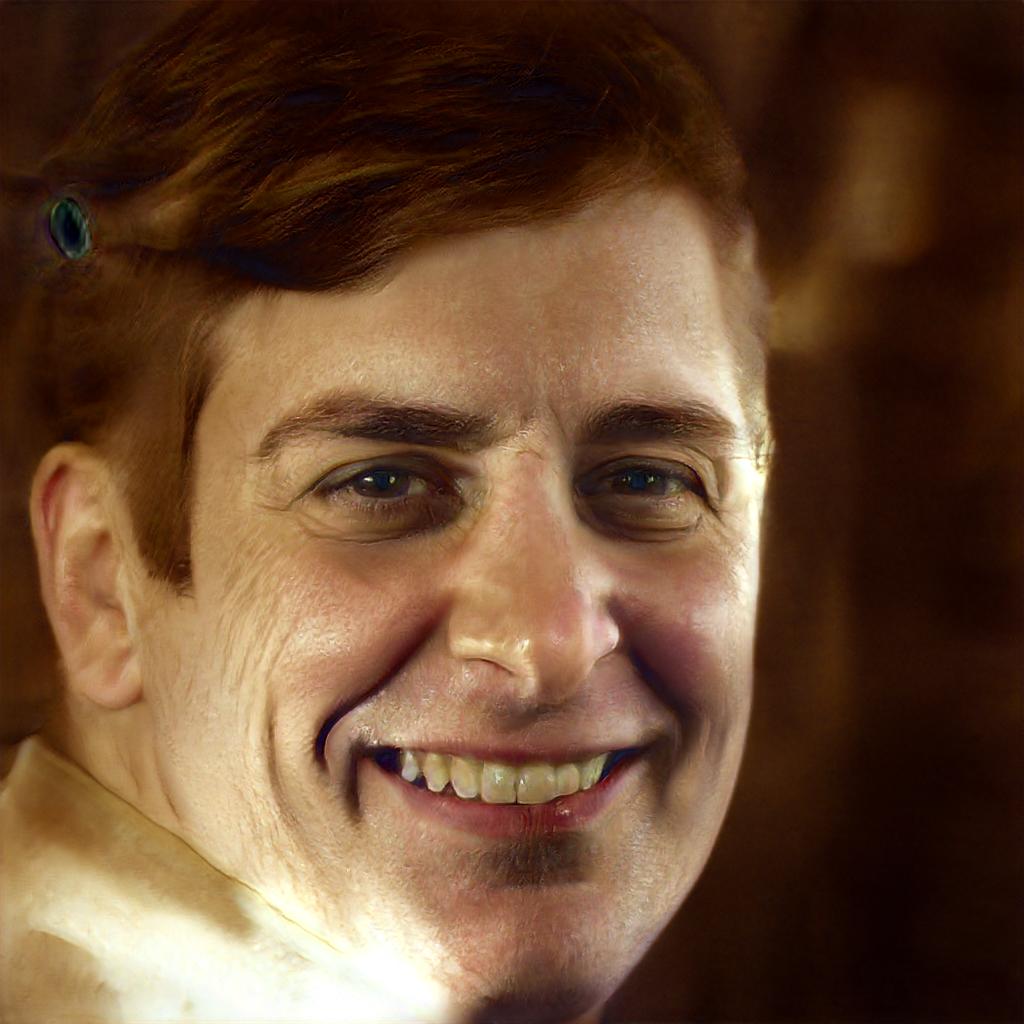} &
		\includegraphics[width=.12\linewidth]{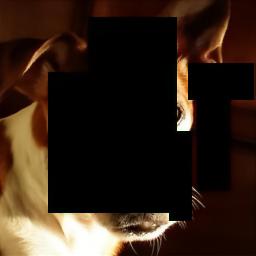} &
		\includegraphics[width=.12\linewidth]{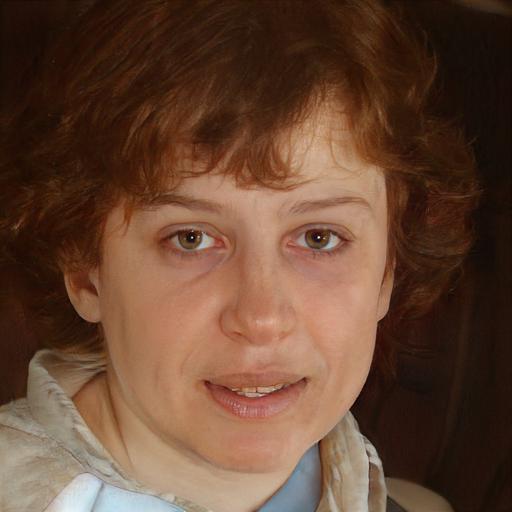}
		\\
		
		Source Image& VQ-I2I & GP-UNIT & StarGAN2 & DiFa & PULSE & DiffusionCLIP & \textbf{UniTranslator}	
	\end{tabular}
	\caption{A comparison of AFHQ-dog$\to$FFHQ in degradation scenarios.}
	\label{fig:super-resolution}
\end{figure*}

\begin{figure}[t]
	\centering
	\setlength{\abovecaptionskip}{0cm}
	\centering
	\setlength{\tabcolsep}{0.05em}
	\setlength{\fboxrule}{1pt}
	\setlength{\fboxsep}{0pt}
	\begin{tabular}{cc cc}
		\includegraphics[width=0.24\linewidth]{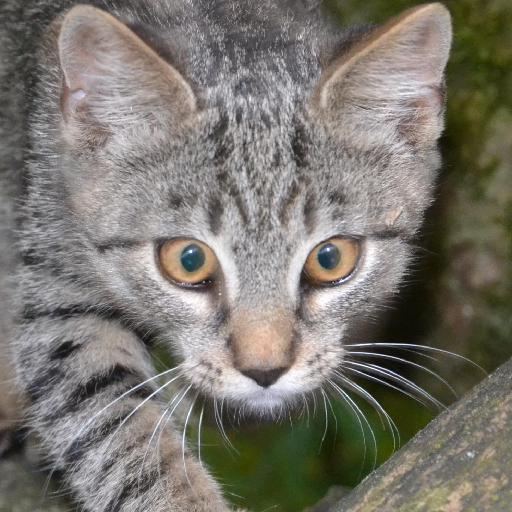} &
		\includegraphics[width=0.24\linewidth]{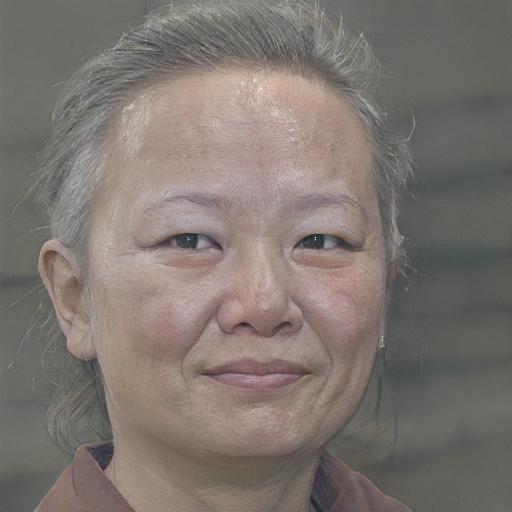} &
		
		\includegraphics[width=0.24\linewidth]{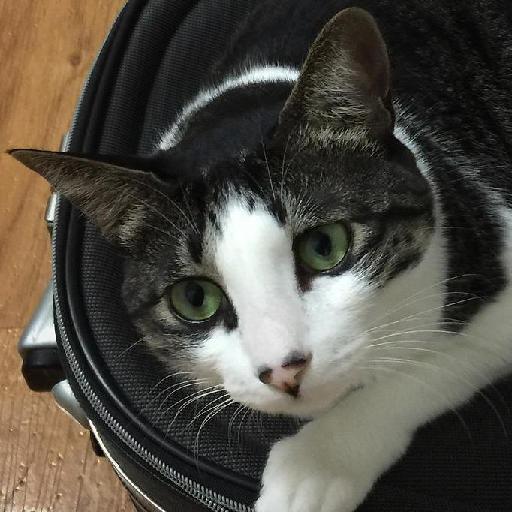} &
		\includegraphics[width=0.24\linewidth]{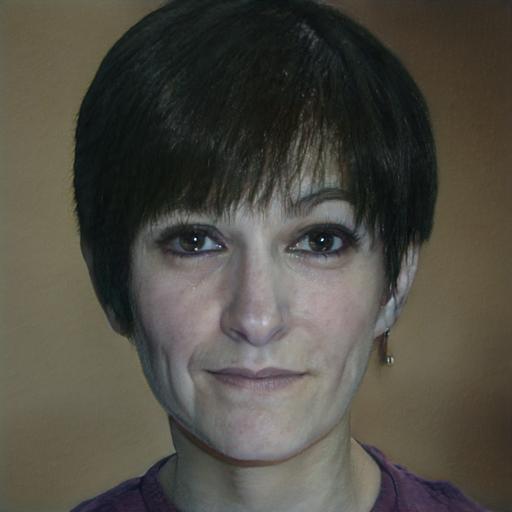}\\	
		
		\multicolumn{2}{c}{AFHQ-cat$\Rightarrow$FFHQ} &
		\multicolumn{2}{c}{AFHQ-cat$\Rightarrow$FFHQ}\\
		
		\includegraphics[width=0.24\linewidth]{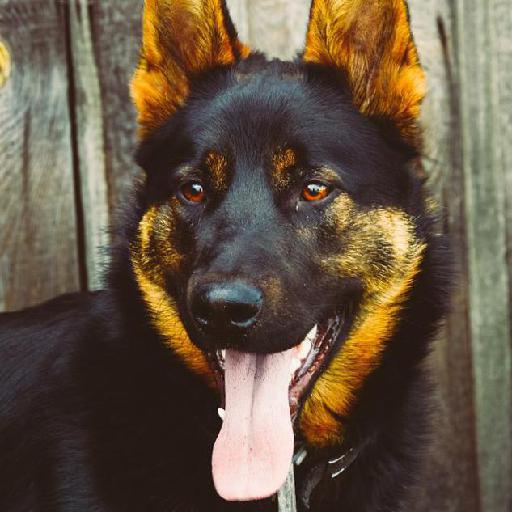} &
		\includegraphics[width=0.24\linewidth]{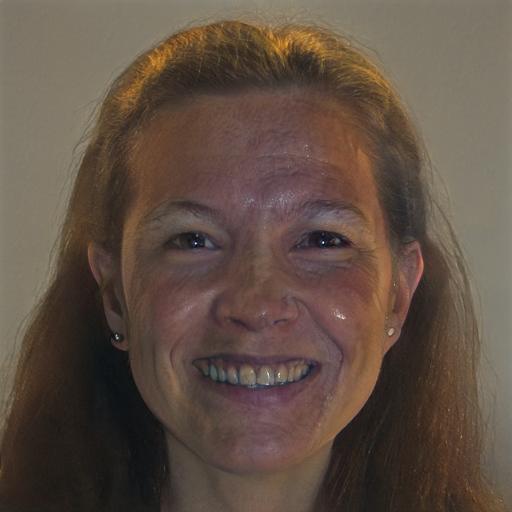} &
		
		\includegraphics[width=0.24\linewidth]{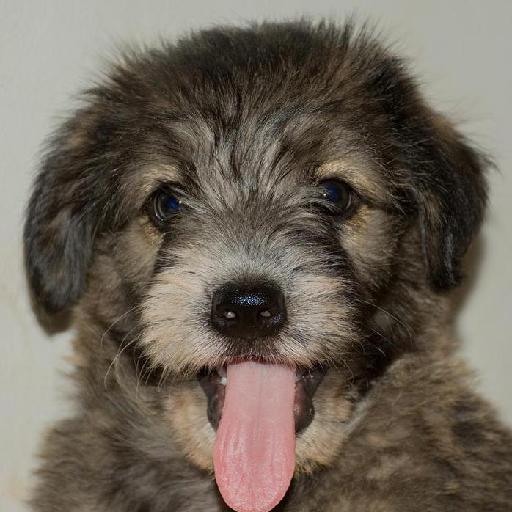} &
		\includegraphics[width=0.24\linewidth]{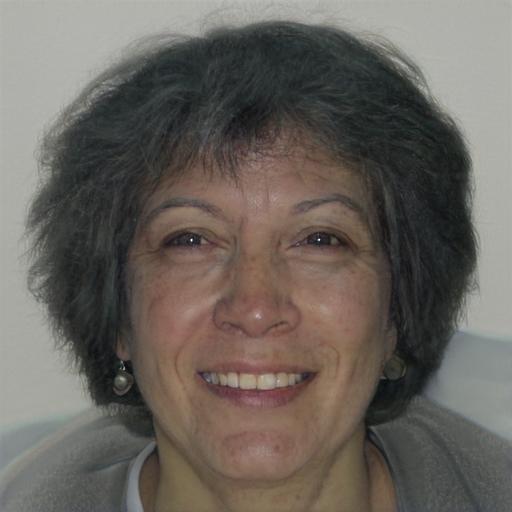}\\	
		
		\multicolumn{2}{c}{AFHQ-dog$\Rightarrow$FFHQ} &
		\multicolumn{2}{c}{AFHQ-dog$\Rightarrow$FFHQ}
		
	\end{tabular}
	\caption{Some failure cases of our method.}
	\label{fig:failure cases}
\end{figure}

\begin{figure}[t]
	\centering
	\setlength{\abovecaptionskip}{0cm}
	\centering
	\setlength{\tabcolsep}{0.05em}
	\setlength{\fboxrule}{1pt}
	\setlength{\fboxsep}{0pt}
	\begin{tabular}{c cc cc}
		\includegraphics[width=0.19\linewidth]{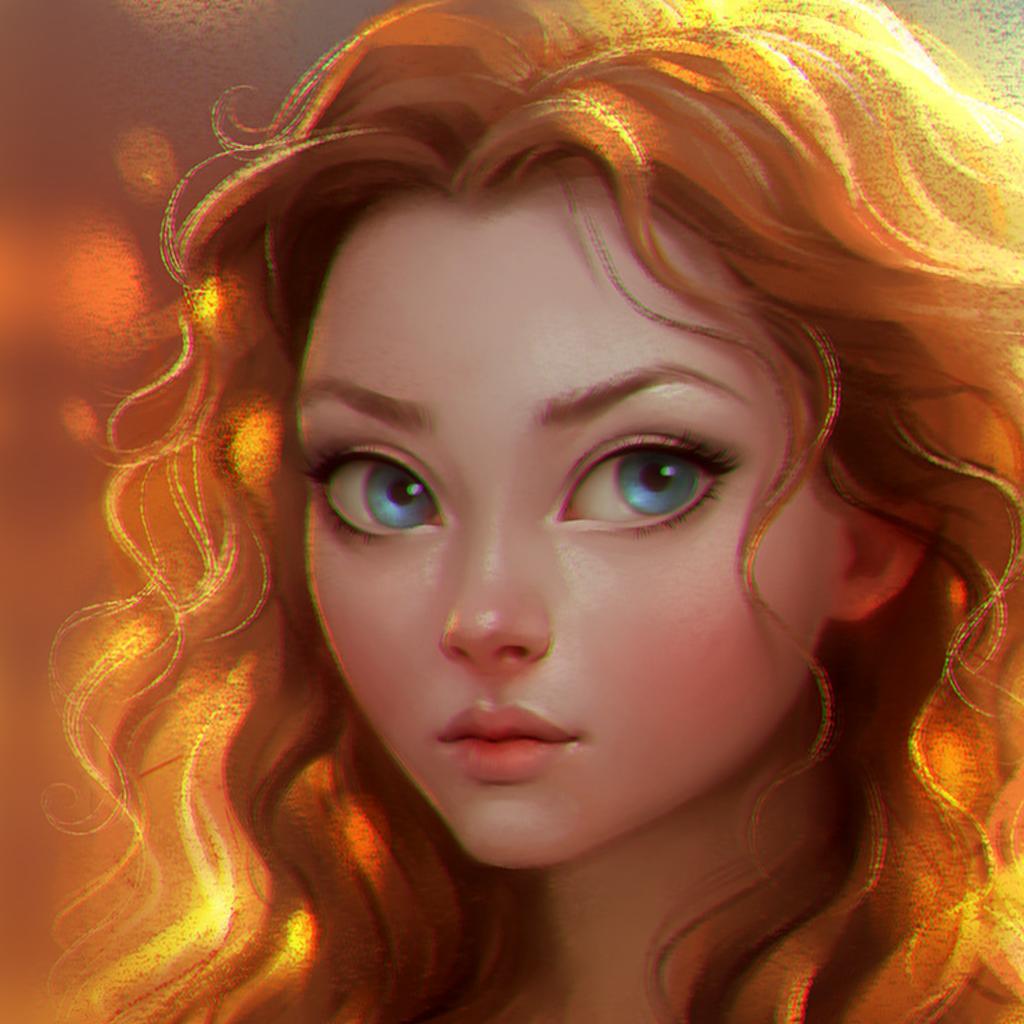} &
		\includegraphics[width=0.19\linewidth]{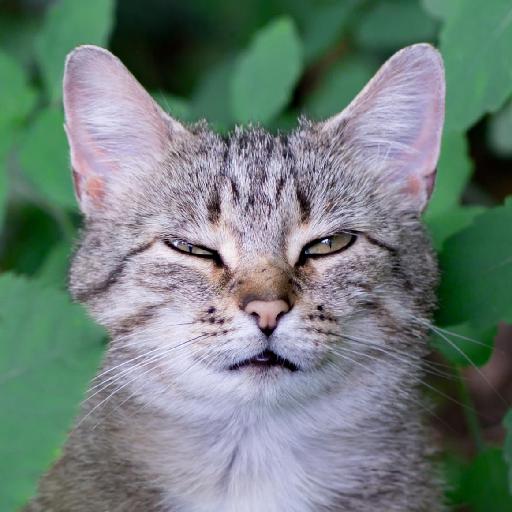} &
		\includegraphics[width=0.19\linewidth]{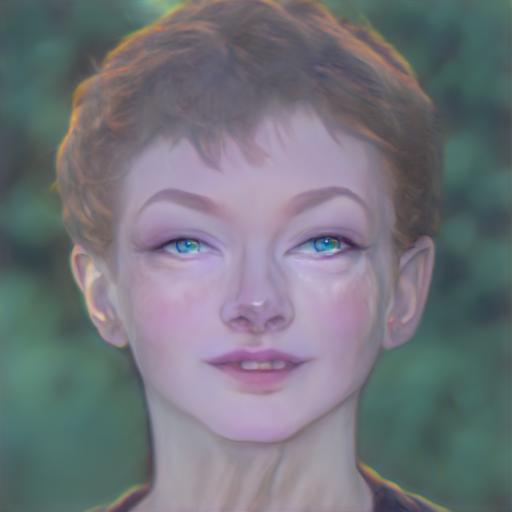} &
		
		\includegraphics[width=0.19\linewidth]{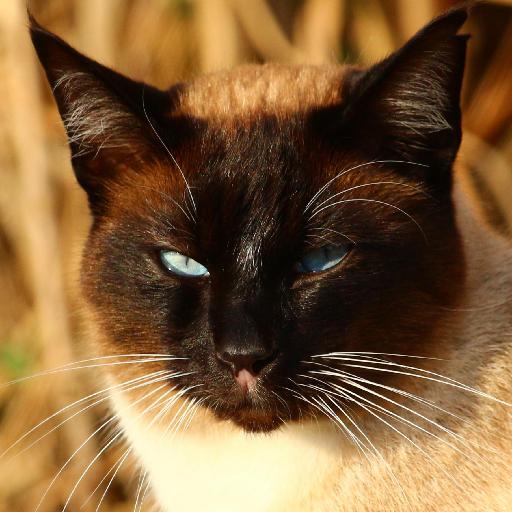} &
		\includegraphics[width=0.19\linewidth]{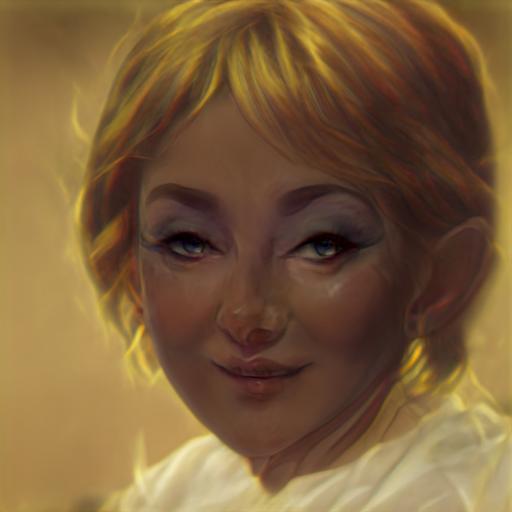}\\	
		
		\multicolumn{1}{c}{Target} &
		\multicolumn{4}{c}{AFHQ-cat$\Rightarrow$Elena Berezina's style} \\
		
		\includegraphics[width=0.19\linewidth]{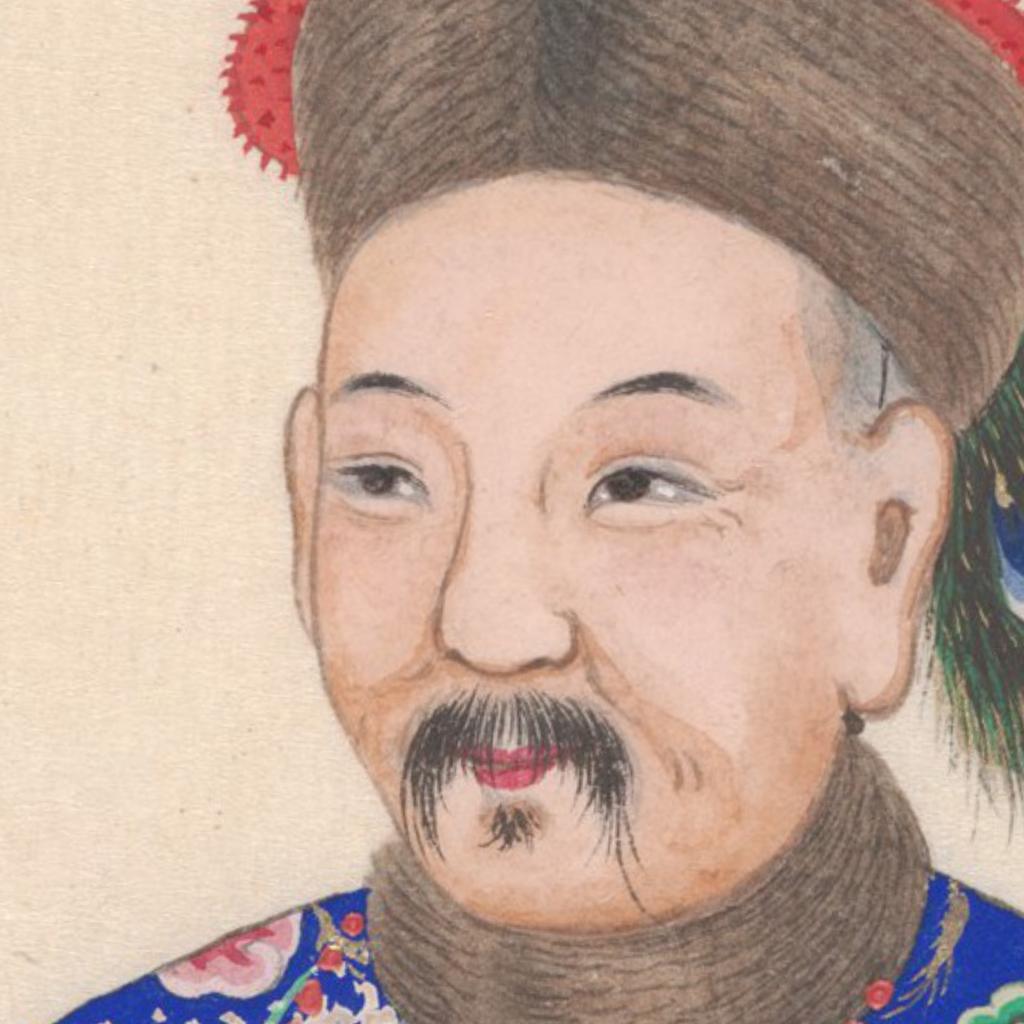} &
		\includegraphics[width=0.19\linewidth]{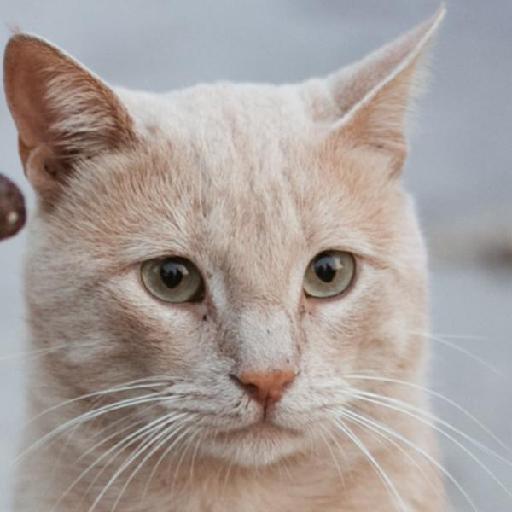} &
		\includegraphics[width=0.19\linewidth]{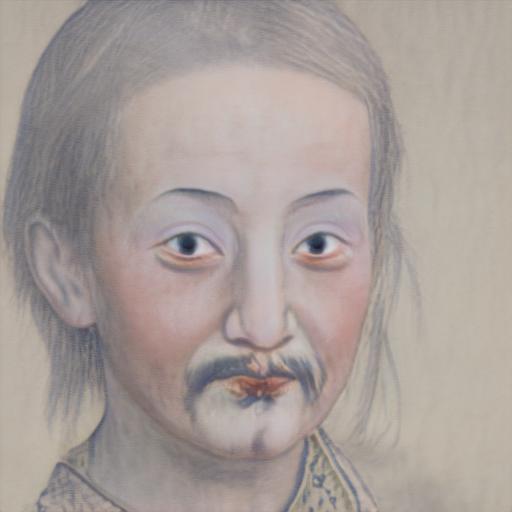} &

		\includegraphics[width=0.19\linewidth]{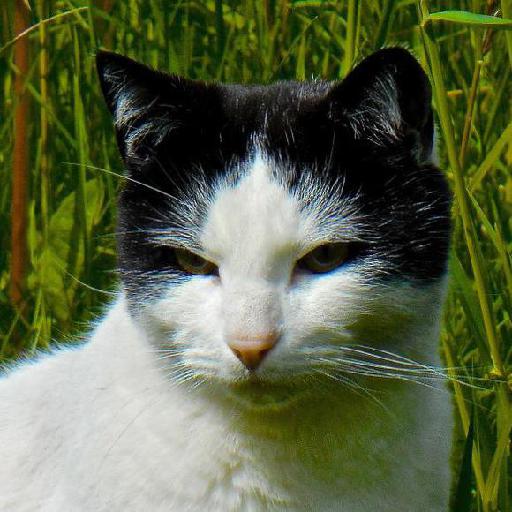} &
		\includegraphics[width=0.19\linewidth]{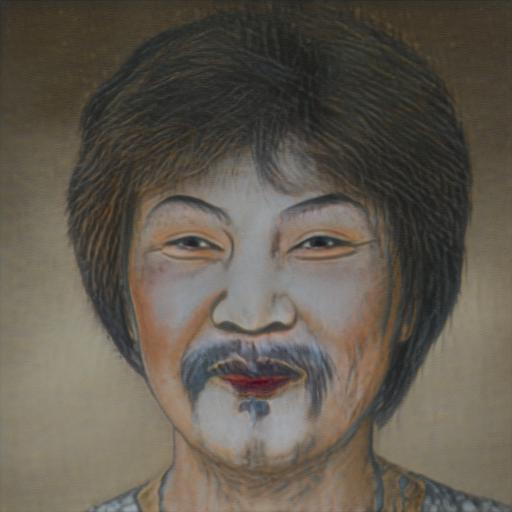}\\	

		\multicolumn{1}{c}{Target} &
		\multicolumn{4}{c}{AFHQ-cat$\Rightarrow$Chinese meticulous painting}
		
	\end{tabular}
	\caption{The StyleGAN2 generator is fine-tuned using a one-shot domain adaptation method, DiFa, to obtain the representation of the unseen target domain.}
	\label{fig:adaptation}
\end{figure}

\section{Applications} 
One salient property of UniTranslator is its ability to derive in-domain $w$ latent codes for its results, ensuring that controllability is not sacrificed for expressiveness. We perform smooth interpolation and style mixing experiments to demonstrate the quality of our latent codes. Furthermore, we conduct qualitative experiments involving style transfer and various degradation scenarios to demonstrate the robustness of our method.

\subsection{Controllable Smooth Interpolation}
The original $W$ space of StyleGAN exhibits properties of smoothness and disentanglement. As a result, interpolation between two latent codes can manifest a smooth transition in multi-level attributes. Initially, we select four source images, each from AFHQ-cat, AFHQ-dog, AFHQ-wild, and LSUN-church, considering their diverse backgrounds, poses, and color tones. Utilizing the proposed method, we transform these images into the FFHQ domain, resulting in four distinct face latent codes. Interpolation between these latent codes seamlessly transitions multi-level attributes along either direction, as illustrated in Fig.~\ref{fig:interpolation}. These results demonstrate that our approach can generate high-quality latent codes within the target domain, preserving the controllability of the $W$ space.

\subsection{Style Mixing}
We use style mixing~\cite{Kang_2021_ICCV} to validate the quality of our latent codes, as shown in Fig.~\ref{fig:stylemixing}. Initially, we translate images of a lion and a dog into the FFHQ domain and save their respective $w$ codes. Subsequently, we replicate and combine these latent codes. The first 4 layers of the synthesis network are injected with the lion's latent code, and the remaining 14 layers receive the dog's latent code, resulting in a mixed outcome (a). In this image, attributes such as pose and face shape are inherited from the lion, while the color scheme and finer details are drawn from the dog. For instance, the individual in result (a) features an angled pose, primarily derived from the lion's pose. However, the dark hair color and fair skin tone originate from the dog's dark head and white face.

Next, we reverse the latent codes of the lion and the dog to create the mixed result (b). Here, the lion influences the fine structure, and the coarse attributes are derived from the dog. The person in result (b) assumes a forward pose, adapted from the dog's pose, while their yellow skin tone is borrowed from the lion's features. These results demonstrate the capacity of the proposed method to achieve nuanced style mixing while preserving the integrity of the target images.

\subsection{Stylization}
Stylization, as one of the significant applications of image translation, warrants special attention. For the stylization experiments, we collect real portraits from Unsplash and Pexels websites and a set of portraits of the star Taylor Swift. These portraits are then translated into the artistic style of Ilya Kuvshinov, as shown in Fig.~\ref{fig:stylization}, yielding noteworthy results. The proposed method captures intricate details, such as the complex texture of hair colors in real scenarios. The results demonstrate the capability of our method to produce high-quality stylized images, even when dealing with smaller domain gaps. 

\subsection{Robustness in Degradation Scenarios}
A translation application for real-world scenarios necessitates the ability to handle user-provided inputs. In such cases, the source domain is uncertain, and the image quality can also vary significantly. Inputs may include low-resolution or corrupted images, highlighting the importance of algorithmic robustness. Thus, we evaluate our method under two conditions: handling low-resolution inputs and corrupted inputs.

For low-resolution inputs, we experiment with 32$\times$32 LR images as inputs and generate 1024$\times$1024 HR results. PULSE can also generate results with a resolution of 1024 pixels. However, VQ-I2I, StarGAN2, GP-UNIT, and DiffusionCLIP cannot process LR images as inputs. As a result, we resize LR images to 256$\times$256 and generate 256$\times$256 HR results. DiFa, which fine-tunes the AFHQ generator, produces 512$\times$512 images. These are presented in the first two rows of Fig.~\ref{fig:super-resolution}. VQ-I2I, GP-UNIT, and StarGAN2 cannot translate well with low-resolution images. DiFa and DiffusionCLIP manage only to generate blurry images. The results of PULSE retain some undesirable patterns from the source images, such as spots on a dog's face. In contrast, our method can generate high-resolution images, translating them to the target domain while maintaining high quality.

We also simulate corrupted images by randomly adding masks to the input images. Note that during this process, the constraints of $\mathcal{L}_{decoupling}$, $\mathcal{L}_{mse}$, and $\mathcal{L}_{lpips}$ are applied only to regions that are not masked. The results of these tests are presented in the last two rows of Fig.~\ref{fig:super-resolution}. Under these challenging conditions, the proposed method performs adaptively and stably, underscoring its potential to handle diverse and unpredictable real-world inputs.

\section{Limitations}
\label{sec:limitations}
Our UniTranslator aims to transform images between visually distinct domains while maintaining domain correspondence. This translation framework can connect any real-world source domain to a chosen target domain. However, our approach is still limited by the generative capacity of StyleGAN2. Illustrative failure cases are depicted in Fig.~\ref{fig:failure cases}. Certain expressions and poses, such as a dog sticking out its tongue or a cat curling up, are common in specific source domains, but their counterparts in target domains, such as humans, do not naturally assume these poses. Given that these rare expressions and poses were not sufficiently represented during the pre-training phase of StyleGAN2, generating such images remains challenging. Moving forward, our focus will pivot towards large-scale generators such as StyleGAN-XL~\cite{sauer2022stylegan} and GigaGAN~\cite{kang2023gigagan}, which demonstrate superior performance on weakly structured datasets, thereby potentially overcoming the current limitation.

The second limitation is that there may be a lack of large-scale data to train the target domain generator. Fortunately, few-shot domain adaptation methods provide a possible solution. By fine-tuning an off-the-shelf pre-trained generator using a few target samples, it is easy to get the target domain generator. As depicted in Fig.~\ref{fig:adaptation}, we first pre-adapt the FFHQ generator by DiFa~\cite{zhang2022towards} using a sample in artist Elena Berezina's style or Chinese meticulous painting style. Subsequently, UniTranslator transforms a diverse range of source images into this new target domain.

In addition, our current method cannot handle inputs with multiple subjects, as StyleGAN exhibits limitations in generating images with multiple subjects. Furthermore, while the abstract nature of the cross-domain correspondences captured by the decoupling module provides flexibility in managing the scope of source domains, it compromises interpretability to a certain degree, which poses challenges in ensuring individual object correspondences in such cases. Addressing this limitation will be part of our future work.

\section{Conclusion}
In this work, we introduce UniTranslator, a pioneering paradigm that combines the domain-neutral capabilities of CLIP with the practical generative ability of StyleGAN for universal visual domain translation. Using the cutting-edge vision-language model CLIP, we develop a decoupling module that extracts abstract and domain-agnostic semantics from CLIP representations. Furthermore, we introduce CLIP2P mapper, a non-linear mapping technique, to bridge CLIP and StyleGAN's latent spaces, effectively utilizing StyleGAN's generative priors. Extensive experimental results, both qualitative and quantitative, 
demonstrate that the proposed method performs favorably against state-of-the-art models regarding semantic correspondences and visual quality. Finally, we also demonstrate the versatility and robustness of UniTranslator through diverse applications.

\bibliographystyle{IEEEtran}
\bibliography{bibliography}
\clearpage
\appendices
\section{Choice of Evaluation Metrics}
It is crucial to assess three key aspects in cross-domain translation tasks: (1) successful translation into the target domain, (2) the realism of the translated images, ensuring they are free from distortions or artifacts, and (3) the preservation of domain correspondences between inputs and outputs. Finding metrics that comprehensively address all three aspects is challenging. To determine the most appropriate metrics for cross-domain translation, we consider Fr\'echet Inception Distance (FID)~\cite{heusel2017gans}, Inception Score (IS)~\cite{Salimans2016}, Naturalness Image Quality Evaluator (NIQE)~\cite{mittal2012making}, and LPIPS~\cite{zhang2018unreasonable}, analyzing their advantages and limitations and providing the rationale behind our choice.

\begin{figure*}
	\centering
	\includegraphics[width=\linewidth]{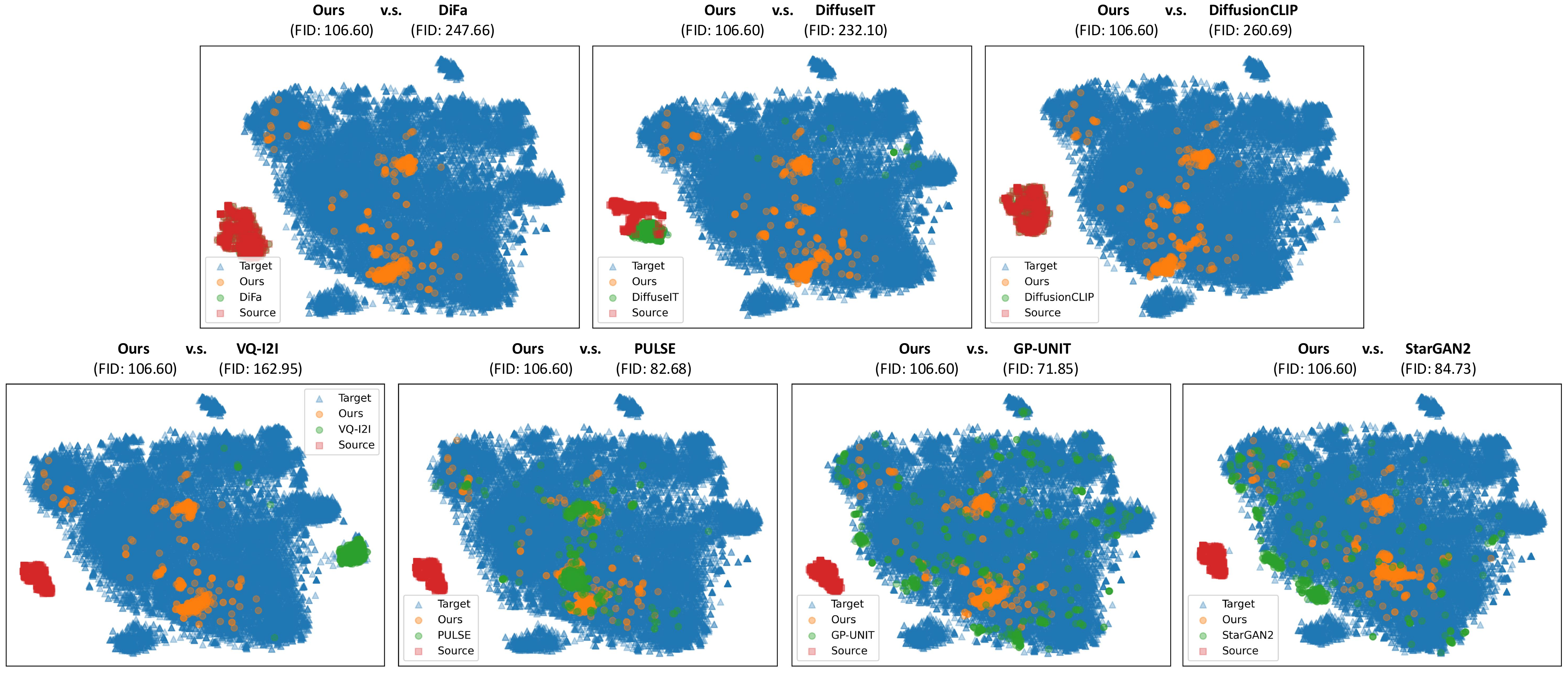}
	\caption{T-SNE visualizations of results for all the comparisons on the AFHQ-Cat (Source) to FFHQ (Target) translation, with FID scores annotated above each subplot.}	
	\label{fig:FID}
\end{figure*}

\begin{figure*}
	\centering
	\includegraphics[width=0.9\linewidth]{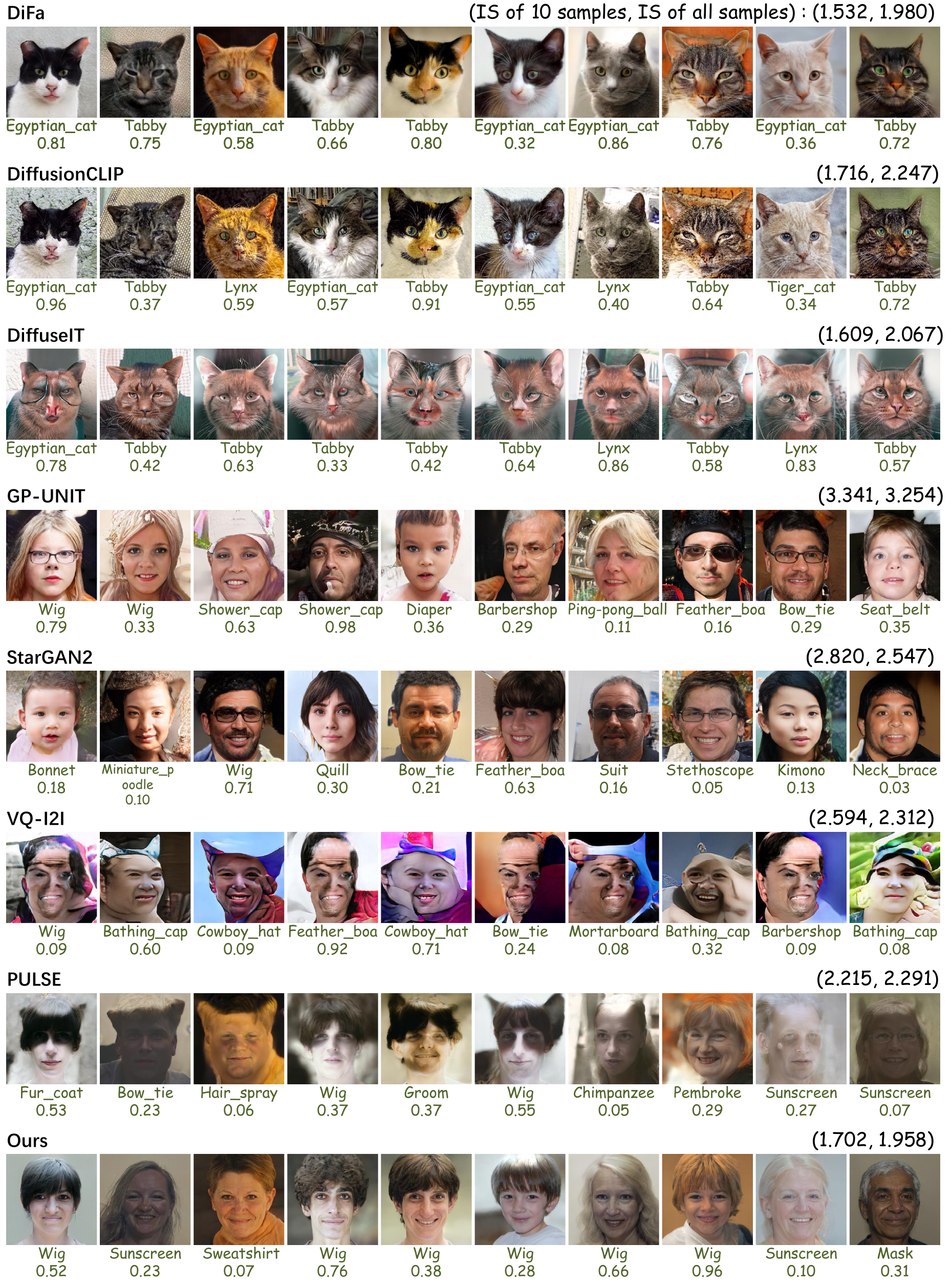}
	\caption{Illustration of representative samples for each method to analyze the IS metric. In each row, the scores for 10 samples and for all samples are annotated in the top-right corner, with the maximum logit value and its corresponding label noted below each sample.}	
	\label{fig:IS}
\end{figure*}

\noindent \textbf{Fr\'echet Inception Distance.} First, we evaluate whether FID is suitable for this task. FID measures the similarity between the feature distributions of the translated results and the target domain, with lower values indicating closer alignment and better quality. We analyze T-SNE visualizations of features generated by InceptionNet~\cite{Szegedy2015}, pretrained on ImageNet~\cite{Russakovsky2015}, showing the source features (red), target features (blue), our method's transformed features (orange), and those of each competitor (green) in the AFHQ-cat to FFHQ translation, as illustrated in Fig.~\ref{fig:FID}.

In these comparisons, methods like Difa, DiffuseIT, and DiffusionCLIP struggle with translating between distant domains, leading to significant overlap between the translated and source distributions, and thus poor FID scores. VQ-I2I achieves some distribution migration but still fails to fully align with the target domain, resulting in suboptimal FID scores. GP-UNIT, StarGAN2, and PULSE generally align with the target domain and produce more dispersed results, leading to better FID scores. However, while our method also aligns results within the target domain, the clustering observed negatively impacts the FID score. This clustering, though, is essential for high-quality cross-domain translation. For instance, in the AFHQ-cat to FFHQ mapping, cats exhibit less individual variation compared to humans, who possess richer identity features. When mapping from a domain with less variation to one with more, ensuring cross-domain correspondences naturally leads to clustering within less varied groups, which impacts the FID score. This issue arises because FID only considers the relationship between the generated results and the target domain distribution, without accounting for the source domain, and thus conflicting with the goal of preserving cross-domain correspondences.

Moreover, FID focuses on distribution differences rather than specific image quality, meaning that results with distortions or artifacts can still achieve a good FID score. This is evident in the qualitative results of Fig. 5 and Fig. 6 of the main paper, where GP-UNIT produces severely distorted images yet outperforms our method in terms of FID.

In summary, while FID can assess alignment with the target domain, it conflicts with the goal of preserving cross-domain correspondences and does not accurately evaluate image quality, making it unsuitable for cross-domain translation tasks.

\noindent \textbf{Inception Score.} Next, we evaluate whether IS is a suitable metric. IS assesses image quality based on two factors: diversity and quality. Diversity is measured by the uniformity of predicted labels across generated images, while quality is reflected by the distinctness of each image’s classification. Higher IS values suggest better quality. Unlike FID, IS cannot be visualized using T-SNE, so we sample ten representative results for each competitor to ensure their IS scores closely reflect overall performance. These visualizations, shown in Fig.~\ref{fig:IS}, include the IS of 10 samples and all samples, with the maximum logit value and label annotated below each sample.

In these comparisons, methods like DiFa, DiffuseIT, and DiffusionCLIP fail to translate successfully, as their generated images are still classified as different categories of cats. However, these methods achieve higher IS scores than ours, despite not successfully translating to the target domain. This indicates that IS does not accurately reflect the success of the translation.

For VQ-I2I, StarGAN2, GP-UNIT, and PULSE, higher IS scores are observed. Due to the absence of a ``face'' label in ImageNet, images from these methods are often misclassified into categories related to human accessories, such as ``Neck\_brace'' and ``Wig'', or unrelated ones like ``Barbershop''. This apparent diversity inflates their IS scores, despite the presence of significant artifacts and distortions. Our results are mostly misclassified as ``Wig'' because the generated human hair resembles animal fur, establishing cross-domain correspondences. However, these results achieve lower IS due to reduced category diversity, despite their realism and lack of artifacts --- qualities not captured by the IS metric.

In conclusion, IS does not effectively evaluate key aspects of universal domain translation. It fails to reflect translation success, does not consider the preservation of cross-domain correspondences, and lacks sensitivity to distortions or artifacts, especially when target categories extend beyond ImageNet. 

\noindent \textbf{NIQE and LPIPS.} Finally, we clarify our choice of NIQE and LPIPS for quantitatively evaluating all competing methods. Unlike FID and IS, NIQE accounts for distortions and artifacts that affect image quality. Our method optimizes the latent code within the target domain's latent space, ensuring that the generated images align well with this domain, typically leading to successful translations. If a translation fails, it indicates that the latent code has deviated from the StyleGAN target manifold, increasing the likelihood of artifacts or distortions, which NIQE can detect. Thus, NIQE provides an indirect but effective measure of translation success for our method.

To assess cross-domain correspondences, we use LPIPS to compare the deep features of input-output image pairs and evaluate their similarity. Strong correspondences are expected to exhibit similar perceptual characteristics.

Despite our top performance in NIQE, which demonstrates our ability to produce high-quality, realistic images and successful translations, we acknowledge that NIQE may not directly reflect translation success for other methods. This issue can impact the evaluation of performance among comparison methods (excluding ours), as methods that generate sufficiently realistic images (\eg, DiffusionCLIP) might achieve better NIQE scores than methods that successfully translate but exhibit distortions and artifacts (\eg, StarGAN2 and GP-UNIT). Furthermore, while our method excels in LPIPS, indicating strong preservation of feature-level correspondences, we acknowledge that LPIPS does not capture pixel-level details. To supplement this, we include a user study and CLIP similarity assessment to further strengthen our evaluations, with details provided in the supplementary materials. Nonetheless, we believe that developing more suitable evaluation metrics in the future will be crucial for advancing cross-domain image translation research.
\begin{table*}[t]
	\setlength{\abovecaptionskip}{0cm}
	\caption{Quantitative comparison of UniTranslator with state-of-the-art methods in terms of CLIP similarity. The best results are highlighted in bold and underlined. A higher value indicates better performance.}
	\centering
	\resizebox{1\textwidth}{!}{
		\begin{tabular}{l||l||c|c|c|c|c|c|c|c}
			\toprule
			Type & Mapping & VQ-I2I & GP-UNIT & StarGAN2& DiFa & PULSE & DiffusionCLIP &DiffuseIT& UniTranslator \\		
			\midrule			
			Adjacent & Metfaces$\to$FFHQ & 0.521 & 0.569 & 0.610 & 0.497 & 0.572 & 0.489  &0.491 &\textbf{\underline{0.623}}\\			
			
			Adjacent & AFHQ-cat$\to$E621Faces & 0.591 & 0.768 & 0.749 & 0.635 &0.692 &0.525 &0.551& \textbf{\underline{0.776}}\\
			
			Far-off & AFHQ-cat$\to$Anime & 0.822 & \textbf{\underline{0.832}} & 0.817  & 0.592 & 0.594 &0.493 &0.550 & 0.817\\
			
			Far-off & AFHQ-cat$\to$FFHQ & 0.524 & 0.582 & 0.612 & 0.570 & 0.569& 0.514 &0.527 & \textbf{\underline{0.628}}\\
			
			Far-off & AFHQ-dog$\to$FFHQ & 0.534 & 0.583 & 0.614 & 0.563 &0.559 &0.487 &0.520 & \textbf{\underline{0.636}}\\
			
			Far-off & AFHQ-wild$\to$FFHQ & 0.532 & 0.585 & 0.612 & 0.542 &0.566 &0.488 &0.503 & \textbf{\underline{0.624}}\\
			
			Intensely far-off& LSUN-church$\to$FFHQ & 0.524 & 0.373 & 0.601 & 0.464 & 0.575 &0.487 &0.409 & \textbf{\underline{0.624}}\\
			
			Intensely far-off& AFHQ-cat$\to$LSUN-church & 0.541 & 0.592 & 0.583 & 0.440 & 0.520 & 0.424 &0.447 & \textbf{\underline{0.593}} \\
			\midrule
			/&Average&0.574&0.611&0.650&0.538&0.581&0.488 &0.500& \textbf{\underline{0.665}}\\			
			\bottomrule
	\end{tabular}}
	\label{tab:comparsion_similarity}
\end{table*}

\section{User Study}
\begin{figure}[h]
	\centering
	\includegraphics[width=\linewidth]{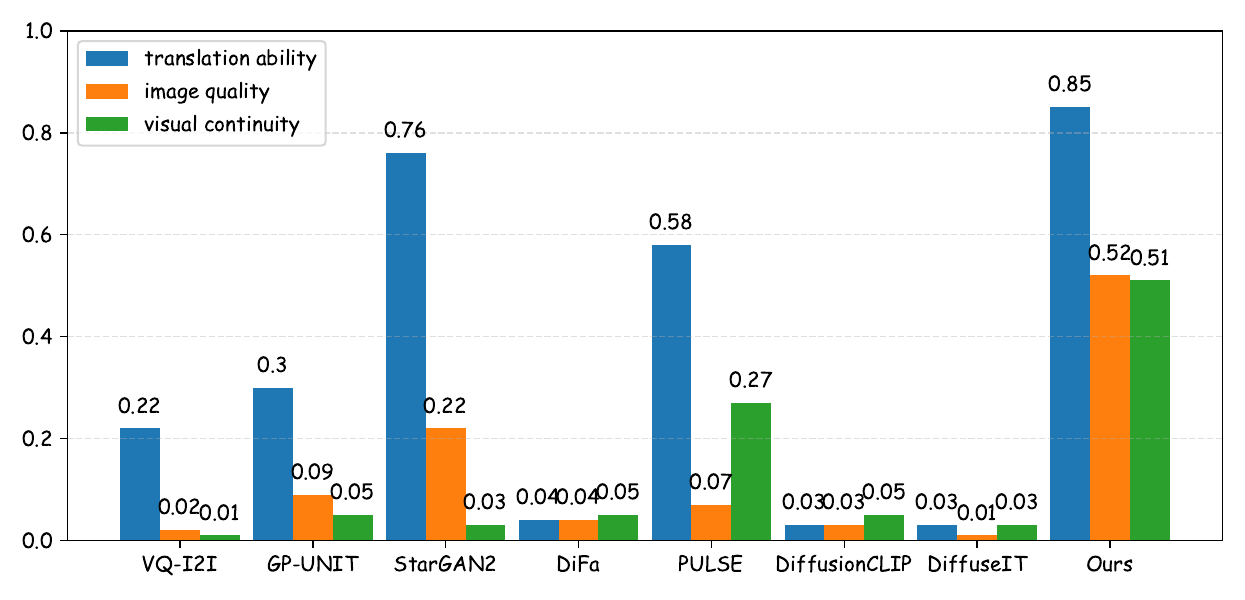}
	\caption{The user study results are analyzed based on three aspects: translation ability, image quality, and visual continuity. The percentage of preferred votes for each method is presented. It should be noted that the question regarding the translation ability is a multiple-choice question, while the other two questions are single-choice questions.}	
	\label{fig:userstudy}
\end{figure}

To assess performance based on human visual perception, we conduct a user study focusing on three key aspects: translation ability, image quality, and visual continuity. In this study, we randomly select 5 sets of images for each source-target translation task, ensuring consistency with the settings used in the quantitative evaluation outlined in the main paper. Each set includes 9 images: one source image, 7 results from comparison methods, and our own generated result. The study utilizes 360 images, with the active participation of thirty subjects who provided valuable insights and feedback.

To evaluate translation ability, we individually display each group of images, excluding the source images. Participants are then instructed to select the images that best represent the target domain, with the option to choose more than one if applicable. For example, when evaluating AFHQ-cat$\to$FFHQ, participants are given a task like: ``Question 1: Please select the face image from the following options. You may choose more than one image.'' This task is designed to assess the effectiveness of the translation methods in accurately mapping an image to the desired target domain.

Regarding image quality, we exclude the source images and those not selected in Question 1 from each group. Participants are then asked to choose the image with the highest quality among the remaining options, with the instruction to select only one image. This approach is adopted because evaluating the quality of unsuccessful results would be meaningless. The participants receive the following instruction: ``Question 2: Please select an image with the highest quality from the following options.'' For each method, we record the frequency of being selected to assess its performance in terms of image quality.

The third question takes visual continuity into account. We remove images not selected in Question 1 and display the remaining images and their corresponding source images in sequential order. Participants are required to consider the high-level correspondences between the source and resulting images: ``Question 3: Please select the image that best demonstrates abstract correspondences with the reference (source) image from the options provided.''

We collect 1200 responses for each question and calculated the average results for all 40 groups. The proportions of users who chose each method are presented in Fig.~\ref{fig:userstudy}. Our method shows favorable translation ability, aligning with the conclusions from our comparative experiments discussed in the manuscript. Most participants, approximately 85\%, consider that our method successfully translates to the target domain.

While many participants also regard StarGAN2 and PULSE as having good translation ability, StarGAN2 struggles with the semantic relationship with the source domain images, and PULSE does not generate high-quality images. Only 3\%  of participants agree with the visual consistency of StarGAN2, and 7\% concur with the quality of PULSE's results. VQ-I2I, GP-UNIT, DiFa, DiffusionCLIP, and DiffuseIT are not favored by participants, with only a few selecting them for any of the three questions.

\section{Quantitative Evaluation on CLIP Similarity}
To comprehensively evaluate UniTranslator's performance, we conduct an additional quantitative experiment to compare how effectively the results generated by various competitors into the target domain. We translate images using the source domain test sets for each mapping task and input them into CLIP's image encoder to obtain a set of CLIP embeddings. Likewise, we process the corresponding target domain test sets to obtain another set of CLIP embeddings. The cosine similarity is calculated between the means of these two sets of CLIP embeddings, and the results of CLIP similarity are summarized in Table~\ref{tab:comparsion_similarity}. Our method performs well across most configurations and secures top-notch average scores in all eight tasks, demonstrating UniTranslator's effectiveness in achieving highly successful translations into the target domain.

\section{Extended Ablation Study on Loss Functions}
\noindent \textbf{$\mathcal L_{mse}$ and Color Consistency.} To evaluate the role of $\mathcal L_{mse}$ in maintaining color consistency, we conduct experiments on the AFHQ-cat (using the entire test set) to LSUN-church translation task, comparing results with and without $\mathcal L_{mse}$. Specifically, we compute per-channel color histograms for both input and generated images and the Bhattacharyya Distance~\cite{bhattacharyya1943measure} is used to quantify histogram similarity. Smaller distances indicate better color matching.

Table~\ref{tab:color} summarizes the results, showing that removing $\mathcal L_{mse}$ weakens the color matching between the input and output images. Additionally, we evaluate the impact of removing other loss functions. Interestingly, the results indicate that, aside from $\mathcal L_{mse}$, the loss term $\mathcal L_{lpips}$ also significantly contributes to maintaining color consistency.

\begin{table}[t]
	\setlength{\abovecaptionskip}{0cm}
	\caption{The role of loss functions in maintaining color consistency is evaluated using the Bhattacharyya Distance (BD), with smaller values indicating better color matching.}
	\centering
	
		\begin{tabular}{l||c|c}
			\toprule
                 \multirow{2}*{Method}&\multicolumn{2}{c}{BD $\downarrow$}\\
                 \cmidrule(lr){2-3}
                       &Bins=16&Bins=32\\
			\midrule
                \textbf{Ours}&\textbf{0.679}&\textbf{0.757}\\
                \midrule
                w/o $\mathcal L_{mse}$&0.788&0.846\\
                w/o $\mathcal L_{lpips}$&0.797&0.852\\
                w/o $\mathcal L_{cycle}$&0.768&0.829\\
                w/o $\mathcal L_{p}$&0.730&0.801\\
                w/o $\mathcal L_{decoupling}$&0.771&0.833\\            
			\bottomrule
	\end{tabular}
	\label{tab:color}
\end{table}

\begin{figure}[h]
	\centering
	\includegraphics[width=0.96\linewidth]{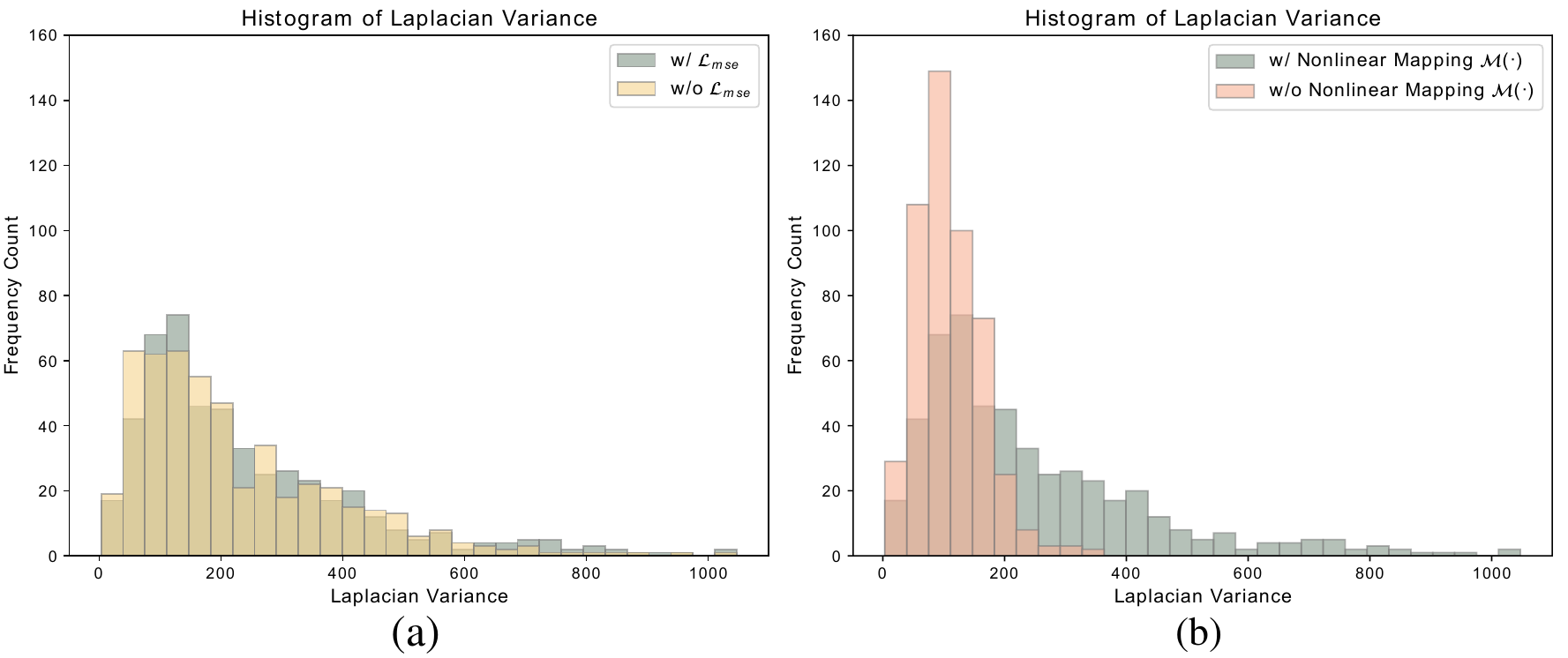}
	\caption{Evaluation of image sharpness across different variants using VoL histograms: (a) with and without $\mathcal L_{mse}$, (b) with and without our nonlinear mapping.}	
	\label{fig:histo}
\end{figure}

\noindent \textbf{$\mathcal L_{mse}$ and Image Sharpness.} $\mathcal L_{mse}$, while optimizing global similarity, carries a risk of introducing blur in the generated results. However, our method mitigates this issue by generating outputs through latent space traversal instead of the conventional feature space-to-output mapping. Consequently, the sharpness of the generated images depends on remaining within the StyleGAN target manifold, regardless of the use of $\mathcal{L}_{mse}$.

We validate this by measuring the Variation of the Laplacian (VoL)~\cite{903548}, a metric for image sharpness where higher values correspond to sharper images. Fig.~\ref{fig:histo} (a) compares VoL frequency histograms for the AFHQ-cat (entire test set) to LSUN-church translation task with $\mathcal L_{mse}$ (green) and without $\mathcal L_{mse}$ (yellow). The comparable distributions indicate that $\mathcal{L}_{mse}$ does not noticeably increase blurring in our method.

Fig.~\ref{fig:histo} (b) further compares VoL histograms when using our nonlinear mapping (green) versus removing it (red), while keeping $\mathcal L_{mse}$. Removing the nonlinear mapping results in significantly blurrier images, demonstrating that our CLIP2P mapper effectively ensures high-quality latent code exploration within the StyleGAN native space. Thus, even with $\mathcal L_{mse}$, our method consistently produces sharp, visually appealing results.

\section{More Visual Results Using UniTranslator}
\begin{figure*}
	\centering
	\setlength{\abovecaptionskip}{0cm}
	\centering
	\setlength{\tabcolsep}{0.05em}
	\setlength{\fboxrule}{1pt}
	\setlength{\fboxsep}{0pt}
	\begin{tabular}{cccccccc}		
		\multicolumn{2}{c}{Vincent van Gogh$\Rightarrow$FFHQ}&
		\multicolumn{2}{c}{Henri Matisse$\Rightarrow$FFHQ} &
		\multicolumn{2}{c}{Henri Matisse$\Rightarrow$FFHQ}&
		\multicolumn{2}{c}{Henri Matisse $\Rightarrow$ FFHQ} \\	
		\includegraphics[width=.12\linewidth]{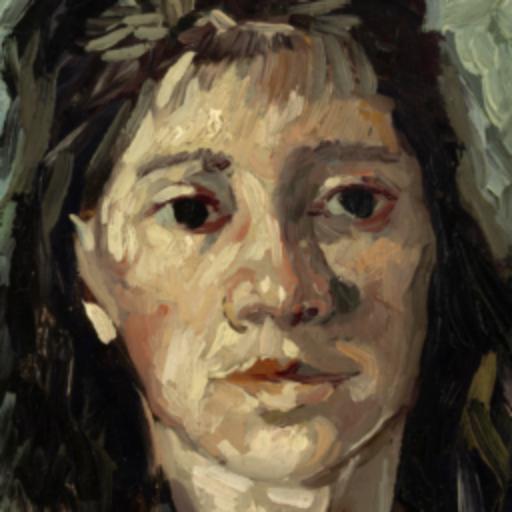}&
		\includegraphics[width=.12\linewidth]{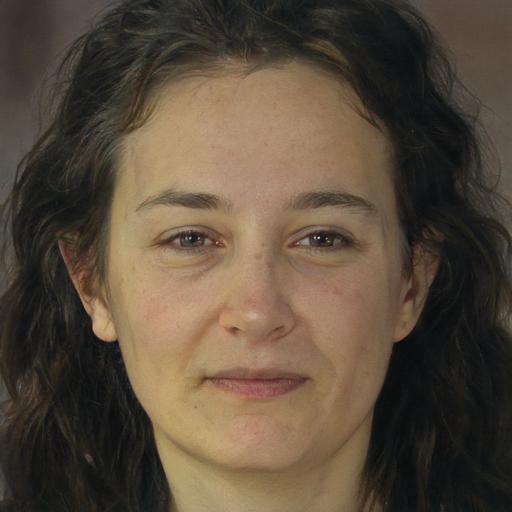}&   	
		\includegraphics[width=.12\linewidth]{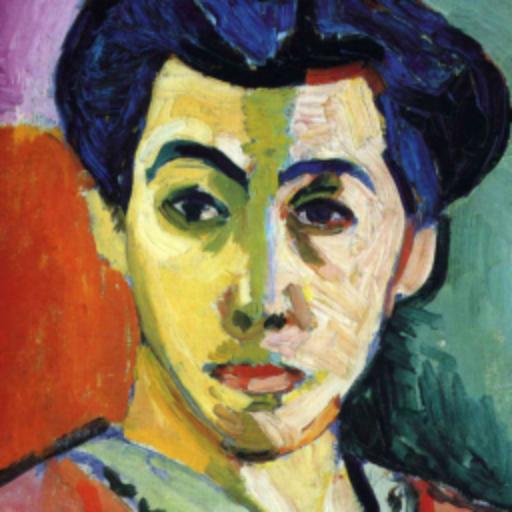}&
		\includegraphics[width=.12\linewidth]{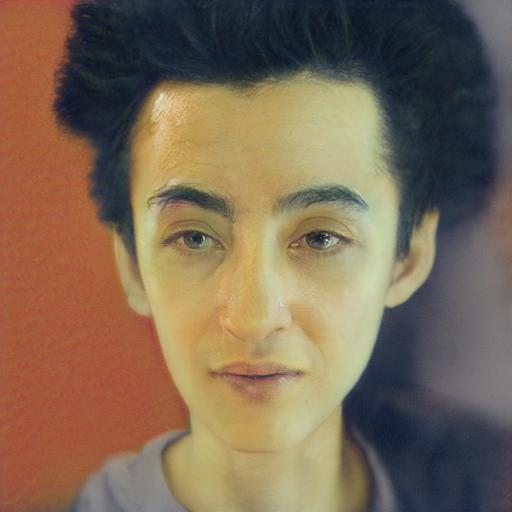}&
		\includegraphics[width=.12\linewidth]{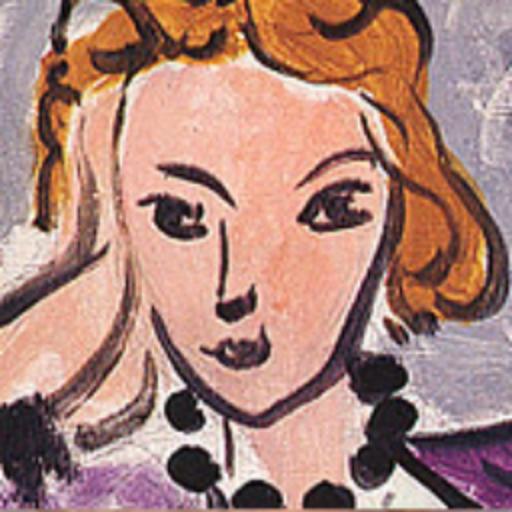}&
		\includegraphics[width=.12\linewidth]{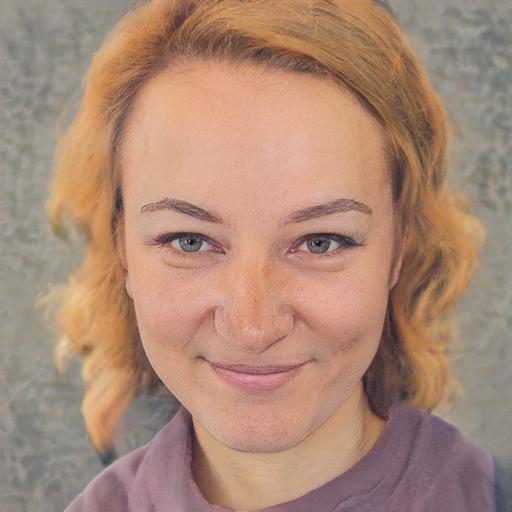}&
		\includegraphics[width=.12\linewidth]{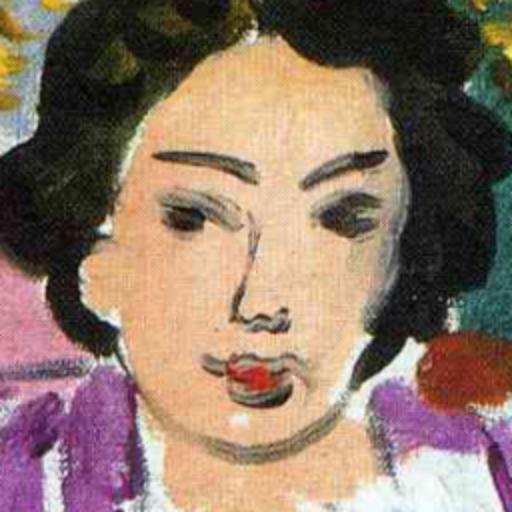}&
		\includegraphics[width=.12\linewidth]{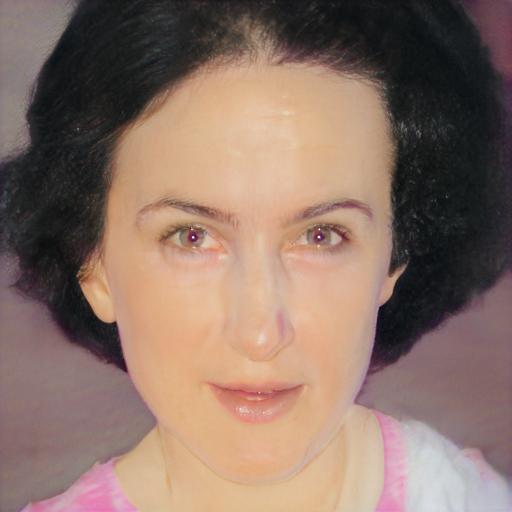}
		\\			
		\multicolumn{2}{c}{Scarlett Johansson$\Rightarrow$AFHQ-cat} &
		\multicolumn{2}{c}{Taylor Swift$\Rightarrow$AFHQ-cat} &
		\multicolumn{2}{c}{AFHQ-cat$\Rightarrow$FFHQ}&
		\multicolumn{2}{c}{AFHQ-dog$\Rightarrow$FFHQ}\\

		\includegraphics[width=.12\linewidth]{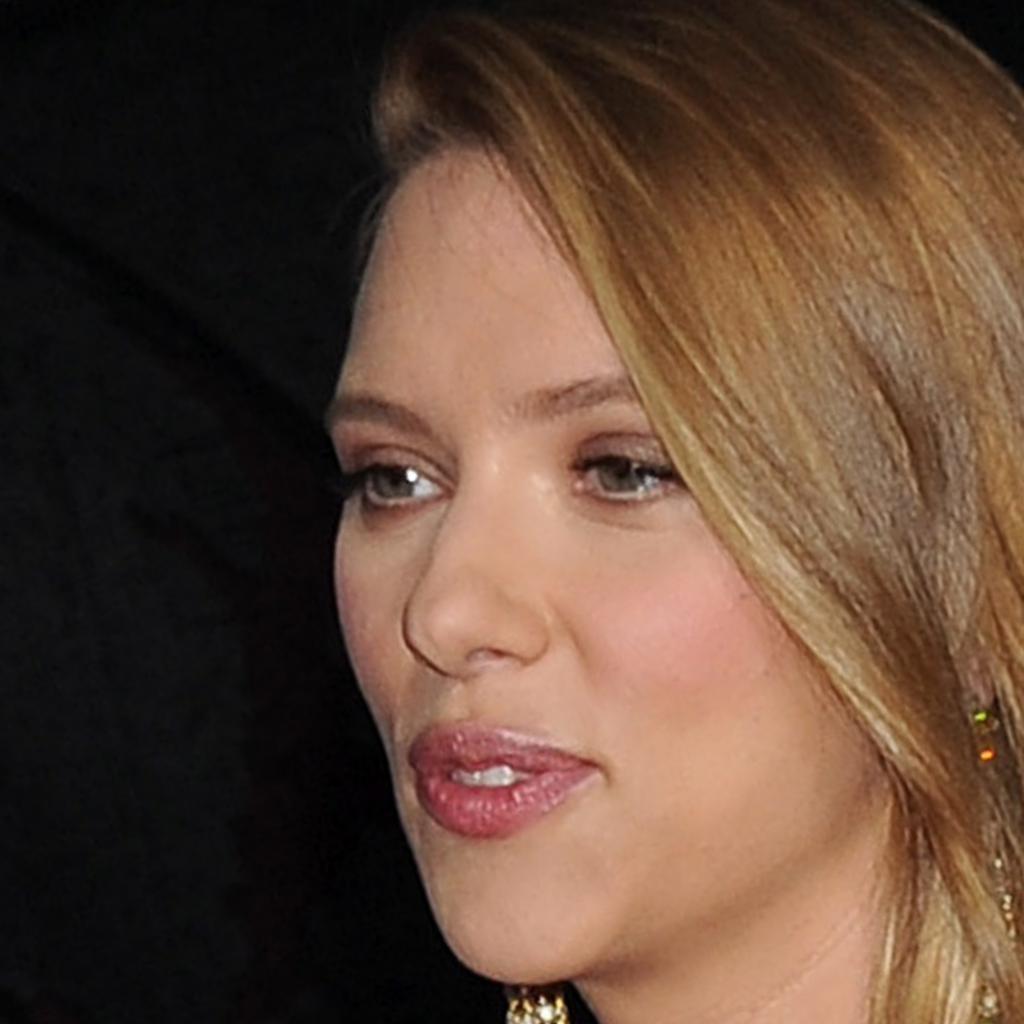}&
		\includegraphics[width=.12\linewidth]{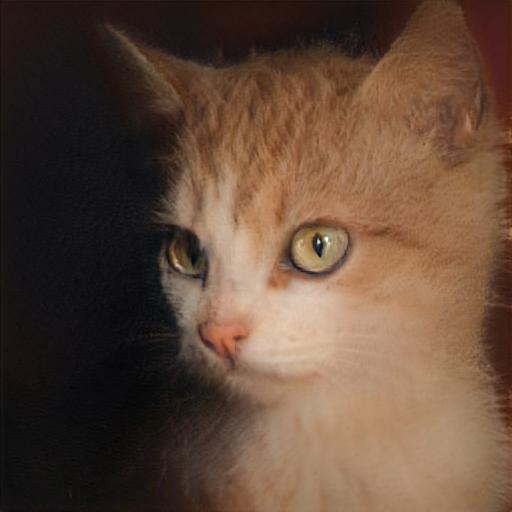}&
		\includegraphics[width=.12\linewidth]{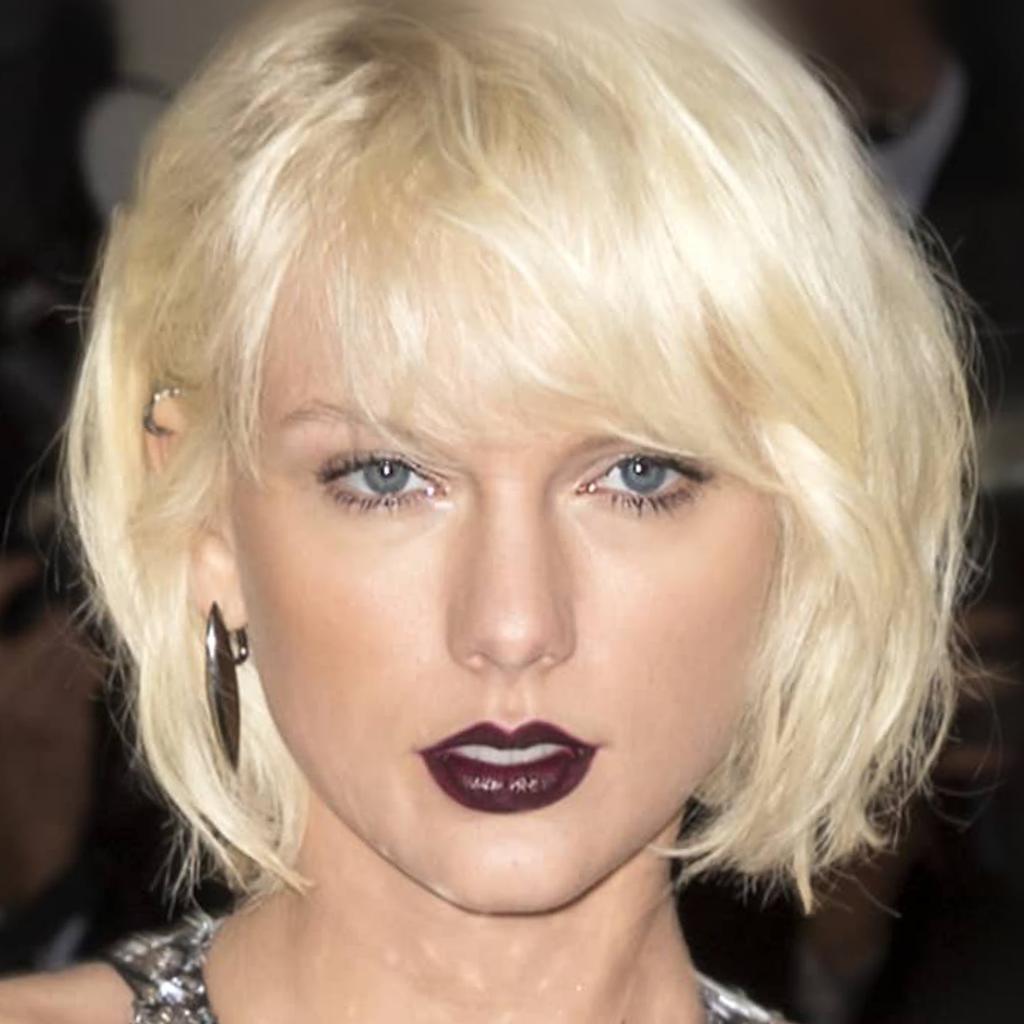}&
		\includegraphics[width=.12\linewidth]{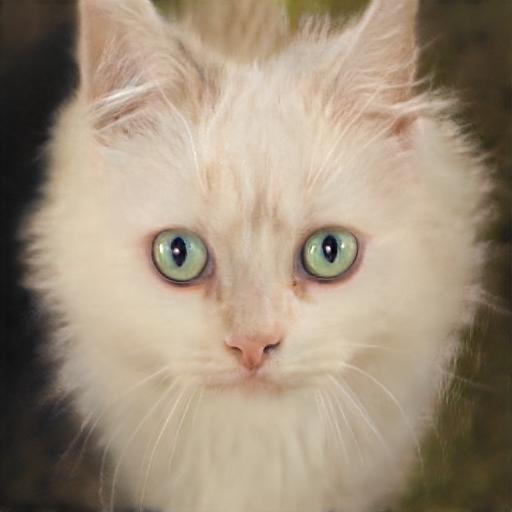}&
		\includegraphics[width=.12\linewidth]{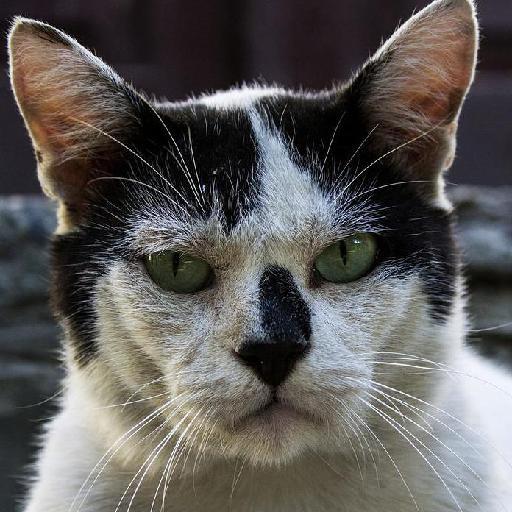}&
		\includegraphics[width=.12\linewidth]{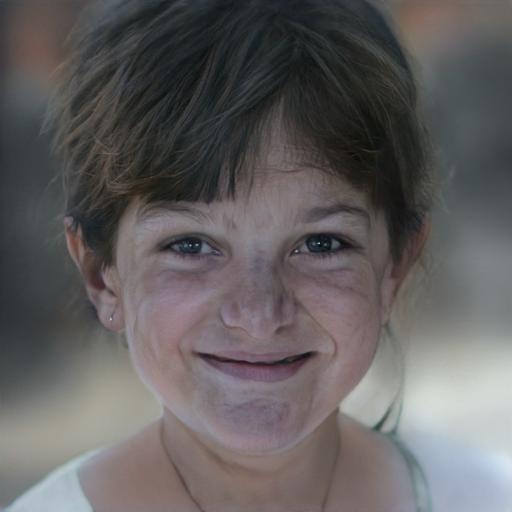}&
		\includegraphics[width=.12\linewidth]{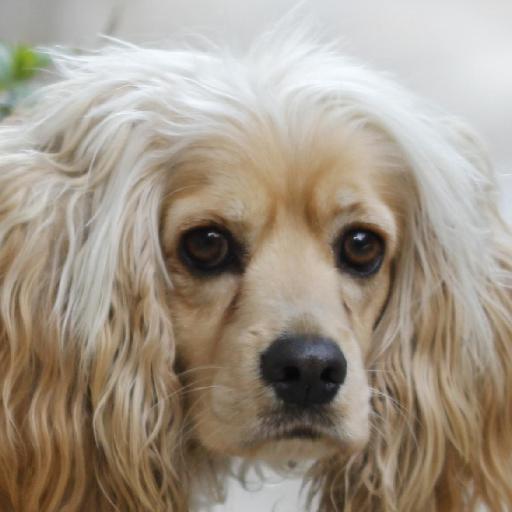}&
		\includegraphics[width=.12\linewidth]{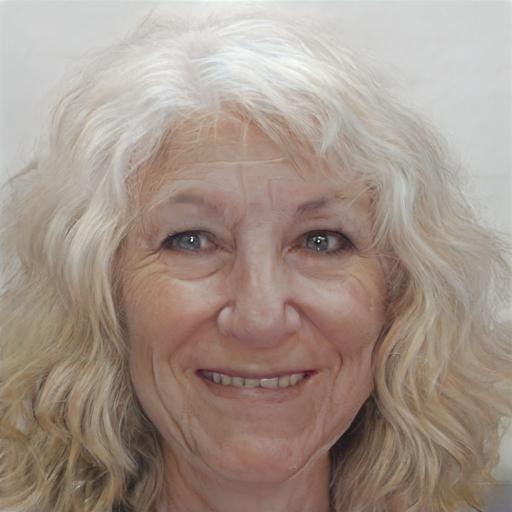}
		\\		
		\multicolumn{2}{c}{AFHQ-cat$\Rightarrow$E621Faces} &
		\multicolumn{2}{c}{AFHQ-cat$\Rightarrow$Anime} &
		\multicolumn{2}{c}{AFHQ-cat$\Rightarrow$E621Faces}&
		\multicolumn{2}{c}{AFHQ-cat$\Rightarrow$E621Faces}\\
		\includegraphics[width=.12\linewidth]{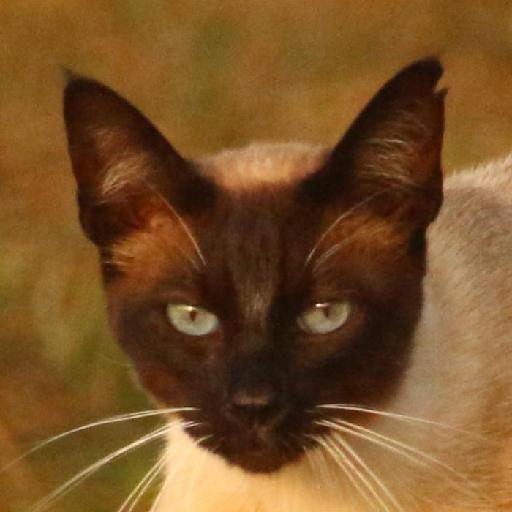}&
		\includegraphics[width=.12\linewidth]{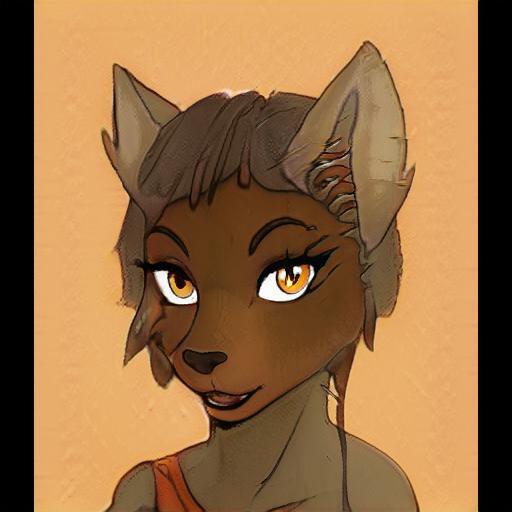}&	
		\includegraphics[width=.12\linewidth]{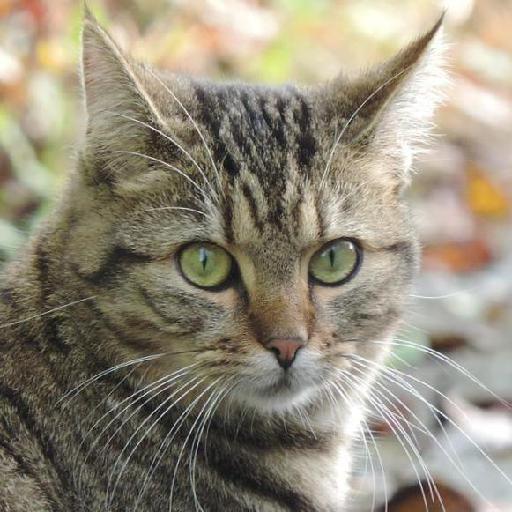}&
		\includegraphics[width=.12\linewidth]{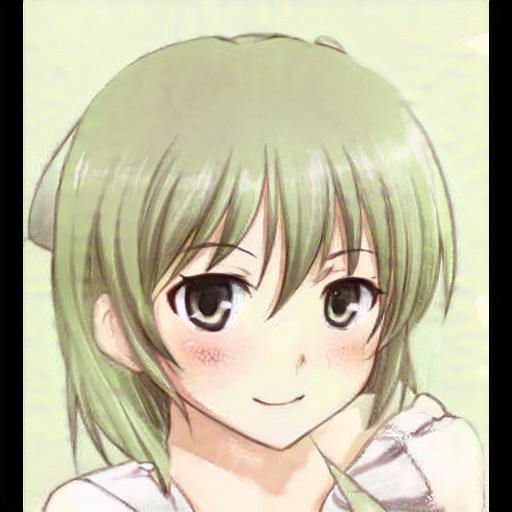}&		
		\includegraphics[width=.12\linewidth]{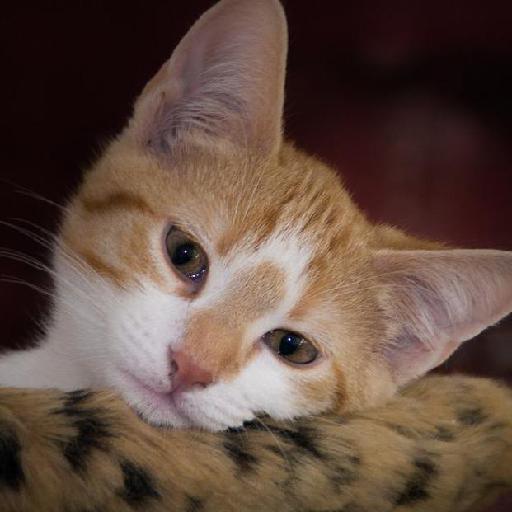}&
		\includegraphics[width=.12\linewidth]{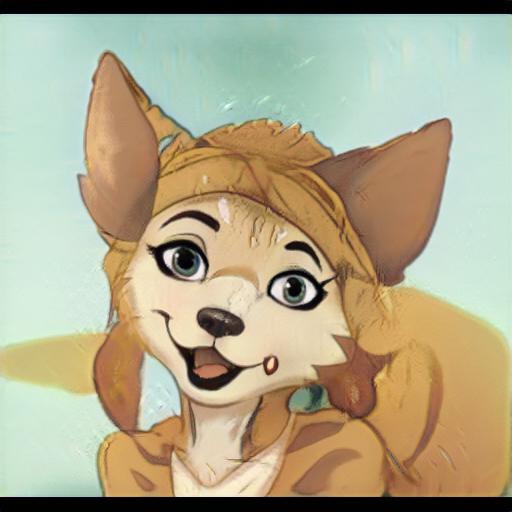}&
		\includegraphics[width=.12\linewidth]{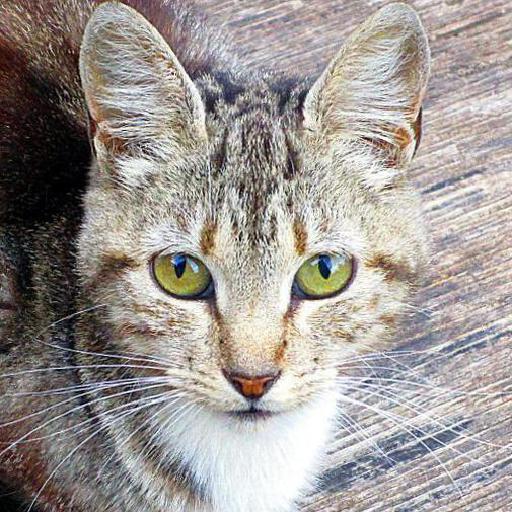}&
		\includegraphics[width=.12\linewidth]{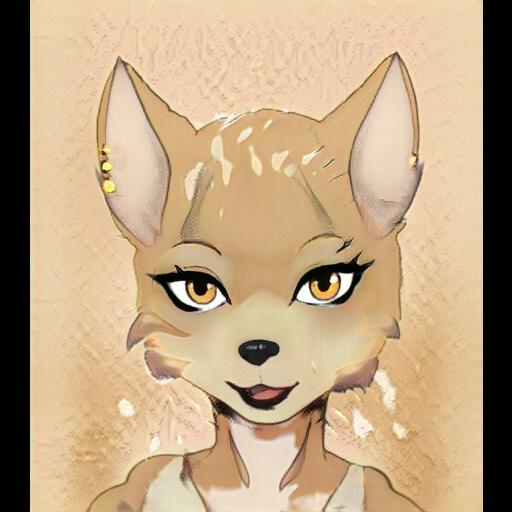}
		\\				
		\multicolumn{2}{c}{ImageNet$\Rightarrow$NABirds} &
		\multicolumn{2}{c}{ImageNet$\Rightarrow$NABirds} &
		\multicolumn{2}{c}{ImageNet$\Rightarrow$NABirds}&
		\multicolumn{2}{c}{Raphael$\Rightarrow$FFHQ}\\
		\includegraphics[width=.12\linewidth]{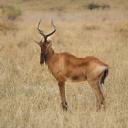}&
		\includegraphics[width=.12\linewidth]{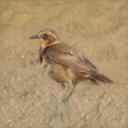}&
		\includegraphics[width=.12\linewidth]{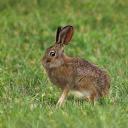}&
		\includegraphics[width=.12\linewidth]{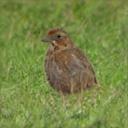}&
		\includegraphics[width=.12\linewidth]{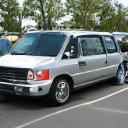}&
		\includegraphics[width=.12\linewidth]{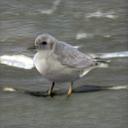}&
		\includegraphics[width=.12\linewidth]{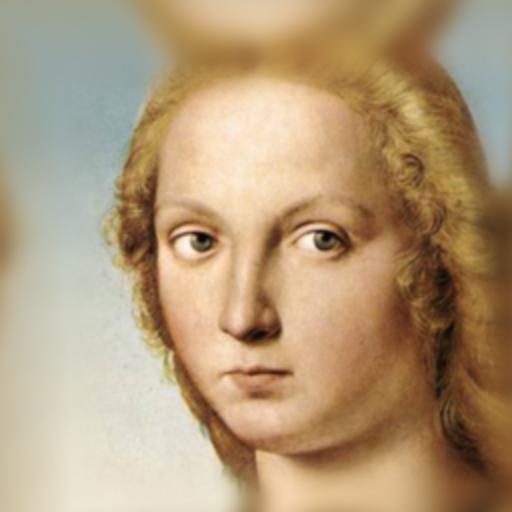}&
		\includegraphics[width=.12\linewidth]{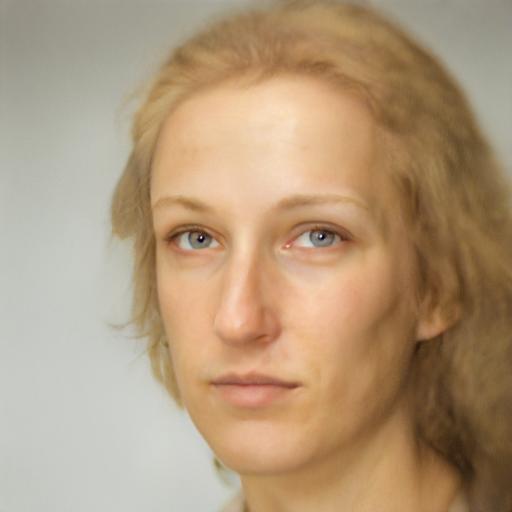}
		\\				
		\multicolumn{2}{c}{AFHQ-cat$\Rightarrow$LSUN-church} &
		\multicolumn{2}{c}{AFHQ-cat$\Rightarrow$LSUN-church} &
		\multicolumn{2}{c}{AFHQ-cat$\Rightarrow$LSUN-church}&
		\multicolumn{2}{c}{AFHQ-cat$\Rightarrow$LSUN-church}\\
		\includegraphics[width=.12\linewidth]{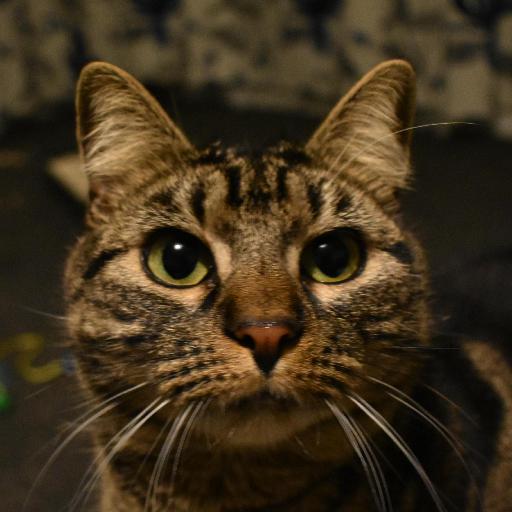}&
		\includegraphics[width=.12\linewidth]{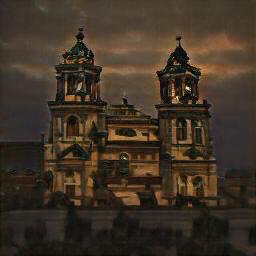}&
		\includegraphics[width=.12\linewidth]{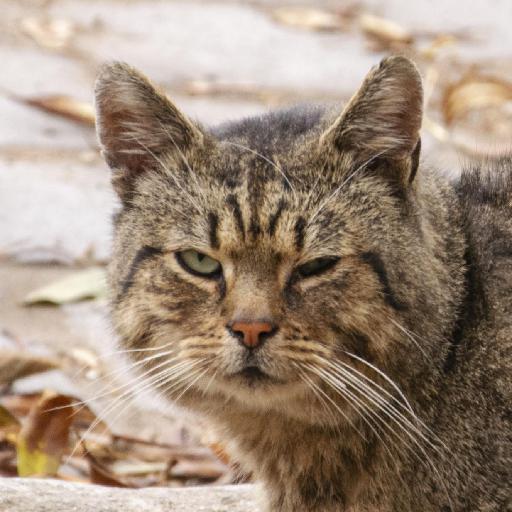}&
		\includegraphics[width=.12\linewidth]{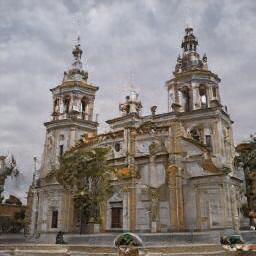}&
		\includegraphics[width=.12\linewidth]{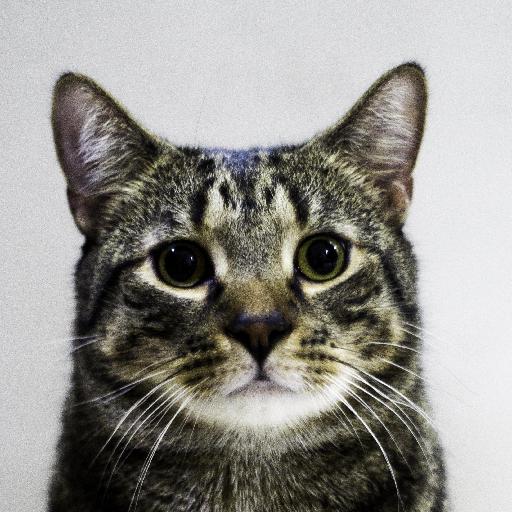}&
		\includegraphics[width=.12\linewidth]{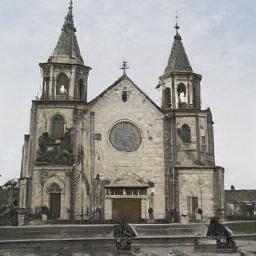}&
		\includegraphics[width=.12\linewidth]{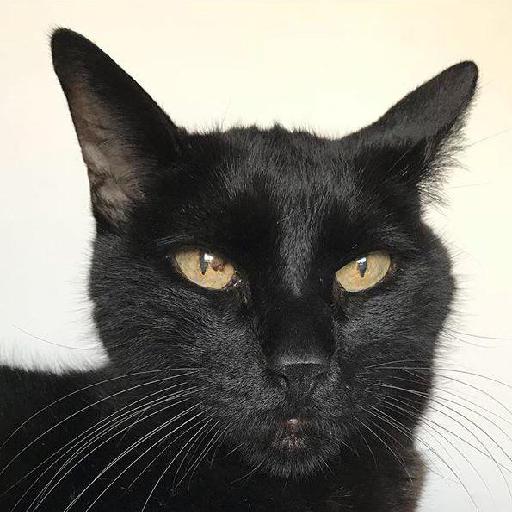}
		&
		\includegraphics[width=.12\linewidth]{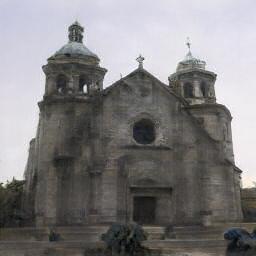}
		\\		
	\end{tabular}
	\caption{More translation results using UniTranslator. Our method can efficiently process any source image and transform it into the given target domain with high quality.}
	\label{fig:display}
\end{figure*}

\begin{figure*}
	\centering
	\setlength{\tabcolsep}{0.05em}
	\begin{tabular}{ccccc}		
		
		\includegraphics[width=.19\linewidth]{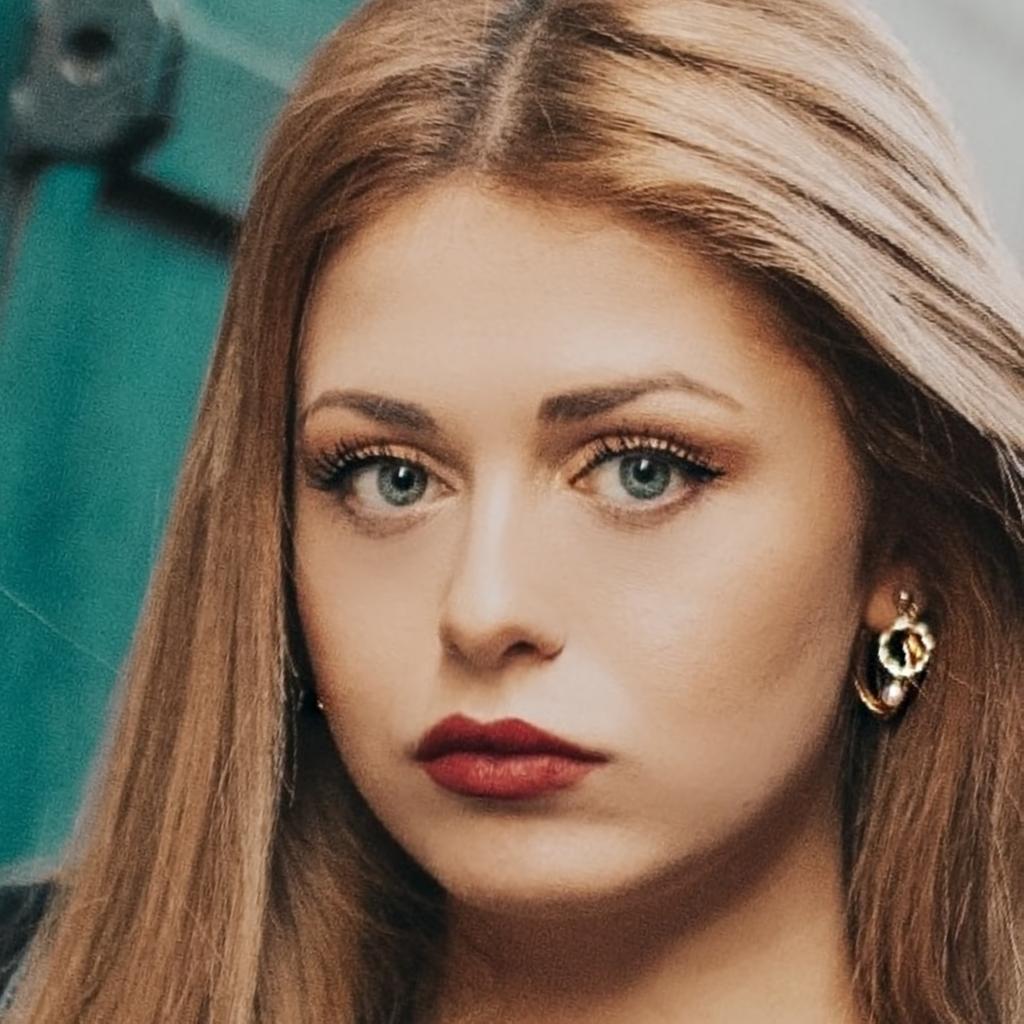} &
		\includegraphics[width=.19\linewidth]{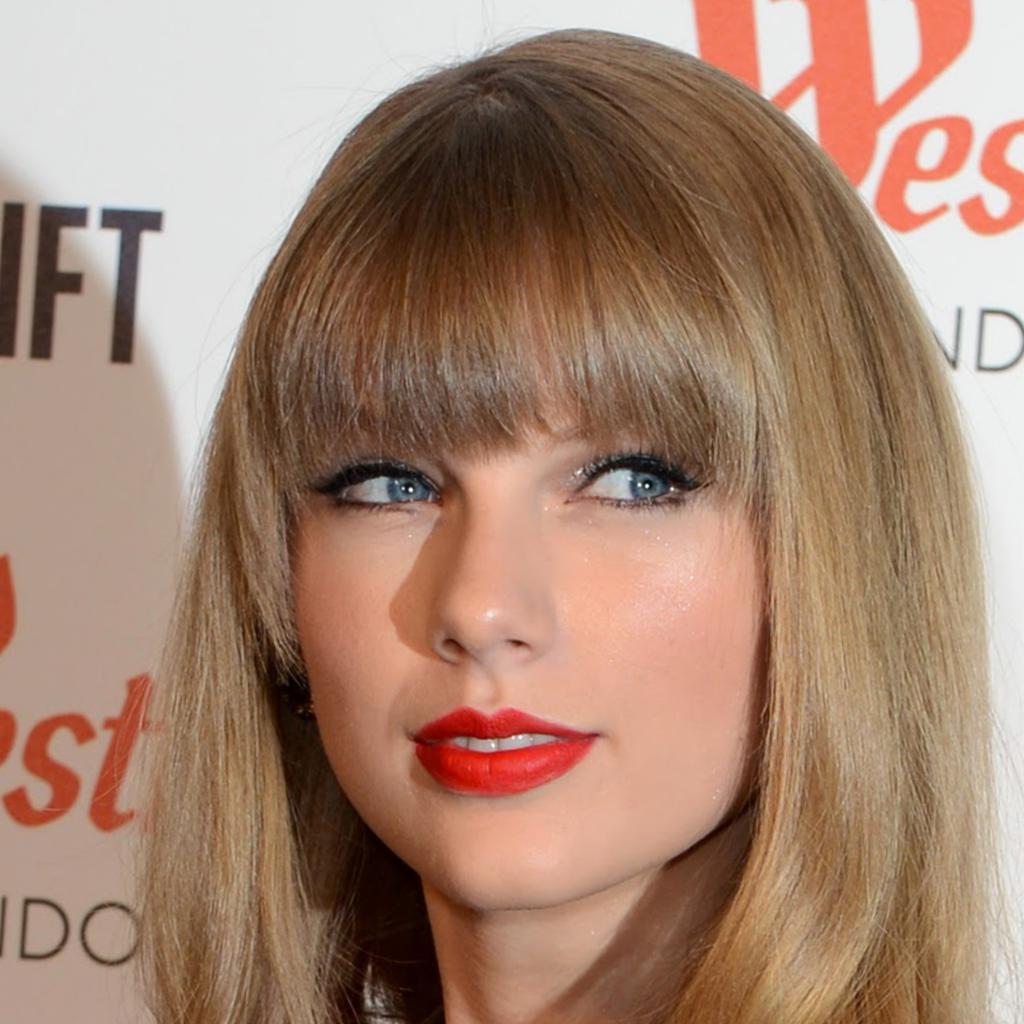} &
		\includegraphics[width=.19\linewidth]{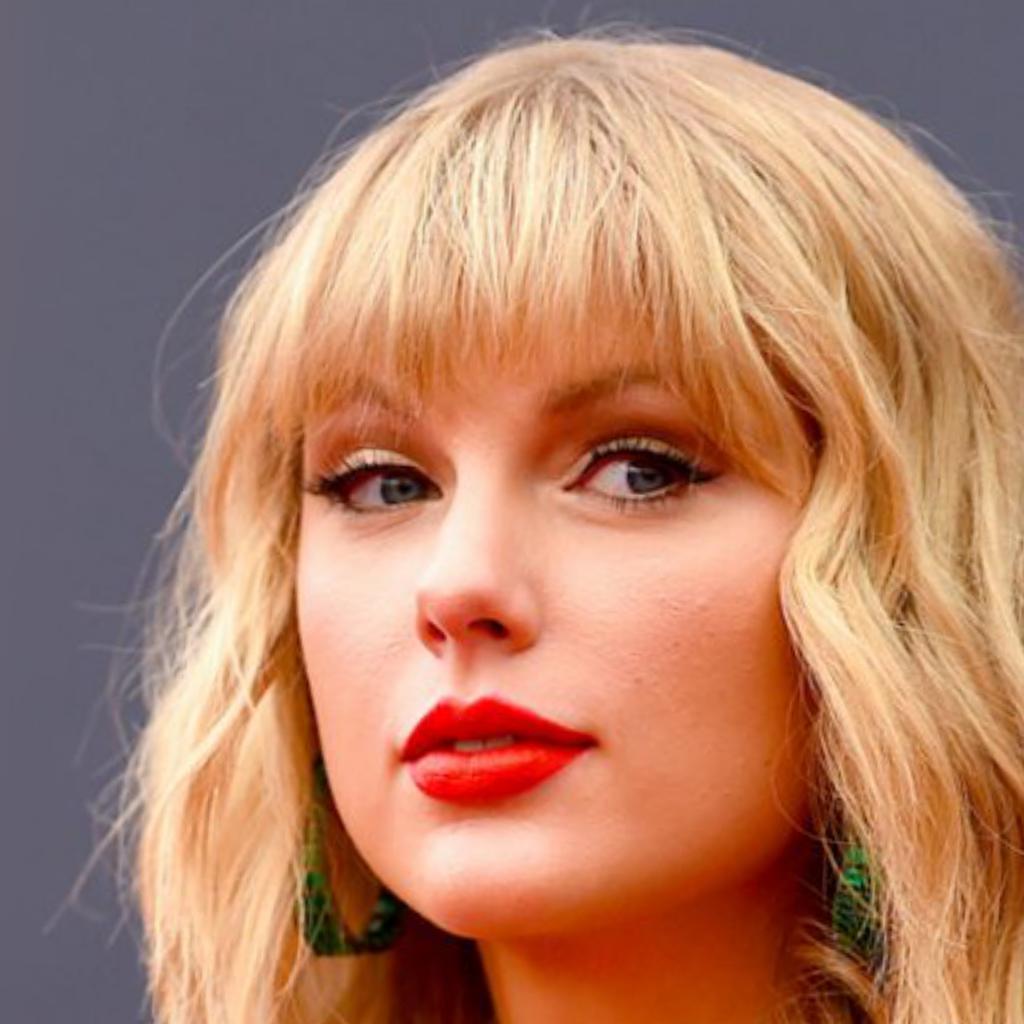} &
		\includegraphics[width=.19\linewidth]{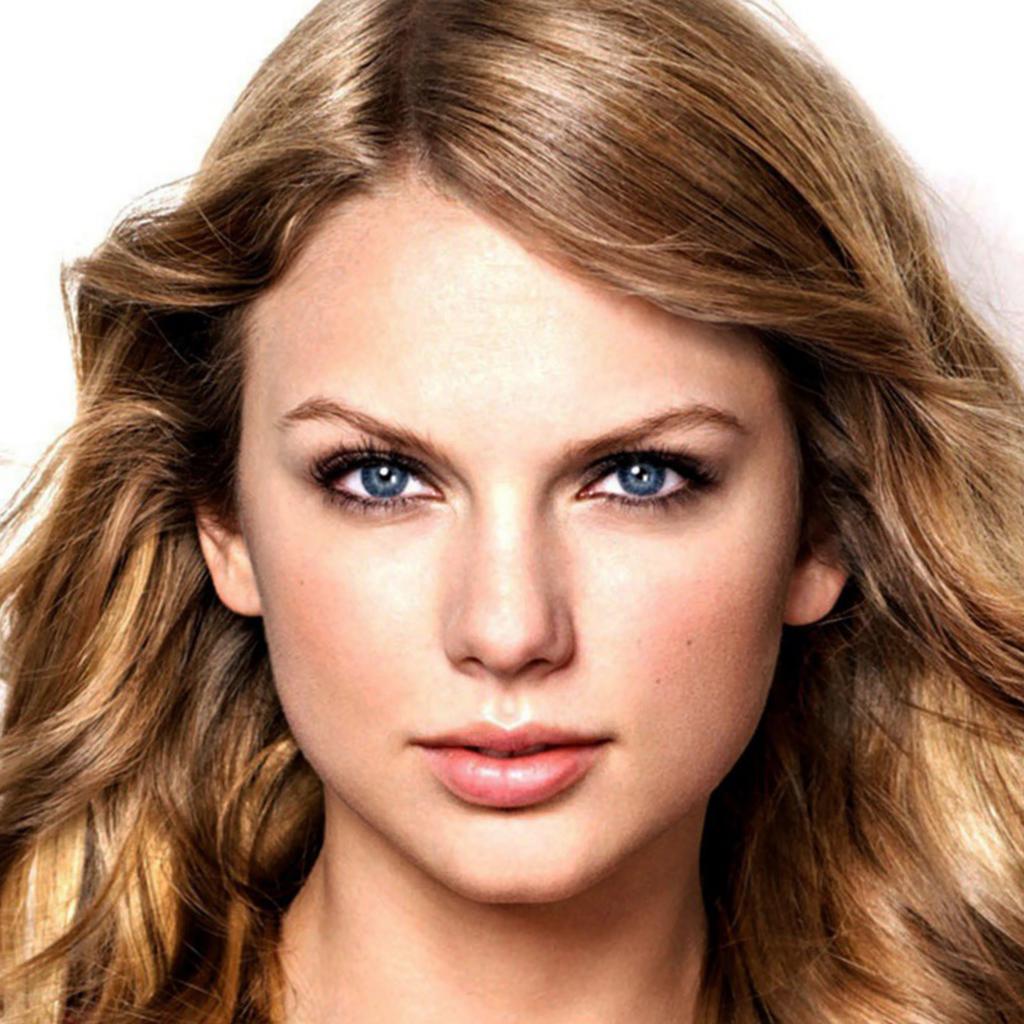} &
		\includegraphics[width=.19\linewidth]{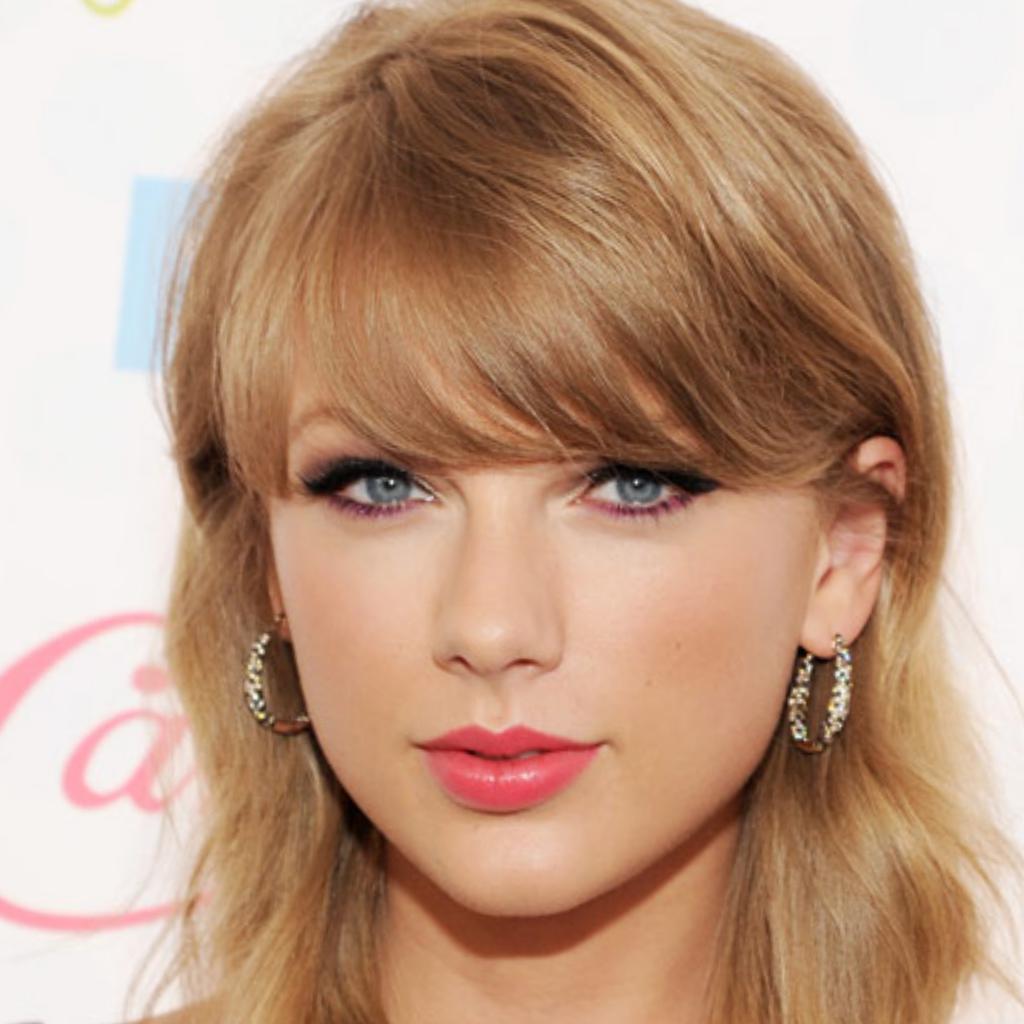} \\
		
		\includegraphics[width=.19\linewidth]{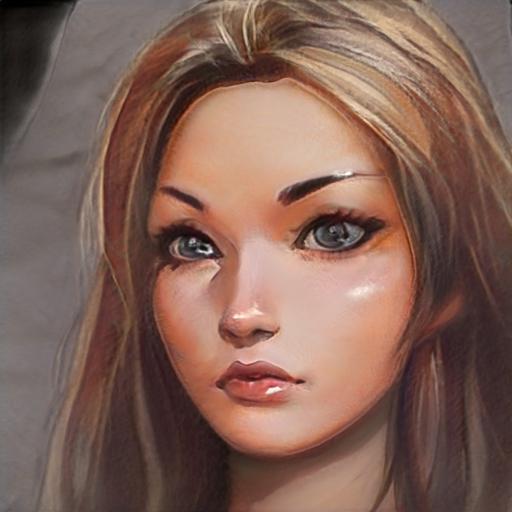} &
		\includegraphics[width=.19\linewidth]{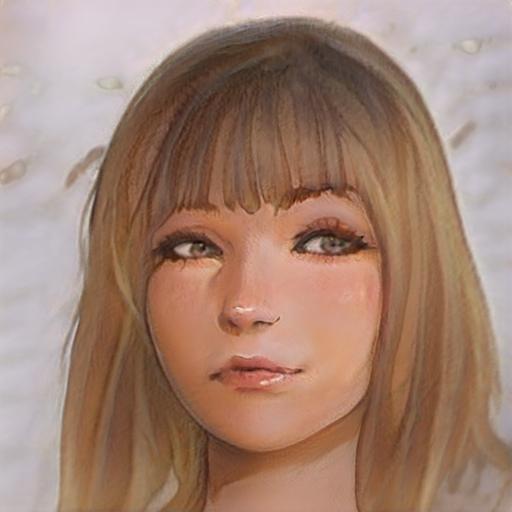} &
		\includegraphics[width=.19\linewidth]{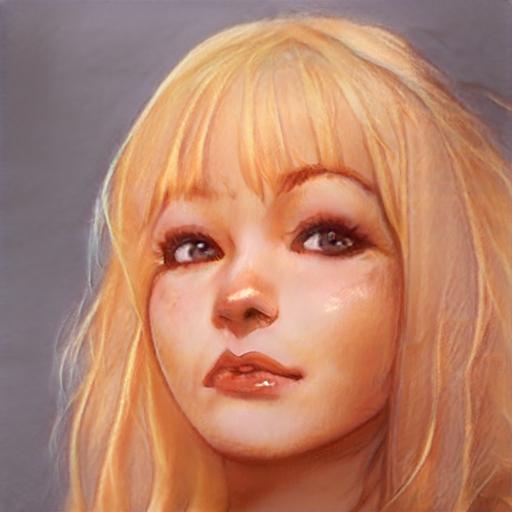} &
		\includegraphics[width=.19\linewidth]{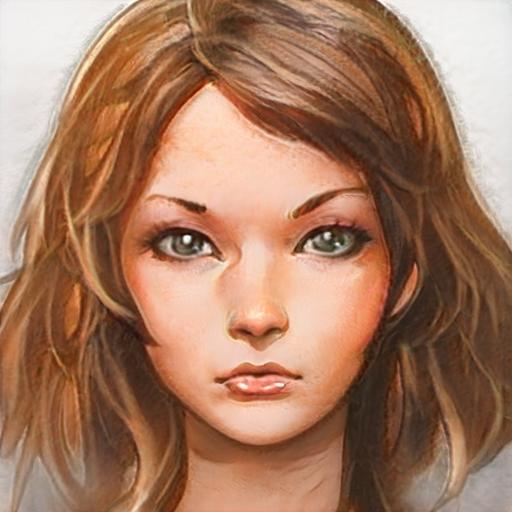} &
		\includegraphics[width=.19\linewidth]{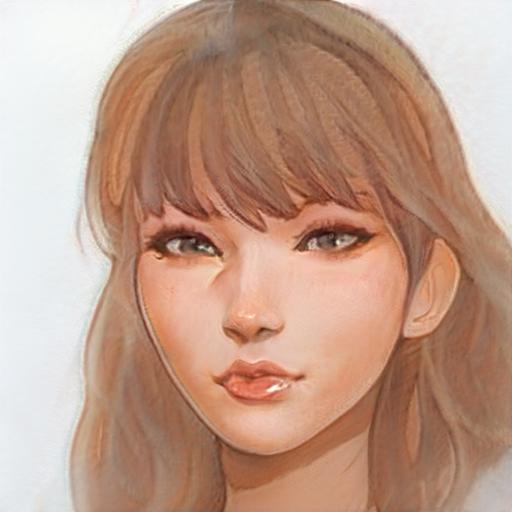} \\
		
		\includegraphics[width=.19\linewidth]{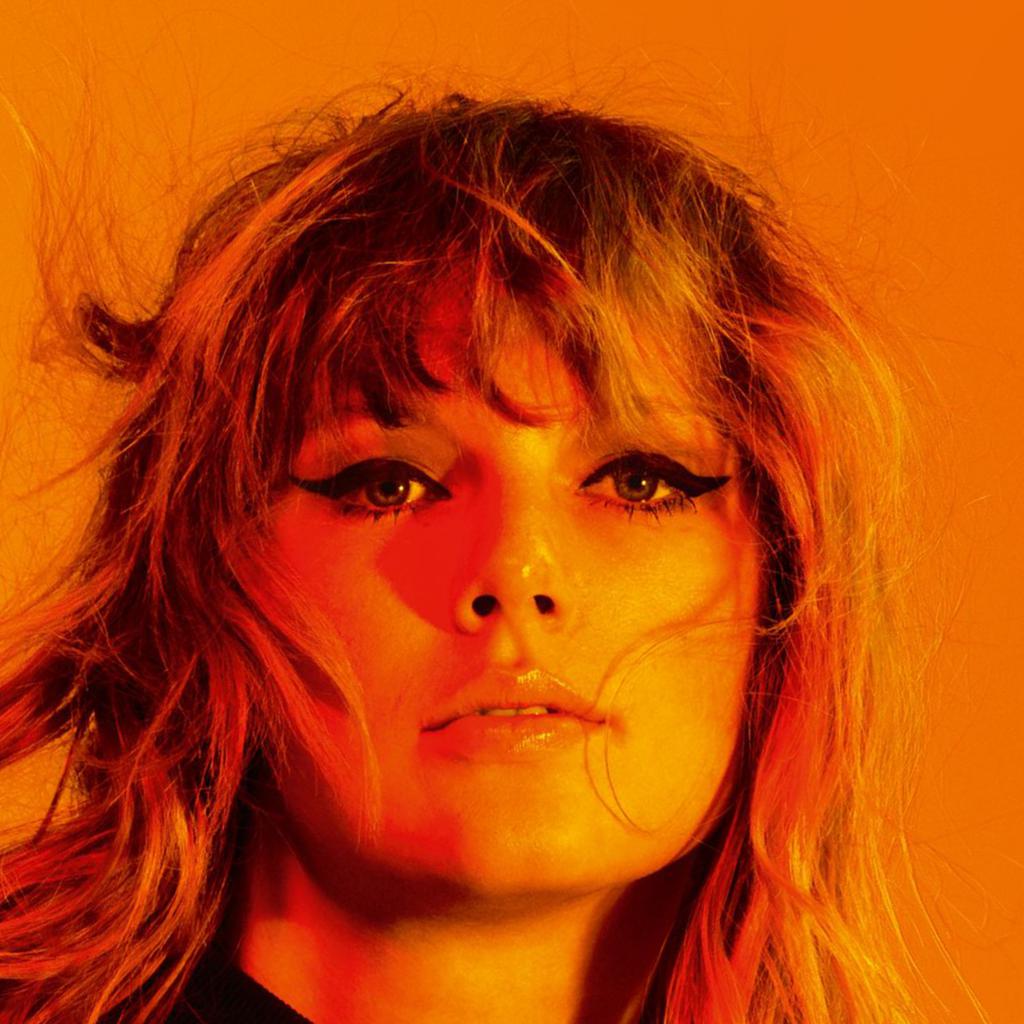} &
		\includegraphics[width=.19\linewidth]{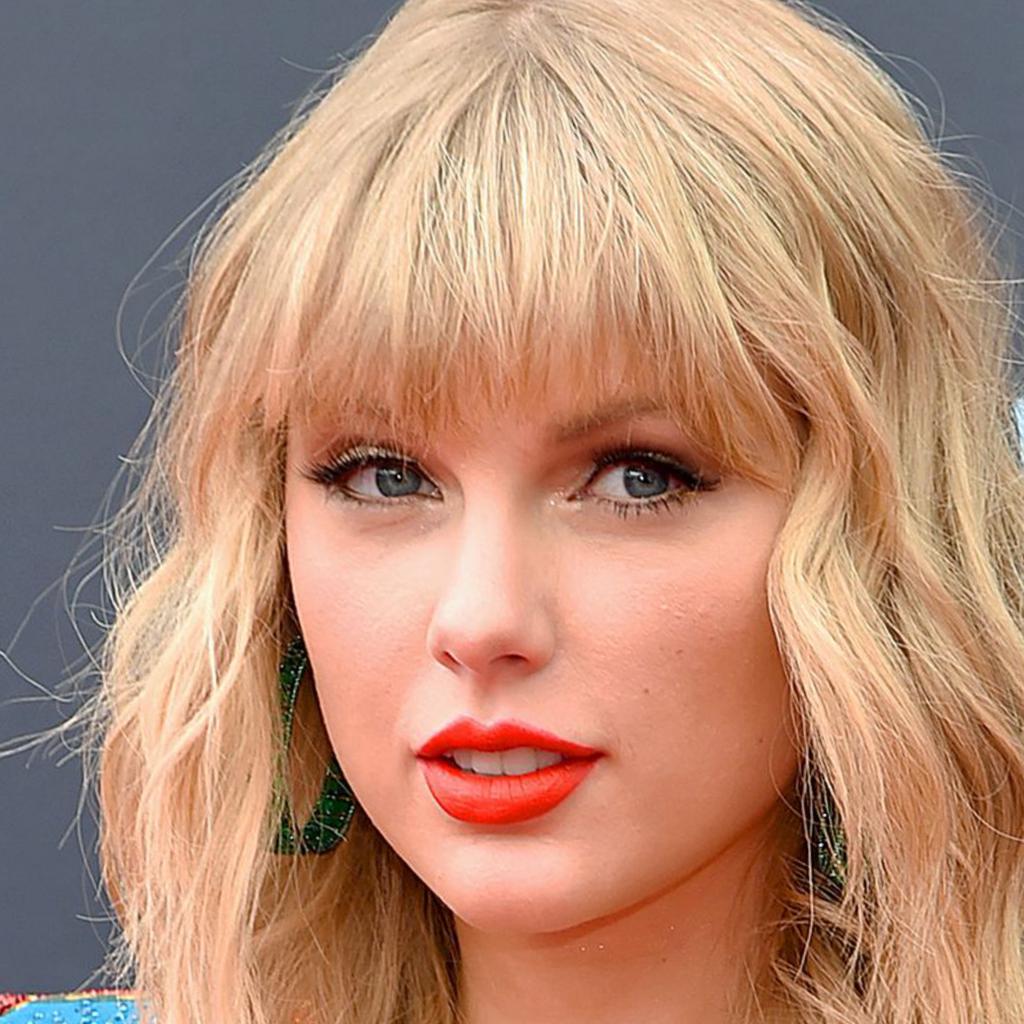} &
		\includegraphics[width=.19\linewidth]{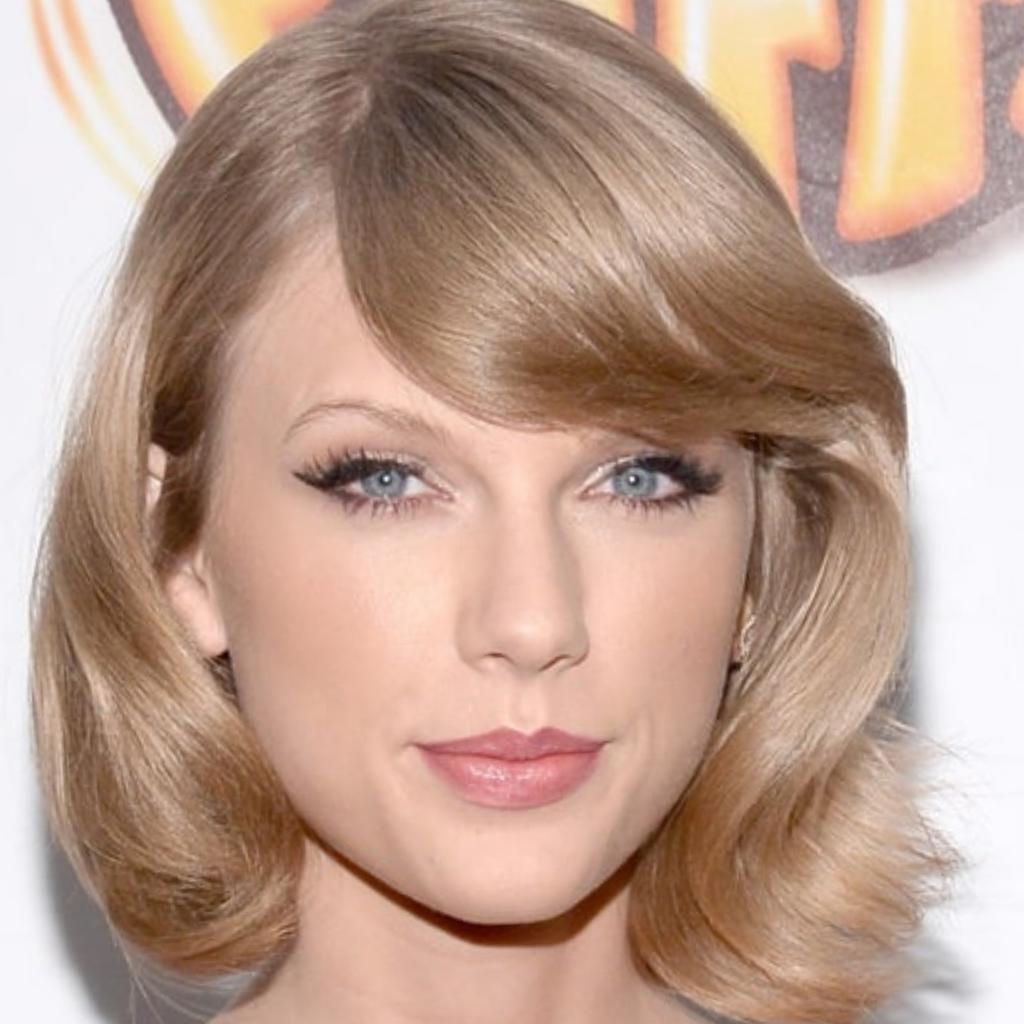} &
		\includegraphics[width=.19\linewidth]{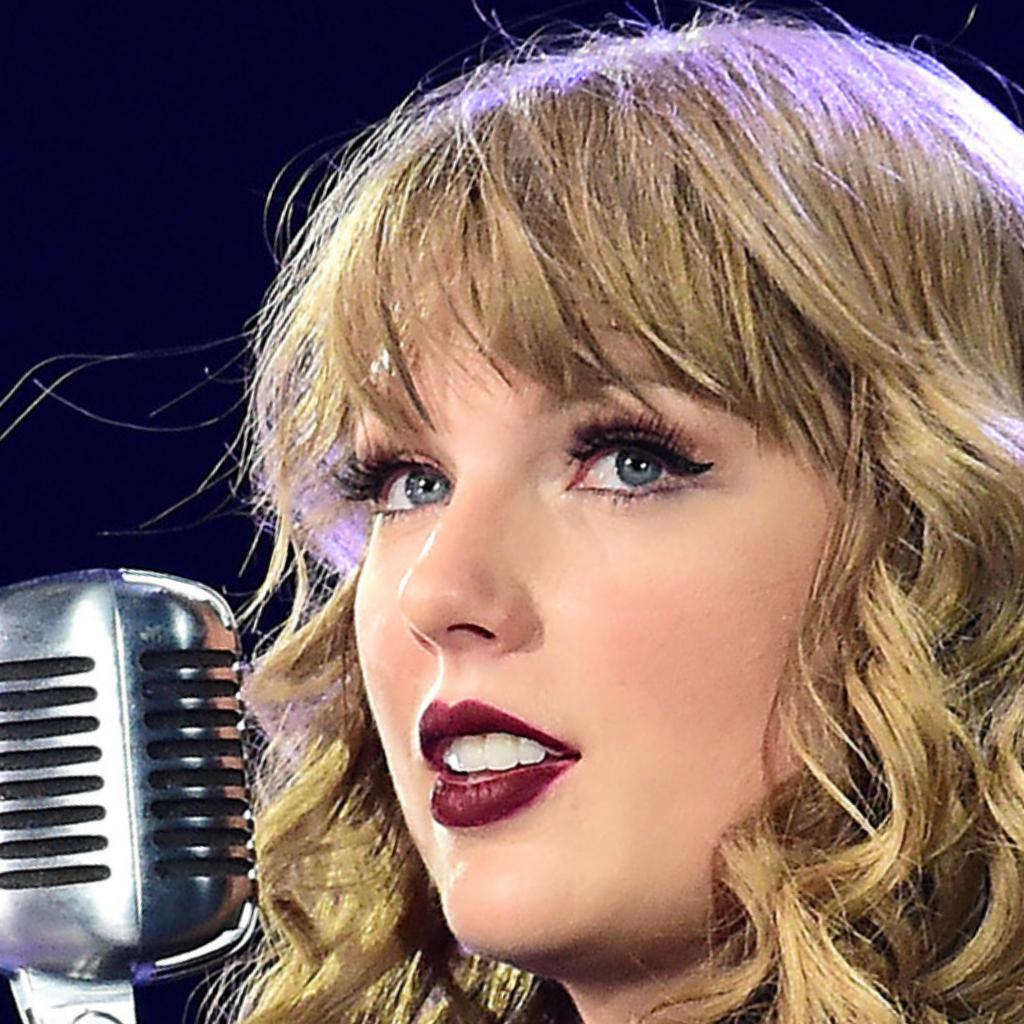} &
		\includegraphics[width=.19\linewidth]{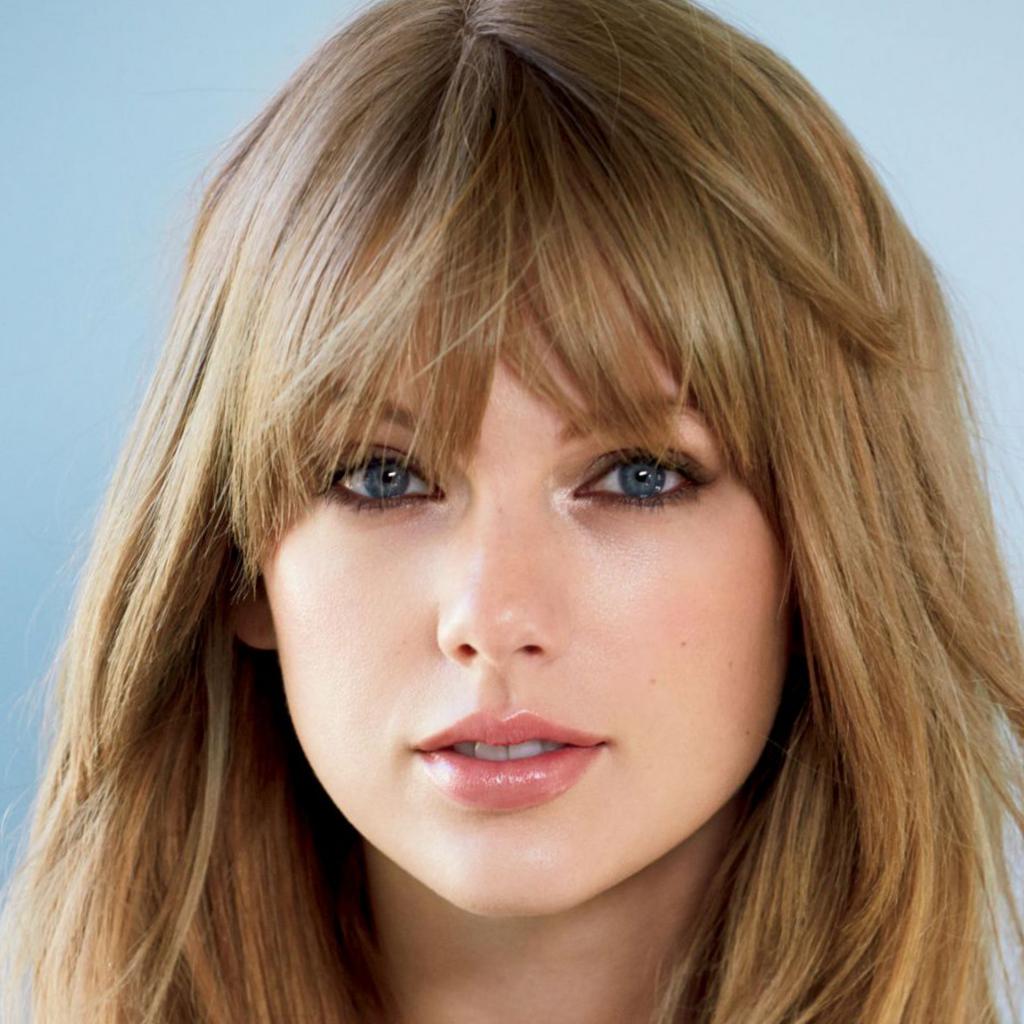} \\
		
		\includegraphics[width=.19\linewidth]{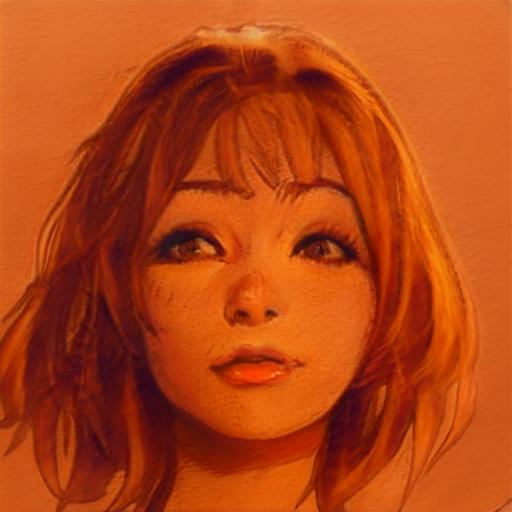} &
		\includegraphics[width=.19\linewidth]{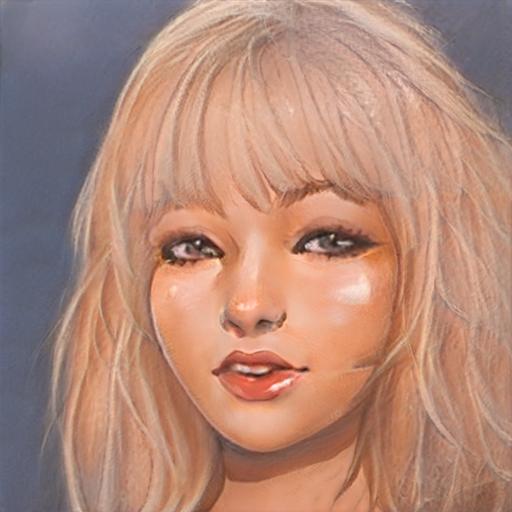} &
		\includegraphics[width=.19\linewidth]{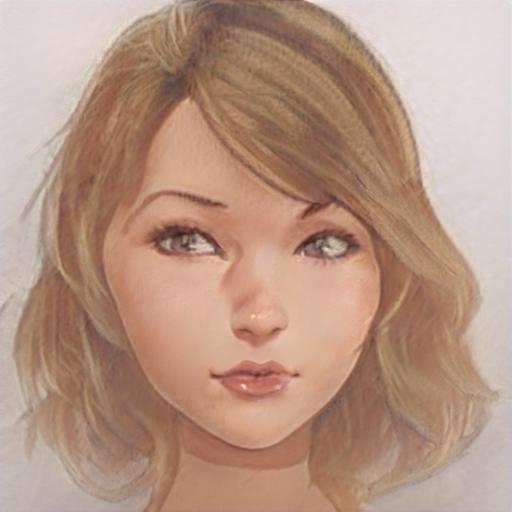} &
		\includegraphics[width=.19\linewidth]{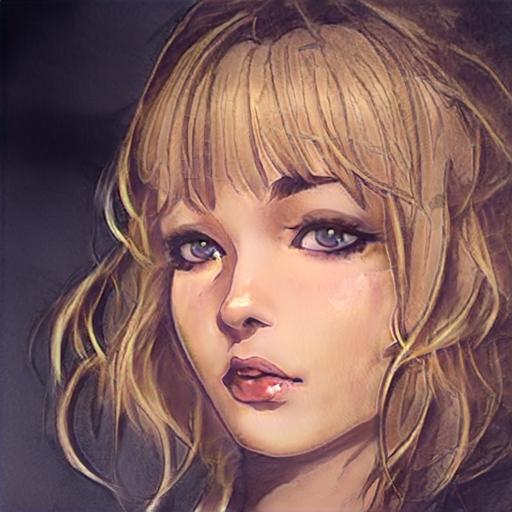} &
		\includegraphics[width=.19\linewidth]{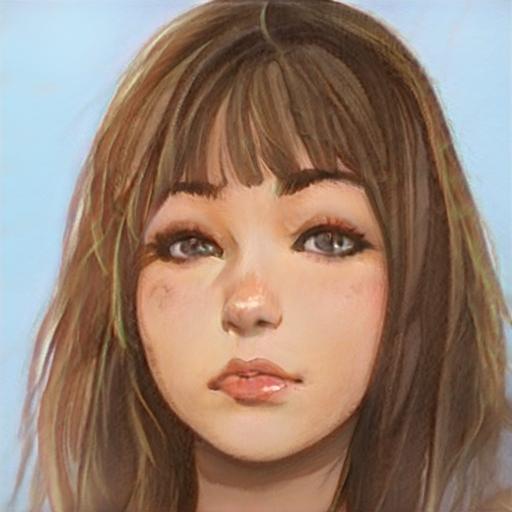} \\
		
		\includegraphics[width=.19\linewidth]{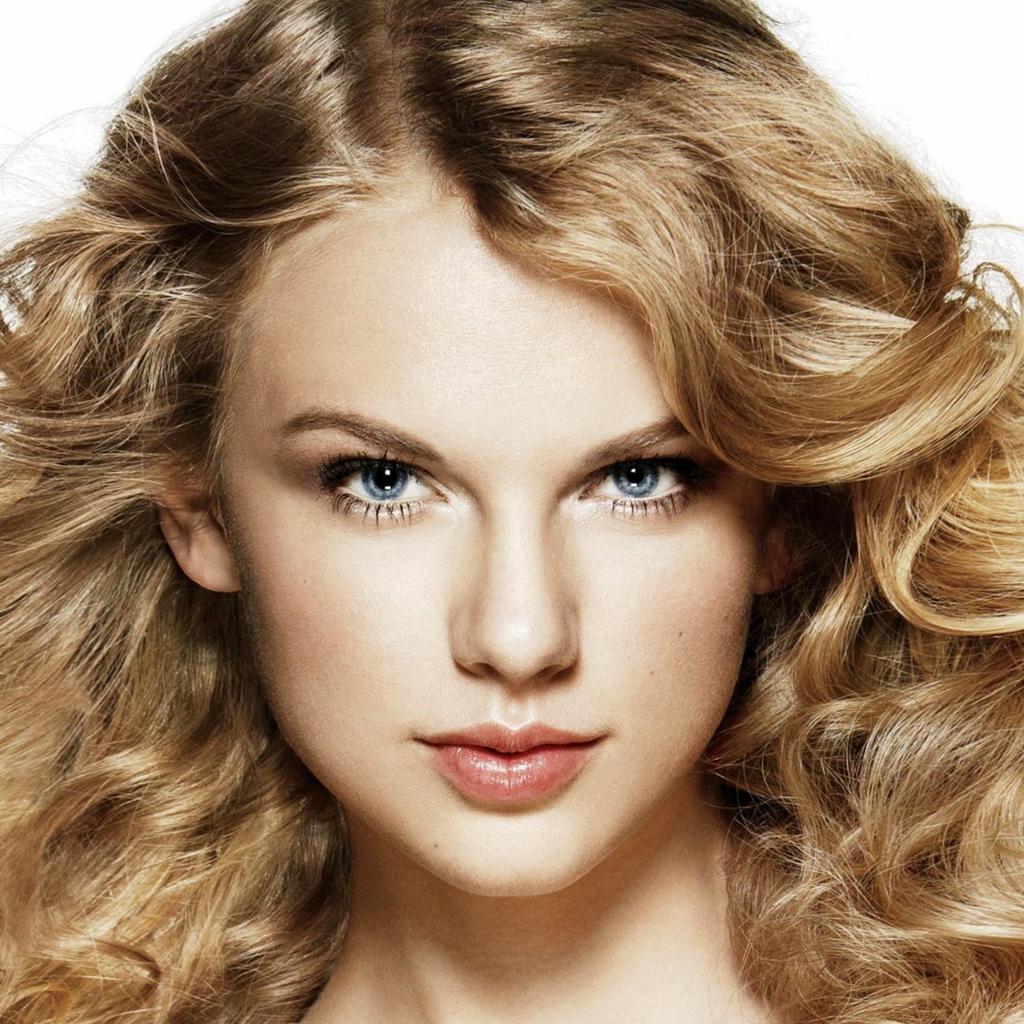} &
		\includegraphics[width=.19\linewidth]{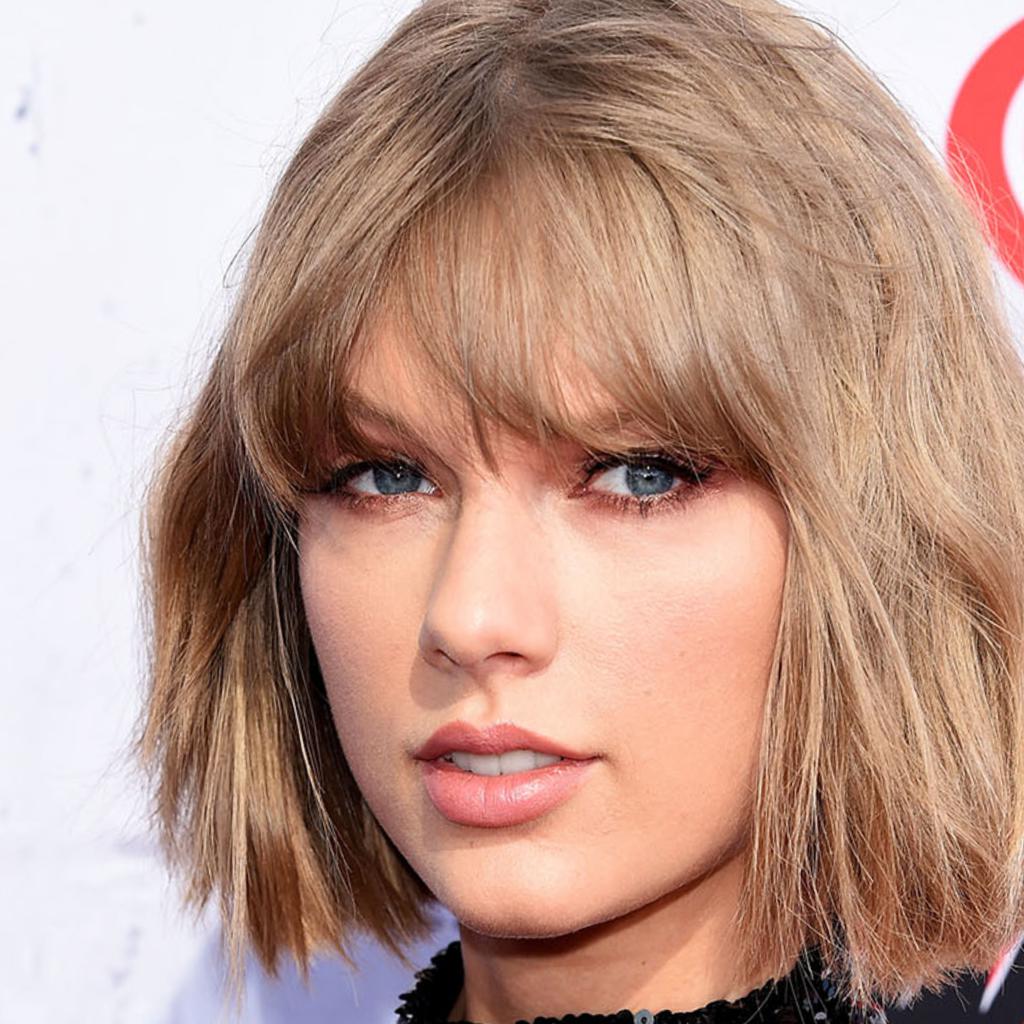} &
		\includegraphics[width=.19\linewidth]{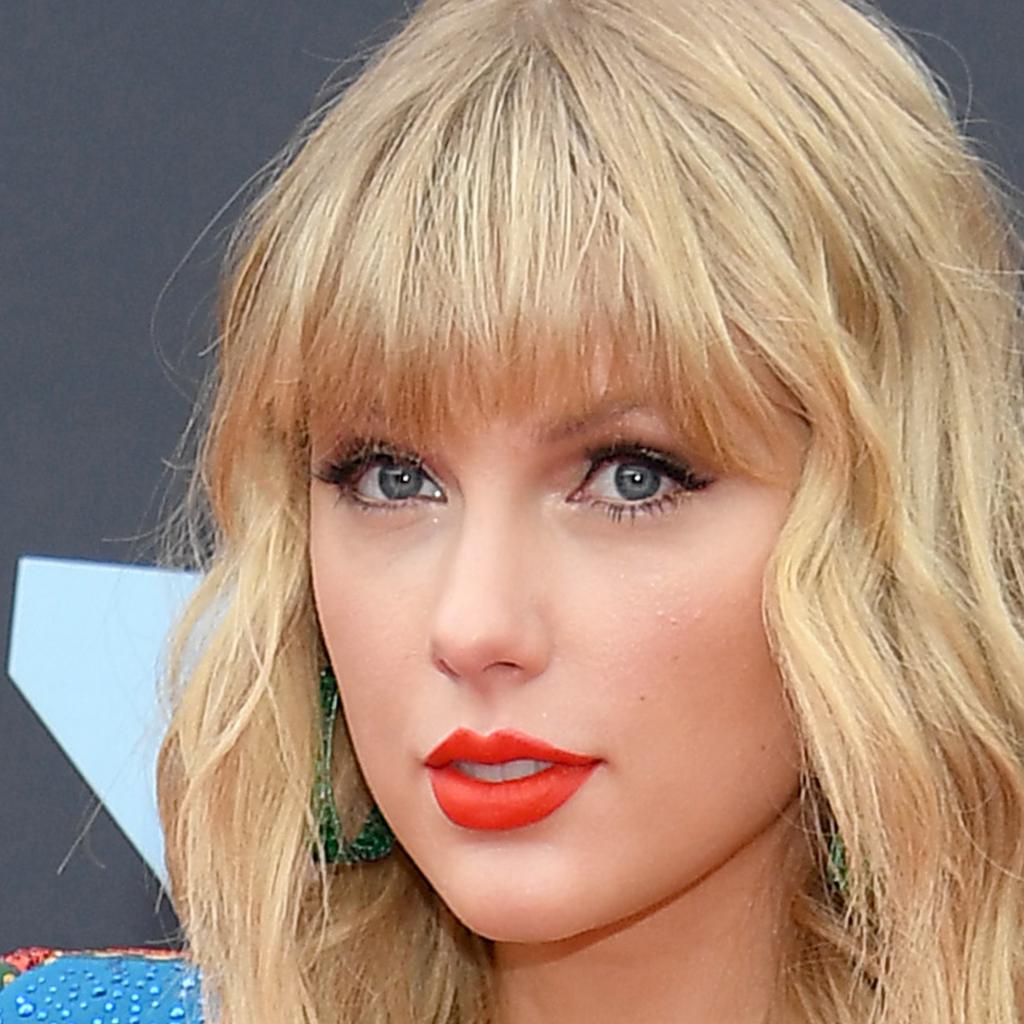} &
		\includegraphics[width=.19\linewidth]{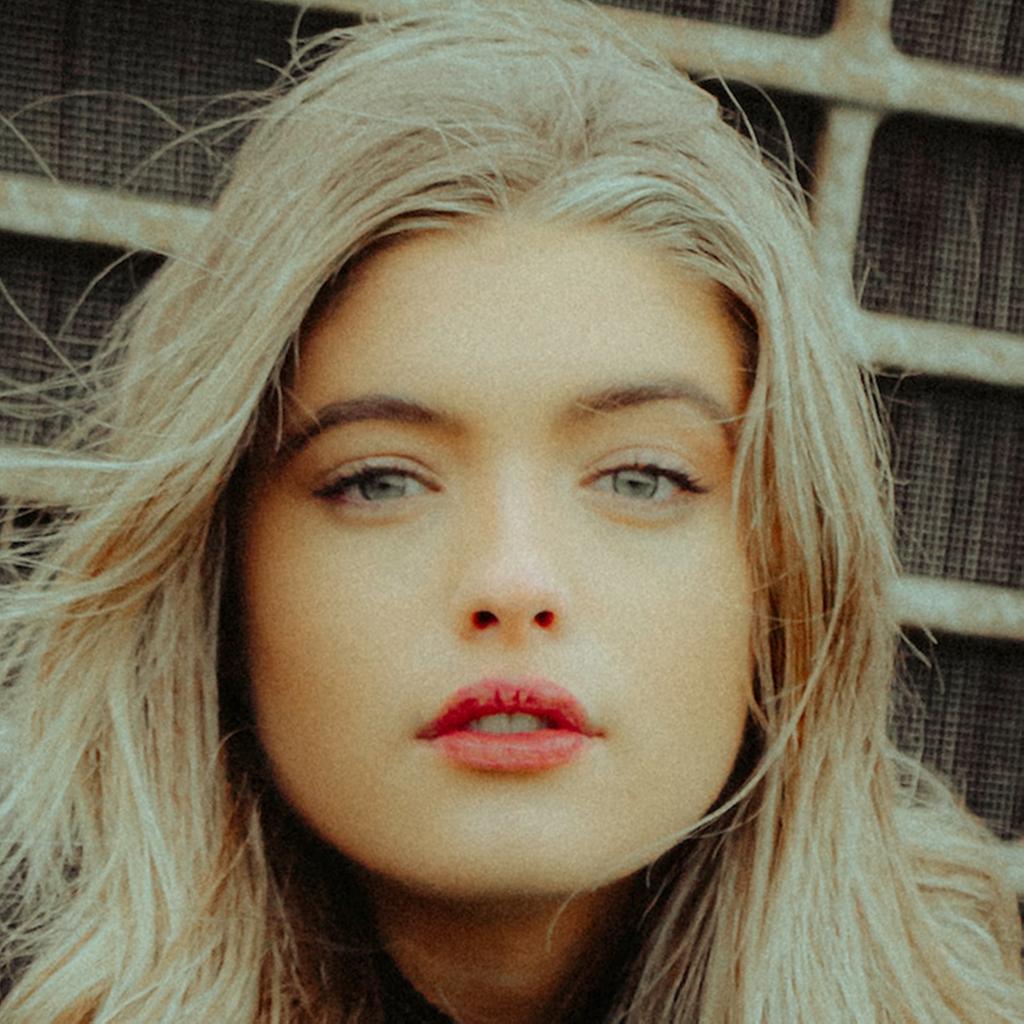} &
		\includegraphics[width=.19\linewidth]{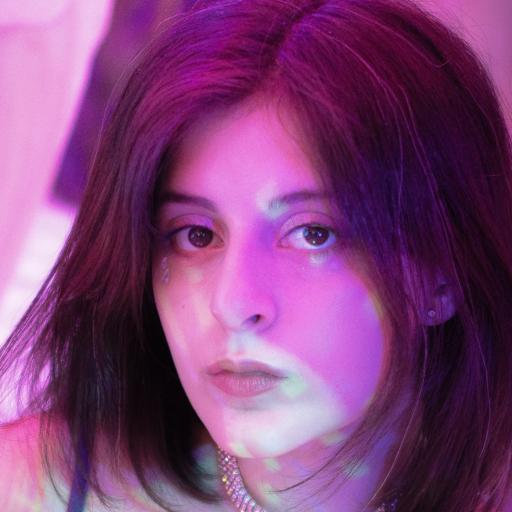} \\
		
		\includegraphics[width=.19\linewidth]{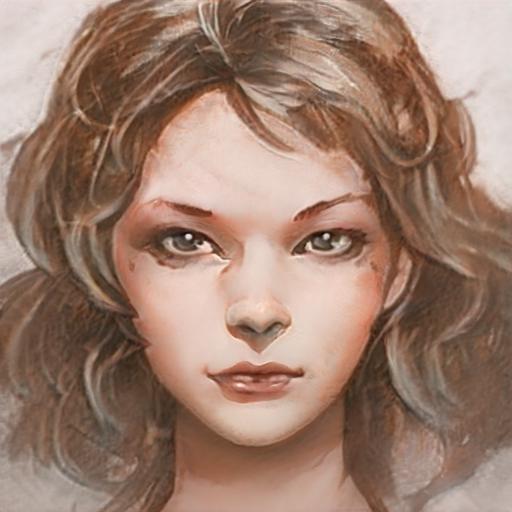} &
		\includegraphics[width=.19\linewidth]{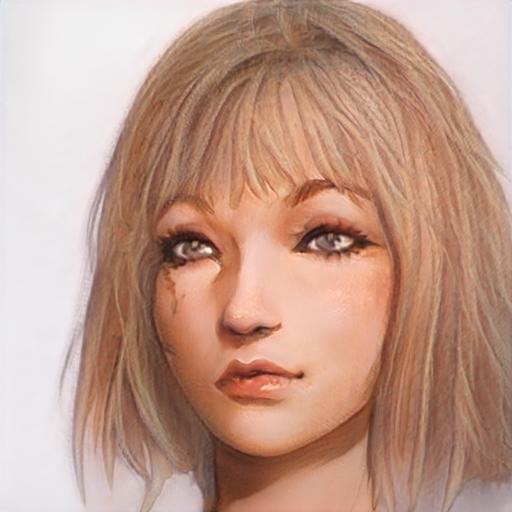} &
		\includegraphics[width=.19\linewidth]{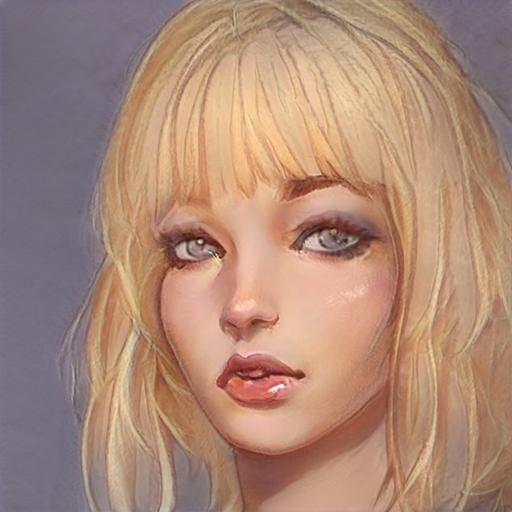} &
		\includegraphics[width=.19\linewidth]{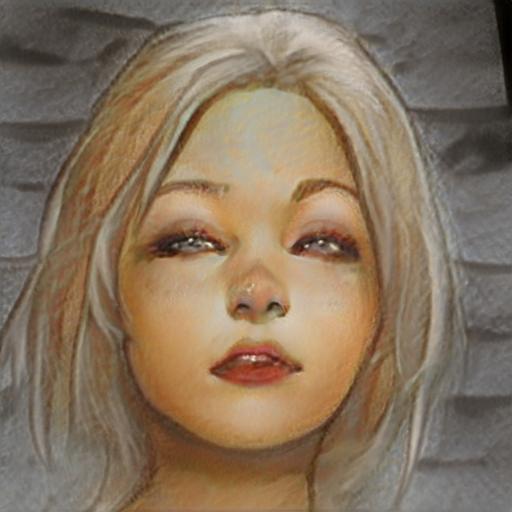} &
		\includegraphics[width=.19\linewidth]{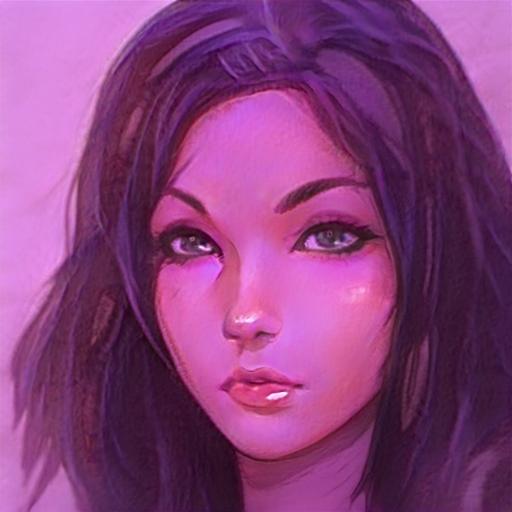} \\
		
	\end{tabular}
	\caption{Stylization of real images. We apply the artistic style of Ilya Kuvshinov to real portraits. The odd-numbered rows display the original images, while the even-numbered rows feature the stylized versions.}
	\label{fig:stylization_supp}
\end{figure*}

\begin{figure*}[t]
	\centering	
	\setlength{\abovecaptionskip}{1mm}
	\centering
	\setlength{\tabcolsep}{0.05em}
	\setlength{\fboxrule}{1pt}
	\setlength{\fboxsep}{0pt}
	\begin{tabular}{ccccccccc}
		\includegraphics[width=.109\linewidth]{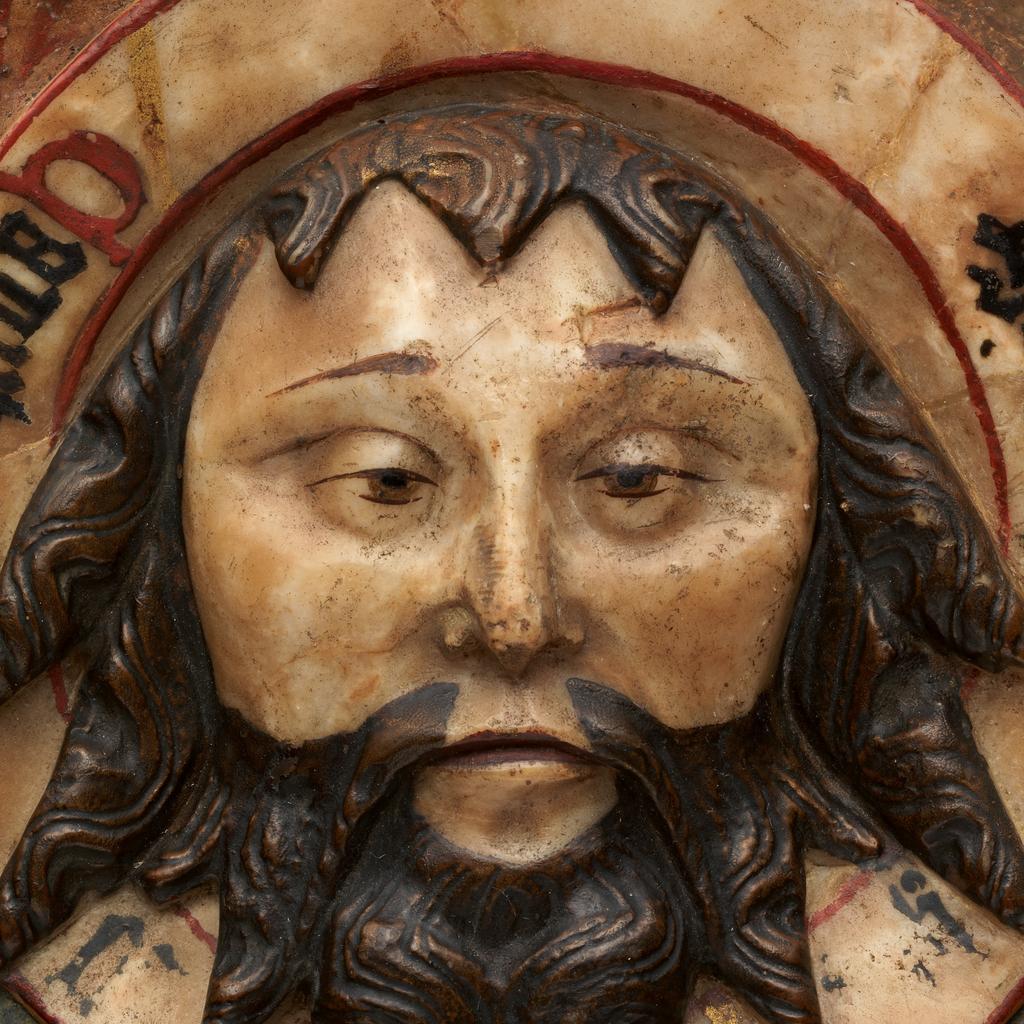} &
		\includegraphics[width=.109\linewidth]{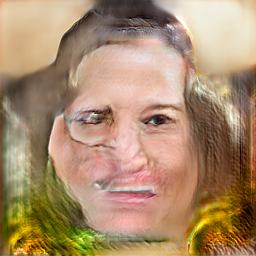} &
		\includegraphics[width=.109\linewidth]{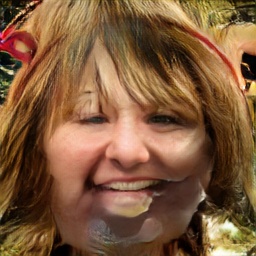} &
		\includegraphics[width=.109\linewidth]{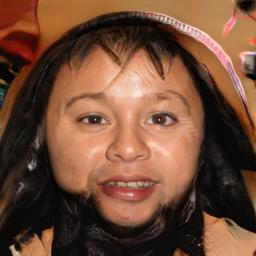} &
		\includegraphics[width=.109\linewidth]{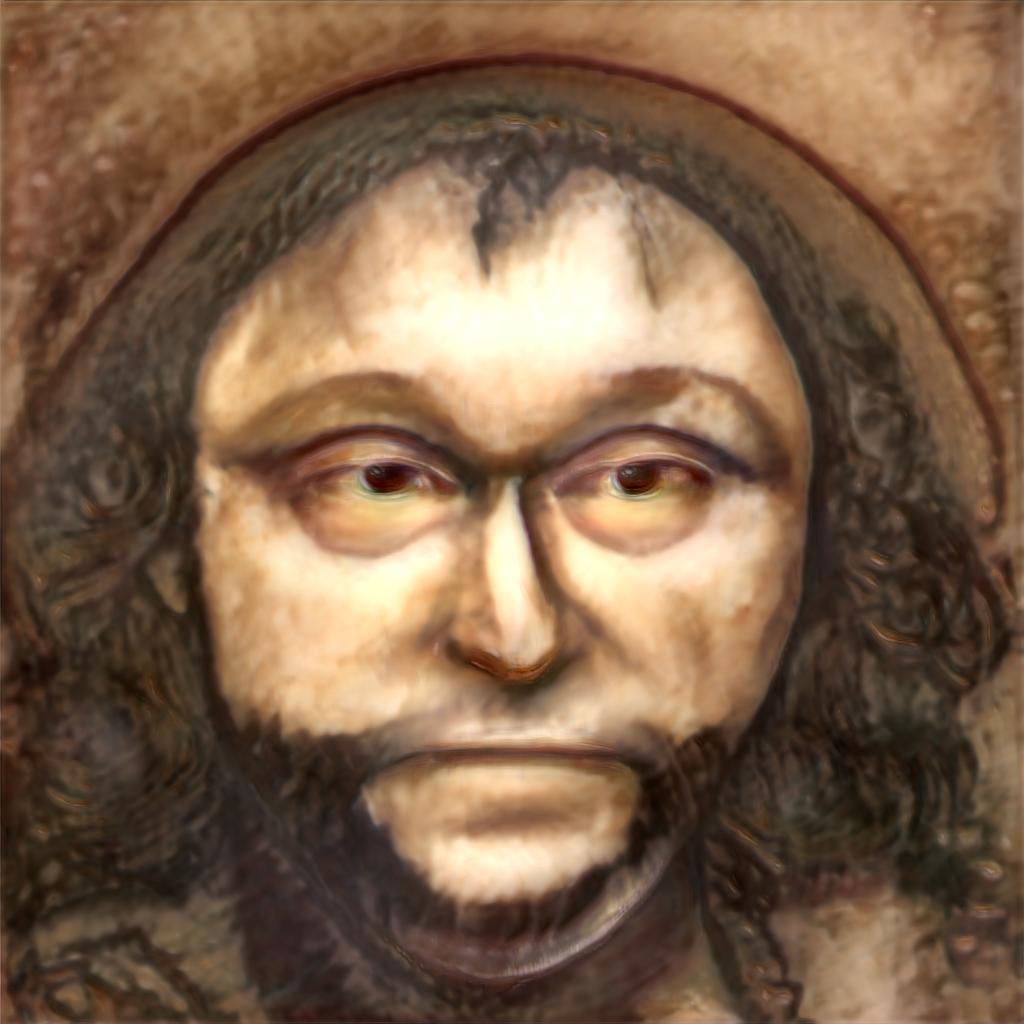} &
		\includegraphics[width=.109\linewidth]{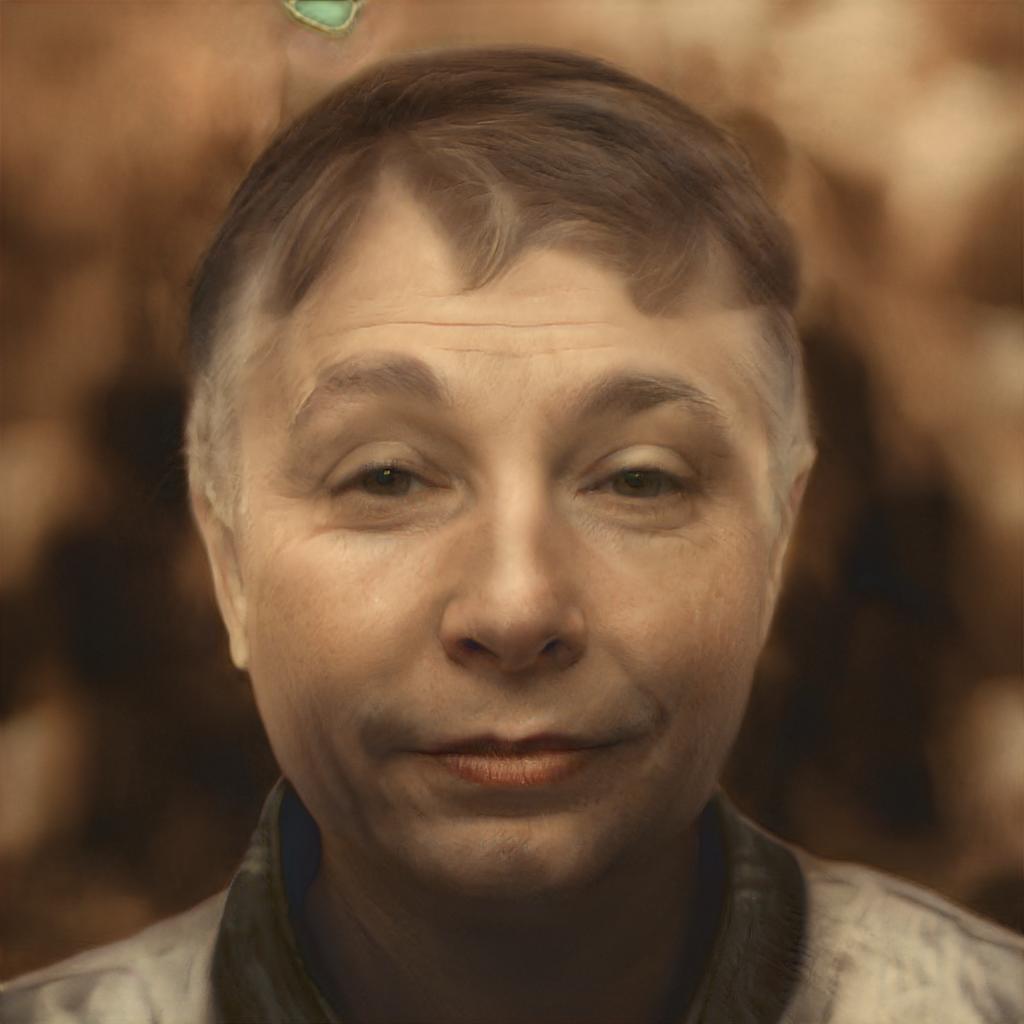} &
		\includegraphics[width=.109\linewidth]{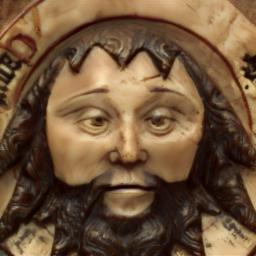} &
		\includegraphics[width=.109\linewidth]{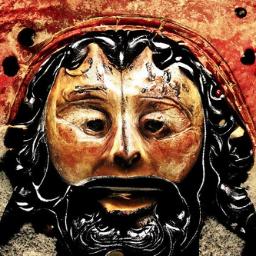} &
		\includegraphics[width=.109\linewidth]{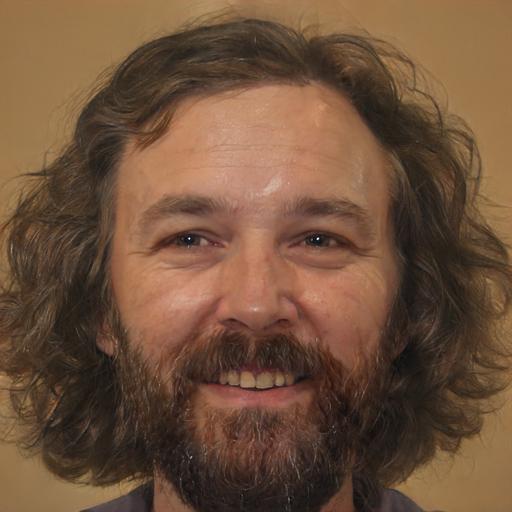}\\
	
		\includegraphics[width=.109\linewidth]{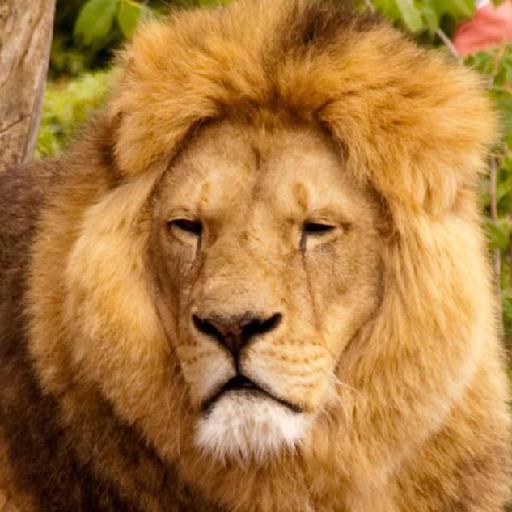} &
		\includegraphics[width=.109\linewidth]{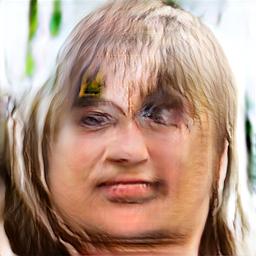} &
		\includegraphics[width=.109\linewidth]{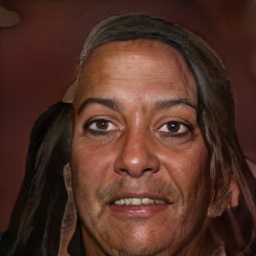} &
		\includegraphics[width=.109\linewidth]{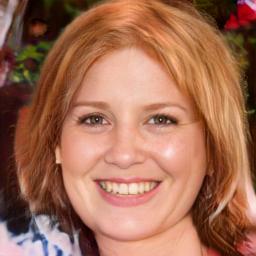} &
		\includegraphics[width=.109\linewidth]{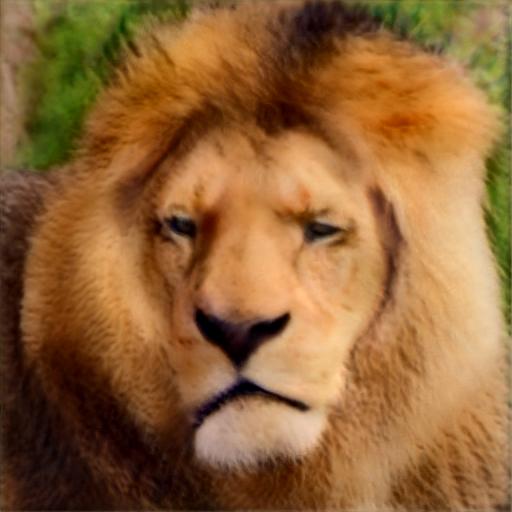} &
		\includegraphics[width=.109\linewidth]{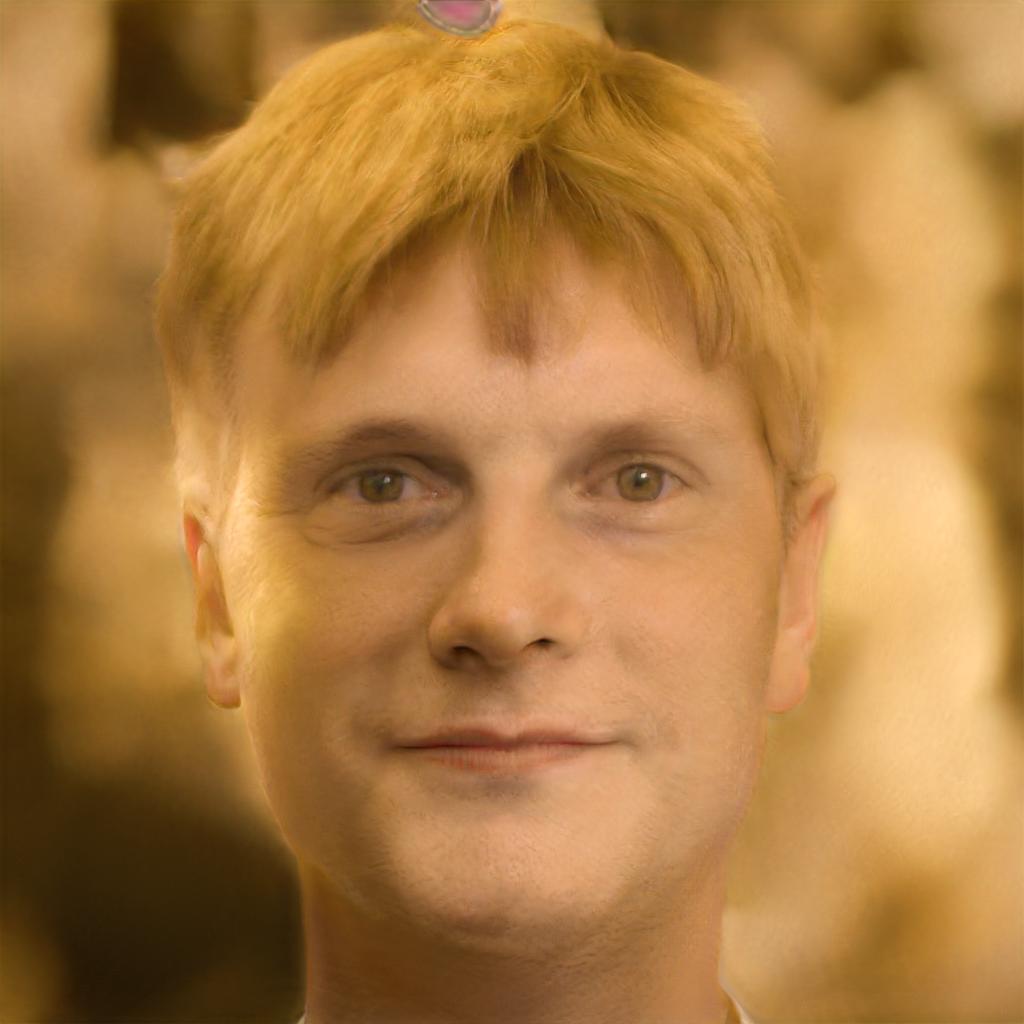} &
		\includegraphics[width=.109\linewidth]{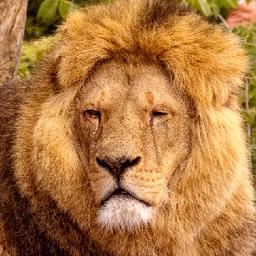} &
		\includegraphics[width=.109\linewidth]{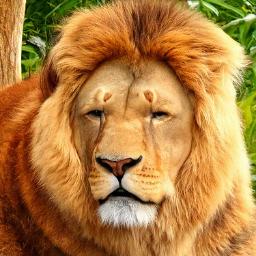} &
		\includegraphics[width=.109\linewidth]{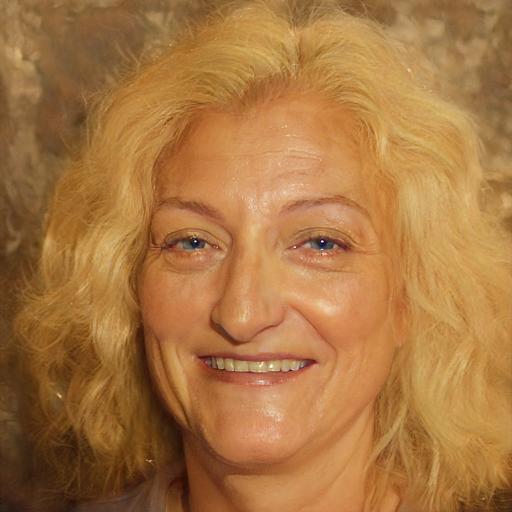}\\

		\includegraphics[width=.109\linewidth]{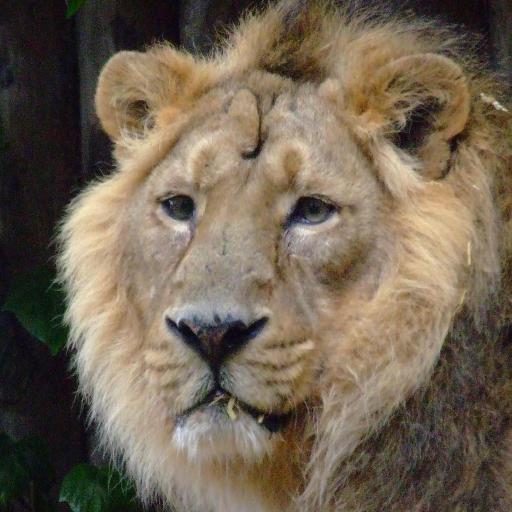} &
		\includegraphics[width=.109\linewidth]{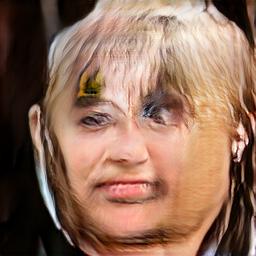} &
		\includegraphics[width=.109\linewidth]{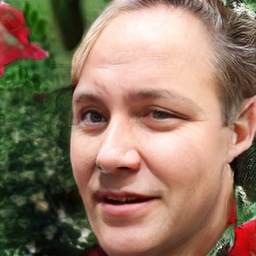} &
		\includegraphics[width=.109\linewidth]{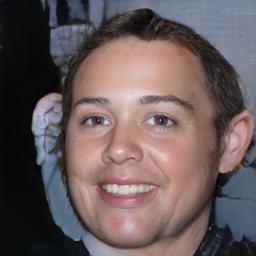} &
		\includegraphics[width=.109\linewidth]{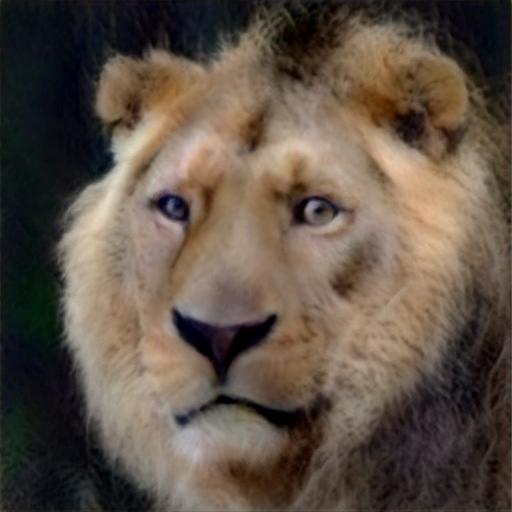} &
		\includegraphics[width=.109\linewidth]{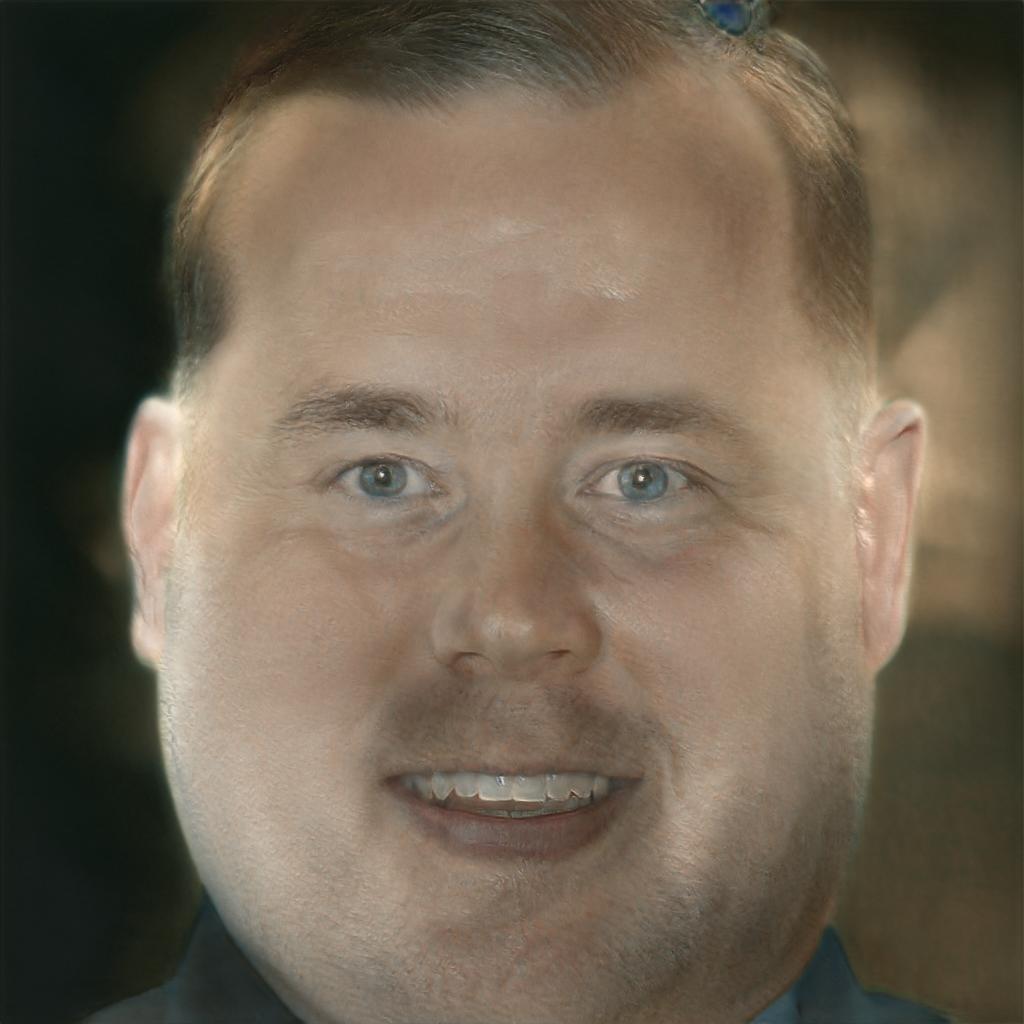} &
		\includegraphics[width=.109\linewidth]{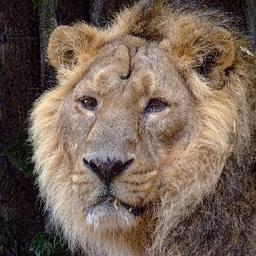} &
		\includegraphics[width=.109\linewidth]{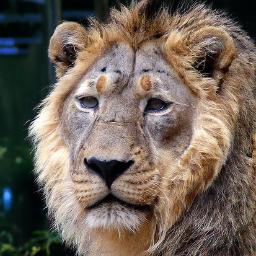} &
		\includegraphics[width=.109\linewidth]{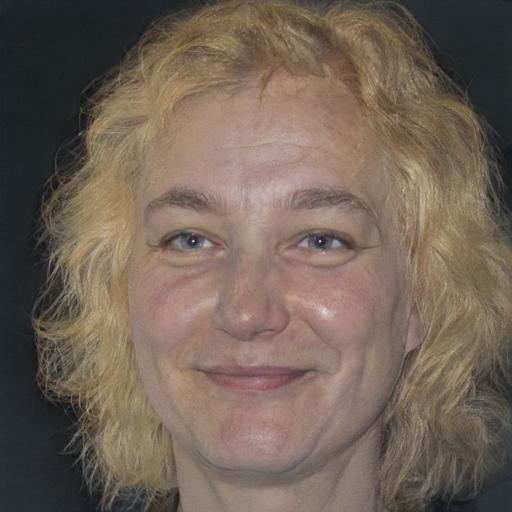}\\

		\includegraphics[width=.109\linewidth]{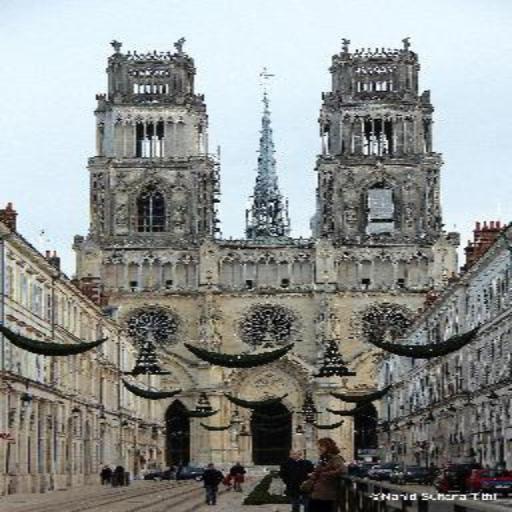} &
		\includegraphics[width=.109\linewidth]{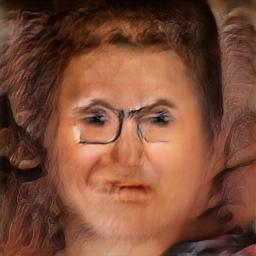} &
		\includegraphics[width=.109\linewidth]{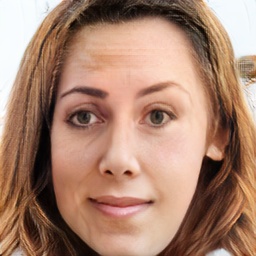} &
		\includegraphics[width=.109\linewidth]{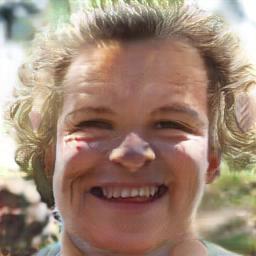} &
		\includegraphics[width=.109\linewidth]{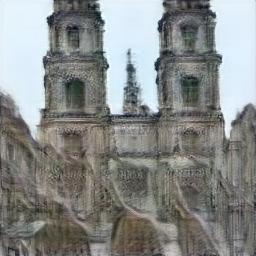} &
		\includegraphics[width=.109\linewidth]{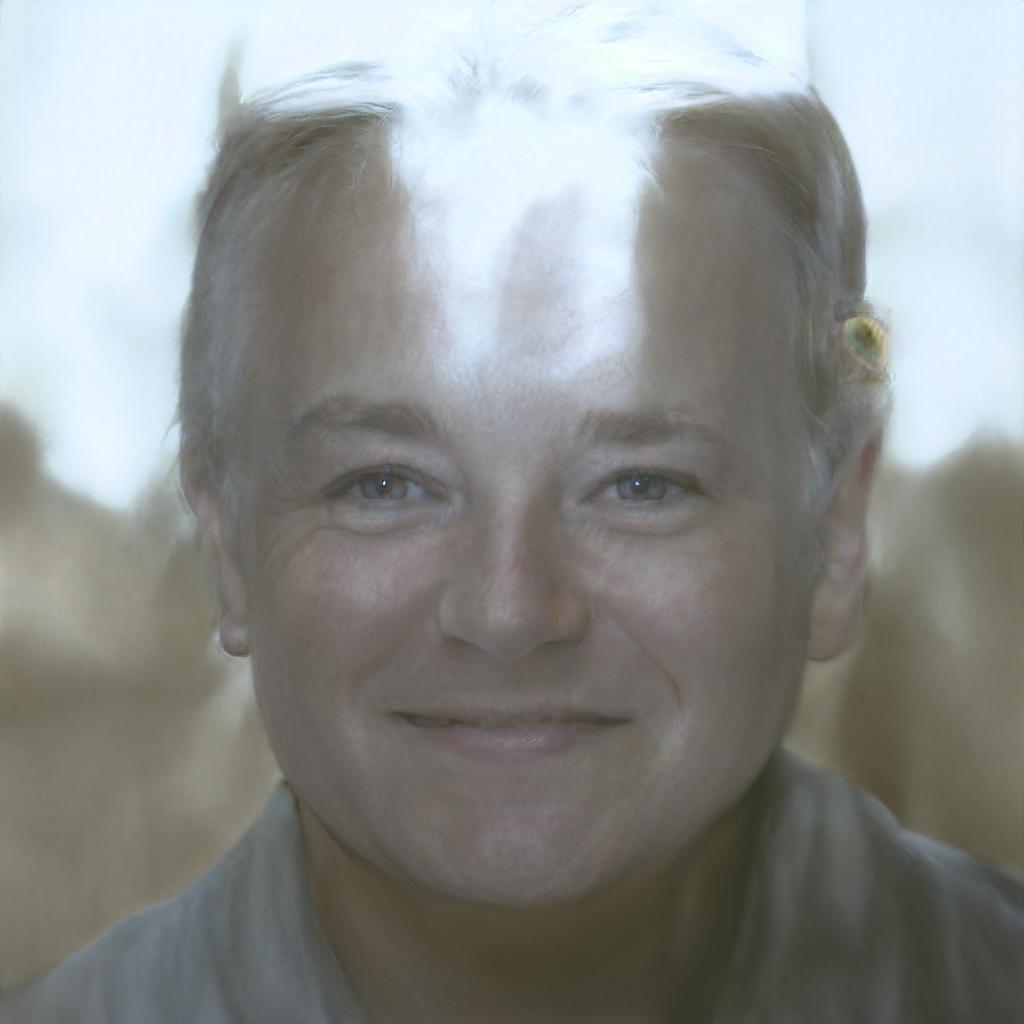} &
		\includegraphics[width=.109\linewidth]{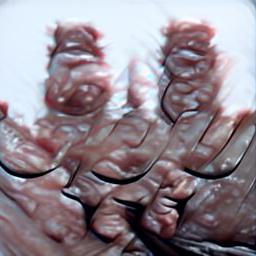} &
		\includegraphics[width=.109\linewidth]{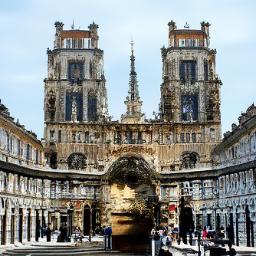} &
		\includegraphics[width=.109\linewidth]{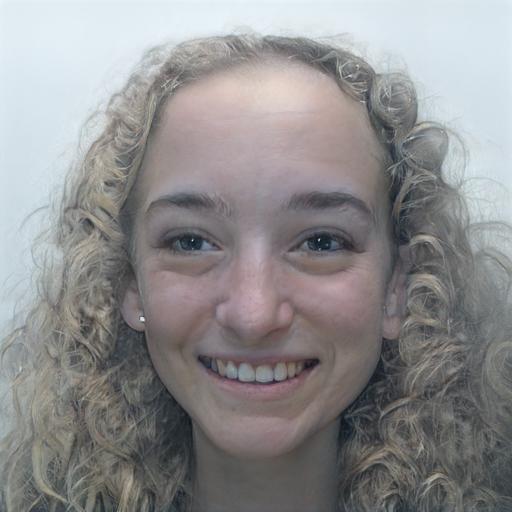}\\

		\includegraphics[width=.109\linewidth]{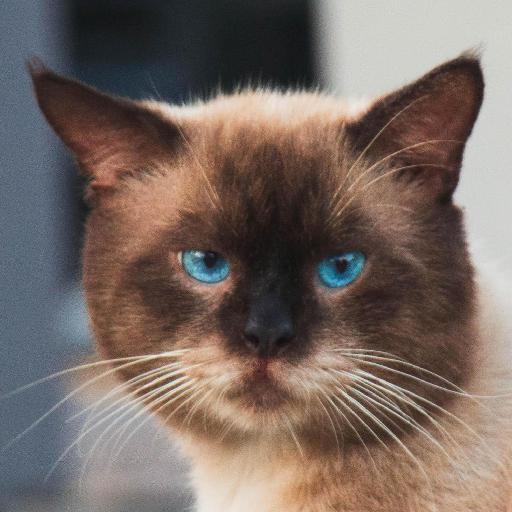} &
		\includegraphics[width=.109\linewidth]{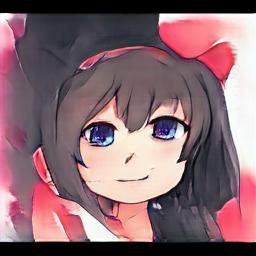} &
		\includegraphics[width=.109\linewidth]{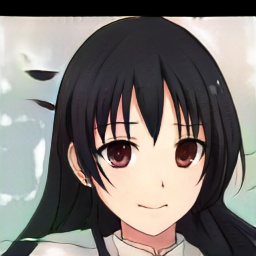} &
		\includegraphics[width=.109\linewidth]{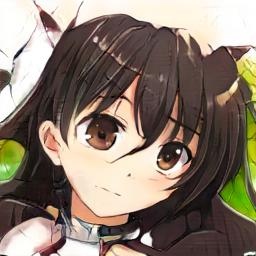} &
		\includegraphics[width=.109\linewidth]{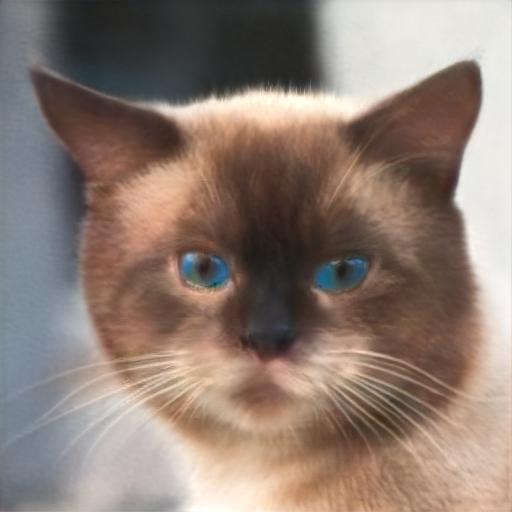} &
		\includegraphics[width=.109\linewidth]{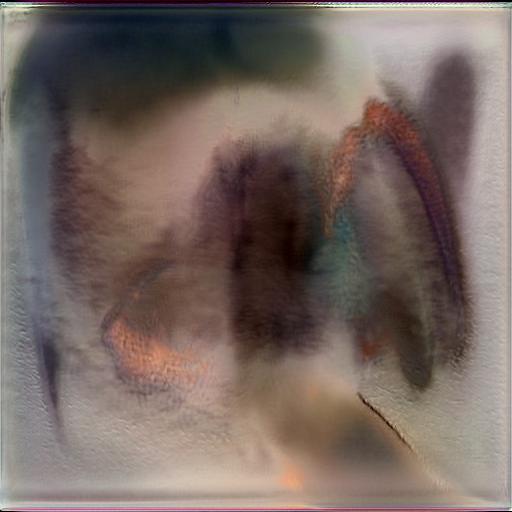} &
		\includegraphics[width=.109\linewidth]{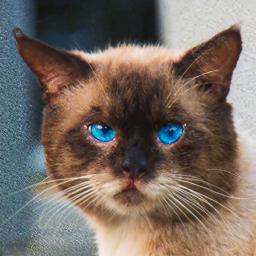} &
		\includegraphics[width=.109\linewidth]{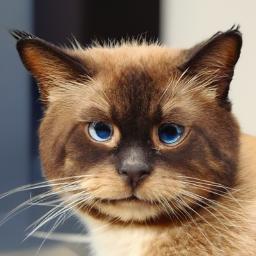} &
		\includegraphics[width=.109\linewidth]{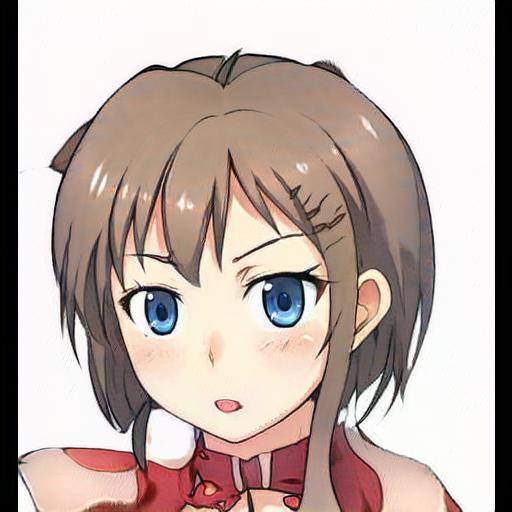}\\
		
		\small{Source} & \small{VQ-I2I} &\small{GP-UNIT}&\small{StarGAN2}&\small{DiFa}&\small{PULSE}&\footnotesize{DiffusionCLIP}&\small{DiffuseIT}&\small{UniTranslator}\\
		\small{Image}&\small{\cite{chen2022eccv}}&\small{\cite{yang2022unsupervised}}&\small{\cite{choi2020stargan}}&\small{\cite{zhang2022towards}}&\small{\cite{menon2020pulse}}&\small{\cite{Kim_2022_CVPR}}&\small{\cite{kwon2022diffusion}}&(\textbf{ours})\\
	\end{tabular}
	\caption{Comparison with the competitors. First row: Metfaces$\to$FFHQ; Second row: AFHQ-wild$\to$FFHQ; Third row: AFHQ-wild$\to$FFHQ; Fourth row: 	LSUN-Church$\to$FFHQ; Fifth row: AFHQ-cat$\to$Anime.}
	\label{fig:comparison}
\end{figure*}

We also present additional visual results generated by UniTranslator. Fig.~\ref{fig:display} showcases the translation results obtained using UniTranslator across various source-target mappings. In Fig.~\ref{fig:stylization_supp} shows the stylization results of real images, where the input images are sourced from the Internet. Fig.~\ref{fig:comparison} compares UniTranslator and state-of-the-art methods.

\begin{figure*}[t]    	
	\centering	
	\setlength{\abovecaptionskip}{0cm}
	\centering
	\setlength{\tabcolsep}{0.05em}
	\begin{tabular}{cccccccccccccccc}

				\includegraphics[width=.059\linewidth]{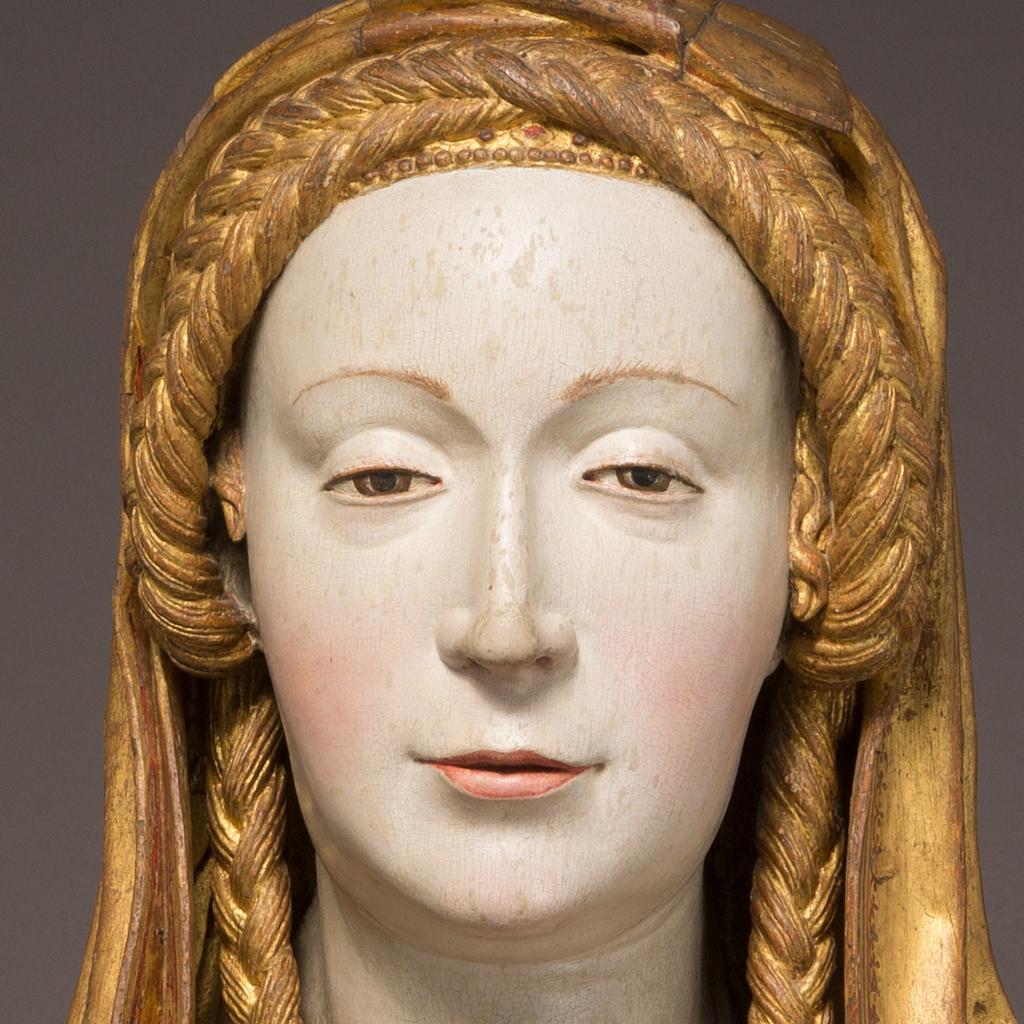} &
				\hspace{0.1mm}
				\includegraphics[width=.059\linewidth]{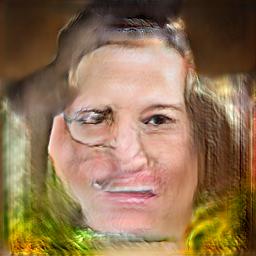} &
				\includegraphics[width=.059\linewidth]{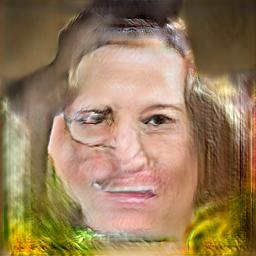} &
				\includegraphics[width=.059\linewidth]{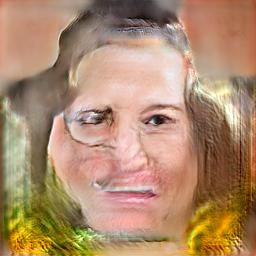} &
				\hspace{0.1mm}
				\includegraphics[width=.059\linewidth]{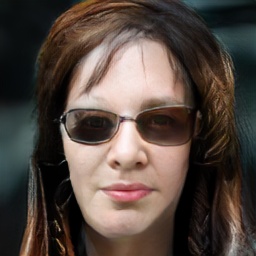} &
				\includegraphics[width=.059\linewidth]{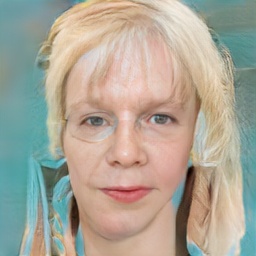} &
				\includegraphics[width=.059\linewidth]{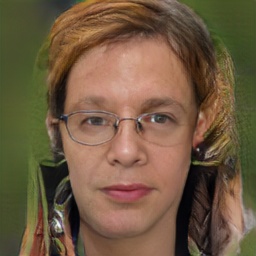} &
				\hspace{0.1mm}
				
				\includegraphics[width=.059\linewidth]{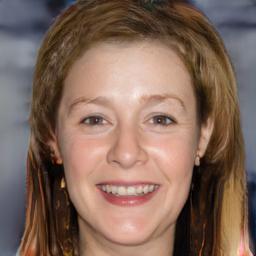} &
				\includegraphics[width=.059\linewidth]{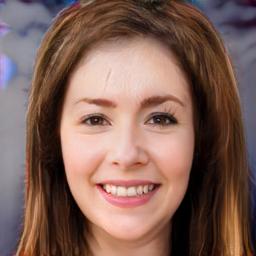} &
				\includegraphics[width=.059\linewidth]{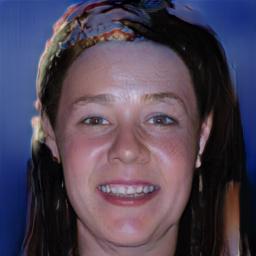} &
				\hspace{0.1mm}	
				\includegraphics[width=.059\linewidth]{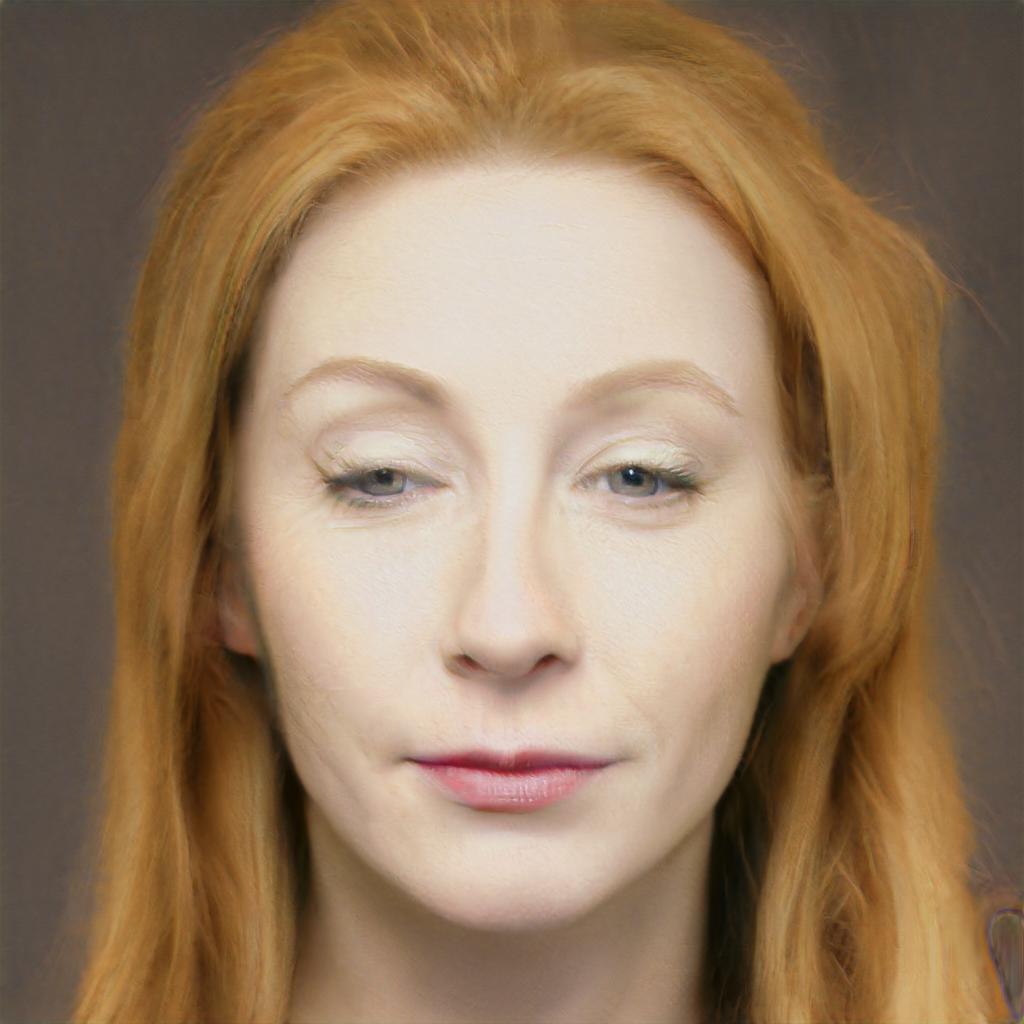} &
				\includegraphics[width=.059\linewidth]{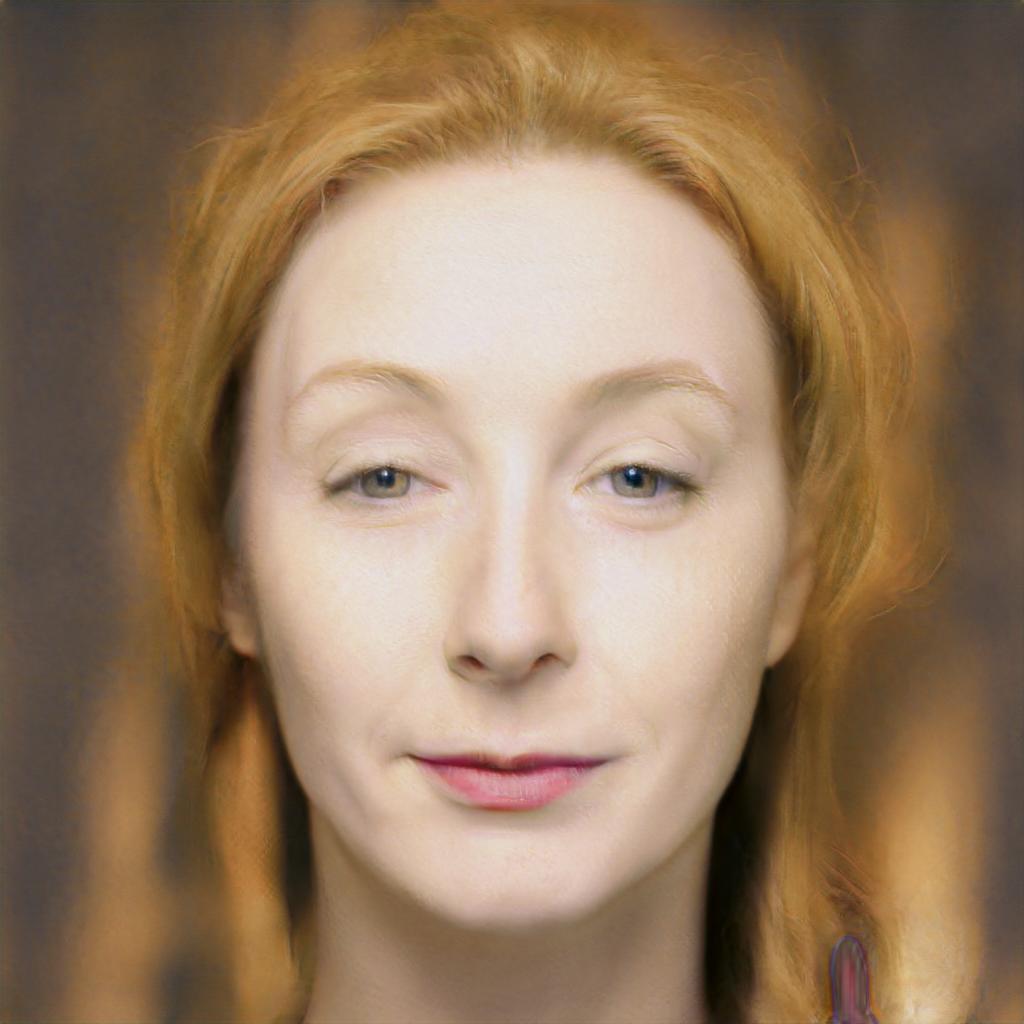} &
				\includegraphics[width=.059\linewidth]{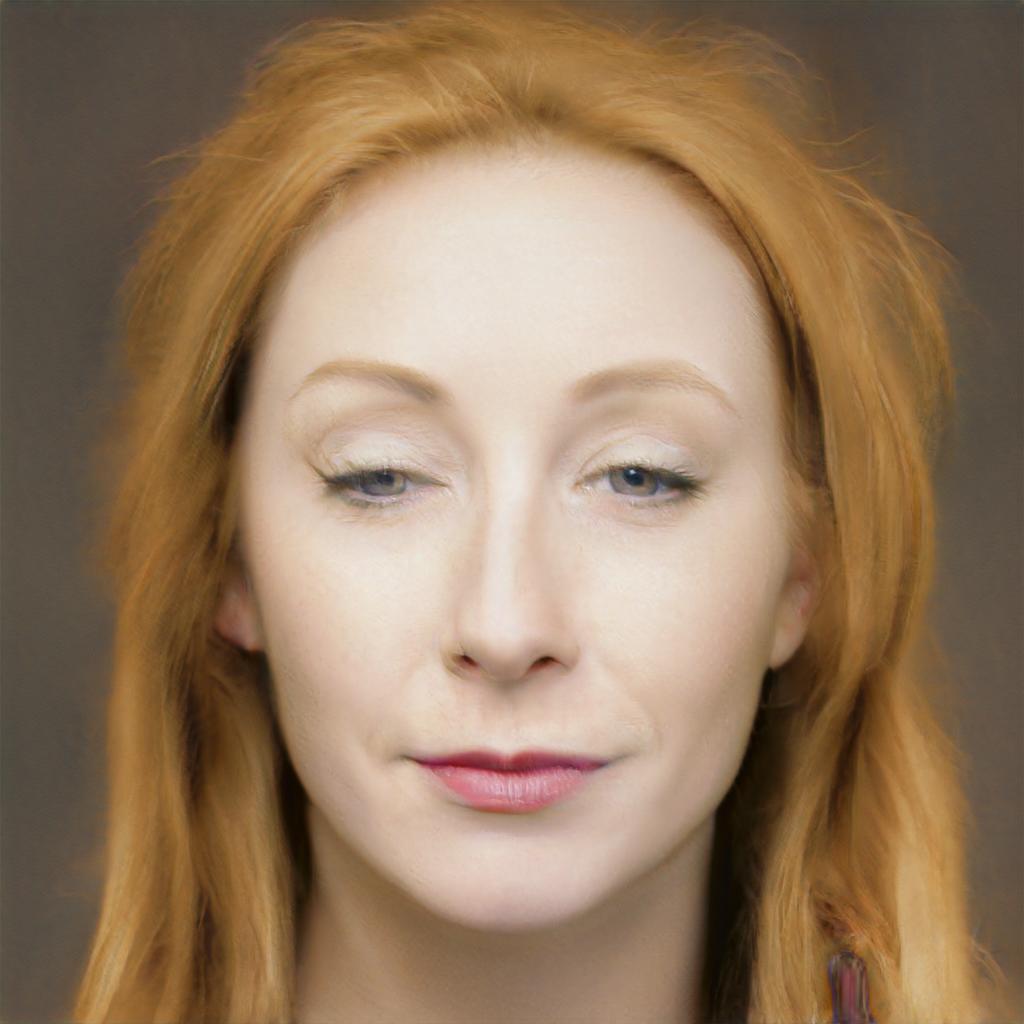} &
				\hspace{0.1mm}
				\includegraphics[width=.059\linewidth]{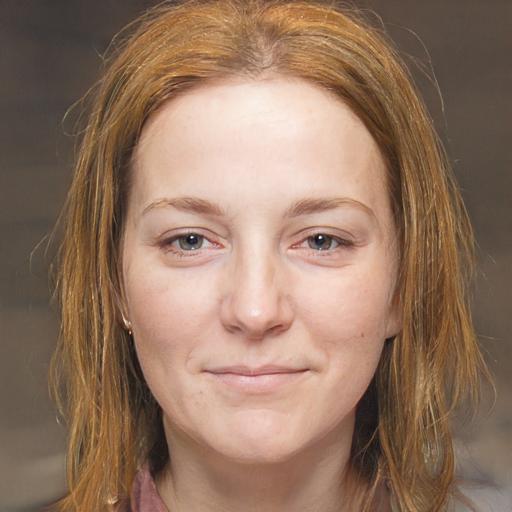} &
				\includegraphics[width=.059\linewidth]{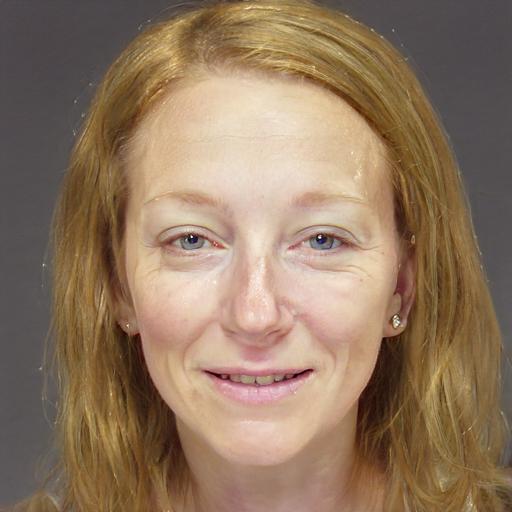} &
				\includegraphics[width=.059\linewidth]{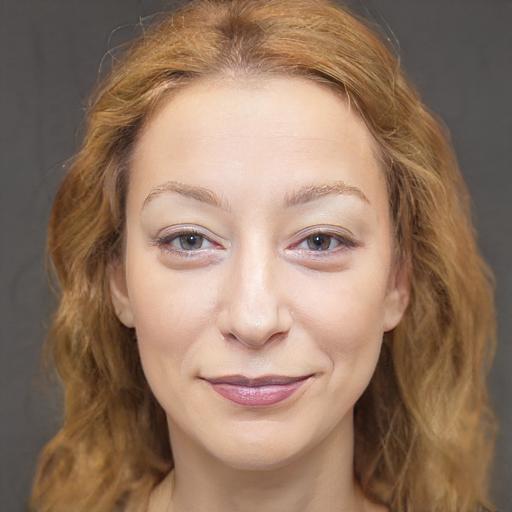}
				\\					
				
				\includegraphics[width=.059\linewidth]{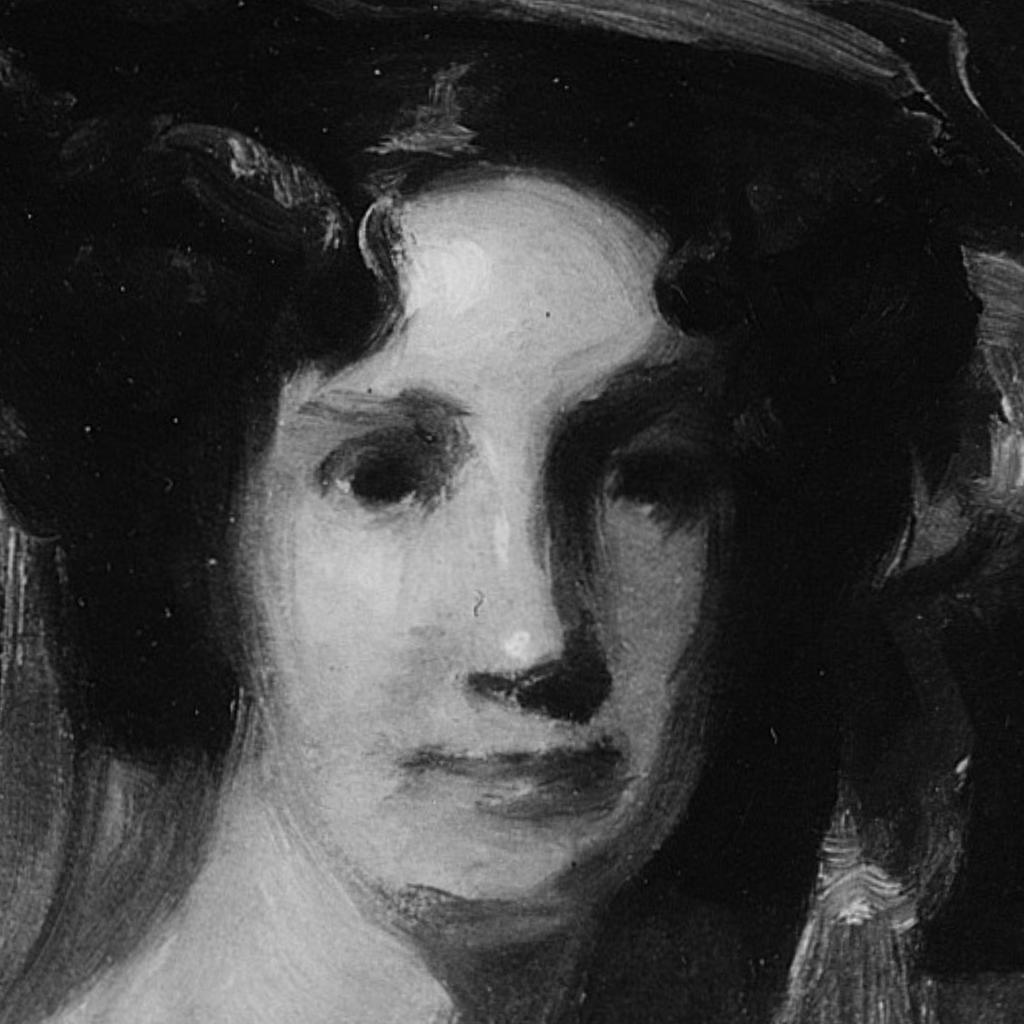} &
				\hspace{0.1mm}
				\includegraphics[width=.059\linewidth]{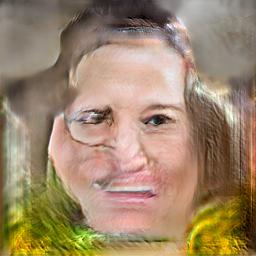} &
				\includegraphics[width=.059\linewidth]{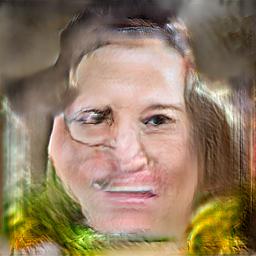} &
				\includegraphics[width=.059\linewidth]{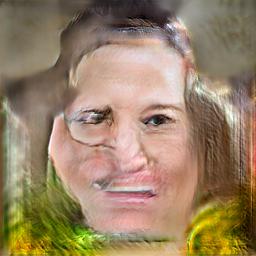} &
				\hspace{0.1mm}
				\includegraphics[width=.059\linewidth]{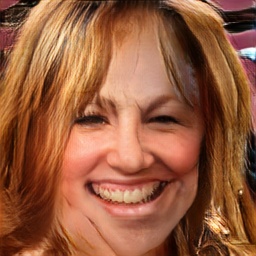} &
				\includegraphics[width=.059\linewidth]{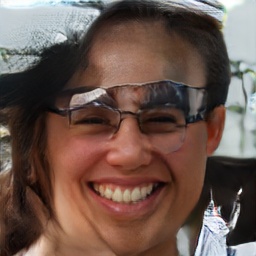} &
				\includegraphics[width=.059\linewidth]{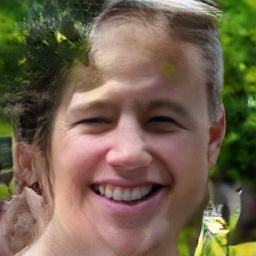} &
				\hspace{0.1mm}
				\includegraphics[width=.059\linewidth]{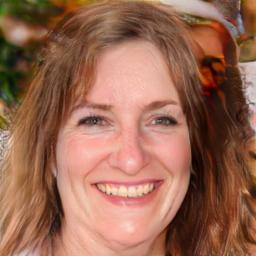} &
				\includegraphics[width=.059\linewidth]{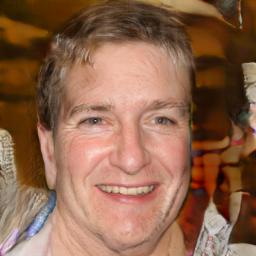} &
				\includegraphics[width=.059\linewidth]{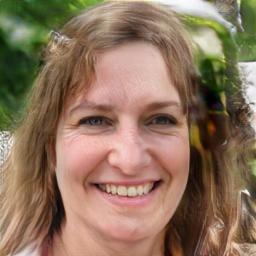} &
				\hspace{0.1mm}	
				\includegraphics[width=.059\linewidth]{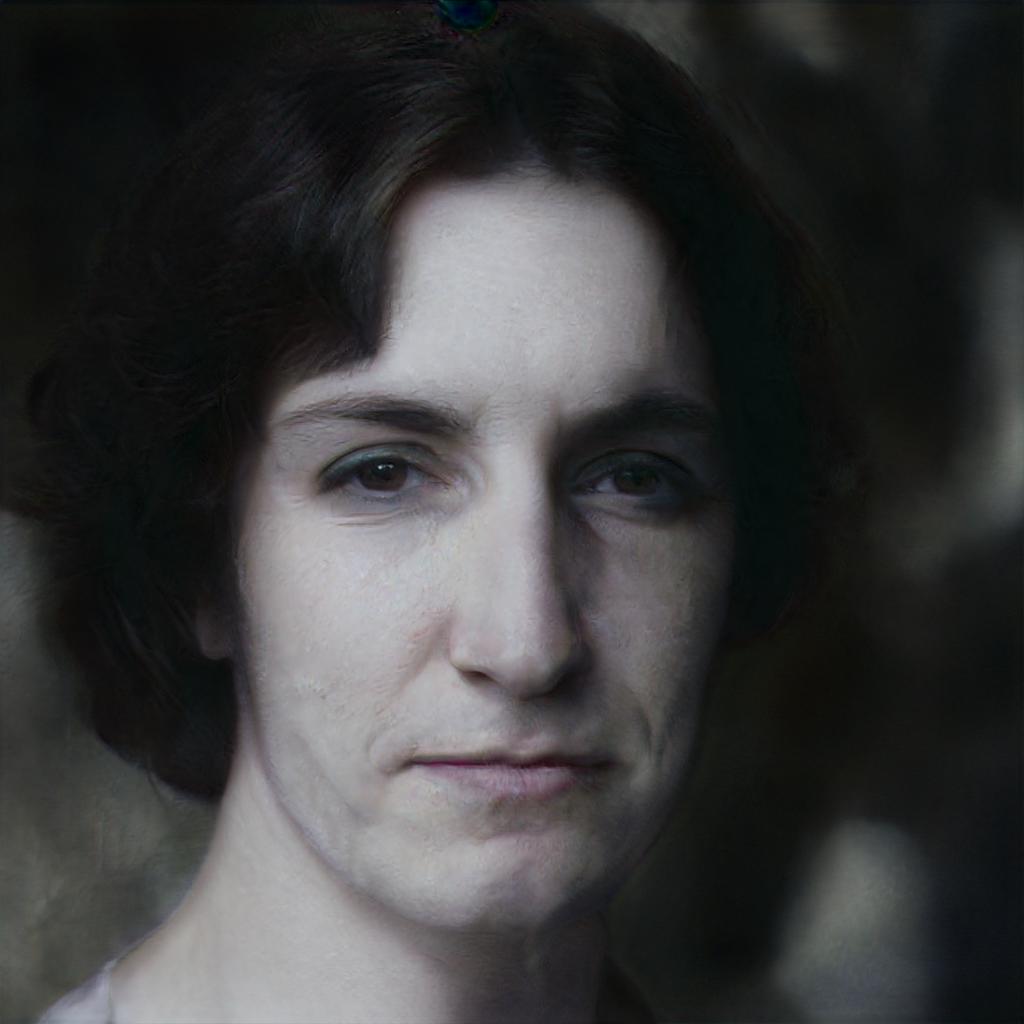} &
				\includegraphics[width=.059\linewidth]{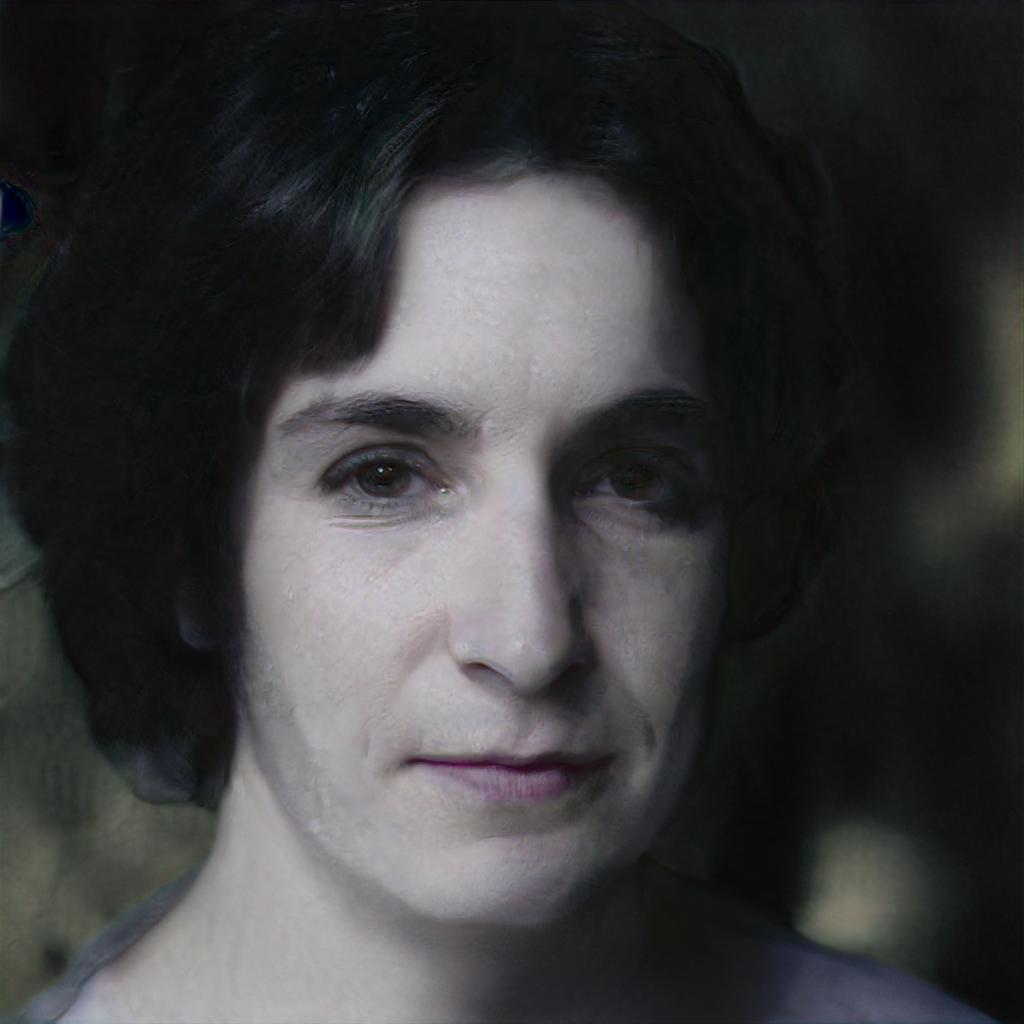} &
				\includegraphics[width=.059\linewidth]{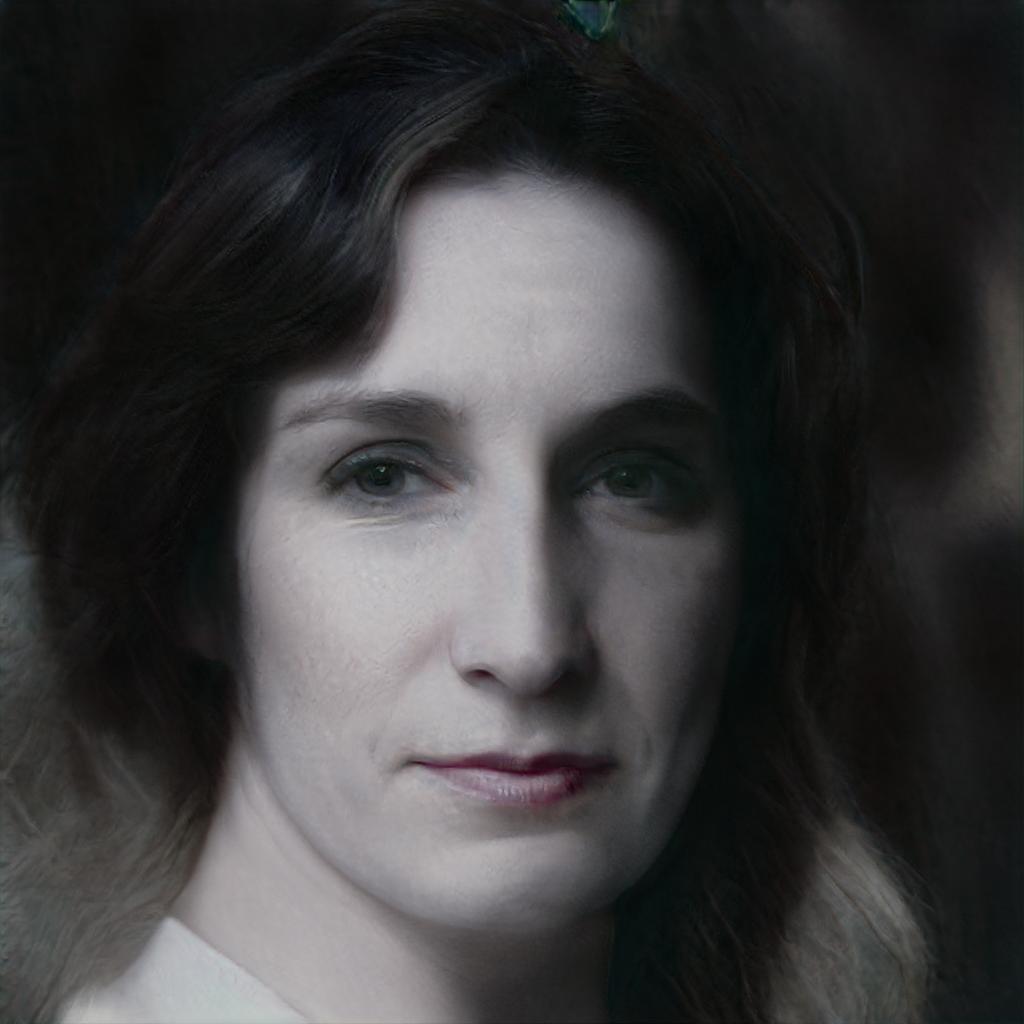} &
				\hspace{0.1mm}
				\includegraphics[width=.059\linewidth]{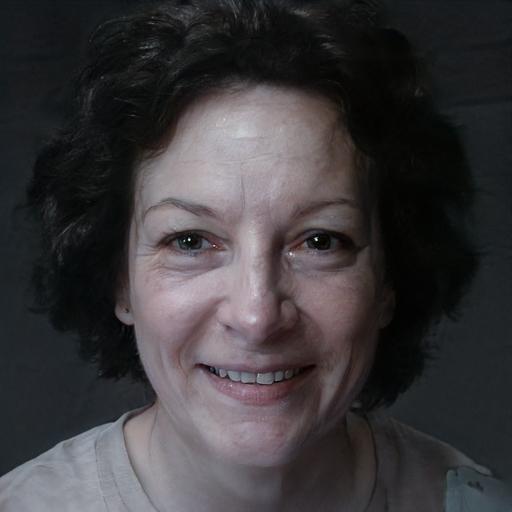} &
				\includegraphics[width=.059\linewidth]{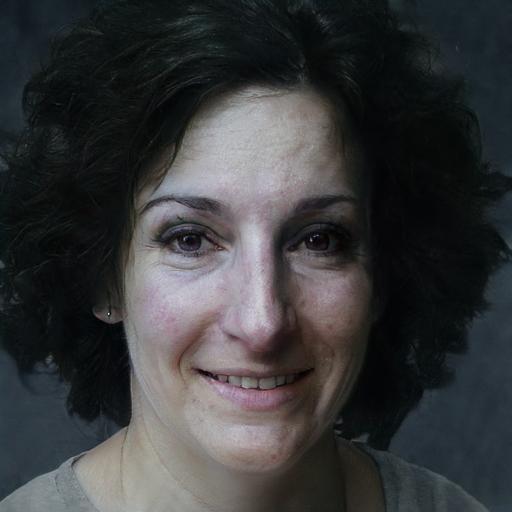} &
				\includegraphics[width=.059\linewidth]{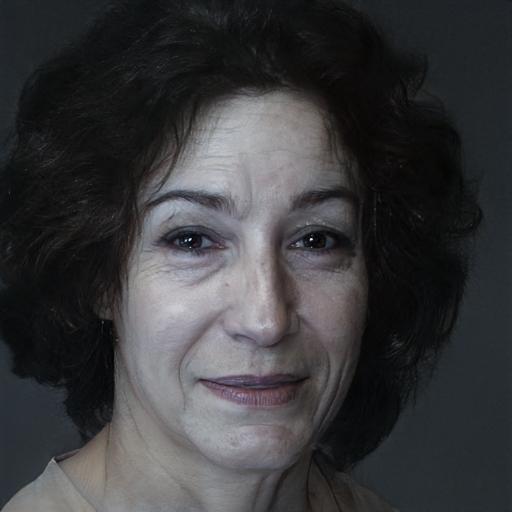}
				\\					
				
				\includegraphics[width=.059\linewidth]{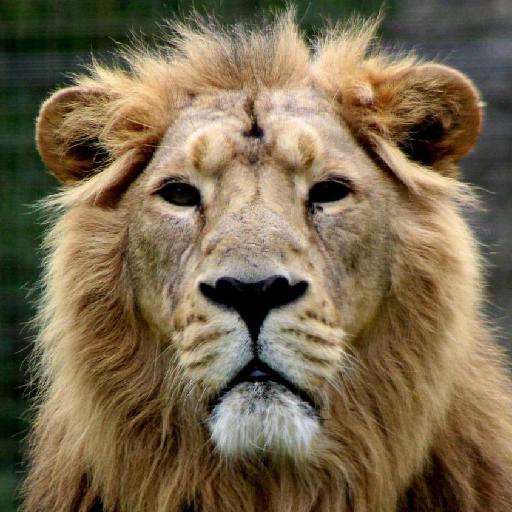} &
				\hspace{0.1mm}
				\includegraphics[width=.059\linewidth]{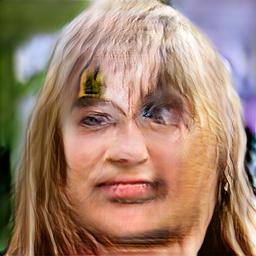} &
				\includegraphics[width=.059\linewidth]{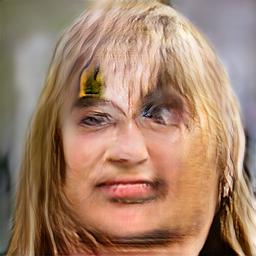} &
				\includegraphics[width=.059\linewidth]{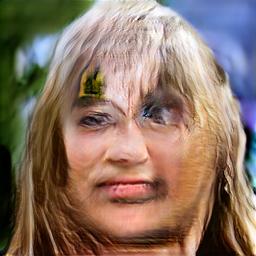} &
				\hspace{0.1mm}
				\includegraphics[width=.059\linewidth]{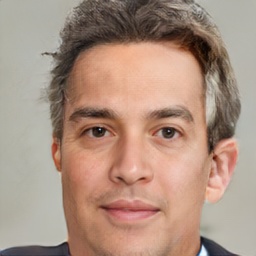} &
				\includegraphics[width=.059\linewidth]{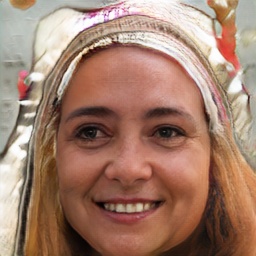} &
				\includegraphics[width=.059\linewidth]{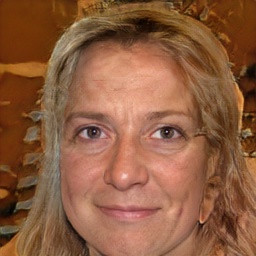} &
				\hspace{0.1mm}
				\includegraphics[width=.059\linewidth]{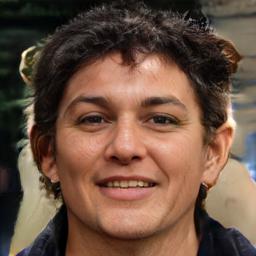} &
				\includegraphics[width=.059\linewidth]{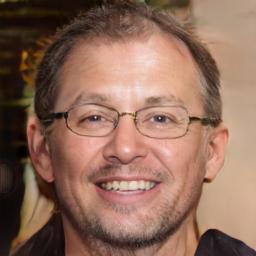} &
				\includegraphics[width=.059\linewidth]{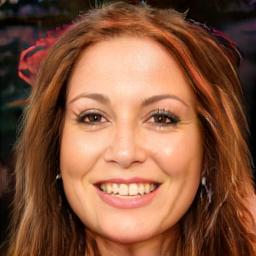} &
				\hspace{0.1mm}
				\includegraphics[width=.059\linewidth]{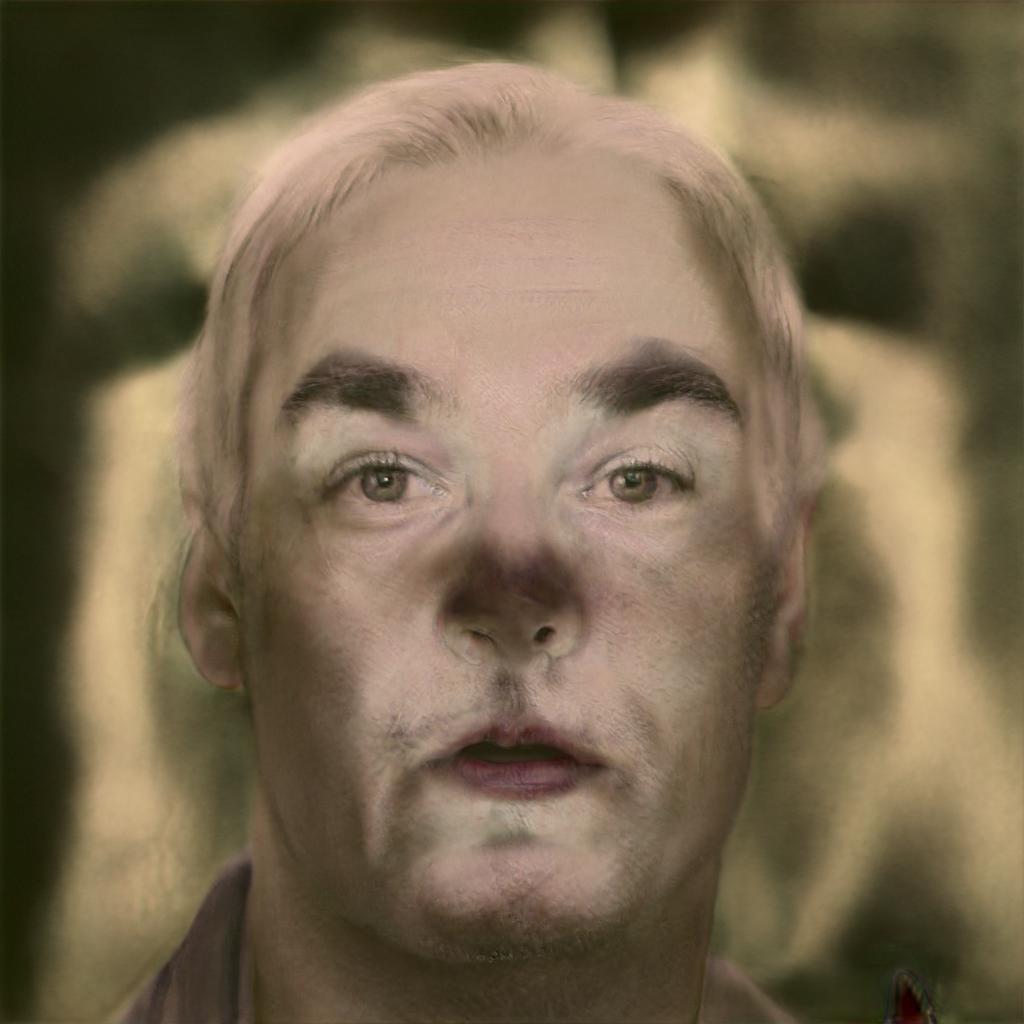} &
				\includegraphics[width=.059\linewidth]{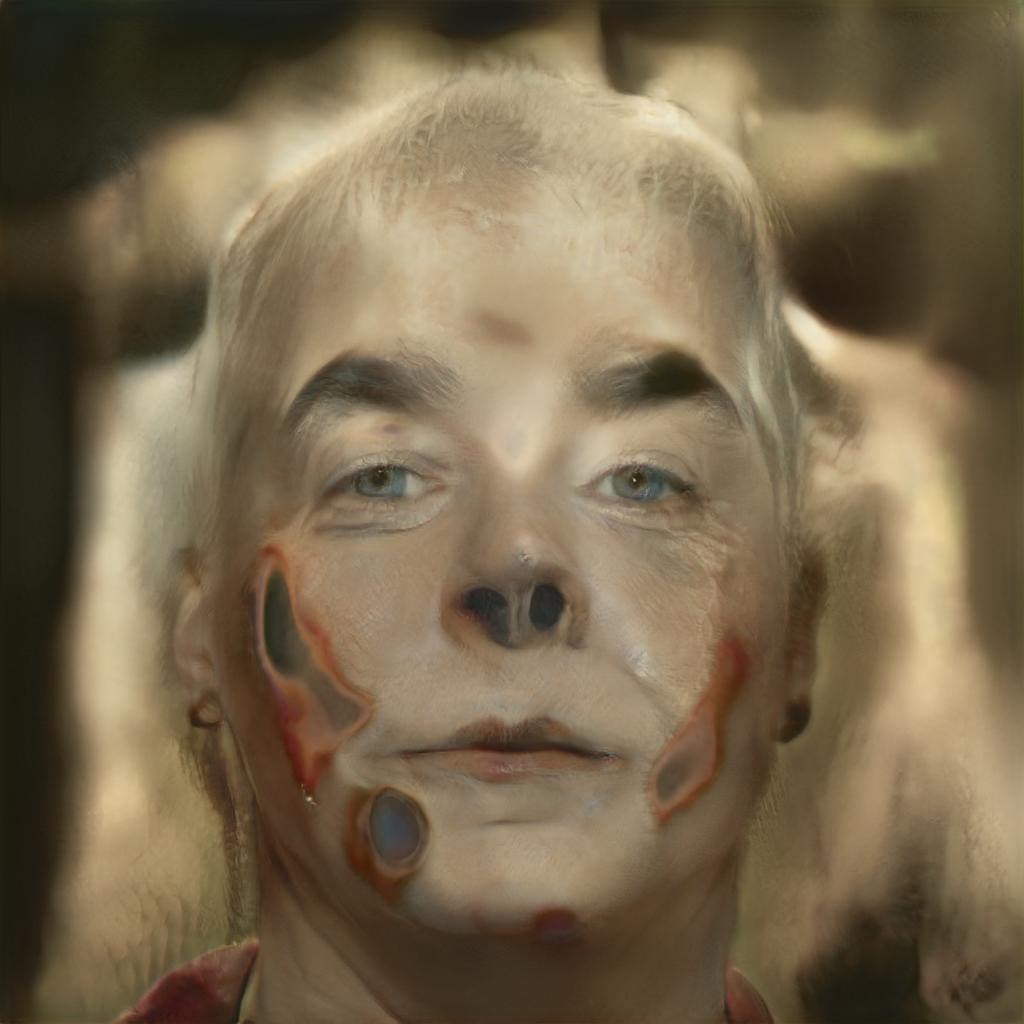} &
				\includegraphics[width=.059\linewidth]{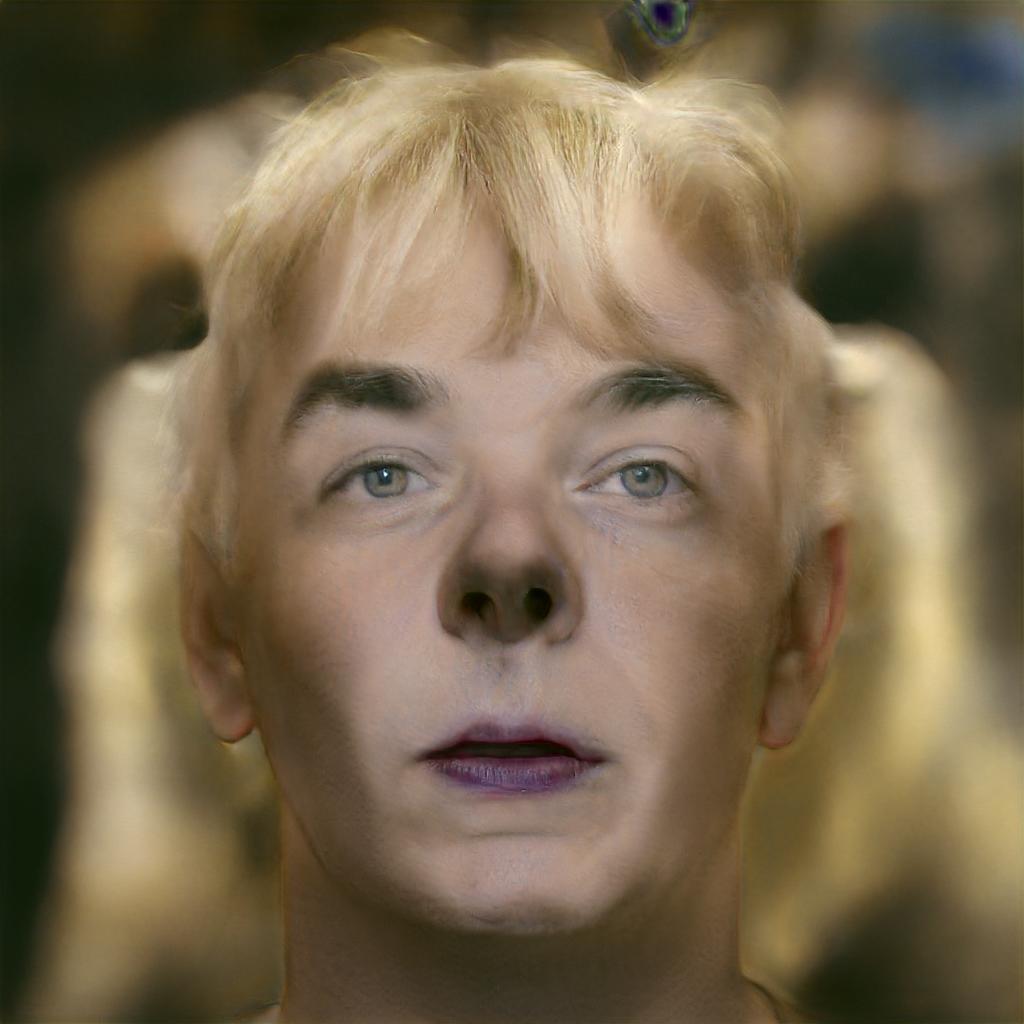} &
				\hspace{0.1mm}
				\includegraphics[width=.059\linewidth]{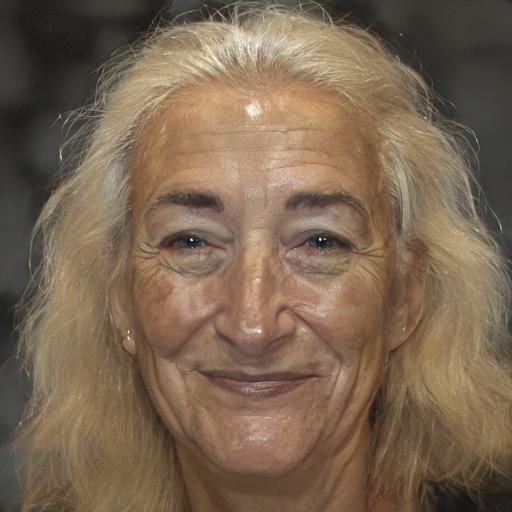} &
				\includegraphics[width=.059\linewidth]{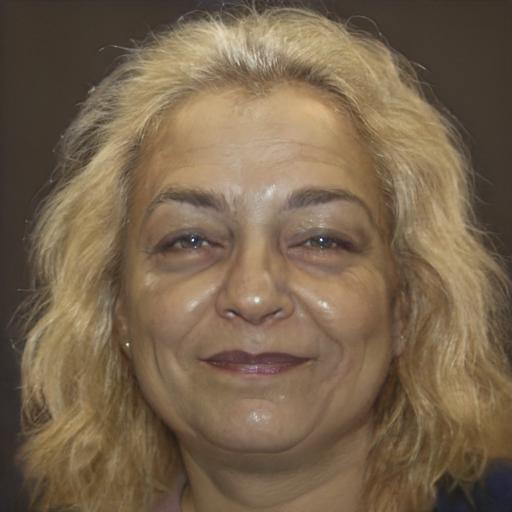} &
				\includegraphics[width=.059\linewidth]{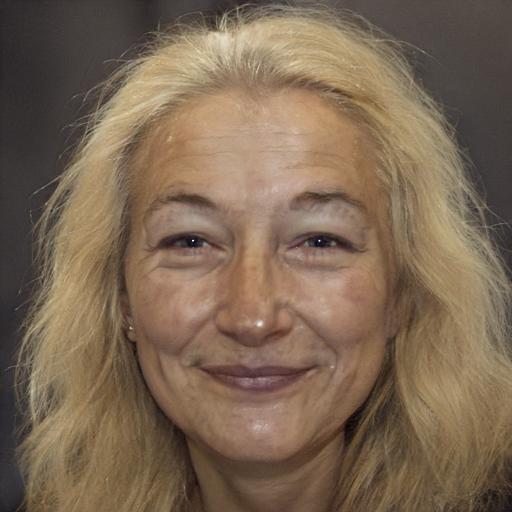}
				\\	
				
				\includegraphics[width=.059\linewidth]{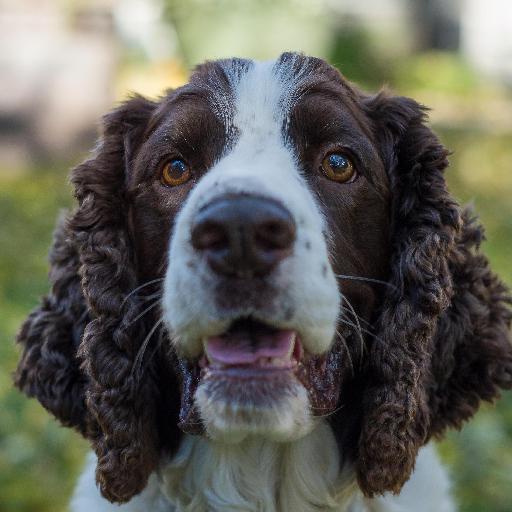} &
				\hspace{0.1mm}
				\includegraphics[width=.059\linewidth]{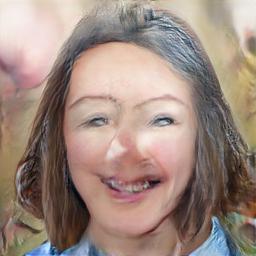} &
				\includegraphics[width=.059\linewidth]{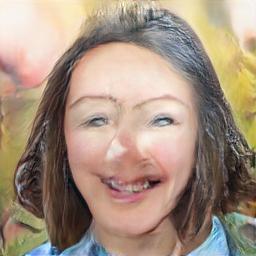} &
				\includegraphics[width=.059\linewidth]{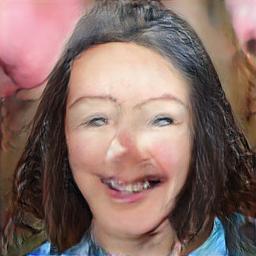} &
				\hspace{0.1mm}
				\includegraphics[width=.059\linewidth]{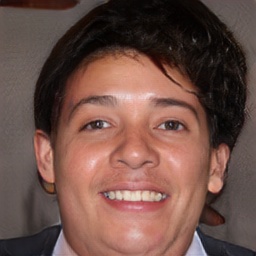} &
				\includegraphics[width=.059\linewidth]{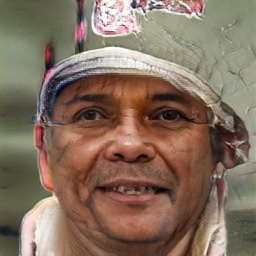} &
				\includegraphics[width=.059\linewidth]{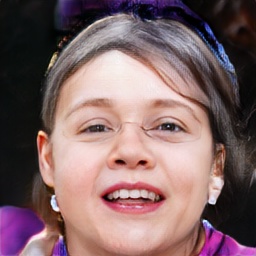} &
				\hspace{0.1mm}
				\includegraphics[width=.059\linewidth]{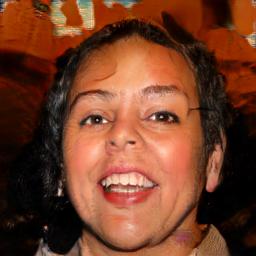} &
				\includegraphics[width=.059\linewidth]{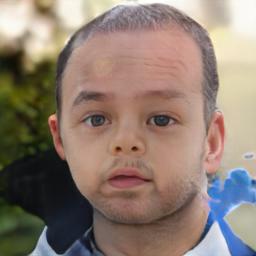} &
				\includegraphics[width=.059\linewidth]{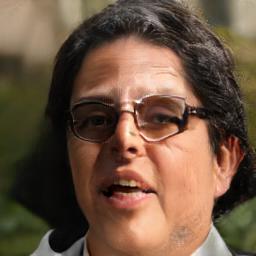} &
				\hspace{0.1mm}
				\includegraphics[width=.059\linewidth]{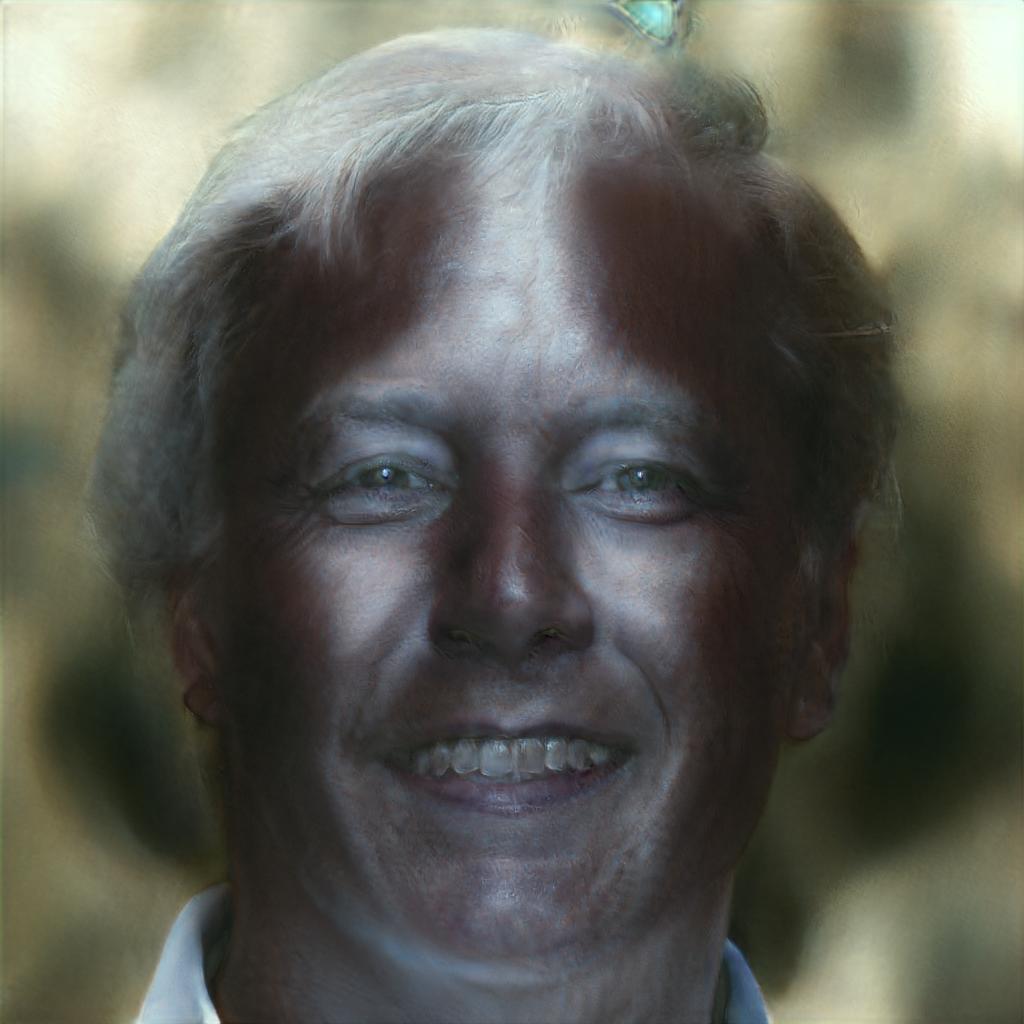} &
				\includegraphics[width=.059\linewidth]{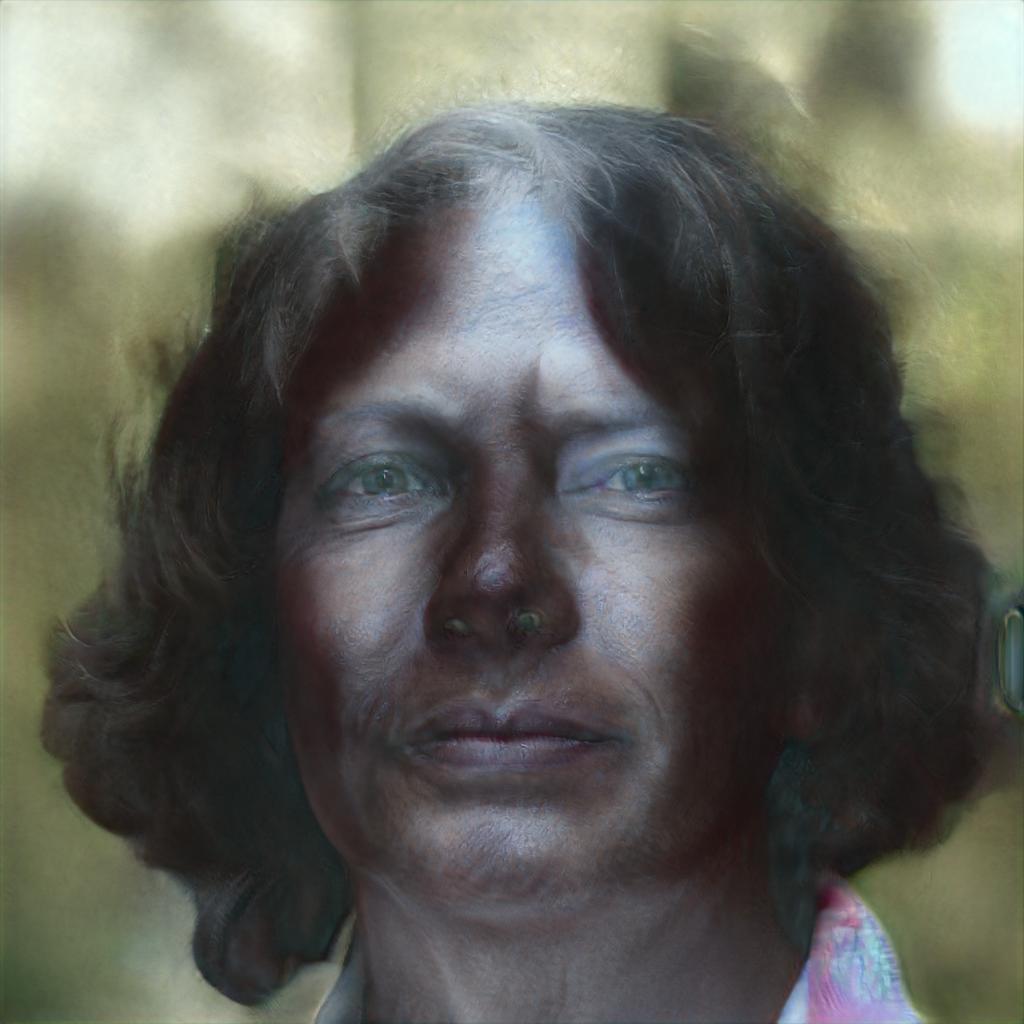} &
				\includegraphics[width=.059\linewidth]{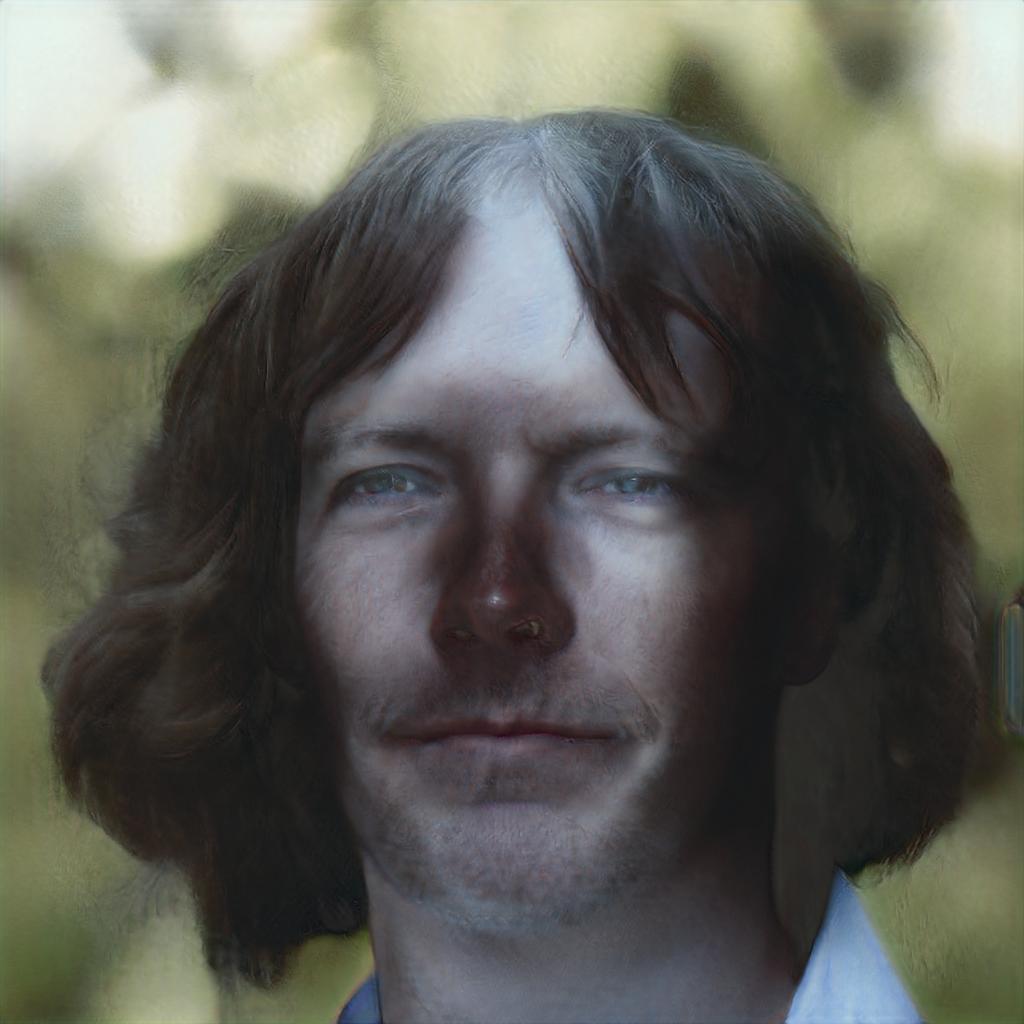} &
				\hspace{0.1mm}
				\includegraphics[width=.059\linewidth]{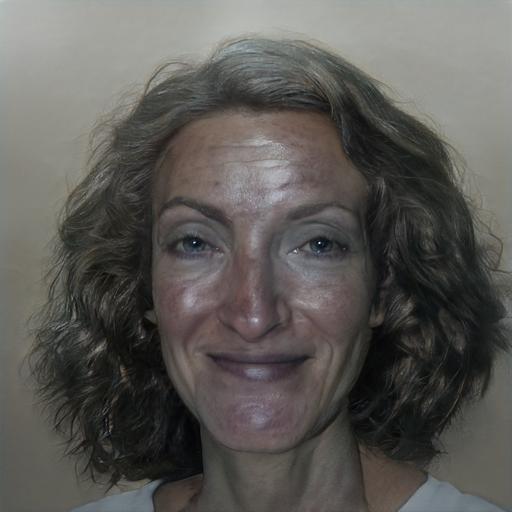} &
				\includegraphics[width=.059\linewidth]{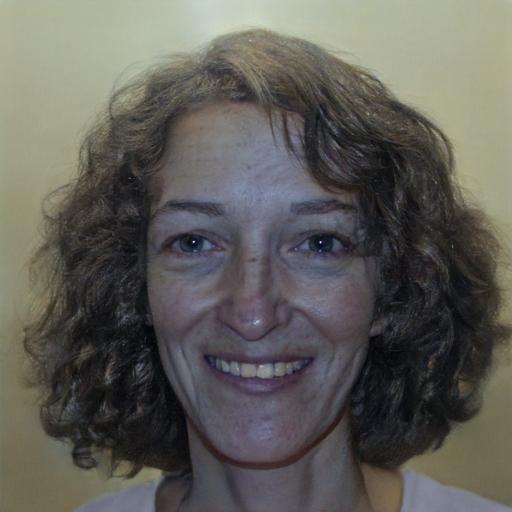} &
				\includegraphics[width=.059\linewidth]{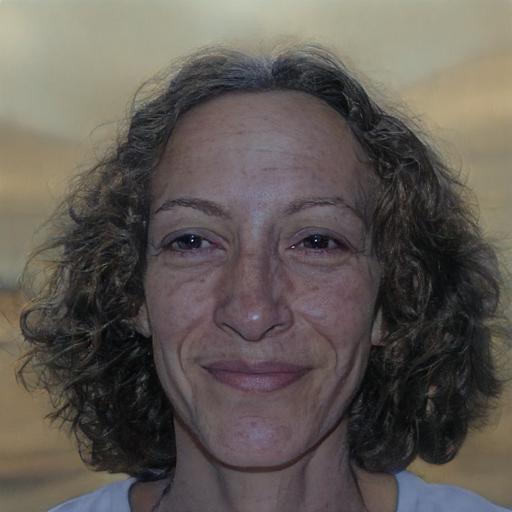}
				\\
								
				\includegraphics[width=.059\linewidth]{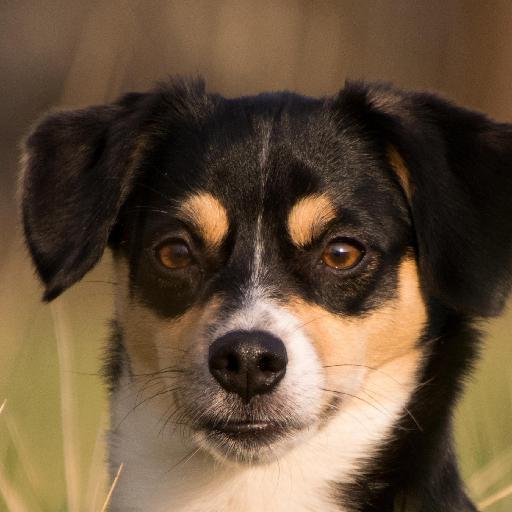} &
				\hspace{0.1mm}
				\includegraphics[width=.059\linewidth]{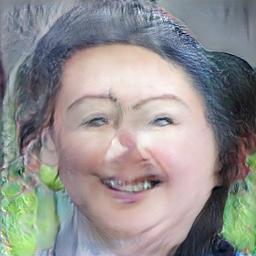} &
				\includegraphics[width=.059\linewidth]{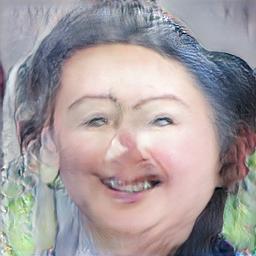} &
				\includegraphics[width=.059\linewidth]{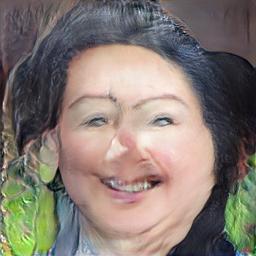} &
				\hspace{0.1mm}
				\includegraphics[width=.059\linewidth]{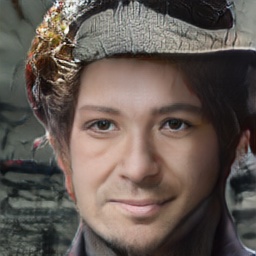} &
				\includegraphics[width=.059\linewidth]{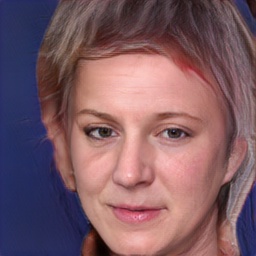} &
				\includegraphics[width=.059\linewidth]{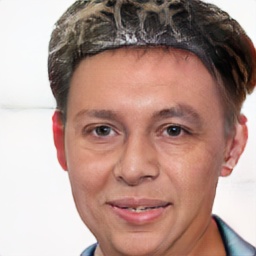} &
				\hspace{0.1mm}
				\includegraphics[width=.059\linewidth]{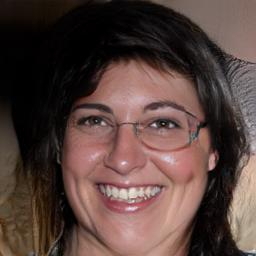} &
				\includegraphics[width=.059\linewidth]{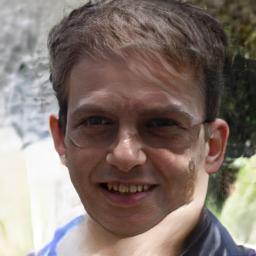} &
				\includegraphics[width=.059\linewidth]{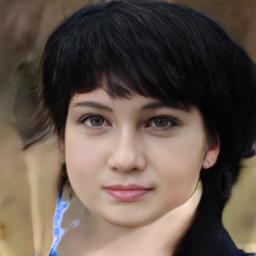} &
				\hspace{0.1mm}
				\includegraphics[width=.059\linewidth]{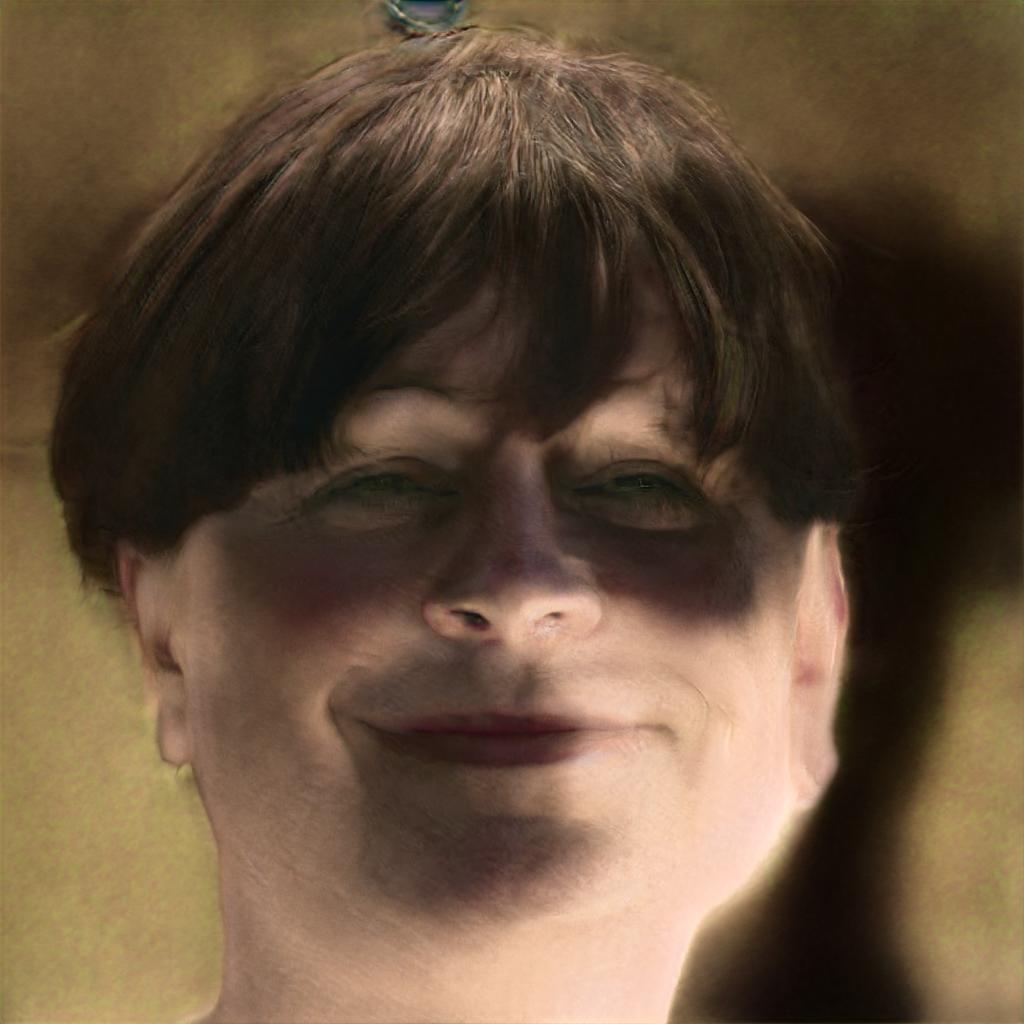} &
				\includegraphics[width=.059\linewidth]{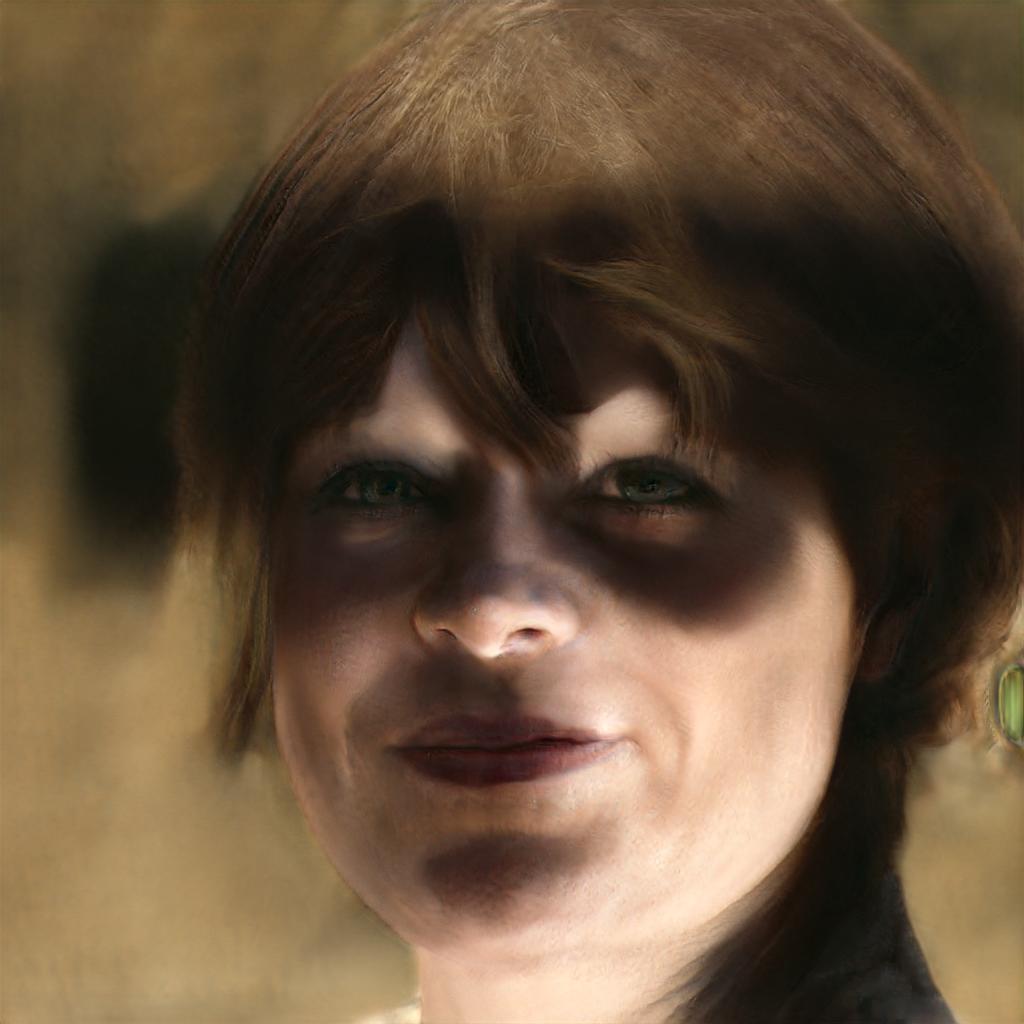} &
				\includegraphics[width=.059\linewidth]{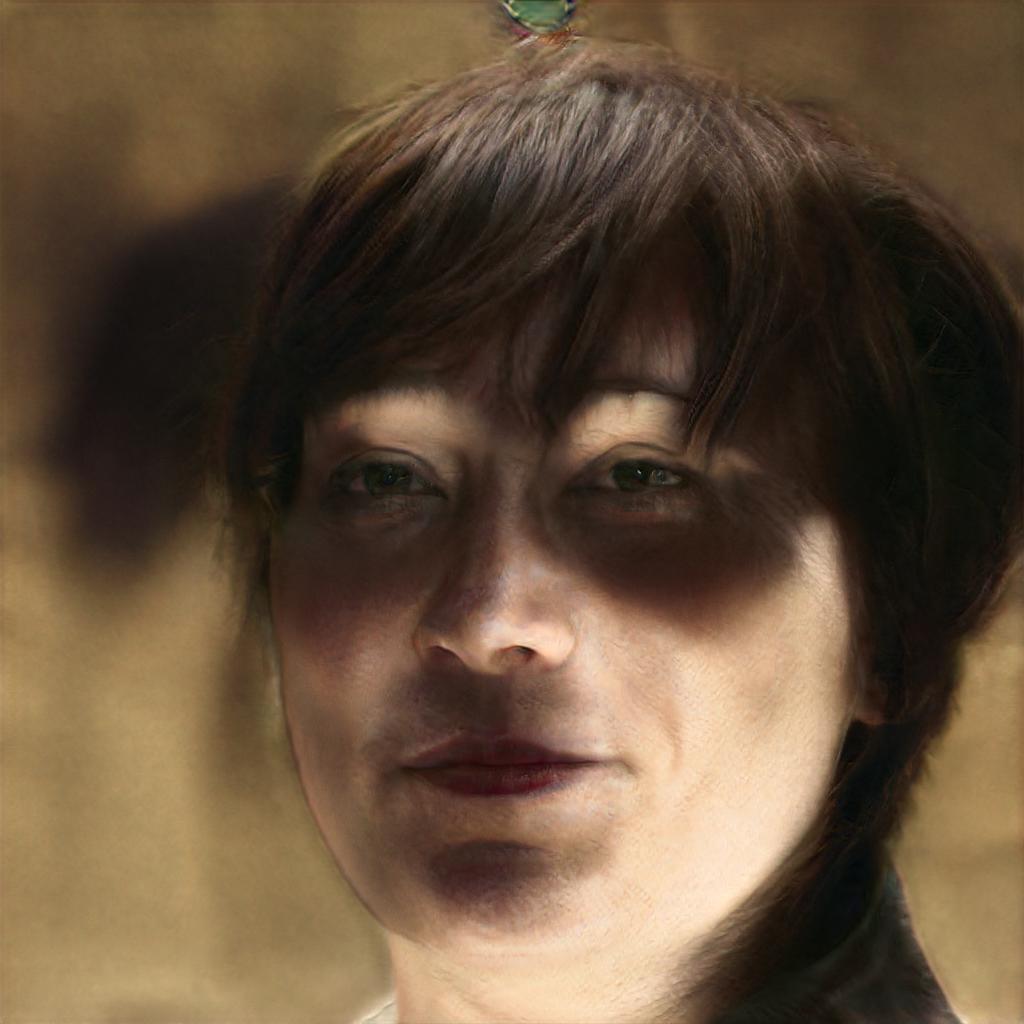} &
				\hspace{0.1mm}
				\includegraphics[width=.059\linewidth]{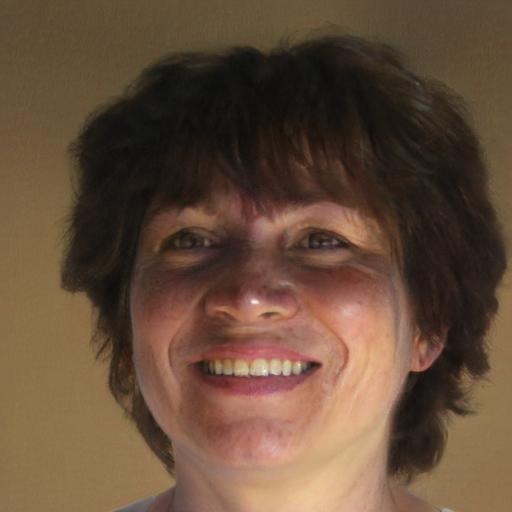} &
				\includegraphics[width=.059\linewidth]{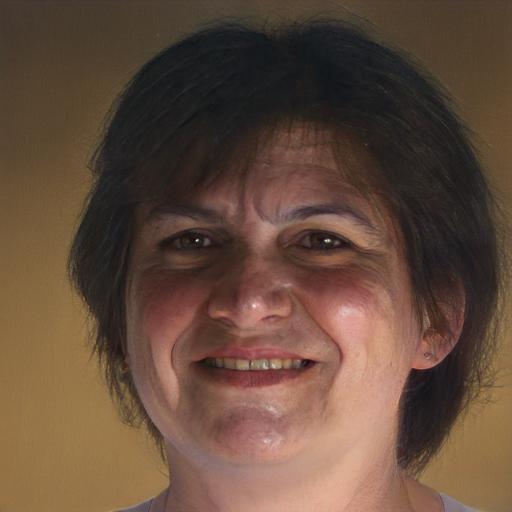} &
				\includegraphics[width=.059\linewidth]{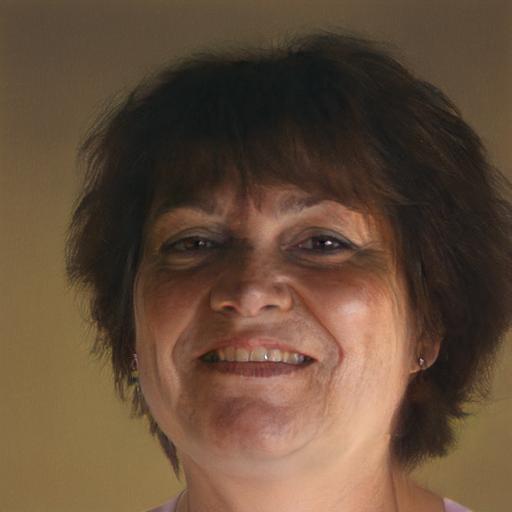}
				\\
							
				\includegraphics[width=.059\linewidth]{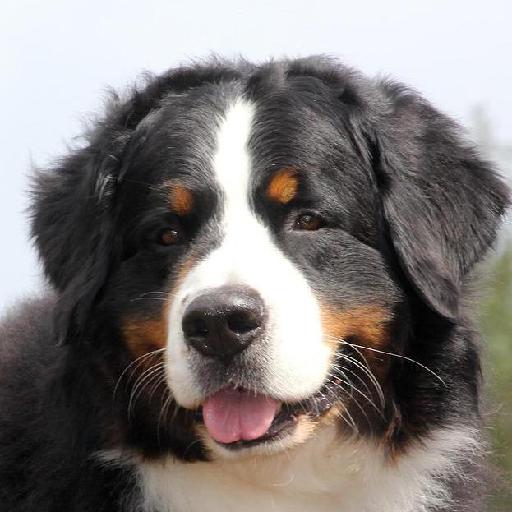} &
				\hspace{0.1mm}
				\includegraphics[width=.059\linewidth]{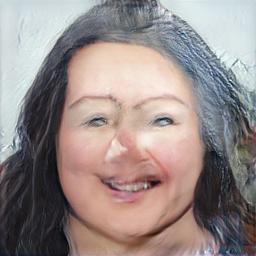} &
				\includegraphics[width=.059\linewidth]{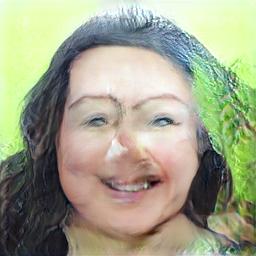} &
				\includegraphics[width=.059\linewidth]{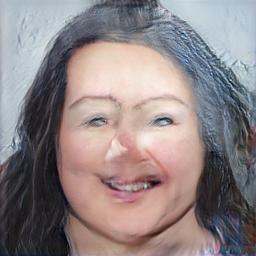} &
				\hspace{0.1mm}
				\includegraphics[width=.059\linewidth]{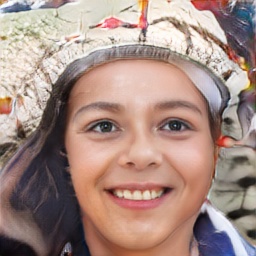} &
				\includegraphics[width=.059\linewidth]{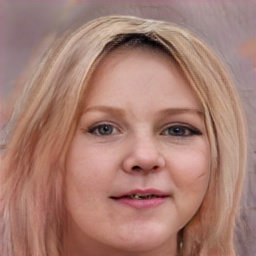} &
				\includegraphics[width=.059\linewidth]{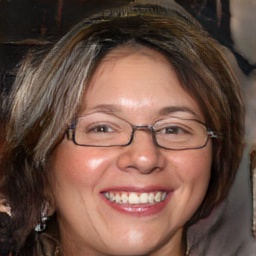} &
				\hspace{0.1mm}
				\includegraphics[width=.059\linewidth]{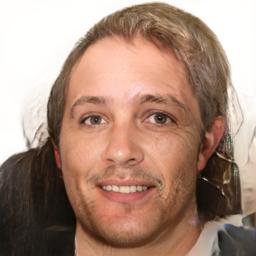} &
				\includegraphics[width=.059\linewidth]{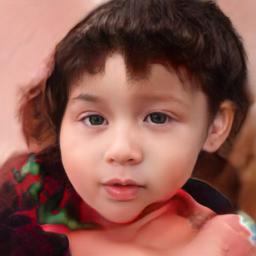} &
				\includegraphics[width=.059\linewidth]{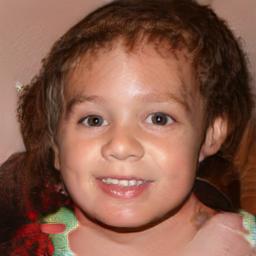} &
				\hspace{0.1mm}
				\includegraphics[width=.059\linewidth]{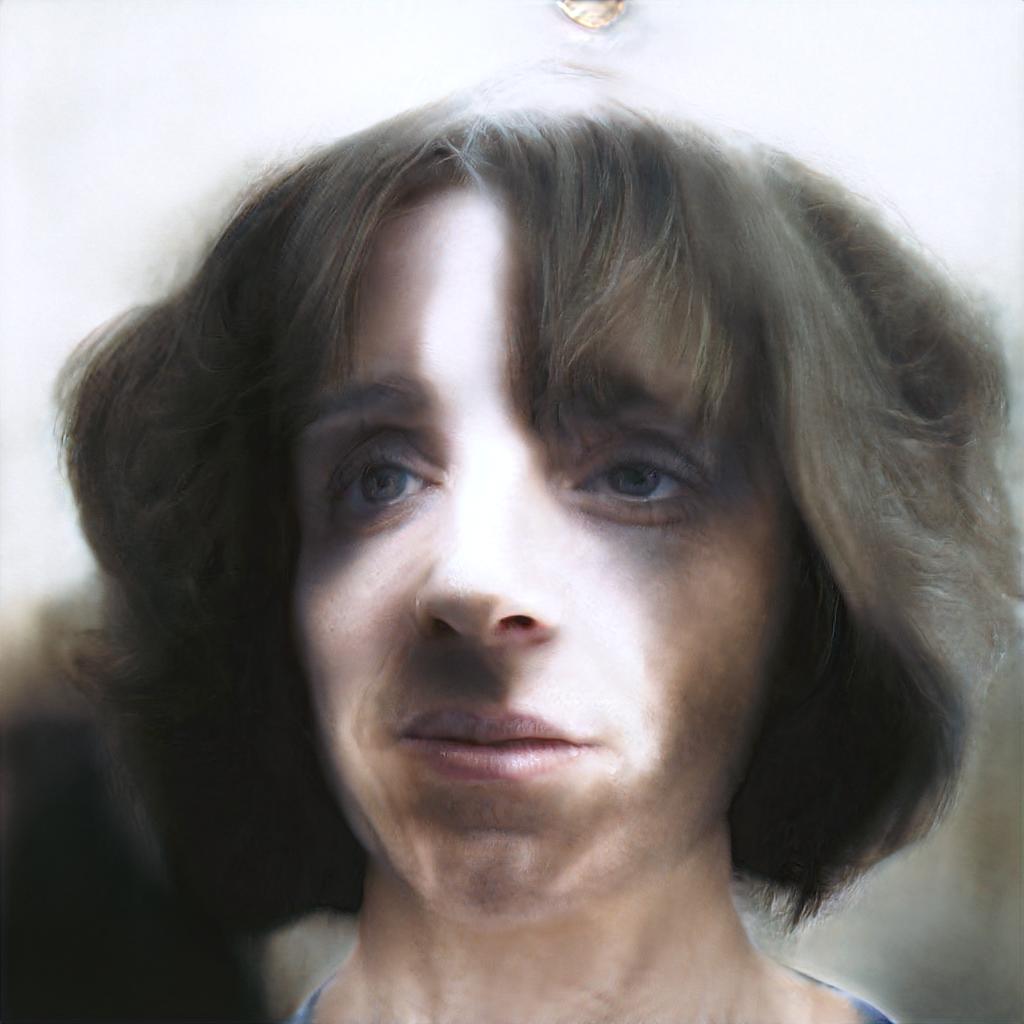} &
				\includegraphics[width=.059\linewidth]{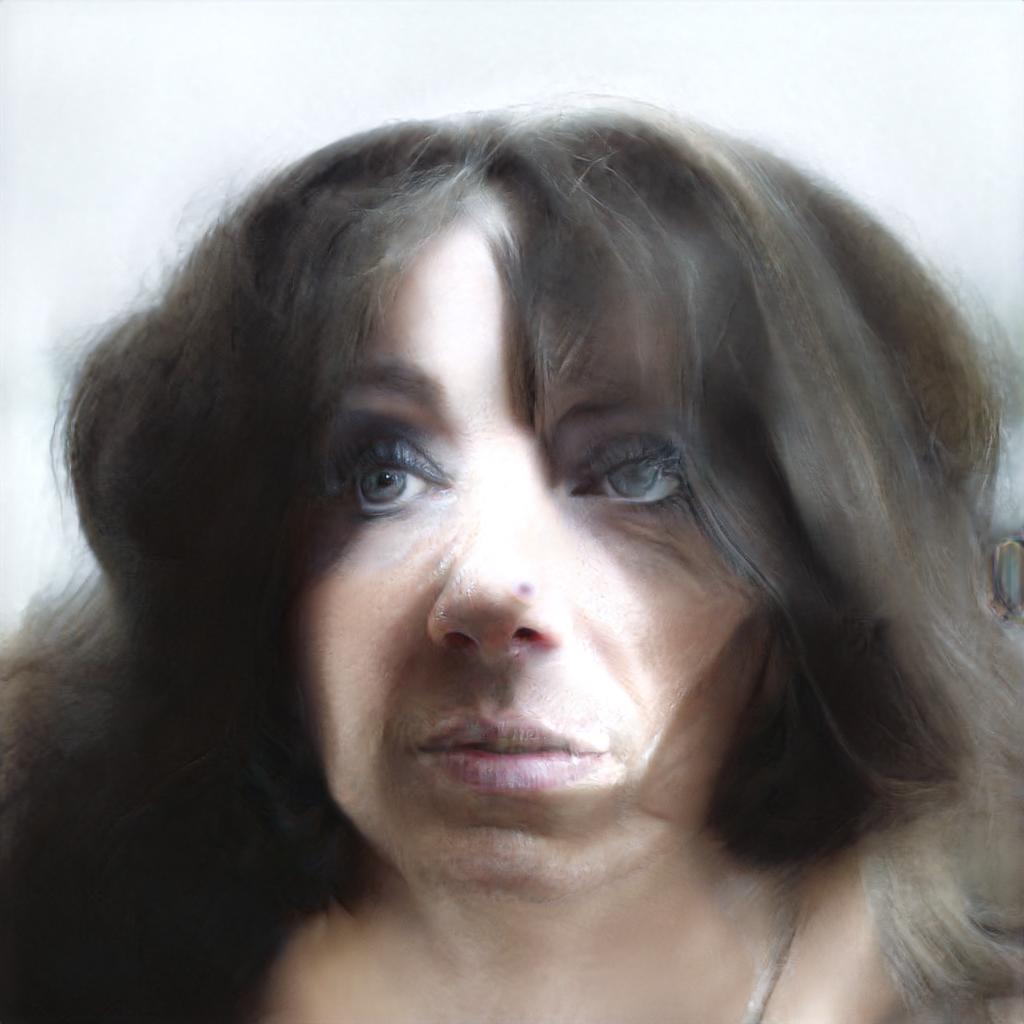} &
				\includegraphics[width=.059\linewidth]{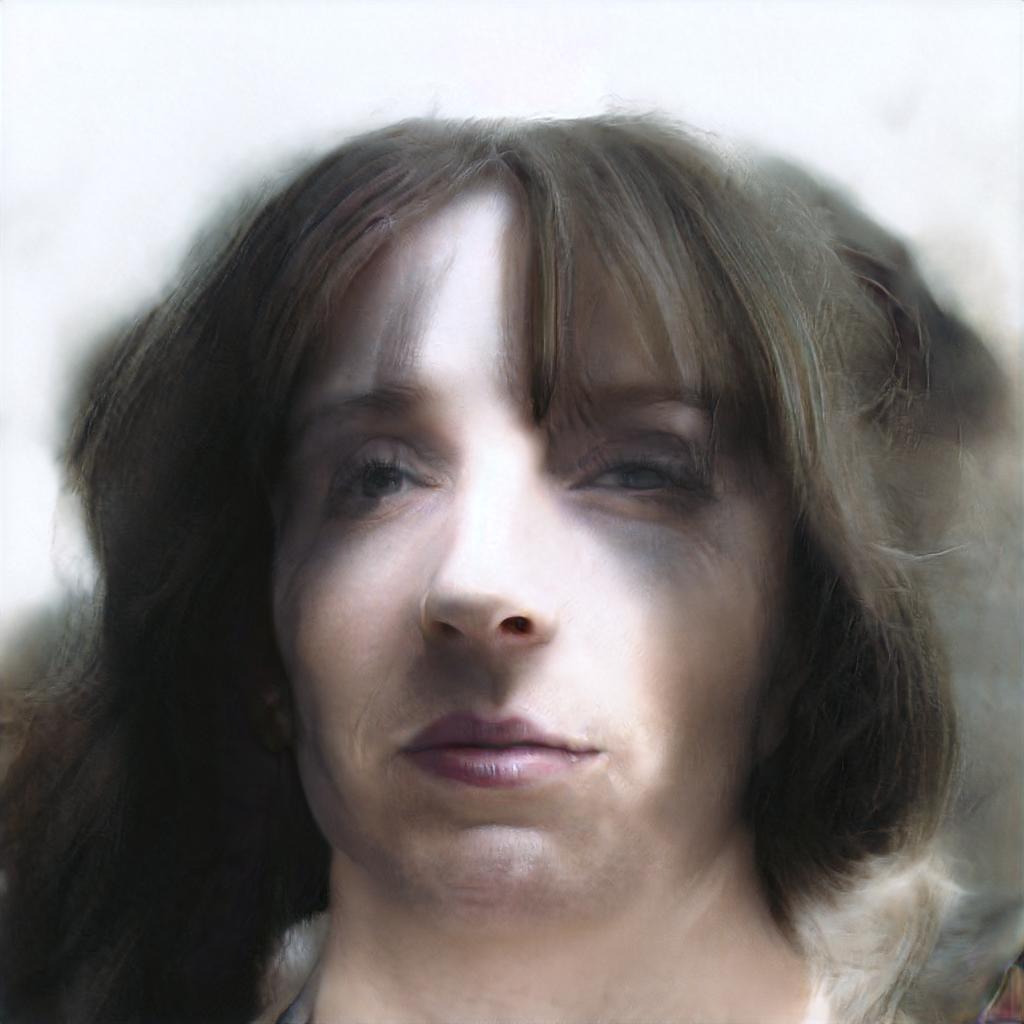} &
				\hspace{0.1mm}
				\includegraphics[width=.059\linewidth]{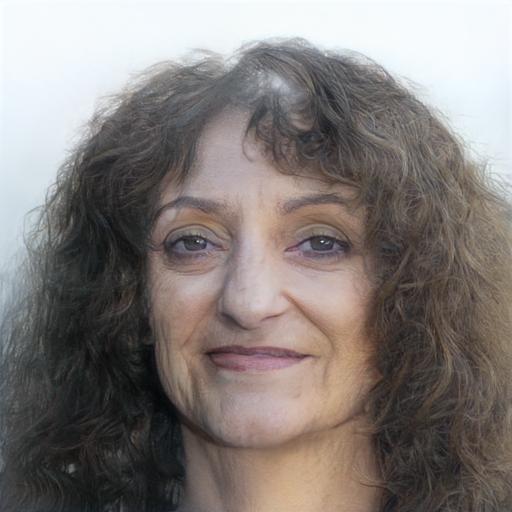} &
				\includegraphics[width=.059\linewidth]{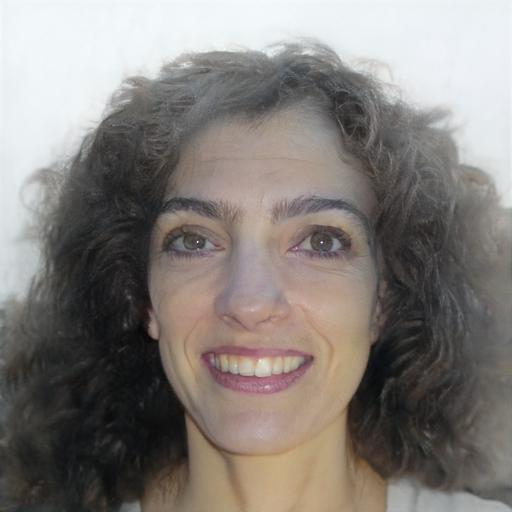} &
				\includegraphics[width=.059\linewidth]{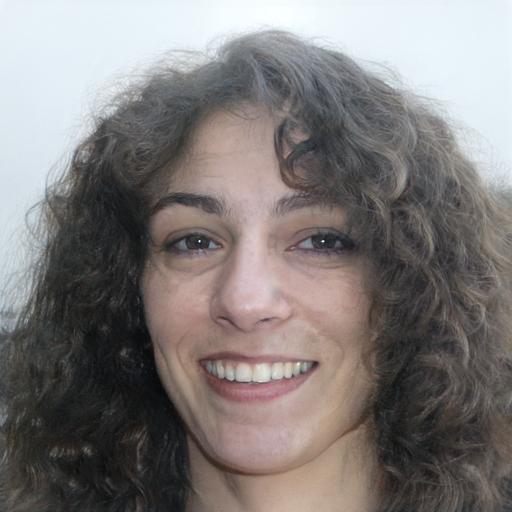}
				\\
								
				\includegraphics[width=.059\linewidth]{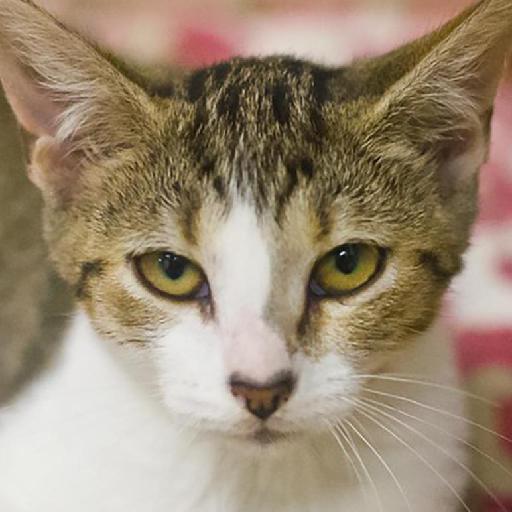} &
				\hspace{0.1mm}
				\includegraphics[width=.059\linewidth]{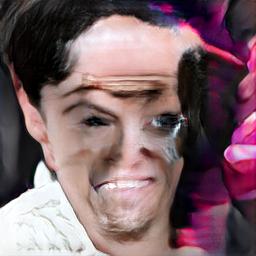} &
				\includegraphics[width=.059\linewidth]{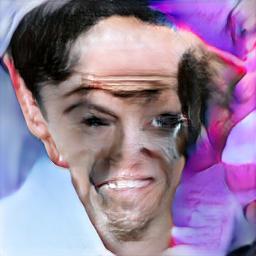} &
				\includegraphics[width=.059\linewidth]{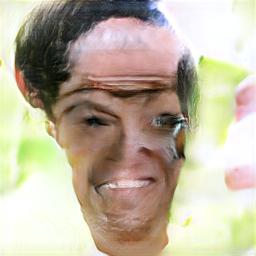} &
				\hspace{0.1mm}
				\includegraphics[width=.059\linewidth]{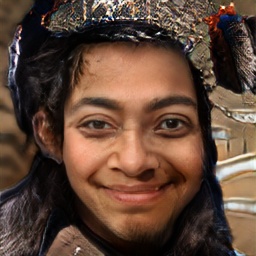} &
				\includegraphics[width=.059\linewidth]{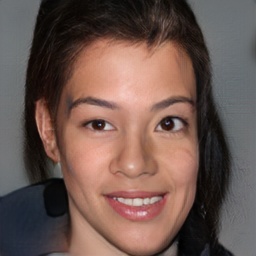} &
				\includegraphics[width=.059\linewidth]{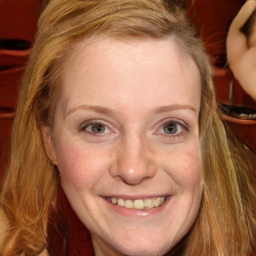} &
				\hspace{0.1mm}
				\includegraphics[width=.059\linewidth]{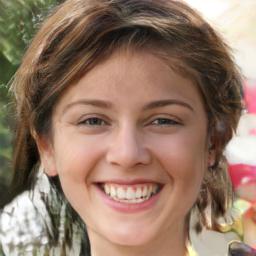} &
				\includegraphics[width=.059\linewidth]{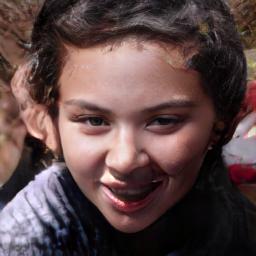} &
				\includegraphics[width=.059\linewidth]{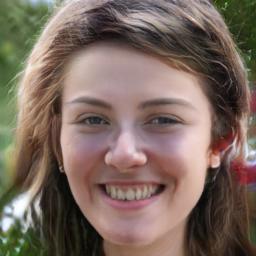} &
				\hspace{0.1mm}
				\includegraphics[width=.059\linewidth]{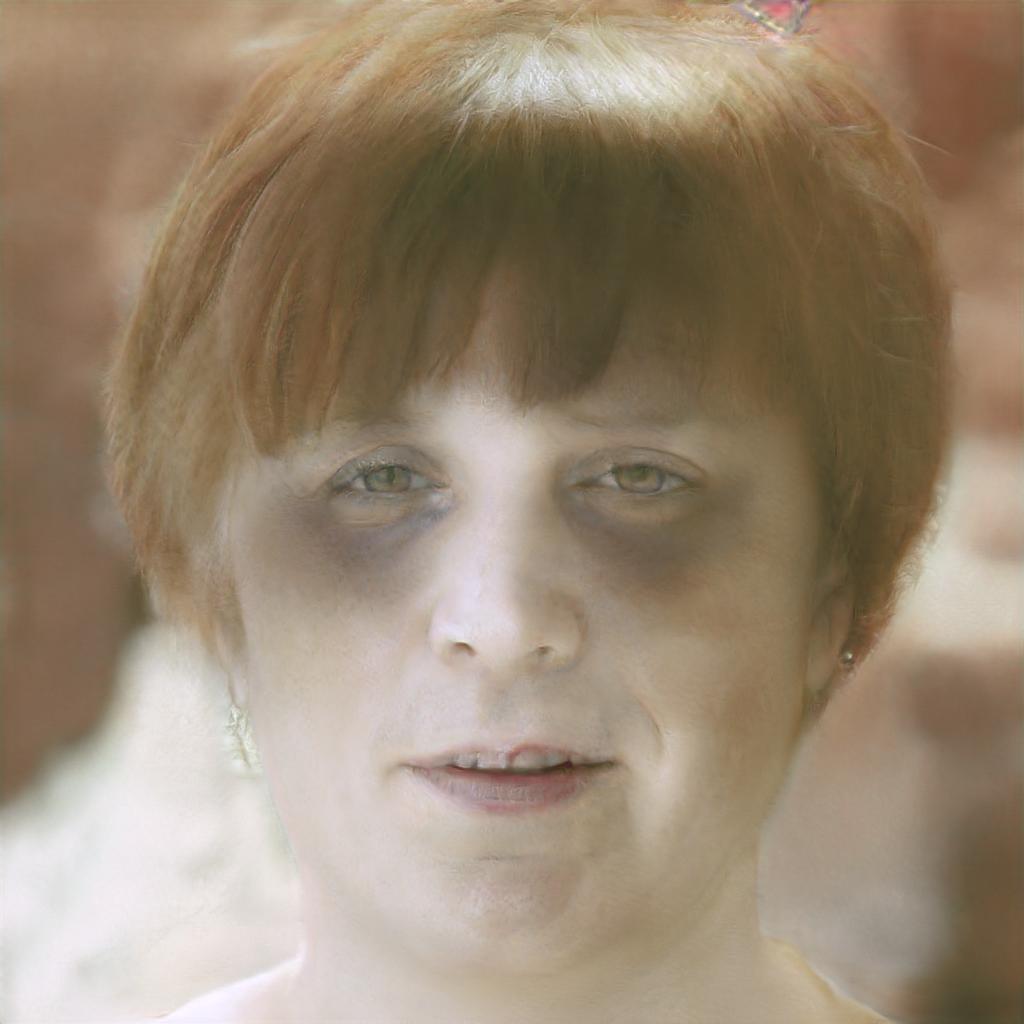} &
				\includegraphics[width=.059\linewidth]{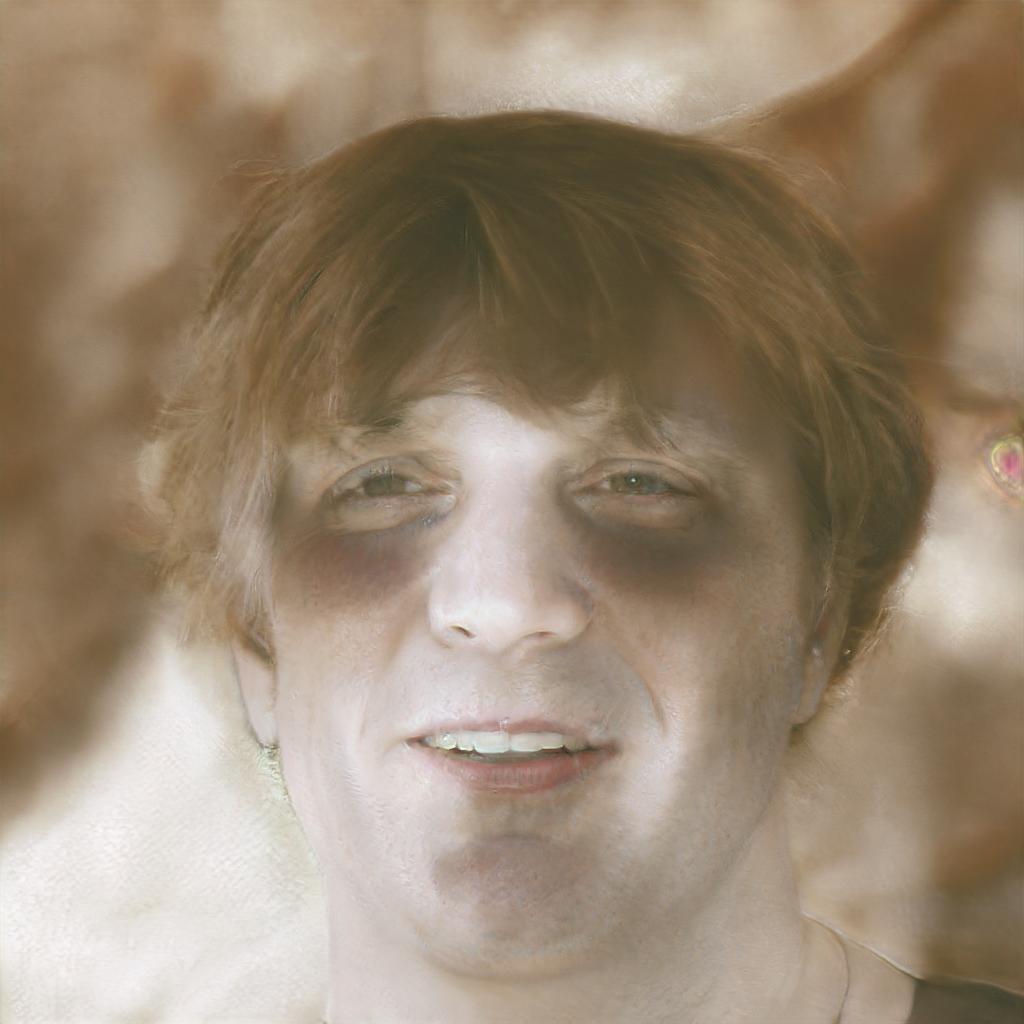} &
				\includegraphics[width=.059\linewidth]{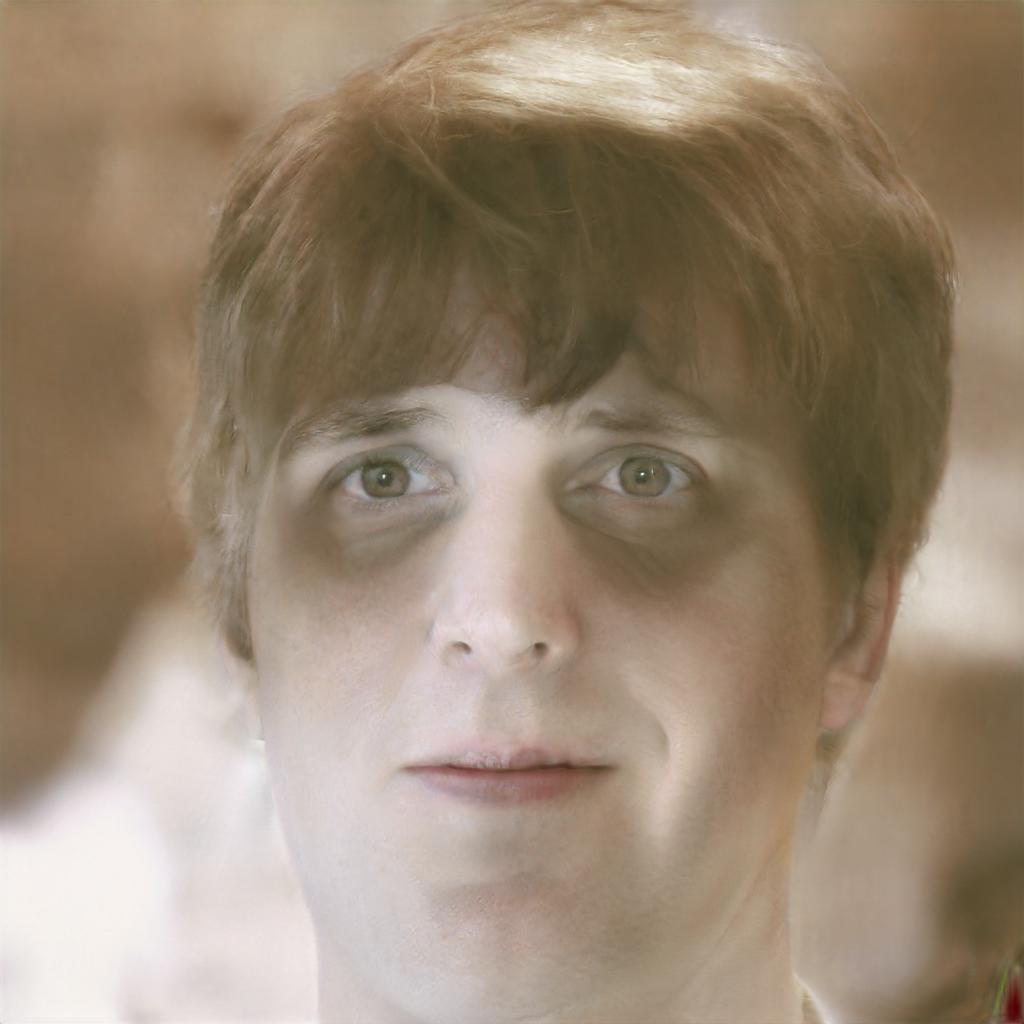} &
				\hspace{0.1mm}
				\includegraphics[width=.059\linewidth]{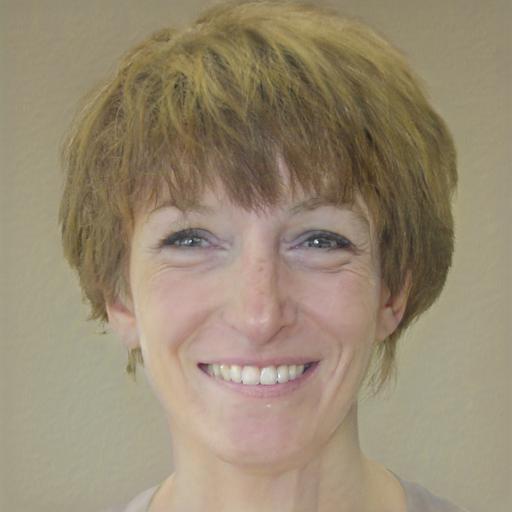} &
				\includegraphics[width=.059\linewidth]{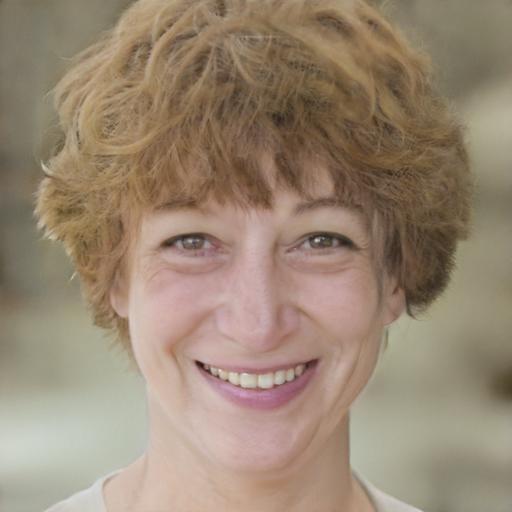} &
				\includegraphics[width=.059\linewidth]{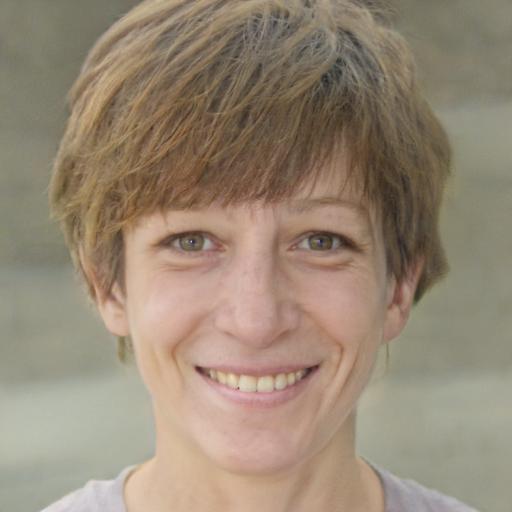}
				\\
		Source & \multicolumn{3}{c}{VQ-I2I} 
		&\multicolumn{3}{c}{GP-UNIT} 
		&\multicolumn{3}{c}{StarGAN2}	
		& \multicolumn{3}{c}{PULSE}
		& \multicolumn{3}{c}{UniTranslator (\textbf{ours})}\\
		Image&\multicolumn{3}{c}{\cite{chen2022eccv}}&\multicolumn{3}{c}{\cite{yang2022unsupervised}}&\multicolumn{3}{c}{\cite{choi2020stargan}}&\multicolumn{3}{c}{\cite{menon2020pulse}}
	\end{tabular}
	\caption{Diversity comparison. UniTranslator generates diverse and sensible results through multiple inferences with a single input, while alternative methods often produce single-modal or lower-quality outputs. First row: Metfaces$\to$FFHQ; Second row:  Metfaces$\to$FFHQ; Third row: AFHQ-wild$\to$FFHQ; Fourth row: AFHQ-dog$\to$FFHQ; Fifth row: AFHQ-dog$\to$FFHQ; Sixth row: AFHQ-dog$\to$FFHQ; Seventh row: AFHQ-cat$\to$FFHQ.}	
	\label{fig:diversity_supp}
\end{figure*}

\begin{figure*}[t]
	\centering
	\setlength{\abovecaptionskip}{0cm}
	\centering
	\setlength{\tabcolsep}{0.05em}
	\setlength{\fboxrule}{1pt}
	\setlength{\fboxsep}{0pt}
	\begin{tabular}{cccccccc}
		
		\includegraphics[width=.12\linewidth]{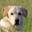} &
		\includegraphics[width=.12\linewidth]{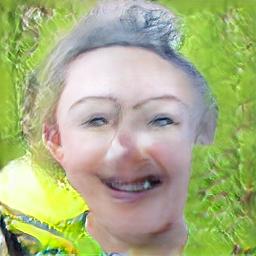} &
		\includegraphics[width=.12\linewidth]{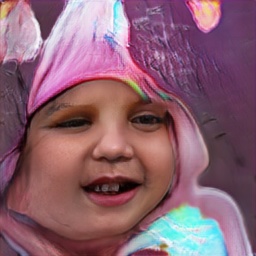} &
		\includegraphics[width=.12\linewidth]{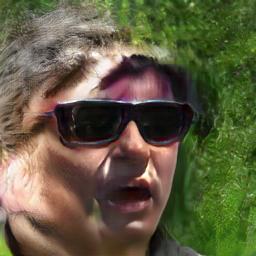} &
		\includegraphics[width=.12\linewidth]{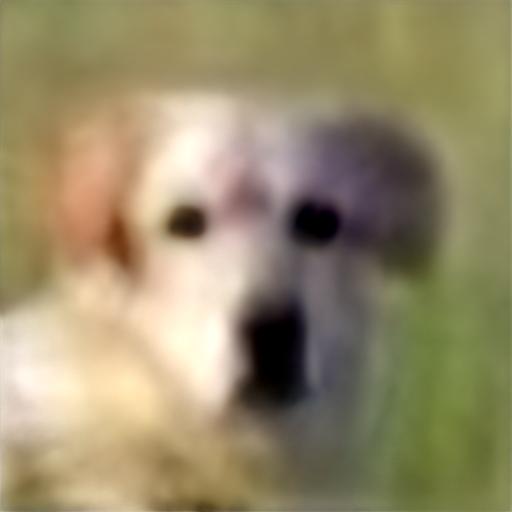} &
		\includegraphics[width=.12\linewidth]{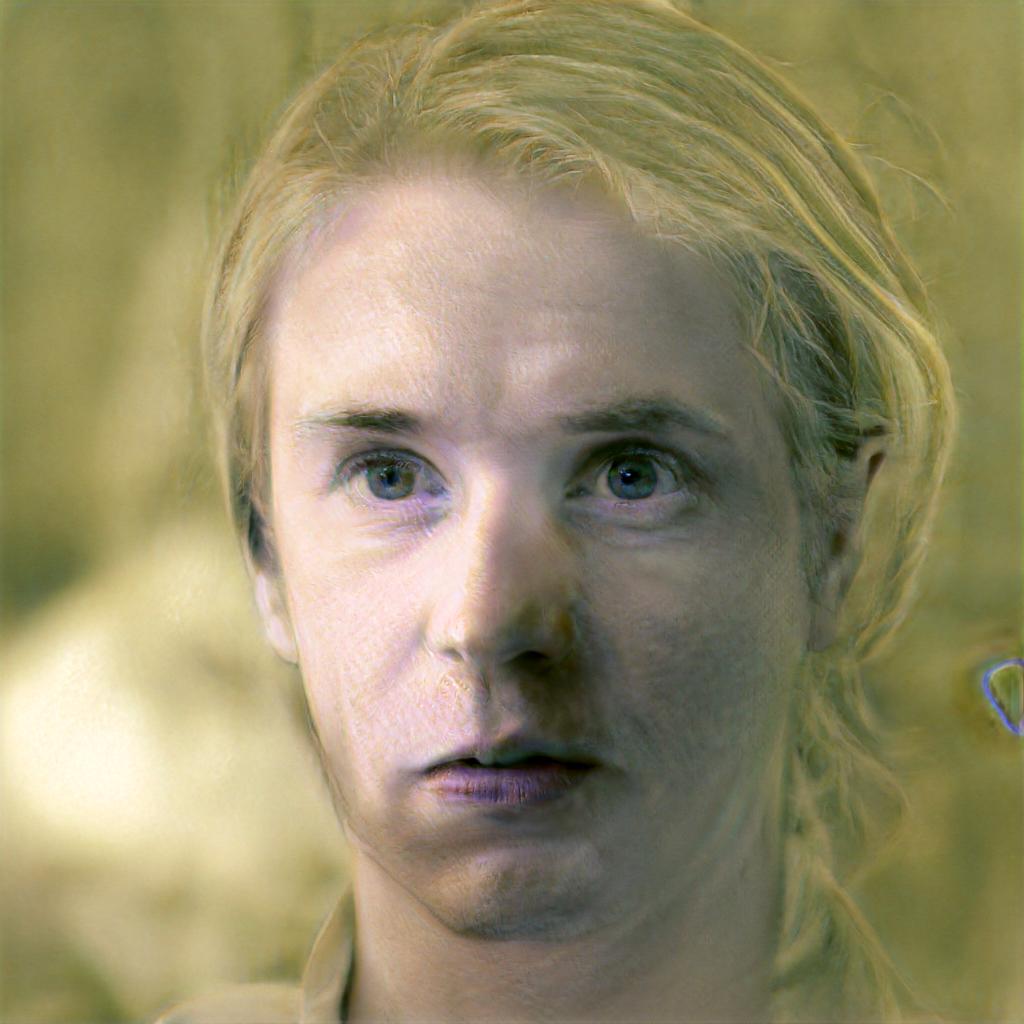} &
		\includegraphics[width=.12\linewidth]{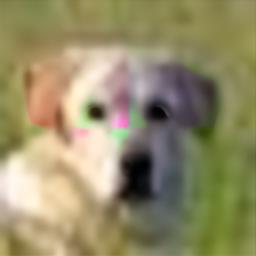} &
		\includegraphics[width=.12\linewidth]{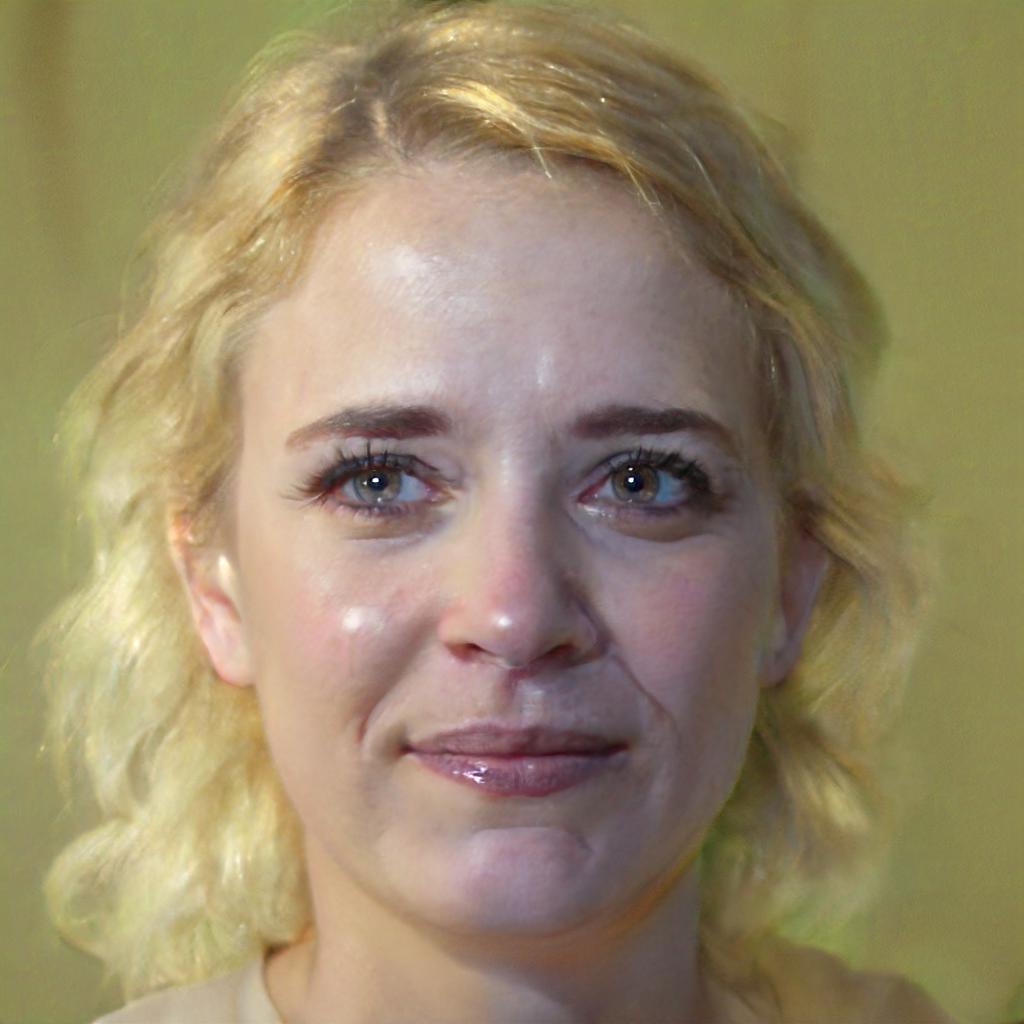}\\	
		
		\includegraphics[width=.12\linewidth]{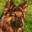} &
		\includegraphics[width=.12\linewidth]{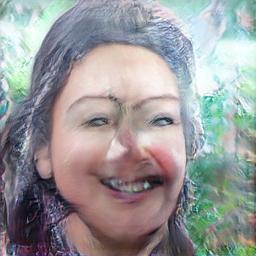} &
		\includegraphics[width=.12\linewidth]{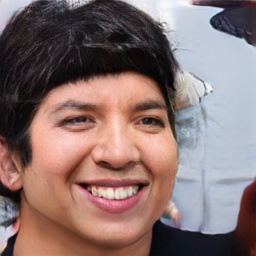} &
		\includegraphics[width=.12\linewidth]{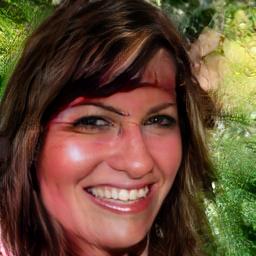} &
		\includegraphics[width=.12\linewidth]{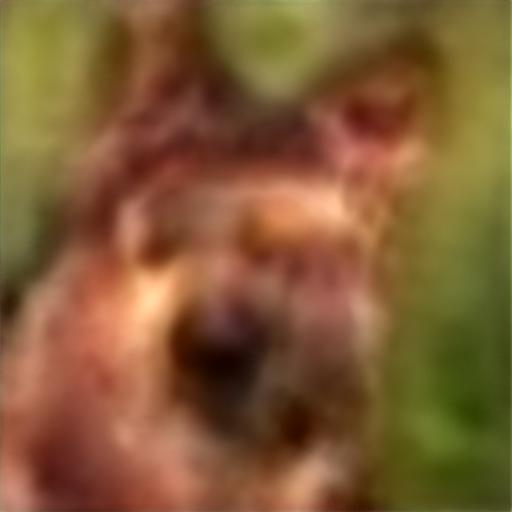} &
		\includegraphics[width=.12\linewidth]{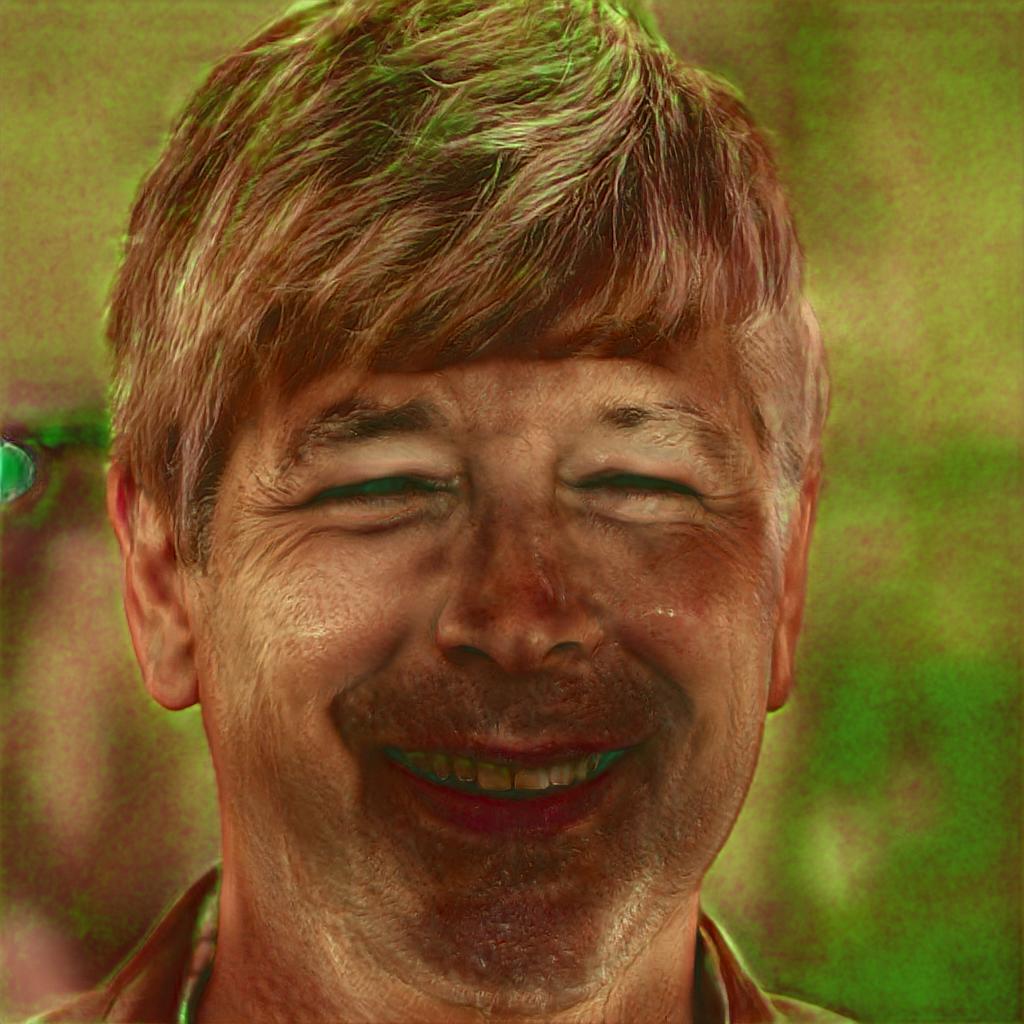} &
		\includegraphics[width=.12\linewidth]{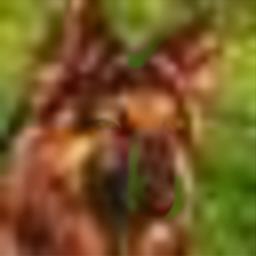} &
		\includegraphics[width=.12\linewidth]{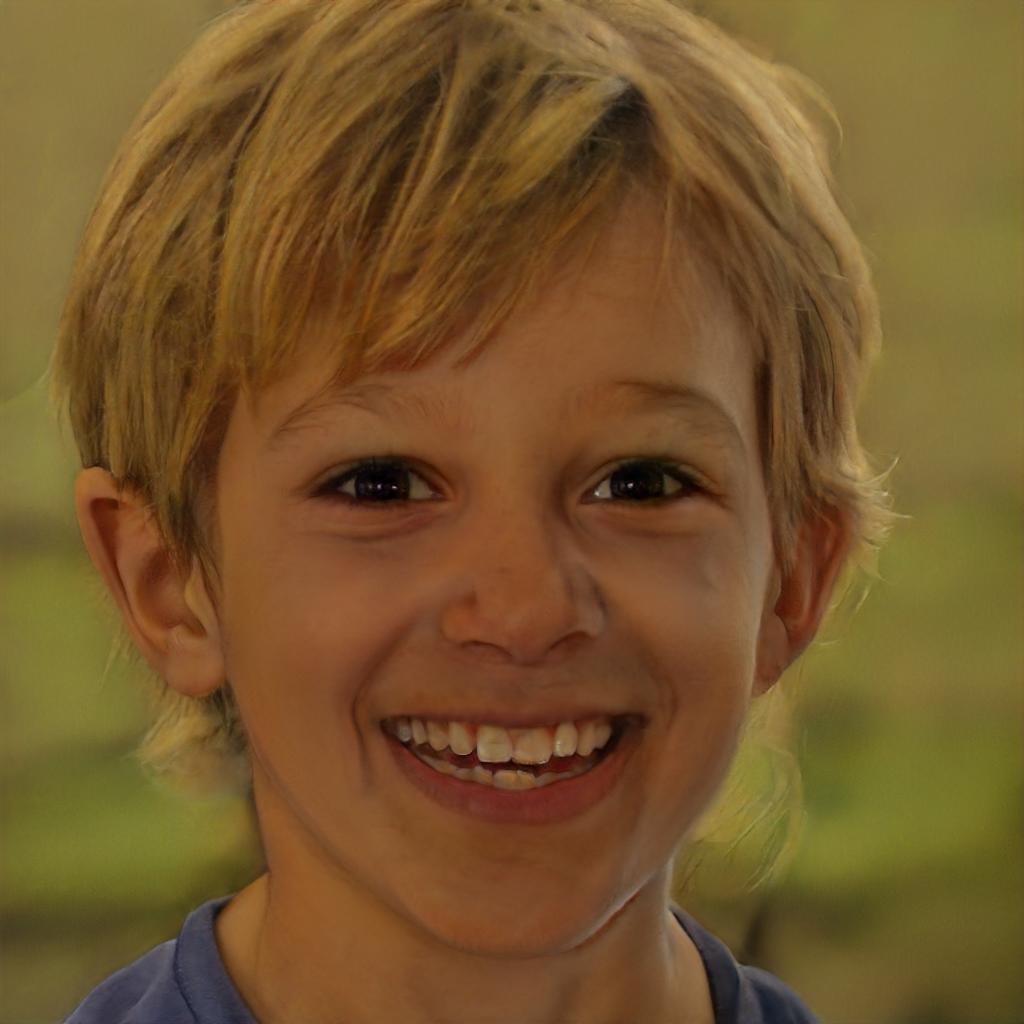}\\	
		
		\includegraphics[width=.12\linewidth]{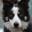} &
		\includegraphics[width=.12\linewidth]{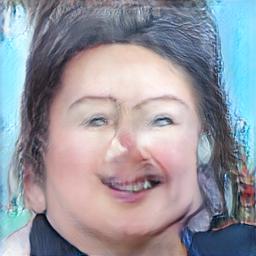} &
		\includegraphics[width=.12\linewidth]{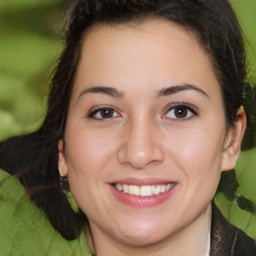} &
		\includegraphics[width=.12\linewidth]{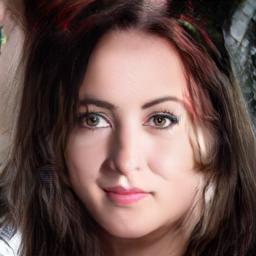} &
		\includegraphics[width=.12\linewidth]{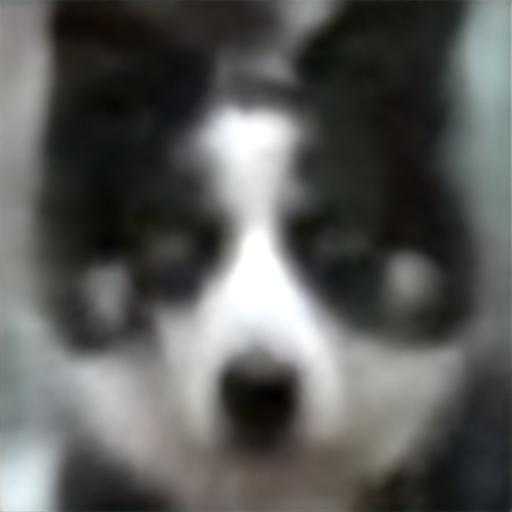} &
		\includegraphics[width=.12\linewidth]{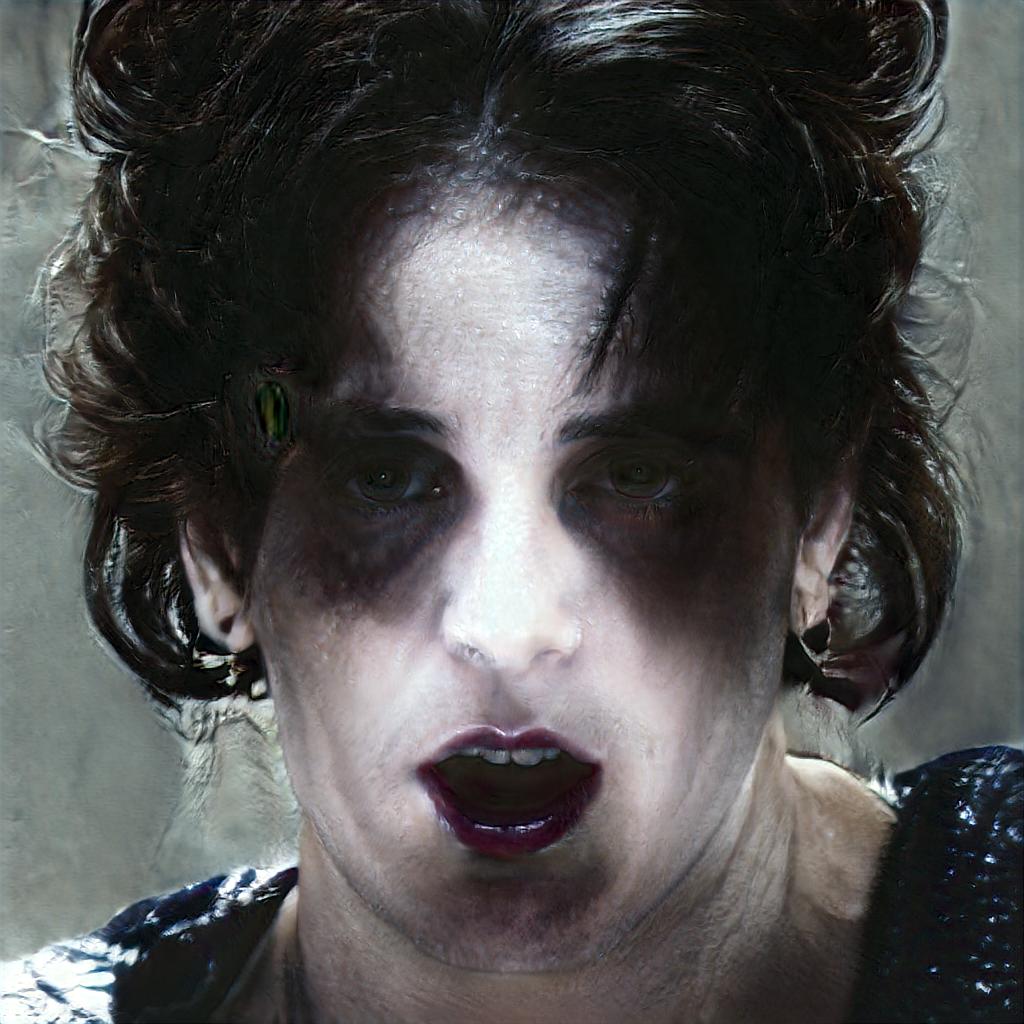} &
		\includegraphics[width=.12\linewidth]{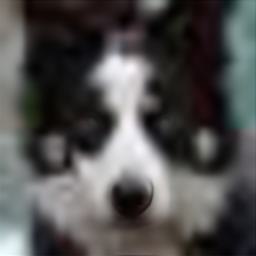} &
		\includegraphics[width=.12\linewidth]{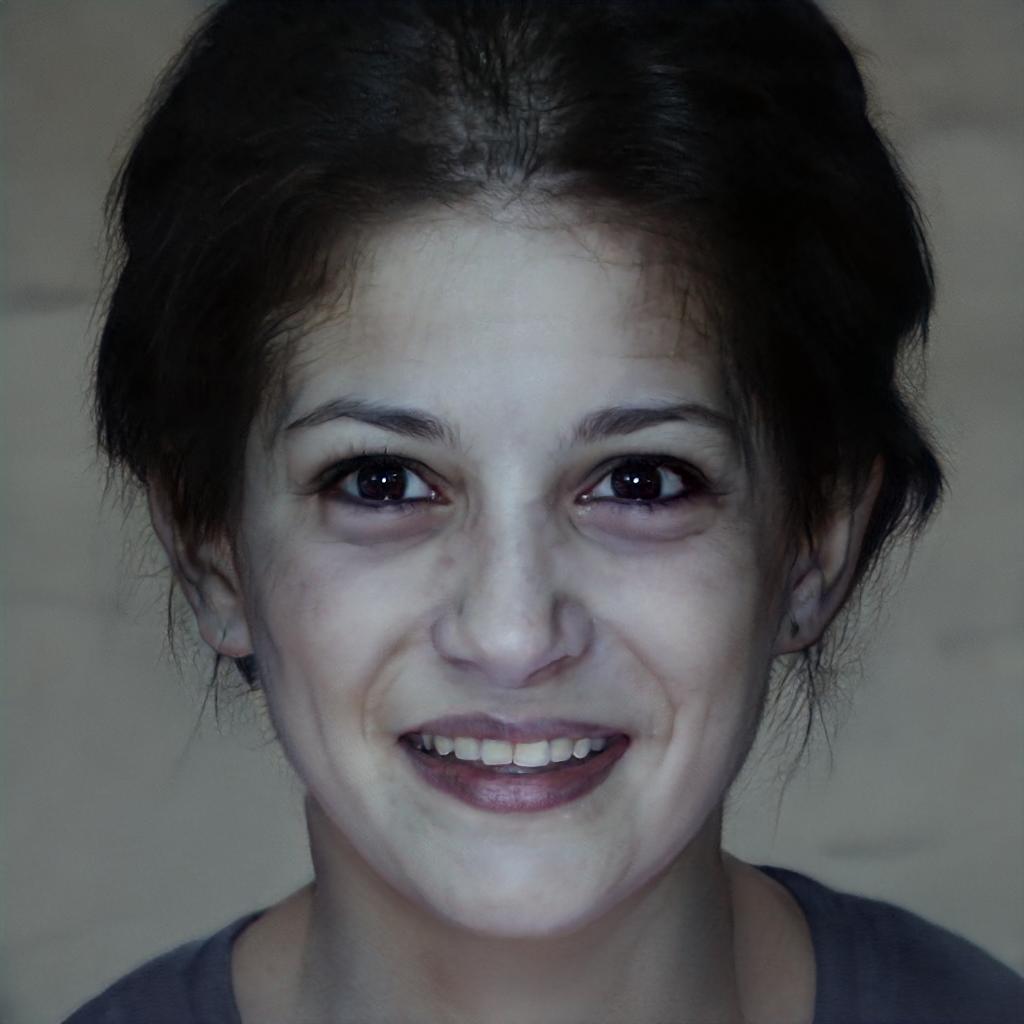}\\

		\includegraphics[width=.12\linewidth]{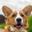} &
		\includegraphics[width=.12\linewidth]{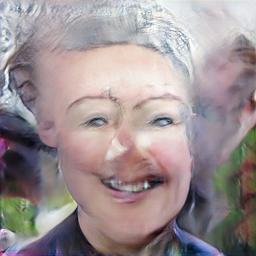} &
		\includegraphics[width=.12\linewidth]{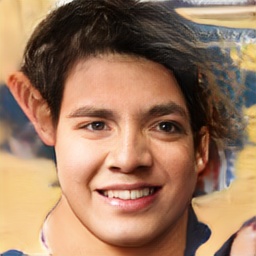} &
		\includegraphics[width=.12\linewidth]{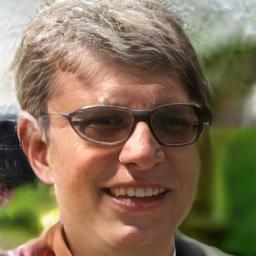} &
		\includegraphics[width=.12\linewidth]{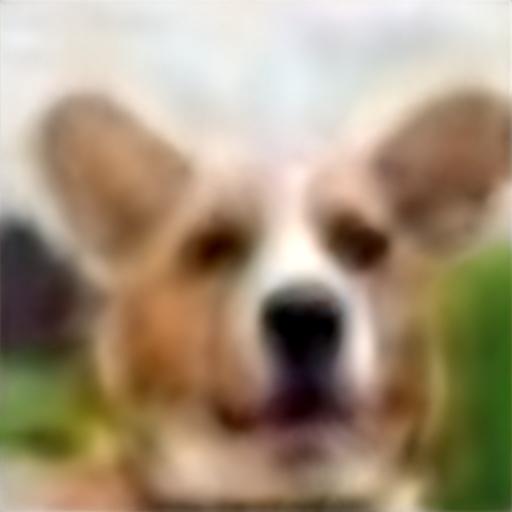} &
		\includegraphics[width=.12\linewidth]{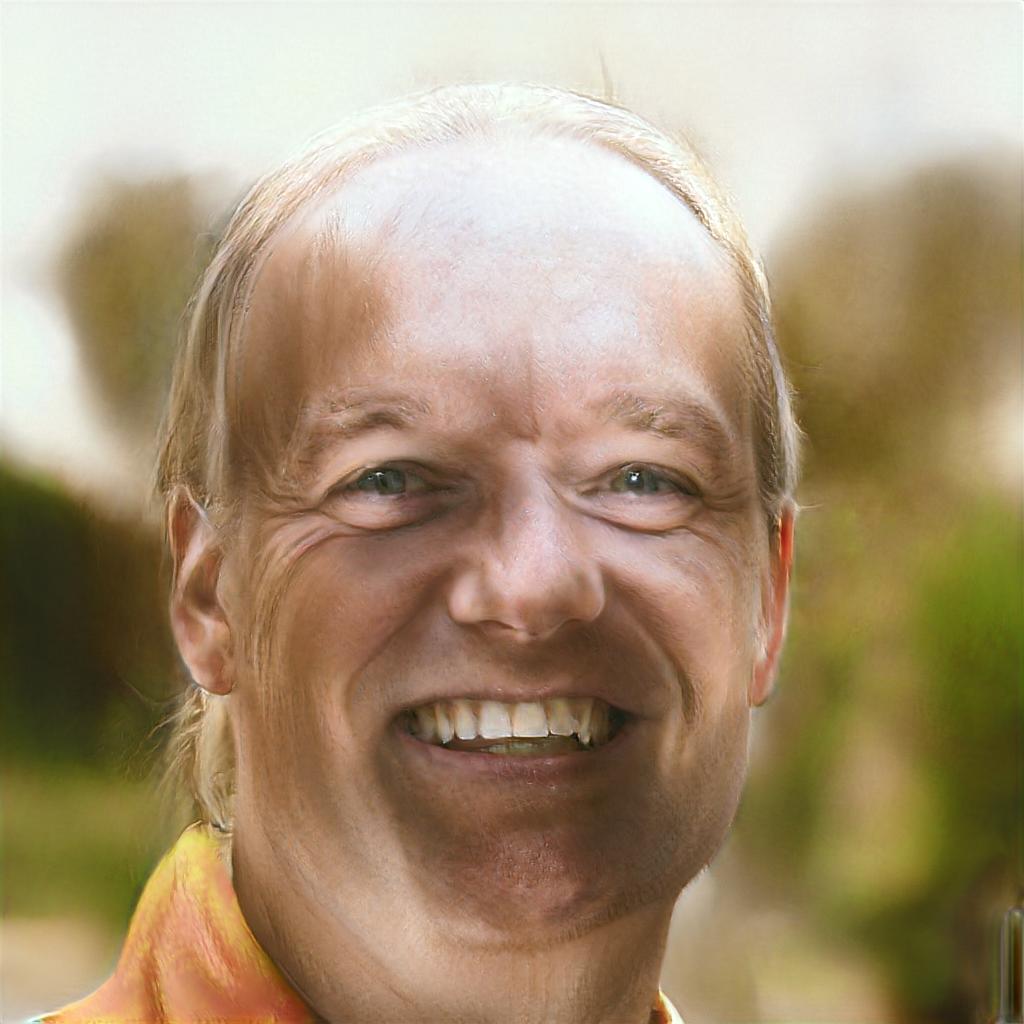} &
		\includegraphics[width=.12\linewidth]{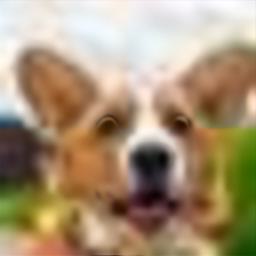} &
		\includegraphics[width=.12\linewidth]{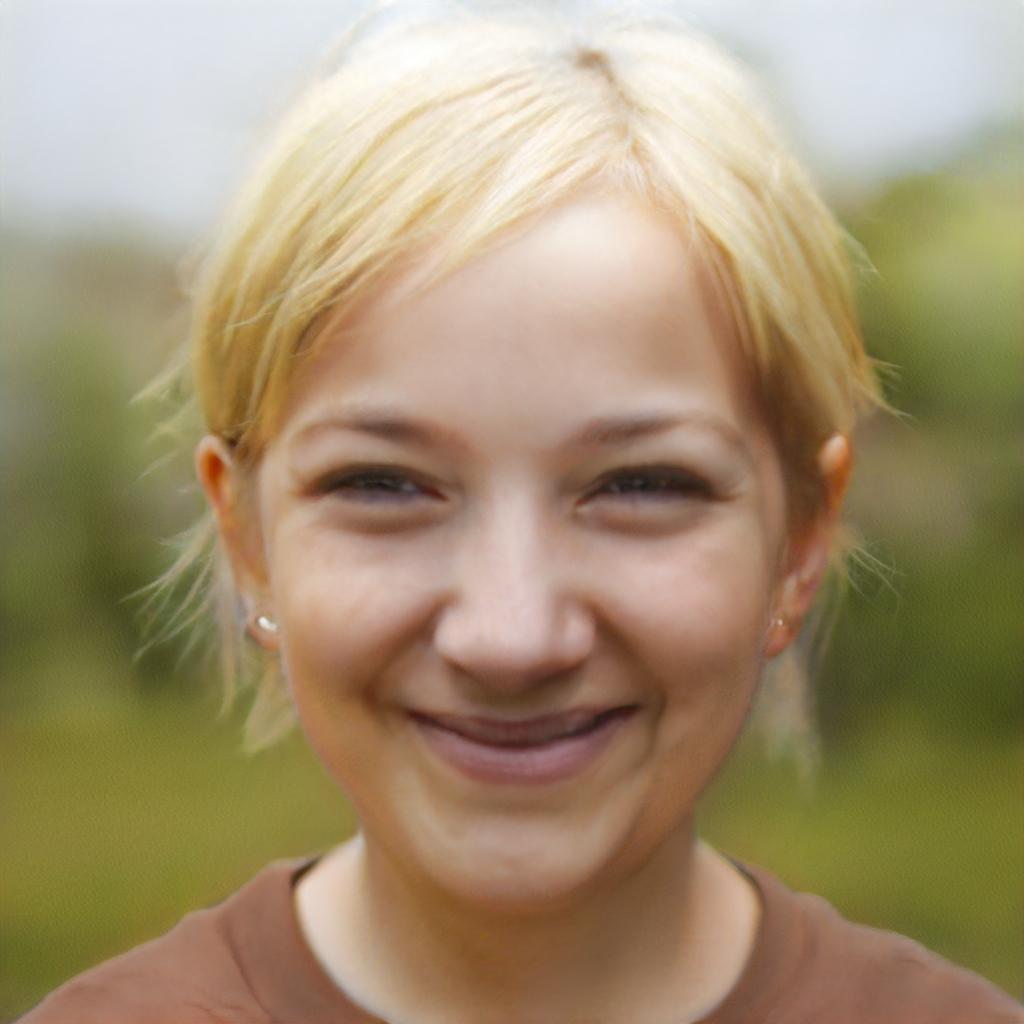}\\	
		
		Source Image& VQ-I2I&GP-UNIT &StarGAN2 & DiFa &PULSE& DiffusionCLIP &UniTranslator\\ 
		LR, 32$\times$32&8$\times$&8$\times$&8$\times$&16$\times$&32$\times$&8$\times$&(\textbf{ours}, 32$\times)$
	\end{tabular}
	\caption{More results of AFHQ-dog$\to$FFHQ with low-resolution input scenarios.	
	}
	\label{fig:super-resolution_supp}
\end{figure*}

\begin{figure*}[t]
	\centering
	\setlength{\abovecaptionskip}{0cm}
	\centering
	\setlength{\tabcolsep}{0.05em}
	\setlength{\fboxrule}{1pt}
	\setlength{\fboxsep}{0pt}
	\begin{tabular}{cccccccc}
				
		\includegraphics[width=.12\linewidth]{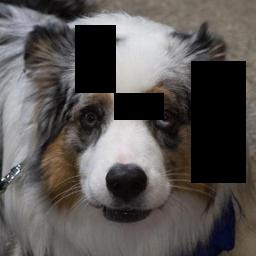} &
		\includegraphics[width=.12\linewidth]{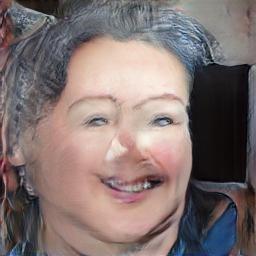} &
		\includegraphics[width=.12\linewidth]{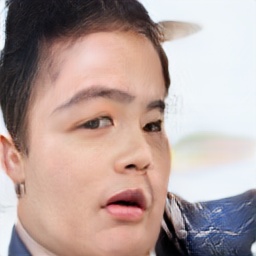} &
		\includegraphics[width=.12\linewidth]{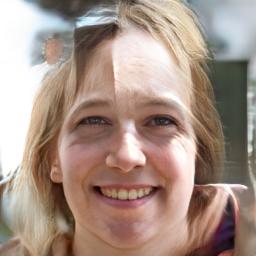} &
		\includegraphics[width=.12\linewidth]{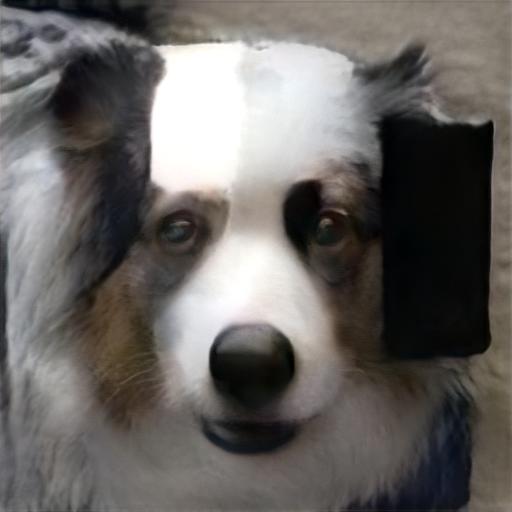} &
		\includegraphics[width=.12\linewidth]{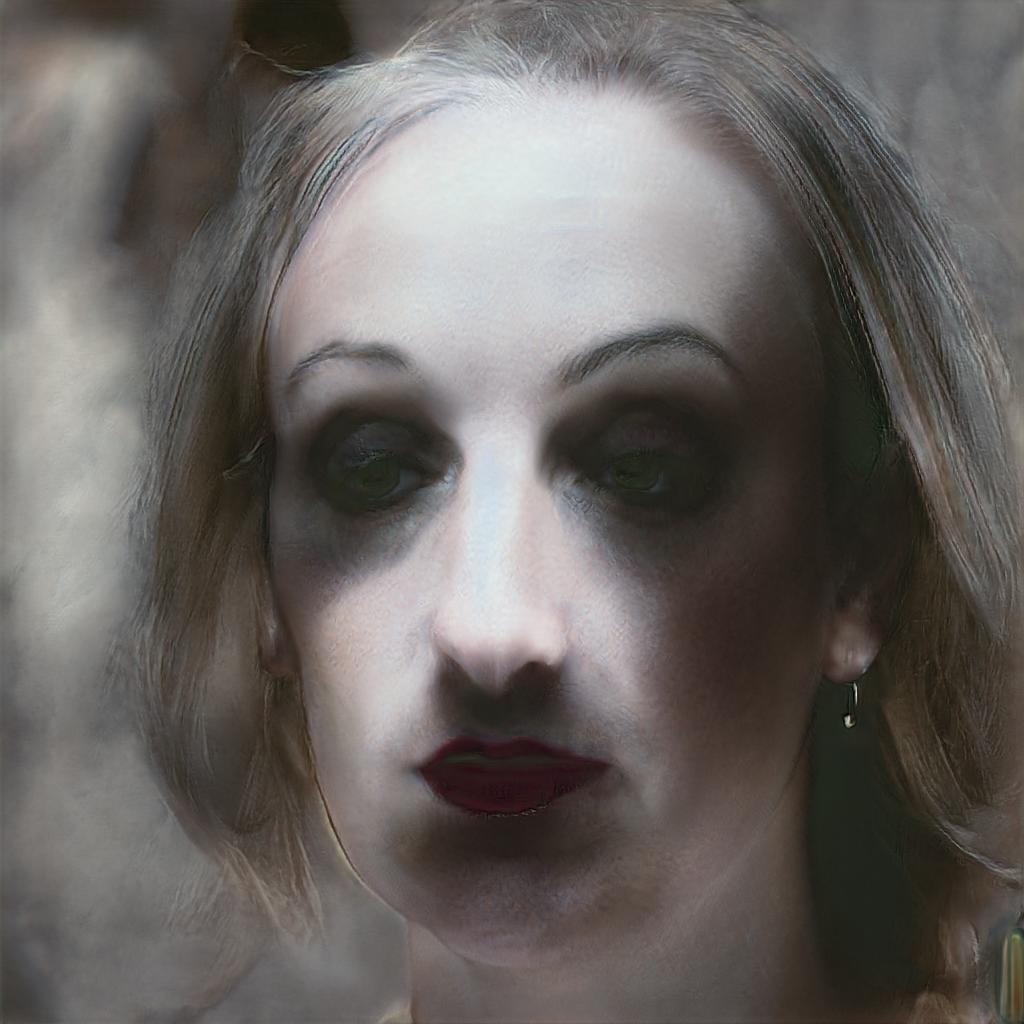} &
		\includegraphics[width=.12\linewidth]{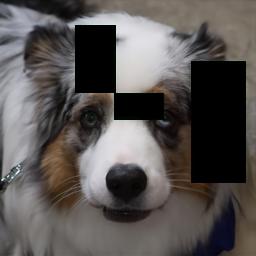} &
		\includegraphics[width=.12\linewidth]{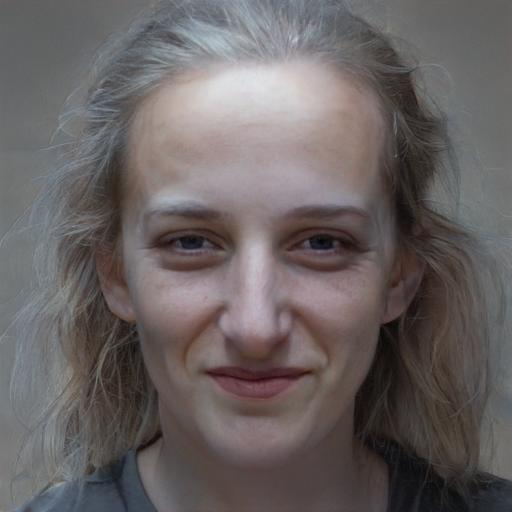}
		\\	
		
		\includegraphics[width=.12\linewidth]{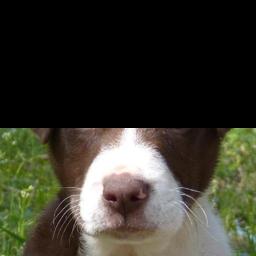} &
		\includegraphics[width=.12\linewidth]{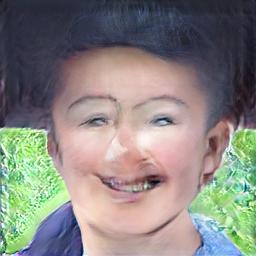} &
		\includegraphics[width=.12\linewidth]{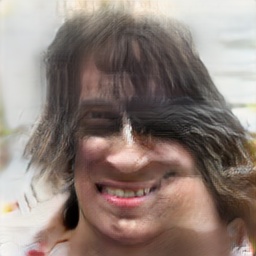} &
		\includegraphics[width=.12\linewidth]{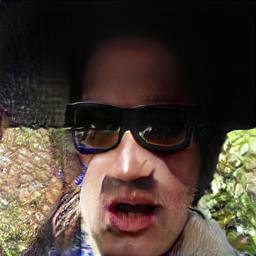} &
		\includegraphics[width=.12\linewidth]{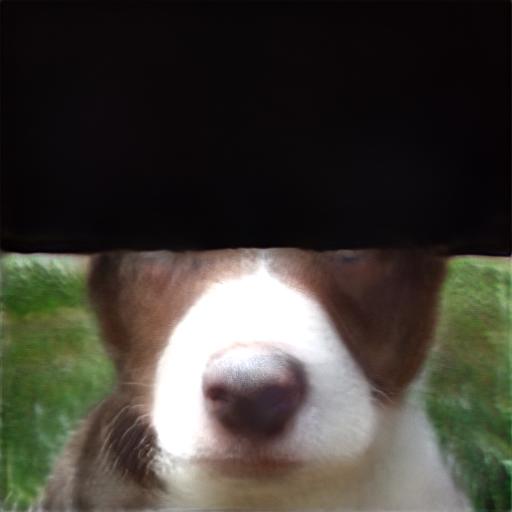} &
		\includegraphics[width=.12\linewidth]{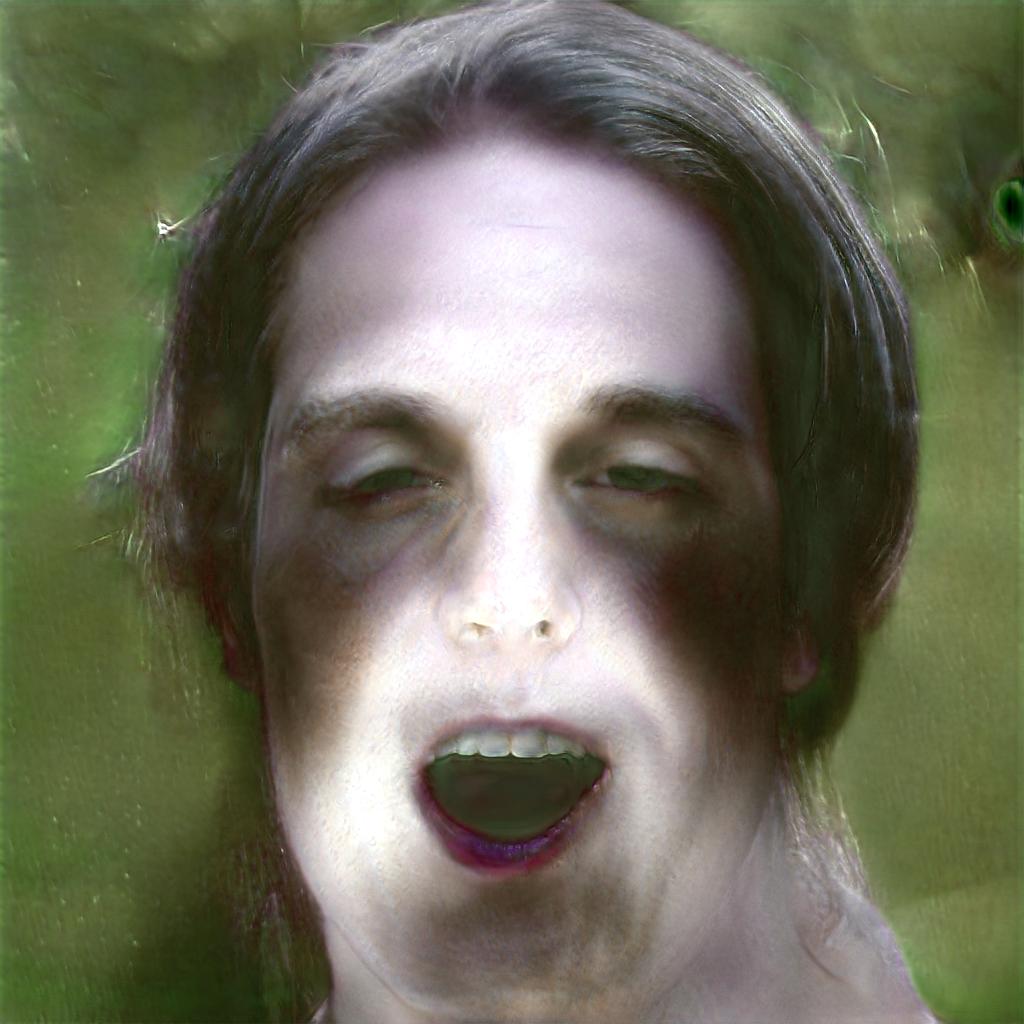} &
		\includegraphics[width=.12\linewidth]{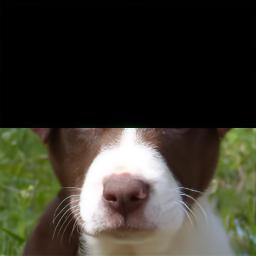} &
		\includegraphics[width=.12\linewidth]{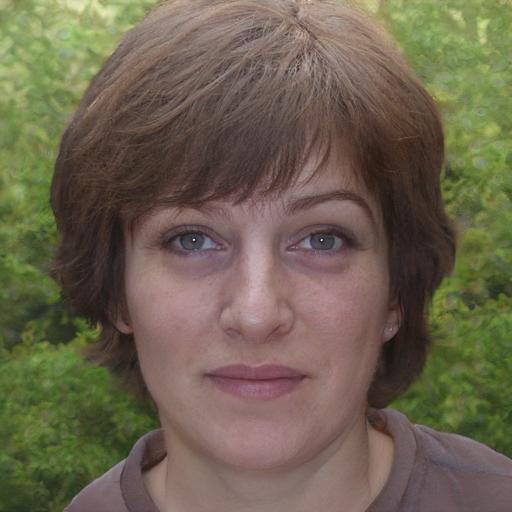}
		\\
		
		\includegraphics[width=.12\linewidth]{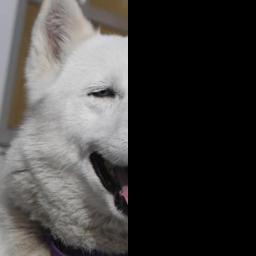} &
		\includegraphics[width=.12\linewidth]{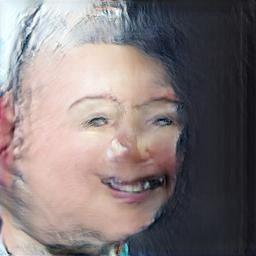} &
		\includegraphics[width=.12\linewidth]{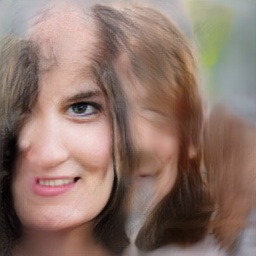} &
		\includegraphics[width=.12\linewidth]{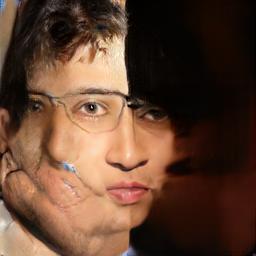} &
		\includegraphics[width=.12\linewidth]{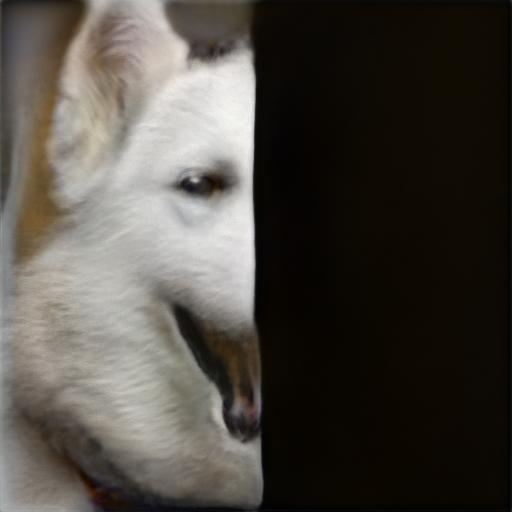} &
		\includegraphics[width=.12\linewidth]{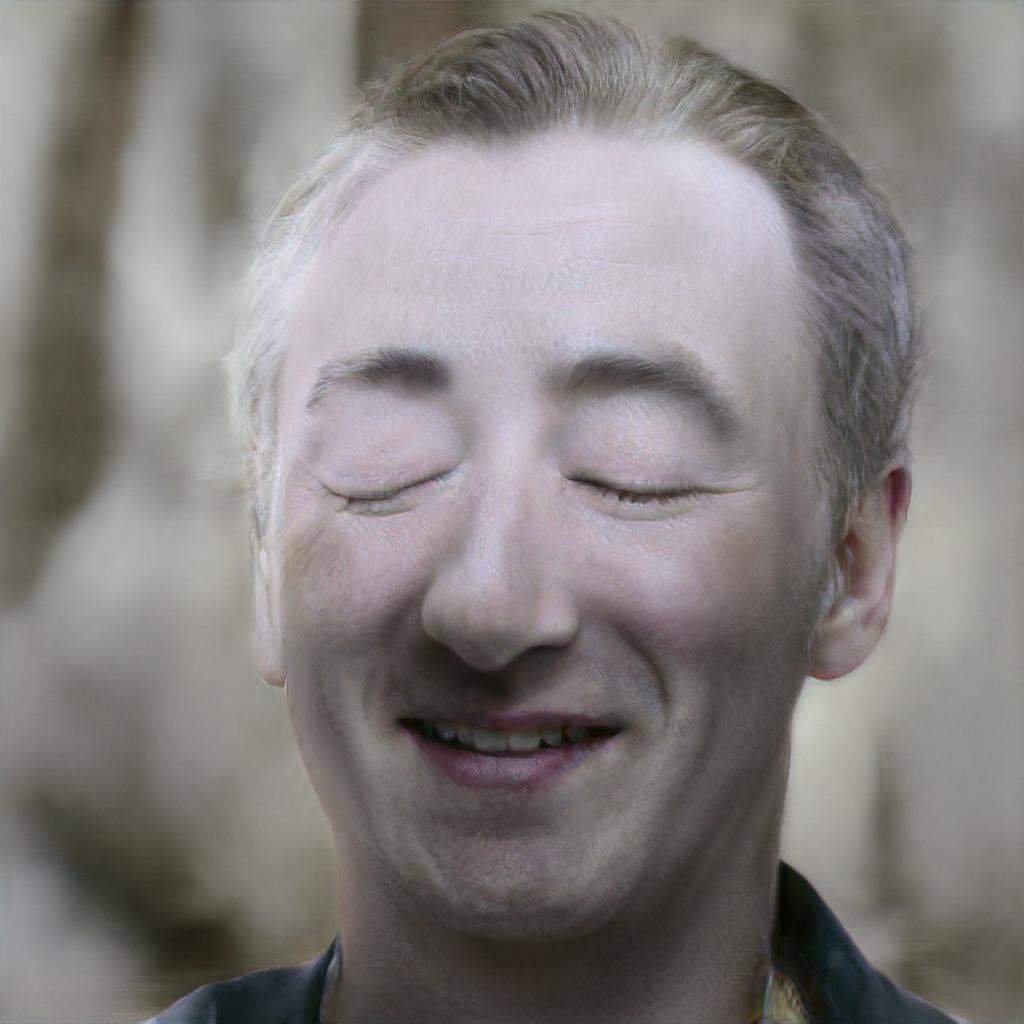} &
		\includegraphics[width=.12\linewidth]{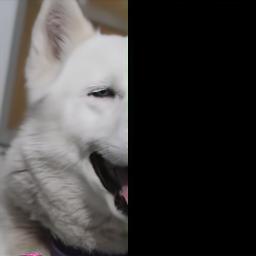} &
		\includegraphics[width=.12\linewidth]{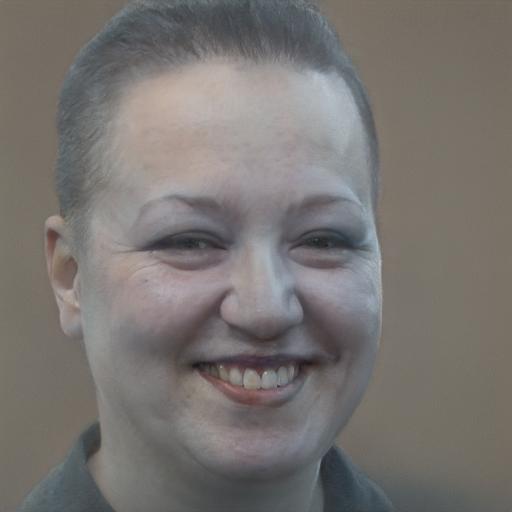}
		\\
		
		\includegraphics[width=.12\linewidth]{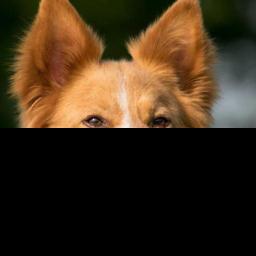} &
		\includegraphics[width=.12\linewidth]{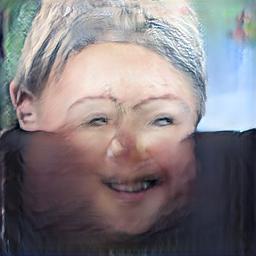} &
		\includegraphics[width=.12\linewidth]{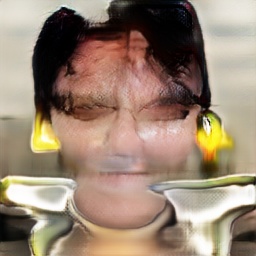} &
		\includegraphics[width=.12\linewidth]{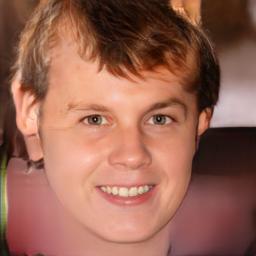} &
		\includegraphics[width=.12\linewidth]{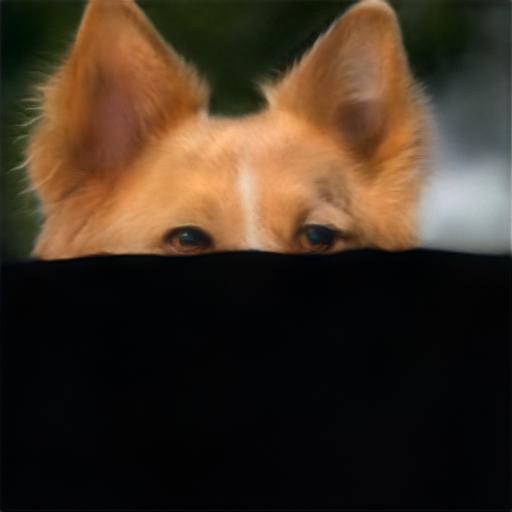} &
		\includegraphics[width=.12\linewidth]{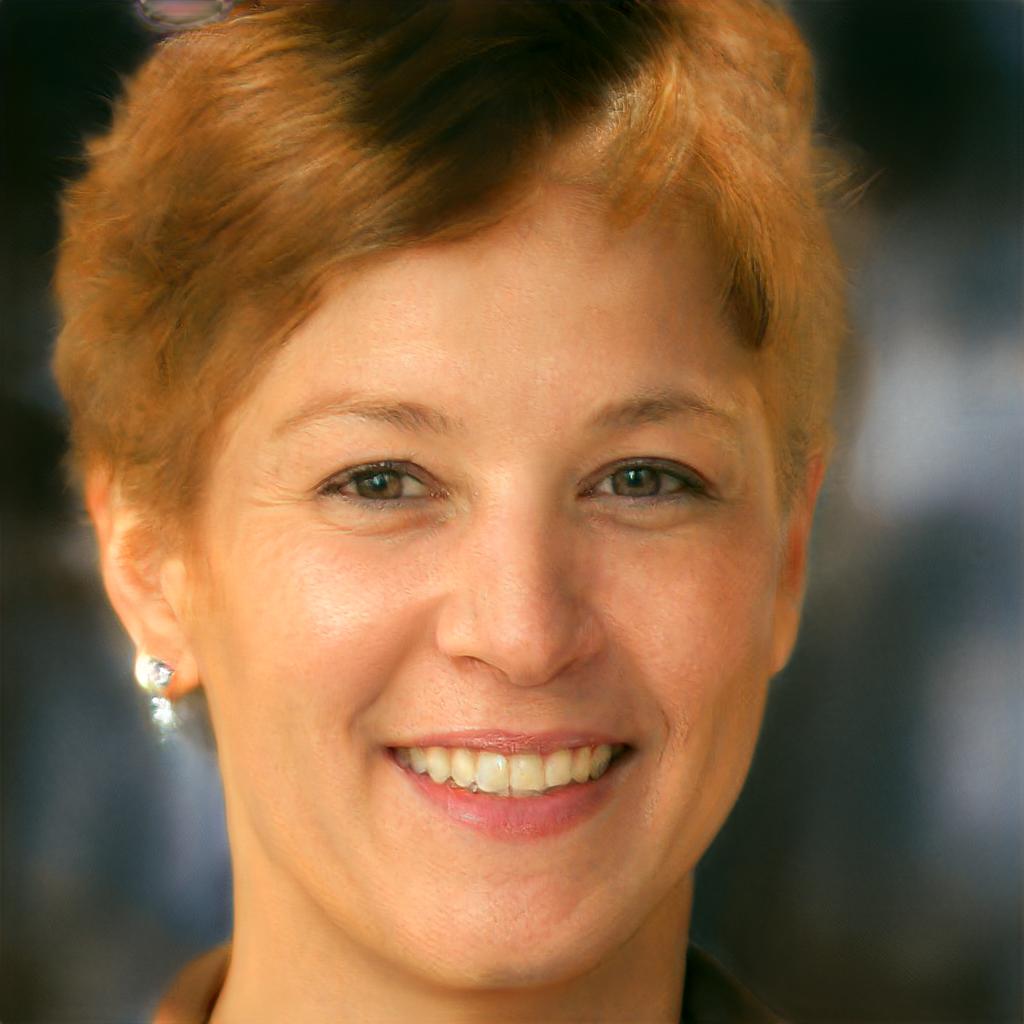} &
		\includegraphics[width=.12\linewidth]{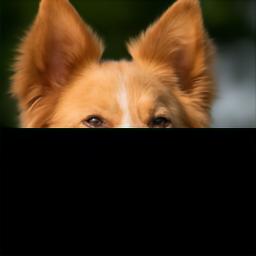} &
		\includegraphics[width=.12\linewidth]{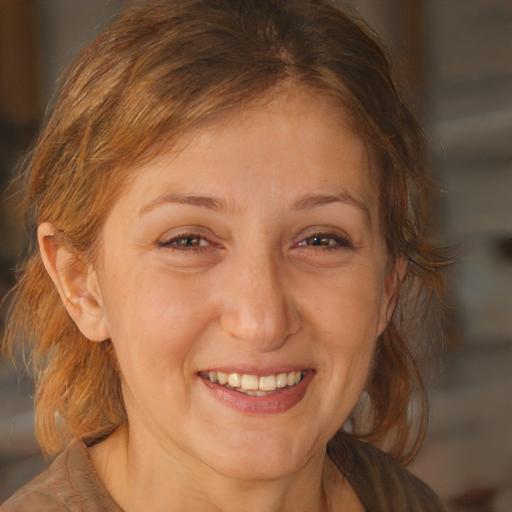}
		\\
		
		Source Image& VQ-I2I & GP-UNIT & StarGAN2 & DiFa & PULSE & DiffusionCLIP & UniTranslator\\
		&&&&&&&(\textbf{ours})
	\end{tabular}
	\caption{More results of AFHQ-dog$\to$FFHQ in corruption scenarios. 		
	}
	\label{fig:inpainting_supp}
\end{figure*}

Fig.~\ref{fig:diversity_supp} shows translated images with different diversity levels. Furthermore, Fig.~\ref{fig:super-resolution_supp} and Fig.~\ref{fig:inpainting_supp} show additional results in degraded scenarios, including low-resolution inputs and corrupted inputs.

\end{document}